\documentclass[12pt,a4paper]{report}
\pdfoutput=1
\usepackage[pdftex]{graphicx}
\usepackage[T1]{fontenc}
\usepackage[utf8]{inputenc}
\usepackage[pdfusetitle]{hyperref}
\usepackage{amsmath, amssymb, subcaption, multirow, breakcites, bm, xcolor}
\usepackage[top=1in, bottom=1in, left=1.25in, right=1.25in]{geometry}
\definecolor{darkblue}{rgb}{0, 0, 0.5}
\hypersetup{colorlinks=true, citecolor=darkblue, linkcolor=darkblue, urlcolor=darkblue}

\newcommand{\figref}[1]{Fig.~\ref{fig:#1}}
\newcommand{\tabref}[1]{Tab.~\ref{tab:#1}}
\newcommand{\equationref}[1]{Eq.~\ref{eq:#1}}
\newcommand{\sectionref}[1]{Section~\ref{sec:#1}}
\newcommand{\chapterref}[1]{Chapter~\ref{ch:#1}}
\newcommand{\appendixref}[1]{Appendix~\ref{ch:#1}}
\newcommand{\longcaption}[2]{\caption[#1]{#1 #2}}
\newcommand{\rot}[1]{\rotatebox{90}{#1}}
\newcommand{\vect}[1]{\mathbf{#1}}

\DeclareMathOperator*{\argmax}{arg\,max}

\DeclareMathOperator{\countfunction}{count}
\DeclareMathOperator{\stopgradient}{stop\_gradient}
\DeclareMathOperator{\roundfunction}{round}
\DeclareMathOperator{\me}{\mathrm{e}}

\graphicspath{{./figures/}}

\title{Vector representations of text data in deep learning}

\author{Karol Grzegorczyk}

\makeatletter
\let \entitle \@title
\let \shortauthor \@author
\makeatother

\def \pltitle {Wektorowe reprezentacje danych tekstowych\\w uczeniu głębokim}

\newif\ifenglish

\renewcommand{\maketitle}{
\begin{titlepage}
\begin{center}

	\vspace*{15mm}

	\includegraphics[width=0.1\linewidth]{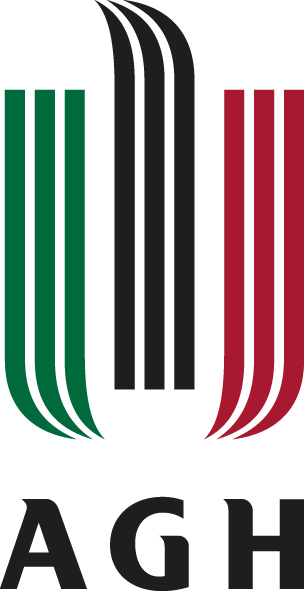}

	\vspace*{10mm}

	{\large
		\ifenglish
			AGH University of Science and Technology
		\else
			Akademia G\'{o}rniczo-Hutnicza im. Stanis\l{}awa Staszica w Krakowie
		\fi
	}

	\vspace*{3mm}

	{\large
		\ifenglish
			Faculty of Computer Science, Electronics and Telecommunications
		\else
			Wydzia\l{} Informatyki, Elektroniki i Telekomunikacji
		\fi
	}

	\vspace*{3mm}

	{\large
		\ifenglish
			Department of Computer Science
		\else
			Katedra Informatyki
		\fi
	}

	\vspace*{25mm}

	{\large
		\ifenglish
			Doctoral dissertation
		\else
			Rozprawa doktorska
		\fi
	}

	\vspace*{9mm}

	{\Large
		\ifenglish
			\entitle
		\else
			\pltitle
		\fi
	}

	\vspace*{9mm}

	{\large \shortauthor}
\end{center}

\vspace*{50mm}

\ifenglish
	Advisor:

	Professor Witold Dzwinel, PhD, DSc

	Co-advisor:

	Marcin Kurdziel, PhD
\else
	Promotor:

	prof. dr hab. in\.{z}. Witold Dzwinel

	Promotor pomocniczy:

	dr in\.{z}. Marcin Kurdziel
\fi

\vspace*{5mm}

\begin{center}

	Krak\'{o}w, 2018

\end{center}
\end{titlepage}
}

\begin{document}

\sloppy  

\englishtrue

\maketitle

\renewcommand{\abstractname}{Abstract}
\begin{abstract}
In this dissertation we report results of our research on dense distributed representations of text data. We propose two novel neural models
for learning such representations. The first model learns representations at the document level, while the second model learns word-level
representations.

For document-level representations we propose Binary Paragraph Vector: a neural network models for learning binary representations of text
documents, which can be used for fast document retrieval. We provide a thorough evaluation of these models and demonstrate that they
outperform the seminal method in the field in the information retrieval task. We also report strong results in transfer learning settings,
where our models are trained on a generic text corpus and then used to infer codes for documents from a domain-specific dataset. Finally,
we propose a model that jointly learns short binary codes and high-dimensional real-valued representations. This model can be used for rapid
retrieval of documents highly relevant to the query. In contrast to previously proposed approaches, Binary Paragraph Vector models learn
embeddings directly from raw text data. Thus far, the most common way of building binary document representations was to use a data-oblivious
locality sensitive hashing method on top of some intermediate text representation.

For word-level representations we propose Disambiguated Skip-gram: a neural network model for learning multi-sense word embeddings.
Representations learned by this model can be used in downstream tasks, like part-of-speech tagging or identification of semantic relations.
In the word sense induction task Disambiguated Skip-gram outperforms state-of-the-art models on three out of four benchmarks datasets.
Our model has an elegant probabilistic interpretation. Furthermore, unlike previous models of this kind, it is differentiable with respect
to all its parameters and can be trained with backpropagation. Disambiguated Skip-gram is parametric, i.e. the number of word senses must
be specified a priori. That said, we describe and evaluate a pruning strategy that discards word senses with low marginal probabilities.
We also introduce a regularization term that influence the expected number of senses. In addition to quantitative results, we present
qualitative evaluation of Disambiguated Skip-gram, including two-dimensional visualisations of selected word-sense embeddings.

The dissertation opens with a review of background works and closes with a summary of our contributions and a discussion of possible
directions for future research. In the appendix we describe datasets and software libraries that were used to conduct the experiments,
as well as works that were carried out for this dissertation but did not yield as strong results as the one described in the core chapters.
\end{abstract}

\renewcommand{\abstractname}{Acknowledgements}
\begin{abstract}
I would like to thank my advisor Professor Witold Dzwinel for overall guidance and support,
my co-advisor Marcin Kurdziel for countless hours spent explaining difficult concepts to me and recommending new research directions,
Professor Krzysztof Zieliński for introducing me to academia
and a fellow PhD candidate Piotr Wójcik for collaboration on a few research papers.

This research was supported by National Science Centre, Poland grant no.~\mbox{2013/09/B/ST6/01549} ``Interactive Visual Text Analytics~(IVTA):
Development of novel, user-driven text mining and visualization methods for large text corpora exploration''.

This research was carried out with the support of the ``HPC Infrastructure for Grand Challenges of Science and Engineering'' Project,
co-financed by the European Regional Development Fund under the Innovative Economy Operational Programme.

Last but not least, I would like to thank my immediate family for their love.
\end{abstract}


\setcounter{page}{1}

\tableofcontents

\listoftables
\addcontentsline{toc}{chapter}{\listtablename}
\clearpage
\listoffigures
\addcontentsline{toc}{chapter}{\listfigurename}
\clearpage

\chapter{Introduction}

Prior to deep learning, machine learning often boiled down to numeric optimization over hand-engineered features. Users of traditional
machine learning systems needed to carefully design or select features, and to do that they needed to deeply understand their data. Feature
engineering was, and sometimes still is, one of the most time-consuming, daunting and tedious tasks in a machine learning pipeline.
Training a state-of-the-art learning algorithm with data represented by a poorly selected set of features most often leads
to poor overall performance. Engineering of features is therefore a bottleneck on a way to achieve satisfying results.
Some researchers go even further and suggest that we cannot talk about true artificial intelligence (AI) when features are handcrafted.

Luckily, due to the recent advancements in neural network research, we can now discover some latent features of data,
effectively enabling learning from raw data. A field of study that revolves around learning rich dense representations of data
is called \emph{representation learning}. It is a growing and fascinating field.
Representation learning took off together with deep learning in the late 2000s. Since then, many rich representations
of images, speech, text and other types of data were proposed.
In this dissertation we focus on learning high-quality representations of text data.

An ultimate goal of AI is to build an AI-complete system, which is a system as intelligent as a human. A key element of such a system
is an ability to fully understand humans, which require, among other, an understanding of natural languages that people use.
This goal if far from being met. Nevertheless, intelligent systems can perform a lot of useful functions without fully understanding
the language, i.e. with just partial understanding. For example, one of the major outcomes of the recent AI revolution is increased
popularity of intelligent personal assistants. Example of them are Apple's Siri, Google Assistant, Facebook M, Amazon Alexa,
Microsoft Cortana, Samsung's Bixby or Yandex's Alisa. Intelligent personal assistants revolutionize the way we interact with
mobile devices and personal computers. Most of them interact with humans using voice. However, in most cases the voice is converted into
text as a first step of a processing pipeline. With an advent of deep learning, accuracy of speech recognition systems improved to the
extend that speech recognition is sometime considered a solved problem (e.g.~\cite{graves2013speech}). Much more difficult is the second
step of the pipeline, namely \emph{natural language understanding}.

The first step towards text understanding is to embed small units of text, often words but also sentence or phrases, into some low-dimensional
vectors space. Those vectorised representations are then used as an entry for downstream NLP techniques, like structure
parsing~\cite{socher2013parsing}, machine translation~\cite{sutskever2014sequence}, question answering~\cite{weston2015towards} or image
captioning~\cite{karpathy2015deep}. Therefore, building rich representations of text data is a key element of modern natural language processing.

\section{Motivation}

Amount of digital text data available globally is increasing rapidly. As a consequence, ability to quickly retrieve relevant information
from massive datasets is becoming more and more important. In many cases quality of search results is more important than retrieval time.
However, in some cases users are willing to compromise on the quality of search results in favor of fast retrieval. In general,
retrieval in these settings can be seen as an instance of \emph{approximate nearest neighbor search}. Such approximation to searching is
often realized with \emph{locality-sensitive hashing} (LSH) methods. The idea is to generate short binary codes for documents that carry
semantic information, i.e., similar documents will end up having similar codes. Having such codes, we can treat them as memory addresses and
quickly retrieve similar documents by generating a hash for a given query and then taking all documents having the same or similar
memory address as the query.

Traditionally, LSH codes were generated from text documents represented by the \emph{bag-of-words} (BoW) representation, which in its simplest form
is just a set of word counters. BoW is a popular representation, often used for text document classification and information retrieval.
Despite its popularity and applicability, it is a limited and simplistic representation: for example, it does not carry word order information.
In the recent years, many dense, high-quality representations of text data were proposed. We describe them in \sectionref{nn_text_rep}.
Many of them can be used to obtain state-of-the-art results in tasks like document classification, sentiment analysis or information retrieval.
All of them are real-valued representations. In order to use them for addressing, one still needs to convert them to binary codes using
some locality-preserving hashing technique. It would be desirable to be able to build a high-quality distributed binary representation
of documents that can be directly used for approximate nearest neighbor search.

Word embedding models are ubiquitous, but most of them have one inherent limitation: each word, even ambiguous one, is placed in one unique spot
in a vector space. One of the implications of this is that some non-related words are `drawn' to each other, e.g. high-tech companies are
`drawn' to fruits, because of the word \emph{apple}. Many solutions were proposed to deal with ambiguity when learning word embeddings.
We review them in \sectionref{multi_sense_word_embeddings}. One of the classification criteria of those methods is a way of
estimation of latent variables and parameters of the model. Some of the multi-sense word embedding models employ \emph{error backpropagation}
while other use variants of the \emph{expectation-maximization} algorithm. To the best of our knowledge,
none of the models trained with backpropagation has a clean probabilistic interpretation. Instead, to discover word senses they employ,
for example, implicit context clustering during training. It would be beneficial to have a clean end-to-end differentiable
probabilistic multi-sense word embedding model.

\section{Contributions}

This dissertation can be divided roughly into two main parts. The first focuses on learning distributed representations of documents
(\chapterref{dsh}). Therein we propose a novel model for learning binary vector representations of text documents,
which can be used for fast information retrieval. To the best of our knowledge, no one proposed a similar model for learning binary vectors
directly from raw text. Existing solutions require a two-step approach, where binary codes are learned from some
intermediate real-valued representation. Our model is simple, has smaller memory requirements than the two-step approach
and produces competitive results. We presented the model at the 2nd Workshop on Representation Learning for NLP~\cite{grzegorczyk2017binary}.

The second major part of this dissertation revolves around dense representations of words (\chapterref{disgram}). We introduce a novel neural
network that is an extension to the popular \emph{skip-gram} model. Our contribution consists of adding a disambiguation
subnetwork to the model. The resulting solution has an elegant probabilistic interpretation. To assure high-quality of word representations
produced by our model we employ some recently introduced deep learning techniques. We test our model against several state-of-the-art models
on a few benchmark test sets, and we demonstrate its superior performance.

The dissertation opens with a review of background works (\chapterref{background}) and closes with a summary of our contributions
and a presentation of some directions for future research (\chapterref{conclusions}). \appendixref{improving_bow} describes
research that we carried out for this dissertation but which did not yield as promising results as the one described in earlier chapters.
Finally, \appendixref{software_datasets} describes datasets and software libraries that we used to conduct experiments.

\chapter{Background and related works} \label{ch:background}

In this chapter we discuss different ways of representing text data as well as various deep learning concepts.
We start with an introduction of selected machine learning (ML) terms and concepts that are frequently used in this dissertation.
A more comprehensive introduction to ML can be found in~\cite{bishop2006pattern, murphy2012machine, abu2012learning, goodfellow2016dl}.
A layperson's overview of ML concepts is presented~\cite{domingos2015master}.

\section{Selected concepts in machine learning}

\begin{figure}[htb!]
  \centering
  \includegraphics[width=0.4\linewidth]{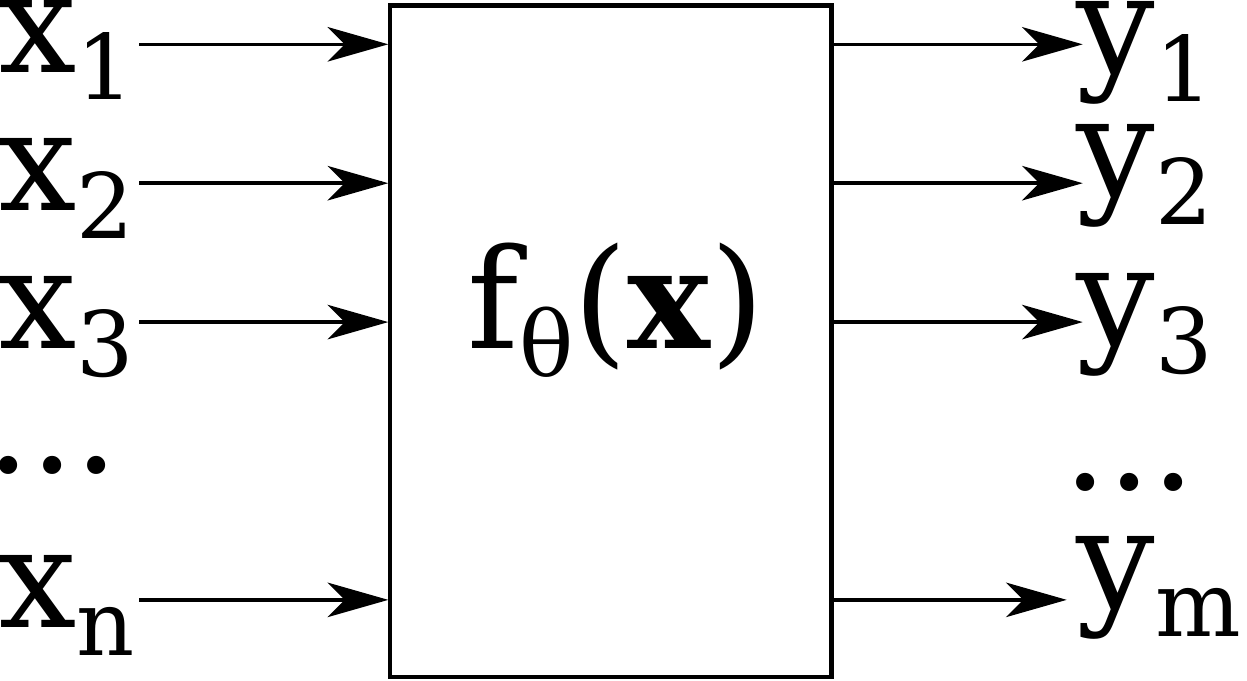}
  \longcaption{A software function model.}{Where $ x_i $ are the inputs, $ \Theta $ are the model parameters and $ y $ is an output.}
  \label{fig:machine_learning}
\end{figure}

In order to deal with complexity of software systems they are often modularized on various levels of abstraction.
Software modules or components can be seen as \emph{black-boxes} that take some input,
do some internal processing and output some results.
One of the low level abstractions in software is a \emph{function}. Function, a concept borrowed from the field of mathematics,
takes some data $ \mathbf{x} $ as an input and produces some data $ \mathbf{y} $ as a result (~\figref{machine_learning}).
In addition, $ \mathbf{\Theta} $ is a set of internal function parameters which influence the output.
Those parameters encapsulate the \emph{knowledge} that is needed to produce accurate outputs.
Traditionally this knowledge was explicitly given to the system by experts.
\emph{Machine learning} (ML) is a family of algorithms that enable computers to obtain the knowledge in an automated way
by learning it from data.
In ML terminology the function $ f $ is often called an ML \emph{model} or a \emph{hypothesis} or en \emph{estimator}
and the goal is to discover or estimate the optimal parameters $ \mathbf{\Theta} $ that produce expected outputs $ \mathbf{y} $.
The model parameters are sometimes called \emph{coefficients}.

A single data item passed to the ML algorithm is called an \emph{example}.
The example has multiple data fields called \emph{features}.
The features can be binary, numerical, textual or categorical.
If all the features are numerical, then the data example can be represented as a vector $ \mathbf{x} $
in an n-dimensional vector space, where $ n $ equals the number of features.

Taxonomy of ML is complex, and we are not going to analyze it here, but probably the two main types of ML are:
\begin{itemize}
\item supervised learning - learning model parameters $ \Theta $ by providing input data $ x $ with desired outputs $ y $,
often called \emph{labels},
\item unsupervised learning - learning model parameters $ \Theta $ without any labeled data, relying exclusively on the input data.
\end{itemize}

The process of estimating optimal values of model parameters $ \mathbf{\Theta} $ is often dubbed \emph{fitting} parameters to the data.
One of the important traits of the supervised ML algorithm is its ability to \emph{generalize}.
A model that generalize well performs well not only on the data on which it was trained, but also on data examples
which were unknown at the training time.
Therefore, examples available in a data set are often split into two subsets.
One is called \emph{training set} and is used to fit the parameters, and the other is called \emph{test set}
and is used to evaluate the model.
It is important that the test set is held out during training and is only used to evaluate the model after training.
If the model performs well on the training set but poorly on the test set,
we say that it is \emph{overfitted} to the training data or that it has high \emph{variance}.
If the model performs badly even on the training data, we say that it is \emph{underfitted} or \emph{biased}.
Sometimes it is said that training the model is a bias-variance tradeoff~\cite[Section 2.2.2]{james2013introduction}.
A good illustration of this dilemma is depicted in~\cite[Fig. 2.12]{james2013introduction}.

There is a multitude of tasks that are solved using ML. Probably the most common supervised learning tasks are:
\begin{itemize}
\item regression - predicting a single continuous output value for a given set of inputs,
\item classification - assigning each data example to appropriate class;
when there are only two possible classes we say that it is a \emph{binary classification} problem;
when there are more than two classes we call it a \emph{multinomial classification} problem.
\end{itemize}
Unsupervised learning tasks include:
\begin{itemize}
\item clustering - separating data examples into distinct groups,
\item dimensionality reduction - expressing data in a lower number of dimensions losing as little knowledge as possible.
\end{itemize}
\subsection{Training a supervised machine learning model}

When machine learning is used for regression or classification,
the performance of a model is measured by a \emph{loss} function, a.k.a. a cost function, often denoted by $ L $.
There are multiple loss functions used in the field of machine learning.
One of the basic loss functions is the \emph{mean squared error}:
\begin{equation}
L( f_{\Theta} ( \mathbf{X} ) , \mathbf{y} ) = \frac{1}{N} \sum_{i=1}^N ( f_{\Theta} ( \mathbf{x}_i ) - y_i )^2,
\end{equation}
where N is a dataset size. It is sometimes written with an Euclidean norm:
\begin{equation}
L( f_{\Theta} ( \mathbf{X} ) , \mathbf{y} ) = || f_{\Theta} ( \mathbf{X} ) - \mathbf{y} ||^2.
\end{equation}
For simplicity, instead of $ L( f_{\Theta} ( \mathbf{X} ) , \mathbf{y} ) $ we often write just $ L(\Theta) $.

Fitting the model parameters $ \mathbf{\Theta} $ boils down to the problem of finding the model parameters
that yield the lowest cost for the training data.
More formally, training can be interpreted as minimizing the cost function $ L $ over the model parameters $ \Theta $.
There is a plethora of numerical optimization methods. One of the most popular optimization methods is Gradient Descent~(GD).
In GD one needs to randomly initialize parameters, and then iteratively update them:
\begin{equation}
\Theta_{s + 1} = \Theta_s + \Delta \Theta_s,
\end{equation}
where $ s $ is a step number.
The parameter update matrix $ \Delta \Theta_s $ is a negative gradient of a loss function, $ - \nabla L $,
multiplied by \emph{learning rate} hyperparameter $ \alpha $:
\begin{equation} \label{eq:gd_update}
\Delta \Theta = - \alpha \nabla L(\Theta).
\end{equation}
Note that since the gradient of the loss function is a set of all possible partial derivatives with respect to model parameters,
the loss function needs to be differentiable with respect to all of them.
In practice, a stochastic variant of Gradient Descent~(SGD) is often used for optimization.
In SGD the gradients are calculated not for the entire training data but for a limited number of sampled examples.

To prevent model from overfitting some \emph{regularization} term is often added to the cost function.
Regularization may \emph{penalize} high values of model parameters during the optimization process
and, as a consequence, cause the hypothesis to be simpler.
Probably the two most common regularization methods are the sum of squares of model parameters and the sum of absolute values.
The first one is called $ L_2 $ regularization and the second $ L_1 $.
The mean squared error with $ L_2 $ regularization takes the form:
\begin{equation}
L( \Theta ) = \frac{1}{N} \sum_{i=1}^N ( f_{\Theta} ( \mathbf{x}_i ) - y_i )^2 + \lambda \sum_{i=1}^M \Theta_i^2,
\end{equation}
where $ \lambda $ is a regularization parameter, sometimes dubbed a \emph{penalty} or \emph{shrinkage},
and $ M $ is a total number of model parameters.

In order to speed up convergence and prevent from getting stuck in a local minima
the \emph{momentum} method~\cite{polyak1964some} is often applied.
The momentum modify \equationref{gd_update} by adding a fraction of updates from the previous step:
\begin{equation}
\Delta \Theta_t = \varepsilon \Delta \Theta_{t-1} - \alpha \nabla L(\Theta),
\end{equation}
where $ \varepsilon $ is a momentum hyperparameter.

One of the drawbacks of the cost function presented above is that it may not be a \emph{convex} function,
i.e. it could get stuck in a local minima during optimization, depending on a modeled function $ f $.
It is not a problem when the modeled function is linear. However, when we want to learn parameters of a nonlinear function $ f $,
then we need to optimize other, more complex cost functions. We will discuss them later on in this chapter.

An alternative approach to fitting model parameters is \emph{black-box optimization}. Methods from this family tune the parameters based only
on analysis of signals exiting the model, regardless of an internal structure of the model. One recent example of black-box numerical
optimization solution is Google Vizier~\cite{golovin2017google}.

\section{Vector representations of text data} \label{sec:text_vector}

One of the main applications of computers is data processing.
Data processing can involve analyzing data, extracting some knowledge from it, converting it into other formats or visualizing it.
In general, we can distinguish two types of data: structured and unstructured.
Structured data is organized and described by same meta-data, and is often stored in relational databases or spreadsheets.
Unstructured data is not organized and is often stored in non-relational databases or directly as raw files in a file-system.
Examples of raw data are: images, videos, sound records, or unstructured text documents.
Processing unstructured data is more challenging than structured data.
However, by and large, there is much more raw data available and it is easier to obtain.

For humans, understanding text data is relatively easy.
Assuming that a text document is written in a natural language native to the reader, they can understand it without effort.
For computers it is much harder to process natural languages. Nevertheless, it is a very important task.
There is proliferation of applications that rely on understating of text data.
Examples of such applications are: information retrieval, sentiment analysis,
question answering, machine translation, text summarization or information extraction.
All those tasks can be classified as Natural Language Processing (NLP).
A comprehensive introduction to NLP can be found in~\cite{manning1999foundations, jurafsky2008speech}.
In this thesis we will focus on one aspect of NLP, which is vector representations of text data.

Text data at different levels can be represented by vectors. A single vector can represent a document, a paragraph, a word, or even a
single character. One of the most popular applications of vector representations of documents is Information Retrieval (IR).
A comprehensive introduction to IR can be found in~\cite{manning2008introduction}.
Below we present just basic IR ideas.

Information Retrieval methods attempt to retrieve a relevant document for a given query.
In practice, instead of a single document a list of candidates ranked according to the relevance is returned.
The simplest ranking model is based on the occurrence of query terms in the ranked document. It is called a \emph{Boolean model}.
Documents matching most of the query terms are placed nn the top of the result list.
However, for short queries there could be multiple documents containing them and,
therefore, the order of result could be ill-defined.

\subsection{Vector Space Model} \label{sec:vsm}

Probably the most popular IR model is Vector Space Model~\cite{salton1975vector} (VSM).
In VSM both the query and the documents are represented as vectors in the same vector space.
Then, the inner product of two vectors $ \mathbf{a} $ and $ \mathbf{b} $ serves as a similarity measure used to rank the results:
\begin{equation}
sim(\mathbf{a}, \mathbf{b}) = \mathbf{a}^\mathrm{T} \mathbf{b} = \sum_{i=1}^M a_i b_i,
\end{equation}
where $ M $ is a number of dimensions in a vectors space.
Often the inner product is normalized by vectors lengths to make the measure independent of them.
This way we obtain \emph{cosine similarity}:
\begin{equation}
sim(\mathbf{a}, \mathbf{b}) = \cos(\mathbf{a}, \mathbf{b}) = \frac{\mathbf{a}^\mathrm{T} \mathbf{b}}{\left | \mathbf{a} \right | \left | \mathbf{b} \right |}.
\end{equation}

The most common way to place documents and a query in a vector space is to represent them as counts of words from a vocabulary.
The resultant number of dimensions of the space equals the vocabulary size.
A simple count is often referred to as frequency $ f_{t,d} $ of a term $ t $ in a document $ d $.
Often sublinear scaling is applied to term frequencies:
\begin{equation}
TF( t , d ) = \left\{
\begin{matrix}
	1 + \log ( f_{t,d} ) & \text{if}~~f_{t,d} > 0, \\
	0                   & \text{otherwise}.
\end{matrix}\right.
\end{equation}
However, this schema does not take into account that some words are statistically more common than the other
and, therefore, values in some dimensions will be much higher than in others.
To solve this issue, term frequencies are often multiplied by \emph{inverse document frequencies}:
\begin{equation} \label{eq:idf}
IDF( t ) = \log \frac{ N }{ n( t ) },
\end{equation}
where $ N $ is the total number of documents in the corpus and $ n(t) $ is a number of documents containing term $ t $.
It is beneficial to smooth $ IDF $ term by adding $ 1 $ to it:
\begin{equation}
IDF_{smoothed}(t) = \log ( 1 + \frac{ N }{ n( t ) } ).
\end{equation}
Resultant combined schema is often called \emph{TF-IDF}.
There are many variations of TF-IDF weighting scheme used in information retrieval and machine learning.
See~\cite[section 6.4]{manning2008introduction} for details.

\subsection{Bag-of-words model}

When documents represented in a form of term frequencies are used for applications other than Information Retrieval,
we often call it the \emph{bag-of-words} (BoW) representation.
BoW research dates back to the 1950s~\cite{harris1954distributional}.
One of the popular applications of BoW is text classification.
For example, given a set of emails we want to be able to tell which of them are unsolicited and which are not.

An inherent limitation of the BoW representation is that the order of words in a document is not preserved.
Phrase ``The Allies defeated the Axis'' and ``The Axis defeated the Allies'' are represented by the same vector.
The other drawback of BoW is that resultant vectors are sparse.
Even if document has thousands unique words, it still is just a fraction of the vocabulary size, which can be in hundreds of thousands.
Another limitation is that multiple senses of polysemous and homonymous words are represented by a single dimension.
For example, a document dealing with river \emph{banks} and a document about the federal \emph{bank}
will both have high value in a dimension associated with a word \emph{bank}.
Conversely, in BoW we have multiple dimensions for synonymous words, which causes some features to be redundant.
We will discuss how to deal with those limitations later in the thesis (\sectionref{wsd}).

One of the generalizations of bag-of-words is to extend the vocabulary
by adding to it combinations of words occurring next to each other in sentences.
This generalization is called \emph{bag-of-n-grams}.
Using the example from previous paragraph we will have separate dimension for \emph{bigrams} (word pairs) ``Allies defeated''
and ``Axis defeated'' and, therefore, those two phrases will be represented by different vectors.
The drawback of bag-of-n-grams is even higher dimensionality and sparsity than in the case of BoW.

At this point it is worth noting that for some algorithms high dimensionality is not a problem while for others is a major issue.
One simple machine learning algorithm which is very scalable and deals well with high dimensionality is \emph{naive Bayes} classifier.
This classifier is based on the \emph{Bayes' theorem}:
\begin{equation}
P(A | B) = \frac{P(B | A) P(A)}{P(B)},
\end{equation}
where $ A $ and $ B $ are some events, $ P(A | B) $ is a \emph{posterior} probability of event $ A $ given event $ B $,
$ P(A) $ is a \emph{prior} probability of event $ A $, $ P(B) $ is an \emph{evidence} and $ P (B | A) $ is \emph{likelihood}.
Employing naive Bayes classifier, the probability of a document $ d $ belonging to a class $ c $ is estimated in the following way:
\begin{equation}
P(c | d) = \frac{P(d | c) P(c)}{P(d)}.
\end{equation}
Classifications boils down to selecting the class with the highest probability:
\begin{equation}
\hat{c} = \argmax_{c \in C} \frac{P(d | c) P(c)}{P(d)},
\end{equation}
where $ \hat{c} $ is the predicted class and $ C $ is a set of all classes.
The evidence $ P(d) $ is constant for all the classes and, therefore, we can eliminate it from the equation:
\begin{equation}
\hat{c} = \argmax_{c \in C} P(d | c) P(c).
\end{equation}
Prior $ P(c) $ can be easily estimated by just counting how many times class $ c $ occurs in the corpus and normalizing
by the total number of classes.
Estimation of the likelihood $ P(d | c) $ is slightly more involved.
Assuming that features in the bag-of-words representation are independent, we can estimate likelihood as:
\begin{equation}
P(d | c) = \prod_{i=1}^{N_d} P(w_i | c),
\end{equation}
where $ N_d $ is a number of words in a document $ d $ and $ w_i $ is a word at position $ i $ in the document.
We can estimate the probability of a word given a class in the following way:
\begin{equation}
P(w_i | c) = \frac{\countfunction(w_i, c)}{\sum_{j=1}^M \countfunction(w_j, c)},
\end{equation}
where $ M $ is a vocabulary size.
Therefore, the total number of parameters of naive Bayes classifier equals the number of words in the vocabulary
(likelihood parameters) summed with the number of classes (prior parameters).
In practice, for text data \emph{multinomial} variant of naive Bayes classifiers is used.

As we demonstrated, the number of parameters of naive Bayes classifier is a liner function of number of features
and, therefore, high-dimensionality of feature space is not a problem.
However, for many other algorithms (e.g. neural networks discussed later on in this chapter) relation between the number of features
and the number of parameters is exponential, which poses high memory or low input dimensionality requirements.


\subsection{Topic modeling} \label{sec:topic_modeling}

For years researches have been trying to build low-dimensional representations of text.
The simplest way to cope with high-dimensionality is to select a limited number of most frequent words from a vocabulary (e.g. 2000)
and represent documents as frequencies of only those selected terms.
Such a simplistic solution is sufficient in some applications but not in many.
One of the more sophisticated approaches is to try to discover latent topics of the documents.
This approach is called \emph{topic modeling}.
Probably the first topic model was Latent Semantic Analysis (LSA)~\cite{deerwester1990indexing}, a.k.a. Latent Semantic Indexing.
LSA attempts to discover topics by decomposing word-document co-occurrence matrix using Singular Value Decomposition:
\begin{equation}
\mathbf{X} = \mathbf{U} \mathbf{\Sigma}  \mathbf{V}^\mathrm{T},
\end{equation}
where each column of $ \mathbf{X} $ is the bag-of-words representation of a single document,
each column of $ \mathbf{U} $ is a distribution of words in a single topic
and each row of $ \mathbf{V} $ is a distribution of topics in a single document.
$ \mathbf{\Sigma} $ is a diagonal matrix whose diagonal elements are called \emph{singular values}.
As a result of decomposition we get documents represented as distributions of topics.
In addition, we obtain \emph{definitions} of topics in a form of distributions of words.
For example, a topic regarding Middle East issues will probably have high values for words like `Israel', `Arab' or `Palestine'.
 
More recent topic model is latent Dirichlet allocation (LDA)~\cite{blei2003latent}.
LDA makes a very crude but useful assumption that documents are generated randomly by sampling words from sampled topics.
As a consequence, all the documents in the collection share the same set of topics,
but each document exhibit those topics in different proportions.
In practice those distributions need to be inferred from training data.

One limitation of LDA is that all topics are independent. In reality some topics can be highly correlated with other topics.
To address this limitation Lafferty \& Blei proposed correlated topic models (CTM)~\cite{lafferty2006correlated},
which explicitly model correlations between topics.

From a probabilistic point of view topic models can be seen as directed probabilistic graphical models,
where documents point to topics, which subsequently point to words. Topic models can also be seen as mixture distributions,
i.e. each document is represented as a mixture of topics, where topics are probability distributions over words.
One problem with mixtures is that they are linear combinations of random variables.
Therefore they cannot take into account non-linear relationships between variables.
We can imagine that some topic has high probability of existence of some combination of words but low probability of occurrence
of those words in isolation.
Later we will show how to tackle this problem.

As we mentioned above, one of the drawbacks of the bag-of-words representation is its inability to cope with polysemous words.
Ambiguity is one of the biggest challenges of natural language understanding.
In the following subsection we discuss ways to disambiguate polysemous words
and to embed this information in representations of text data.

\subsection{Word Sense Disambiguation} \label{sec:wsd}

Word Sense Disambiguation (WSD) is a problem studied for many years
in the field of Natural Language Processing~\cite{lesk1986automatic,yarowsky1995unsupervised,schutze1998automatic}.
The problem boils down to determining which meaning of a given ambiguous word should be selected in a given context.
Ambiguity is formalized by two concepts: polysemy and homonymy. Polysemy is the coexistence of many possible meanings of a single word.
Homonymy is when multiple words have the same spelling and pronunciation by just mere linguistic coincidence.
Important difference is that in the case of homonymy there are multiple words with separate lemmas while in the case of polysemy there is just one lemma.
Examples of polysemous words are: mouse, apple, fox, crane, window, plant or palm. Examples of homonymous words: bank, rock, taxi, bear or check.
When natural language is processed by computers it often does not matter whether ambiguous word is a polyseme or a homonym.

There are three main approaches to WSD, namely supervised, knowledge-based and unsupervised.
In a supervised approach machine learning model is trained on a large number of sense-annotated sentences.
Knowledge-based methods rely on an external lexical database like WordNet~\cite{miller1995wordnet},
DBpedia~\cite{lehmann2014dbpedia}, BabelNet~\cite{navigli2012babelnet} or ConceptNet~\cite{speer2017conceptnet}.
The most popular method from this family is a classic Lesk algorithm~\cite{lesk1986automatic}.
Finally, unsupervised methods neither require sense-annotated corpora nor knowledge bases.
In this family of methods, one needs to discover possible word senses prior to disambiguation.
Therefore, this approach is often called Word Sense Induction.
The seminal work that goes in this direction is~\cite{schutze1998automatic},
where authors propose to discover senses by clustering occurrences of ambiguous words.

When we are able to disambiguate polysemous words we can apply this skill to crate a document representation
that has separate dimensions for separate word senses.
Specifically, Reisinger et al.~\cite{reisinger2010multi} proposed a multi-prototype vector-space model,
where each word is represented by multiple word vectors.
They discover senses using word sense discrimination~\cite{schutze1998automatic}, i.e. by clustering word occurrences.
Resultant vectors for a given word not only represent different word senses but also different word usages.
This method is generic, i.e. any embedding method and clustering algorithm can be used.

Similar approach is adopted by Huang et al. in ~\cite{huang2009clustering}.
They propose a \emph{bag-of-concepts} document representation, where each dimension corresponds to one abstract concept,
which can be described by multiple words.
For example, all three: \emph{the Earth}, \emph{the world}, and \emph{the globe}, will be represented by just one dimension.
To build this representations the authors rely on an external knowledge base.
Specifically, they analyze anchor text in Wikipedia hyperlinks and observe that multiple different anchor texts point
to a single wiki page.







In this section we discussed a basic vector space model and standard extensions to it. A more comprehensive survey is presented in~\cite{turney2010frequency}.

\section{Neural network-based text representations} \label{sec:nn_text_rep}

Learning high-quality distributed representations of text data is a complex task. Due to their high capacity, neural networks are an obvious choice
for doing this. In the following subsections we present selected neural network-based text representations. We start with an introduction
of selected concept in neural network models.

\subsection{Artificial neural networks} \label{sec:neural_networks}

\begin{figure}[htb!]
  \centering
  \includegraphics[width=0.5\linewidth]{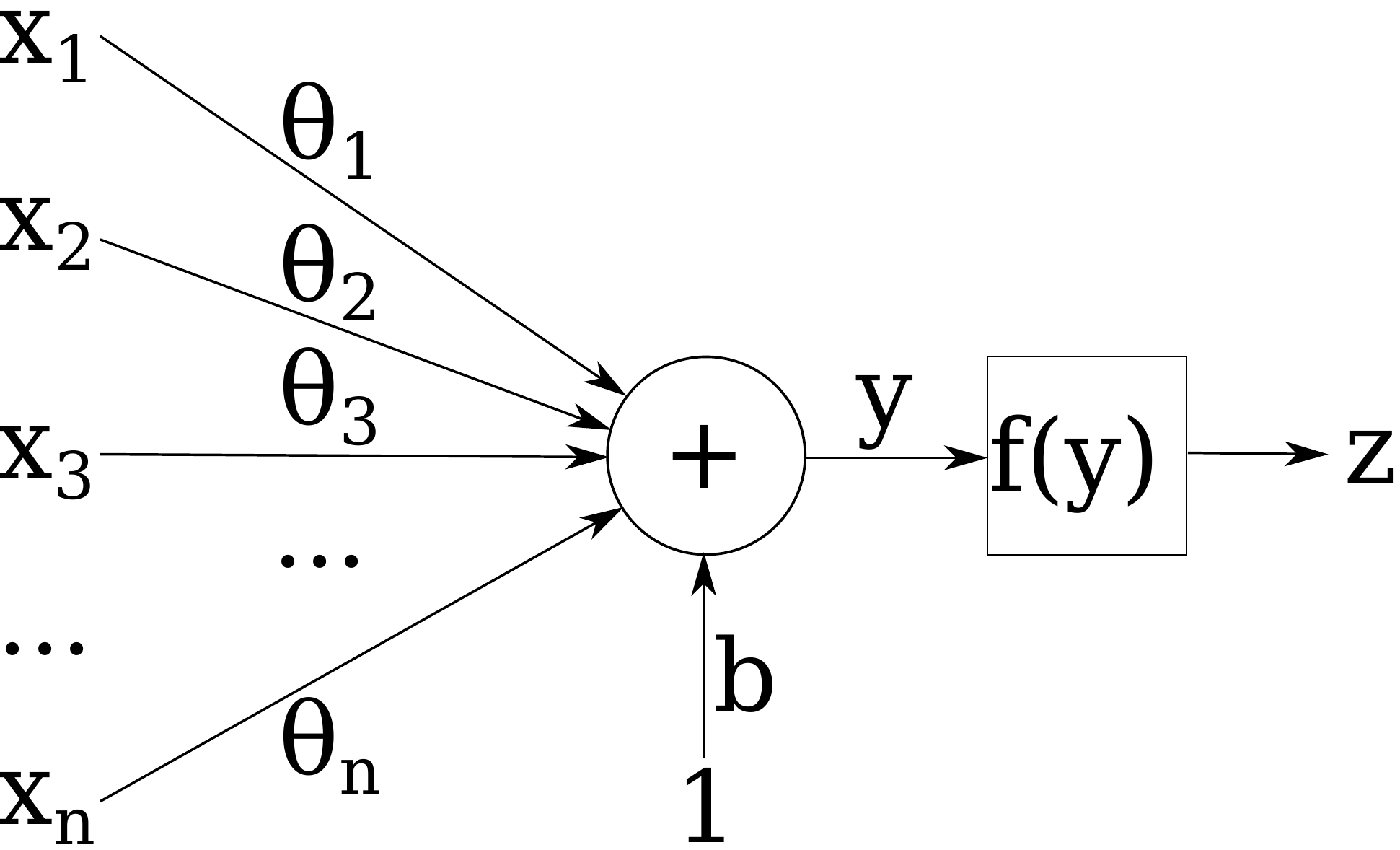}
  \longcaption{An artificial neuron model:}{$ x_i $ are inputs, $ \theta_i $ are weights (a.k.a. parameters),
               $ b $ is a bias term, $ y $ is a weighted sum of the inputs and the bias, $ f( y) $ is an activation function
               (a.k.a. transfer function) and $ z $ is the output.}
  \label{fig:artificial_neuron}
\end{figure}

Artificial neural networks are a family of learning algorithms loosely inspired by the human brain.
The building block of a neural network is an artificial neuron.
The first models of the artificial neuron (\figref{artificial_neuron})
were proposed in the 1940s~\cite{mcculloch1943logical}.
The neuron has multiple inputs $ x_i $ and one output $ z $. Internally, a weighted sum of inputs and a bias term is calculated:
\begin{equation}
y = b + \sum_{i=1}^{n} x_{i} \theta_{i},
\end{equation}
where $ \theta_i $ are input weights, a.k.a. parameters. For convenience, the bias term is often appended to the weights the input vector
is appended with a fixed value of $ 1 $.
Then we can calculate $ y $ as a dot product of input and weight vectors:
\begin{equation}
y = \vect{x}^\mathrm{T} \vect{\theta}.
\end{equation}
This weighted sum, often called neuron pre-activation, is then passed to an activation function.
The simplest form of the activation function is a step function:
\begin{equation}
f( y ) = \left\{
\begin{matrix}
  1 & \text{if}~~y > 0, \\
  0 & \text{otherwise.}
\end{matrix}\right.
\end{equation}
When the Heaviside step function is used as an activation function of an artificial neuron,
then resultant model is called \emph{perceptron}~\cite{rosenblatt1958perceptron}.
In practice, continuous functions are often used as activators.
One popular activation function is the \emph{logistic} function, often dubbed a \emph{sigmoid} function due to its shape:
\begin{equation}
\sigma ( x ) = \frac{1}{1 + e^{-x}}.
\end{equation}
An important trait of this function is that the output is always between 0 and 1
and, therefore, we can use it to model probabilities.

Manual setting of neuron parameters would make wider adoption of those models infeasible.
Therefore, we need to be able to learn neuron weights automatically.
As in the case of other machine learning methods, we do this by minimizing a cost function.
In the case of a neuron with a logistic activation function the cost function $ L $ of model weights $ \vect{\Theta} $ takes the form:
\begin{equation} \label{eq:logistic_cost}
L( \vect{\Theta} ) = - \frac{1}{N} \sum_{i=1}^N [ y_i \log \sigma ( x_i ) + (1 - y_i) \log (1 - \sigma (x_i)) ],
\end{equation}
where N is a dataset size. This function can be derived from a \emph{maximum likelihood principle}.

An artificial neural network (NN) is a composition of multiple artificial neurons.
The most popular type of NN is a \emph{feedforward} NN. Sometimes they are also called multilayer perceptrons (MLPs).
However, this name should be used with caution, since activations in MLPs are rarely step functions.
Moreover, activations are often not only continuous but also nonlinear. Otherwise, the network could be reduced to just one big neuron,
since a function which is a combination of linear functions is still a linear function.
\begin{figure}[htb!]
  \centering
  \includegraphics[width=0.4\linewidth]{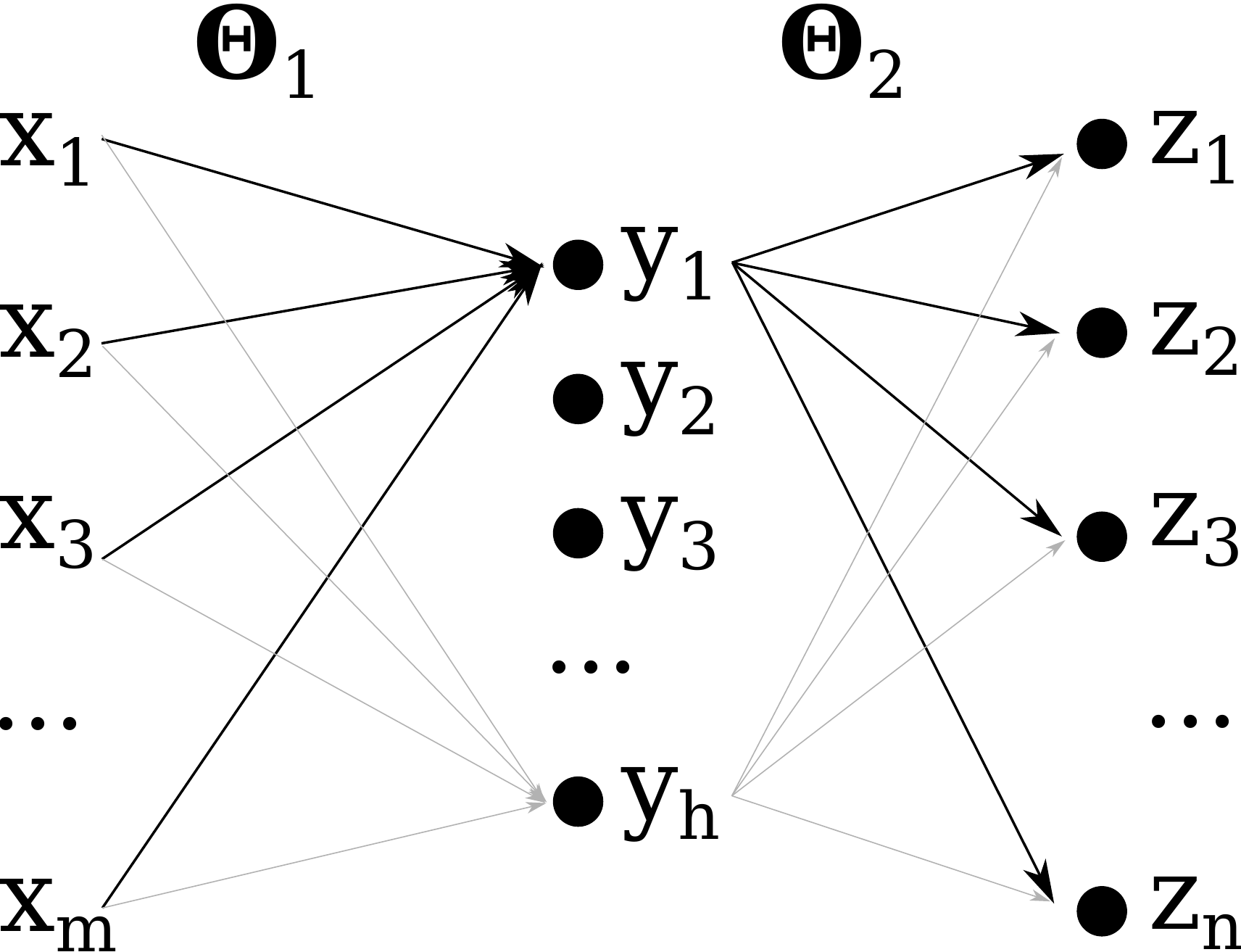}
  \longcaption{A simple feedforward neural network with one hidden layer.}{There are $ m $ input signals, $ h $ hidden neurons
               and $ n $ output neurons. Weights for each neuron are rows in weight matrices $ \vect{\Theta}_1 $ and $ \vect{\Theta}_2 $.
               For simplicity, bias terms are not presented.}
  \label{fig:neural_network}
\end{figure}

The most appealing feature of a feedforward network is that it is, theoretically, sufficient to approximate any continuous
function~\cite{cybenko1989approximation}.
In practice, however, to model any function a single-layer network would have to have so many hidden neurons that its training
would be infeasible. Ability to model different functions is often called model \emph{capacity}. A network with higher capacity
is able to model more complicated functions. To increase the model capacity one could add multiple hidden layers with relatively
small numbers of neurons, instead of adding many hidden units to a single hidden layer. Networks with many hidden layers will be described
later in this chapter.

The cost function of the feedforward neural network is the cost function of the output layer.
If there is only one sigmoid neuron in the output layer, the cost function could be defined by \equationref{logistic_cost}.
However, it is more common to use multiple output neurons.
For example, when the network is used for classification, we want as many neurons in the output layer as there are classes.
We want the correct neuron to output high value (high probability) and the other neurons to output low values (low probabilities).
To this end, we often use the \emph{softmax} activation function, which is a generalization of the logistic function:
\begin{equation} \label{eq:softmax}
p_i( \vect{x} ) = \frac{e^{x_i}}{\sum_{k=1}^K e^{x_k}},~~\text{for}~~i=1, \cdots ,K,
\end{equation}
where K is the number of output neurons.
Softmax ensures that the values in the output layer sums up to one
and, therefore, can be interpreted as probabilities assigned to given classes.

When NN is used with the softmax output layer, we got a vector of probabilities as an output,
which we need to compare with a label (represented using one-hot encoding) in order to calculate the loss.
When we want to compare two probability vectors $ \vect{a} $ and $ \vect{b} $, it is good to use the \emph{cross-entropy} function:
\begin{equation}
S( \vect{a} , \vect{b} ) = - \sum_{i=n}^K a_i \log( b_i ),
\end{equation}
where K is a number of dimensions. The Cross-entropy is a special case of \emph{entropy} function defined for a vector of probabilities $ \vect{p} $:
\begin{equation} \label{eq:entropy}
S( \vect{p} ) = - \sum_{i=n}^K p_i \log( p_i ),
\end{equation}
In order to calculate the training loss we need to average the cross-entropy over the entire training set:
\begin{equation}
L( f_{\vect{\Theta}} ( \vect{X} ), \vect{Y} ) = \frac{1}{N} \sum_{i=1}^N S( f_{\vect{\Theta}} ( \vect{x}_i ), \vect{y}_i ),
\end{equation}
where $ N $ is dataset size, and $ f_{\vect{\Theta}} $ is the function that describes the whole neural network.

\subsection{Neural network training}

In order to minimize a neural network cost function using a gradient-based optimization method, like Gradient Descent,
one need to calculate the derivative of a loss function with respect to all the weights $ \vect{\Theta} $ in all the layers of the network.
The problem is that we calculate error at the output of the network,
but then we need to calculate partial derivatives of the error with respect of the previous layers' weights.
The lack of a fast and easy method to do this delayed applications of NN for many years.
The revival of interest in NN started in mid-eighties with a discovery of backpropagation algorithm~\cite{rumelhart1988learning},
which enabled calculation of the gradients in all hidden layers. 
Backpropagation can be interpreted as an implementation of the chain rule for computing the derivative of the composition of functions.
The chain rule states that the derivative of a composed function is a product of derivatives of the outer and the inner function.
In practice, we compute gradients in four steps.
First, we do forward propagation and obtain a vector of values $ \hat{\vect{y}} $ at the output of the network.
Next, we compare this output with expected labels $ \vect{y} $ to compute a loss or an \emph{error}, denoted by $ \bm{\delta}_{output} $.
\begin{equation}
\bm{\delta}_{output} = \hat{\vect{y}} - \vect{y}.
\end{equation}
Then, we recursively compute errors for each layer $ l $ in the network:
\begin{equation}
\bm{\delta}_l = \vect{\Theta}_l^\mathrm{T} \bm{\delta}_{l+1} \circ f'_l(\vect{z}_l),
\end{equation}
where, $ \circ $ is the Hadamard product (element-wise multiplication)
and $ f'_l(\vect{z}_l) $ is a derivative of an activation function of layer $ l $ evaluated for the input of that layer $ \vect{z}_l $.
Finally, we can calculate a vector of partial derivatives of the cost function $ L $ with respect to the weights of layer $ l $:
\begin{equation} \label{eq:partial_derivative_single_batch}
\frac{\partial L}{\partial \vect{\Theta}_l} = \bm{\delta}_{l+1} f_l^\mathrm{T}(\vect{z}_l).
\end{equation}
In practice, neural networks are often trained in mini-batches
and, therefore, \equationref{partial_derivative_single_batch} takes the form:
\begin{equation}
\frac{\partial L}{\partial \vect{\Theta}_l} = \frac{1}{M} \sum_{i=0}^M \bm{\delta}_{l+1}^i (f_l^i(\vect{z}_l))^\mathrm{T},
\end{equation}
where $ M $ is a total number of training examples in a mini-batch.
Having partial derivatives of the cost function with respect to the weights one can use gradient descent,
or other gradient-based optimization algorithm, to find the optimal set of network weights.

One of the key features of the backpropagation is that an error for a given layer
is computed in terms of an error of the preceding layer (looking from the back of the network).
This has one important implication. If an error, for some reason, become very small in one layer,
then an error in a subsequent layer (again looking from the back) will be also small, or even smaller (if weights are small).
This phenomenon is known as the \emph{vanishing gradient problem}.
Related to it one is the \emph{exploding gradient problem}, where errors become bigger and bigger in subsequent layers.
Those two problems prevented practical use of deep neural networks,
i.e. networks with more than one hidden layer\footnote{In recent years people tend to use term \emph{deep learning}
to describe all neural networks, even shallow ones}, in the early years of backpropagation based neural models.

In the 1990s, neural networks were overshadowed by support vector machines (SVM) \cite{cortes1995support}.
SVM tries to find a hyperplane that separates classes with as wide margin as possible.
The margin is often a \emph{soft} margin, which is immune to the outliers and, therefore, generalizes well.
By definition, hyperplane can separate only classes that are lineary separable.
To separate non-lineary separable classes dedicated \emph{kernel} functions are used.
One of the factors which enabled SVMs to flourish is their relatively low computational and memory requirements.

Current renaissance of neural networks started in 2006
with a proposal of methods that enable training of NN with more than one hidden layer.
We will discuss deep neural networks later in this chapter.



\subsection{Undirected topic models} \label{sec:undirected_topic_models}

As we mention in \sectionref{topic_modeling}, classic topic models are unable to capture convoluted, non-linear relationships between
word distributions in topics. To solve this problem, Hinton \& Salakhutdinov proposed the \emph{replicated softmax}~\cite{hinton2009replicated}
binary topic model. The model is a special variant of the restricted Boltzman machine~(RBM)~\cite{smolensky1986information},
two-layer undirected generative model, which is often trained with the Contrastive Divergence~(CD)~\cite{hinton2002training} algorithm.
Original CD assumes a model with binary input and output units. However, in the case of topic modeling input should model word counts.
To this end, Hinton \& Salakhutdinov had to modify CD algorithm to account for word counts. The authors demonstrated that
replicated softmax generalize better than LDA, i.e. produces better topic distributions for unseen documents. Moreover,
since RBM is an undirected graphical model, not only word distributions in a document are conditioned on topic distributions
but also topic distributions are conditioned on word distributions.

\subsection{Word embeddings}\label{sec:word_embeddings}

As we wrote in \sectionref{text_vector}, an inherent limitation of the bag-of-words representation
is that each word is assigned to a separate dimension, which causes sparseness and high-dimensionality.
If we were to use this model to create vector representations of words, we would end up with a one-hot encoding.
The one-hot encoding conveys no information about meanings of words.
In particular it does not reflect whether given words are similar to each other or completely different.
Such a representation is called a \emph{discrete} or \emph{local} representation.

Alternatively, we can encode words using a \emph{distributed} representation~\cite{hinton1984distributed, hinton1986learning},
which describes each word using a vector from a relatively low-dimensional continuous vectors space.
Since words are \emph{embedded} in a low-dimensional vectors space, those representation are often called word \emph{embeddings}.
Embeddings capture semantic and syntactic relationships between words.
When using embeddings, semantically or syntactically similar words are represented by `similar' vectors,
i.e. vectors having low cosine distance.
Individual dimensions in embedded space do not have any specific interpretation.
It is only the distance between points in a vector space that is meaningful.

Word embeddings are often learned by taking advantage of the \emph{distributional hypothesis}~\cite{harris1954distributional}.
According to this hypothesis words that occur in the same contexts often have similar meanings.
One of the limitations of relying on co-occurrence is that antonymous words can sometimes be placed near each other in the vectors space
For example words \emph{good} and \emph{bad} often occur in similar contexts and therefore could end up having similar vectors,
which in turn would make them useless in some downstream tasks, like sentiment analysis.

In practice word embeddings are often trained using neural networks. Probably the first significant
neural network-based word embedding model was proposed by Bengio et al. in the form of a \emph{neural probabilistic language
model}~\cite{bengio2003neural}. Neural probabilistic language model is a simple feedforward NN with a linear input layer and a non-linear
hidden layer, similar to the one depicted in~\figref{neural_network} on page~\pageref{fig:neural_network}.
The input layer defines \emph{projections} from one-hot encoding of words to low-dimensional vectors.
The network is initialized with random weights and is trained using stochastic gradient descent.

Bengio et al. work inspired several other researches. Among them were the authors of \emph{word2vec}~\cite{mikolov2013efficient} software
library. Word2vec implements two separate embedding algorithms.
They are conceptually different, but similar from a computational point of view.
The first algorithm is called \emph{continuous bag-of-words} (CBOW) and it learns word vectors by trying to predict a word given its context.
To this end, CBOW defines two vector representations for each word $ w $ from the vocabulary $ V $, namely \emph{input embedding vector}
$ \vect{v}_w $ and \emph{output embedding vector} $ \vect{u}_w $. The probability of the center word $ w $ given its context $ C_w $
is defined as:
\begin{equation}
P(w \mid C_w) = \frac{\me^{\vect{v}_w^\mathrm{T} \vect{r}}}{\sum_{w' \in V} \me^{\vect{v}_{w'}^\mathrm{T} \vect{r}}},
\end{equation}
where $ \vect{r} $ is a vector representation of the context $ C_w $, defined as:
\begin{equation} \label{eq:cbow_ctx}
\vect{r} = \sum_{w \in C_w} \vect{u}_w.
\end{equation}
During training CBOW maximizes the log-probability: $ \log P(w \mid C_w) $.
The context $ C_w $ is usually defined as a fixed number of words to the left and to the right of the center word.
Alternatively, the context can be defined as simply a fixed number of preceding words.

The second algorithm, called \emph{skip-gram}, follows a basic structure introduced with CBOW. However, instead of predicting the center
words given their contexts, it predicts the context words $ c $ given the center words $ w $. To this end, it maximizes the log-probability:
$ \log \prod_{c \in C_w}{P(c \mid w)} $, where:
\begin{equation} \label{eq:skip_gram}
P(c \mid w) = \frac{\me^{\vect{v}_w^\mathrm{T} \vect{u}_c}}{\sum_{c' \in V} \me^{\vect{v}_{w}^\mathrm{T} \vect{u}_{c'}}}.
\end{equation}

In practice, both word2vec models are implemented as simple neural networks with just one hidden layer and two weight matrices.
Skip-gram network looks like the one presented in~\figref{neural_network}, where the weight matrix $ \vect{\Theta}_1 $ contains input
embedding vectors $ \vect{v}_w $, the weight matrix $ \vect{\Theta}_2 $ contains output embedding vectors $ \vect{u}_w $ and
the the output activation function is softmax (\equationref{softmax}). The CBOW model can also be seen as a neural network similar to the
one depicted in~\figref{neural_network}. However this model contains a summation operation (\equationref{cbow_ctx}) between the weight
matrices.

Embeddings in the word2vec models are learned as a side-effect of a multinomial classification.
Therefore, the loss function compares the probability distributions over center words (in the case of CBOW) or context words (in the case of
skip-gram) with a given one-hot encoding of the ground truth.
However, using a standard softmax for predicting a target word would be extremely computationally demanding.
In particular, the softmax normalization factor needs to be computed by summing terms from all vocabulary words.
Therefore, some approximation of the full softmax is needed. In the follow-up paper,
Mikolov et al.~\cite{mikolov2013distributed} suggested using one of the two approximate cost functions,
namely hierarchical softmax~\cite{morin2005hierarchical} or negative sampling.
Hierarchical softmax builds a Huffman binary tree where leaves are all the words from the vocabulary.
In order to estimate the probability of a given word, one traverse the tree from the root to a leaf.
Negative sampling, on the other hand, is a simplification of Noise Contrastive Estimation~\cite{gutmann2010noise} technique.
Thorough analysis of hierarchical softmax and negative sampling loss functions,
as well as derivation of gradients for both word2vec algorithms, is presented in~\cite{rong2014word2vec}.


Measuring the quality of word vectors is not an easy task. To do this, Mikolov et al. created
Semantic-Syntactic Word Relationship test set\footnote{Available at \url{https://github.com/dav/word2vec/tree/master/data}},
which contains almost 20k semantic and syntactic questions for words and almost 3k for phrases. The questions are in form:
\emph{X is to Y as Z is to what?} For example: \emph{Poland is to Polish as England to what?}
Mikolov et al. demonstrated that both skip-gram and CBOW outperform earlier word vector models,
especially the one introduced in~\cite{bengio2003neural}, on both semantic and syntactic questions.
On semantic questions higher accuracy was obtained using skip-gram model while on syntactic questions CBOW performed better.

One of the limitations of skip-gram model is that morphology of words is ignored.
Two words sharing some common lemma are treated as separate entities.
In many cases the algorithm will learn similar vectors for those words because they occur in similar contexts.
However, if some variant of a given word is rare, the vector learned by the model could be placed
far away in the vector space from the vector of the main form of the word.
This is particularly important in the case of natural languages with rich morphology, like Finnish or German.
To overcome this limitation Bojanowski et al.~\cite{bojanowski2016enriching}
enriched skip-gram with subword information.
To this end, they condition the probability of context words not on a center word vector but on a \emph{sum} of
the center word vector and its subword vectors. In their experiments they consider character n-grams of size 3, 4, 5 and 6.
Since the number of all posible character n-grams is huge, the authors place them in some fixed-size hash table (e.g. $ 10^6 $ elements) and
embeddings are learned for hashes instead of n-grams.
Bojanowski et al. report results superior to the original skip-gram both on word similarity and analogy tasks.
Their extension was initially implemented as a part of the \emph{fastText}\footnote{Available at
\url{https://github.com/facebookresearch/fastText}} software library.

Recently Wu et al.~\cite{wu2017starspace} proposed a \emph{StarSpace} model\footnote{Reference implementation available at
\url{https://github.com/facebookresearch/StarSpace}}, which has a different training objective than skip-gram and fastText.
Instead of learning to predict context words based on the center word it learns to compare words,
i.e. to accurately tell whether two word embeddings are similar, or dissimilar, given some similarity function (e.g. cosine).
To this end, the loss function is computed on a set of sampled positive and negative examples.
The model is able to learn not only word embeddings but also embeddings for other types of inputs. To this end, Wu et al. introduce a notion of \emph{entity}
represented by a set of discrete features. StarSpace is able to embed different types of entities in the same space. This is useful for
document classification (classify by finding nearest labels to a given document) or recommendations (recommend items to a given user
by finding nearest items). Wu et al. report state-of-the-are results in a variety of tasks. Their model is also very fast, partially due to
the use of Hogwild~\cite{recht2011hogwild} optimizer, which is a parallel asynchronous version of stochastic gradient descent.

As proved by Levy \& Goldberg~\cite{levy2014neural}, when skip-gram is optimized with negative sampling, it is implicitly factorizing
a word-context co-occurrence matrix. To be more specific, this factorization can be seen as truncated Singular Value Decomposition.
Similar observation was made by Pennington et al.~\cite{pennington2014glove}.
Moreover, the authors propose their own embedding method, dubbed \emph{GloVe},
where they explicitly create co-occurrence matrix $ \vect{X} $ for all vocabulary words.
Each cell $ x_{i,j} $ of the matrix represents the number of times words $ i $ occurs in the same context as word~$ j $.
This defines the probability that word $ j $ appears in the same context as word $ i $:
\begin{equation}
P_{i,j} = \frac{x_{i,j}}{x_i},
\end{equation}
where $ x_i $ is the number of occurrences of the word $ i $.
Non-zero elements of this sparse co-occurrence matrix are passed as an input to the GloVe learning algorithm.

More recent embedding methods were proposed by Shazeer et al.~\cite{shazeer2016swivel} and by Xun et al.~\cite{xun2017collaboratively}.
The first of those methods is called \emph{Swivel} and was designed to work in a distributed environment
and be trained with larger text corpora than word2vec or GloVe. The second method assumes that embeddings are learned not from a continuous
corpora but from a set of text documents. This allows it to leverage both local and global contexts when learning
embeddings. Local context is interpreted as co-occurrence matrix while global context is a topic model for an enclosing document

\subsection{Applications of word embeddings to non-NLP domains}

Interestingly, language modeling can be applied also to non-NLP domains.
This is possible because words in sentences can be treated as just some identifiers in some sequences.
For example, Perozzi et al.~\cite{perozzi2014deepwalk} apply word2vec to learn distributed representations of vertices in social networks.
In order to achieve this, they apply truncated random walks in networks and they treat the walks as sentences.
To resemble sentences, walks need to start from random places in the graph and be truncated after passing a few nodes.
Embeddings are learned as an auxiliary task of predicting vertex given previously visited vertices.
They call their method \emph{DeepWalk}. The authors demonstrate state-of-the-art results in multi-label classification tasks.

An extension of DeepWalk is \emph{node2vec}~\cite{grover2016node2vec}.
Instead of learning representations of vertices, the authors suggest learning representations of nodes in a graph.
node2vec outperforms DeepWalk in the multi-label classification tasks.

Word embedding models were also adapted for recommendation systems. For example, Spotify\footnote{One of the biggest music streaming services:
\url{https://www.spotify.com/}} uses word2vec to learn 40-dimensional song embeddings. To this end, users' play lists
(or play queues) are treated as sentences. The assumption is that songs occurring close to each other on play lists
are similar. Embeddings are learned only for songs which have been played at least 500 times.
This way, embeddings are learned for approximately $ 4 \times 10^6 $ tracks.
Feature vectors for newly added songs, or songs rarely played, are discovered using convolutional neural network
based on only audio signals\footnote{For details see \url{https://youtu.be/ZOyBfpcFFVE?t=2488}}.

\subsection{Multi-sense word embeddings} \label{sec:multi_sense_word_embeddings}

\begin{figure}[htb!]
  \centering
  \includegraphics[width=0.4\linewidth]{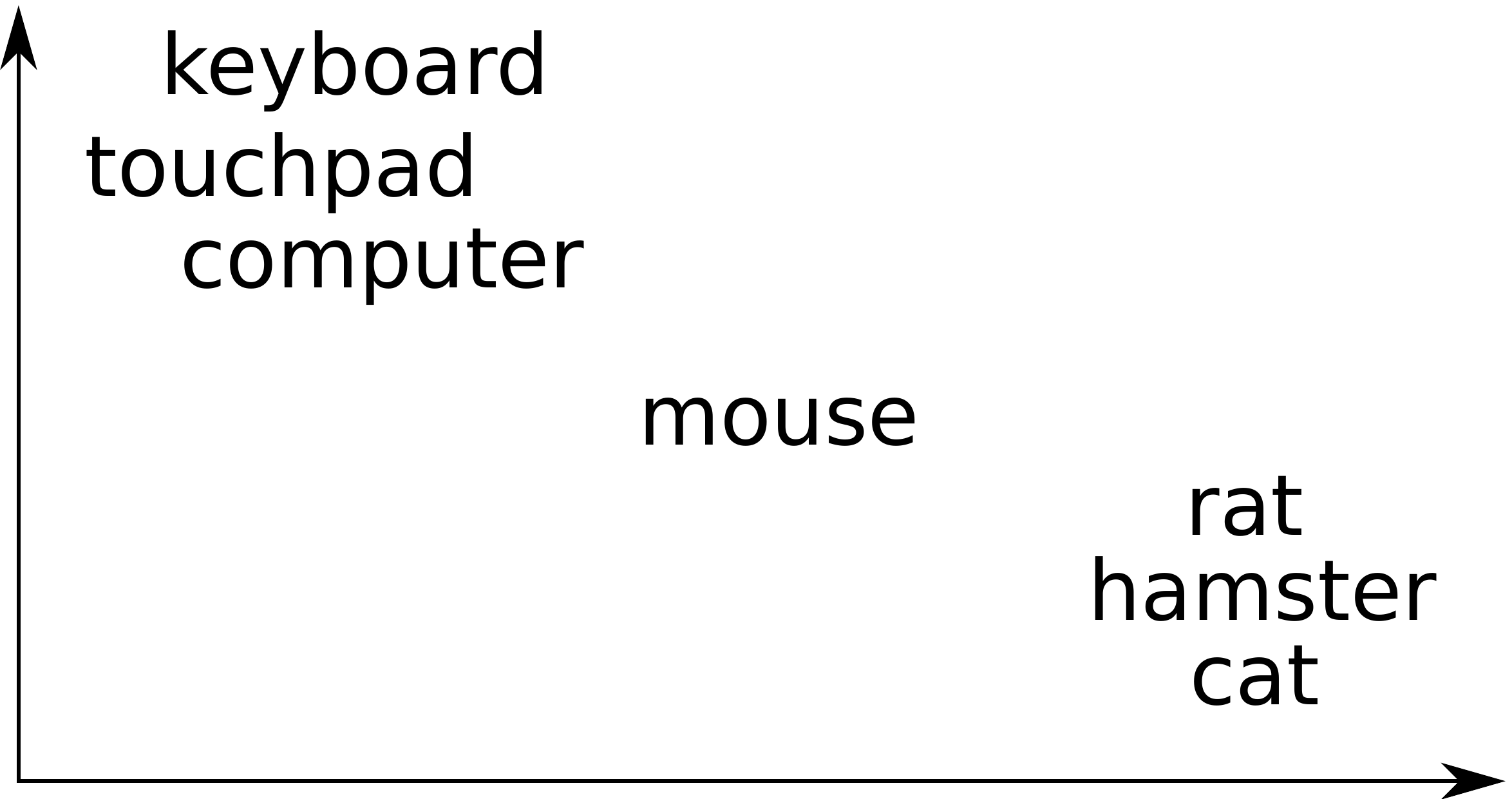}
  \caption{Polysemous word \emph{mouse} pulls clusters of computer parts and cluster of domestic animals to each other.}
  \label{fig:triangle_inequality}
\end{figure}

Word embeddings are extremely useful but they still have one limitation:
they represent each word, even polysemous or homonymous one, by a single vector.
Another problem with ambiguous words is that they pull clusters of separate word domains to each other.
This is depicted in~\figref{triangle_inequality}.
We can alleviate this problem by crating a multi-prototype vector-space model, as described in \sectionref{wsd}.
Recently researches have been trying to tailor the multi-prototype vector-space model to word embeddings.
For example in~\cite{huang2012improving} the authors present a new neural network model that learns word embeddings
by predicting words based on their contexts (as in case of CBOW model) and on a \emph{global context}.
The global context is a weighted average of all word vectors in a given document.
Then, they carry out context clustering as in~\cite{reisinger2010multi}.
Since the clustering need to be performed separately for each word form the dictionary,
this method does not scale well to very big dictionaries.

Even more revolutionary work is~\cite{neelakantan2015efficient},
where the authors propose a multi-sense skip-gram (MSSG) model. It is an extension of skip-gram
that directly learns multiple sense vectors of words.
As in the case of skip-gram, they train a neural network to predict context words
and they get embeddings as a side effect of optimization.
However, instead of conditioning prediction on a center word vector
they condition it on a center word sense vector.
To this end, for each training example they predict a sense of the center word, prior to predicting a context words.
They do this by first measuring similarity between a context vector and cluster centers for all learned senses
and then selecting a sense (cluster) that is nearest to the context.
They build context vector by averaging \emph{global} vectors of context words. Global vectors are trained in addition to sense vectors
and are used exclusively to build vectorized representations of contexts.
Cluster centers are stored for all senses of all vocabulary words and are updated after each training example.
Neelakantan et al., evaluate the MSSG model in the word similarity task, with a special accent on contextual word similarity.
The model outperforms state-of-the-art models, in particular~\cite{huang2012improving, mikolov2013efficient}.
One advantage of~\cite{neelakantan2015efficient} over~\cite{huang2012improving}
is that the former does not require explicit context clustering prior to network training.
However, context clustering is still implicitly performed during the network training.
Therefore, this model cannot be seen as purely probabilistic, which can be considered as its limitation.

A probabilistic model for learning multi-sense embeddings was proposed in~\cite{tian2014probabilistic}.
Their model is called multi-prototype skip-gram~(MPSG). 
The authors use finite Gaussian mixture model to model word senses.
Its latent variables are estimated using \emph{expectation-maximization} (EM)
algorithm~\cite{dempster1977maximum}. MPSG gives almost as good results in contextual word similarity task as~\cite{huang2012improving}.
At the same time it is much faster and memory efficient than~\cite{huang2012improving}. MPSG assumes a fixed number of word meanings.
This limitation is addressed by adaptive skip-gram (AdaGram) model proposed in~\cite{bartunov2016breaking}.
This model can be seen as a non-parametric variant of~\cite{tian2014probabilistic}, where a number of senses is
discovered separately for each word. Specifically, AdaGram employs a Dirichlet process to model multiple senses.
Latent variables in AdaGram are also estimated using an instance of the EM algorithm. 
The authors test their solution in a word sense induction task and for that they introduce a new Wikipedia Word-sense Induction (WWSI)
dataset consisting of almost 200 target words and over $ 3.5 \times 10^4 $ contexts.
They also suggest that contextual word similarity task is not a good evaluation method for multi-sense word embeddings models.
This opinion is confirmed by an observation made by~\cite{upadhyay2017beyond}, that achieving good results in the word sense induction task
does not necessarily correlates with good results in the contextual word similarity task.
This also correlates with the observation made by~\cite{faruqui2016problems} that word similarity evaluation is not a good criterion
not only for multi-sense word embeddings but even for single-sense embeddings.

The authors of \cite{qiu2016context} point out that both~\cite{tian2014probabilistic}
and~\cite{bartunov2016breaking} model sense embedding based on single sense word embeddings of context words (i.e. global vectors).
They propose a novel probabilistic model that takes into account the relations between senses of neighboring words.
Specifically, they use a hidden Markov model, where words are observations and senses are hidden states. Latent parameters of their model are estimated
using a variant of the EM algorithm. They report state-of-the-art results on a word sense induction task.

A slightly different approach is presented by~\cite{liu2015topical}, where the authors use topic modeling to discover multiple word senses.
Specifically, they use latent Dirichlet allocation~\cite{blei2003latent}
to model hidden \emph{topics} over all vocabulary words.
In this context the topics are interpreted as senses.
Therefore, a probability distribution over all the senses is associated with each word.

There are also models that learn multi-sense embeddings using external ontology (e.g. WordNet).
Examples are \cite{chen2014unified, jauhar2015ontologically}.
Finally, in~\cite{li2015multi}, the authors discuss usefulness of multi-sense word models.
They show that in some NLP tasks multi-sense embeddings do not outperform single-prototype ones.

\subsection{Paragraph and document embeddings}\label{sec:document_embeddings}

Until now we were discussing distributed representations of words.
However, in many cases it is useful to have distributed representations of paragraphs or even whole documents.
The simplest way to do this is to take a weighted average of embeddings of all words that occur in a given document.
One active field of research is to directly learn distributed representations of groups of words or even whole documents.
Probably the most important work that goes in this direction is the Paragraph Vector~\cite{le2014distributed} model,
commonly known as \emph{doc2vec}.
Paragraph Vector is a simple yet powerful extension to word2vec.
As in case of word2vec, Paragraph Vector is not a single algorithm, but two conceptually different, yet computationally similar, models.
They are Paragraph Vector Distributed Memory (PV-DM) and Paragraph Vector Distributed Bag of Words (PV-DBOW).
PV-DM, like CBOW, tries to predict target word based on context words.
However, unlike CBOW, PV-DM also takes a paragraph embedding (or a document embedding) as an input.
Paragraph embedding can be either concatenated with word embeddings or averaged.
This approach is similar to the one proposed in~\cite{huang2012improving}.
However, in contrast to the model described in~\cite{huang2012improving},
paragraph embeddings in PV-DM model are not weighted average of word embeddings but separate vectors that are learned during training.
As a result, PV-DM learns simultaneously both word embeddings and document embeddings. 

Paragraph Vector Distributed Bag of Words (PV-DBOW) is a simpler model than PV-DM.
It predicts all words in the document based on the document embedding.
Therefore, in contrast to PV-DM, PV-DBOW does not learn word embeddings, but only document (or paragraph) embeddings. 

Both Paragraph Vector models work in two modes, namely training mode and inference mode.
During training the model is fed with training data and document embeddings as well as softmax weights are modified.
Later, when we generate embeddings for new, previously unseen documents,
data is fed to the model in the same way, but only embeddings are modified in the optimization process.
Softmax weights are fixed in the inference phase. The inference phase is not as time-consuming as the training phase.
Nonetheless, the need to iteratively optimize the cost in order to obtain embeddings for new documents is one of the weaknesses
of Paragraph Vector.

Both Paragraph Vector algorithms achieve state-of-the-art results on text classification and sentiment analysis tasks.

\section{Deep learning} \label{sec:deep_learning}

As we mention in \sectionref{neural_networks},
due to the vanishing and exploding gradient issue training neural networks with more than one hidden layer is challenging.
Yet it is believed that the human brain works in a multi-tier fashion, where data is partially processed
by different visual cortices at different abstraction levels.
Therefore, for years researchers were trying to tackle the vanishing and exloding gradient problem.

\begin{figure}[htb!]
  \centering
  \includegraphics[width=0.8\linewidth]{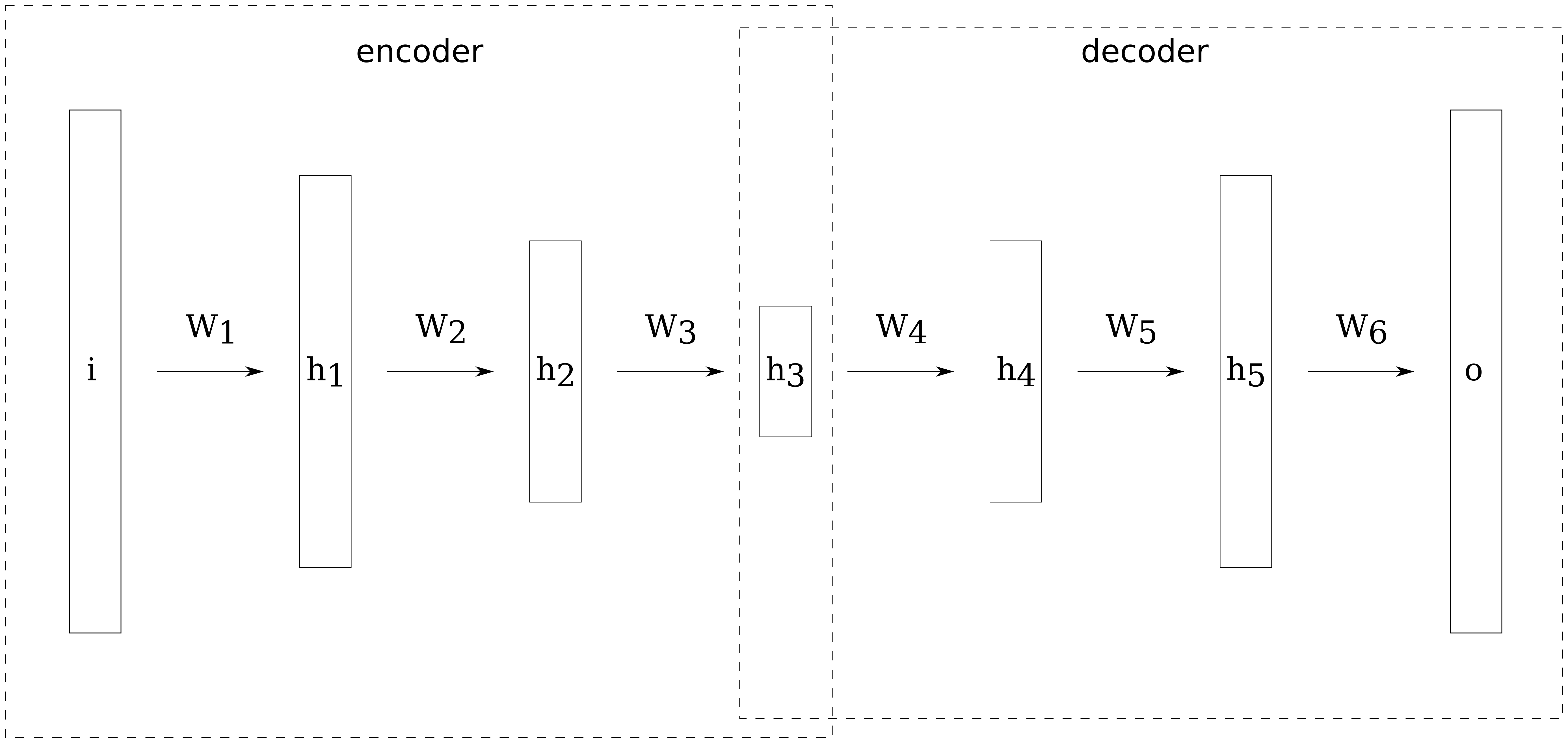}
  \longcaption{A deep autoencoder with hidden layers $ h_1 $ to $ h_5 $ and weight matrices $ W_1 $ to $ W_6 $.}
              {The goal of the training is to restore the output $ o $ that resemble input $ i $ as closely as possible.
               After the training, an encoder part of the autoencoder can be used to generate
               a low-dimensional representation $ h_3 $ of input data $ i $.}
  \label{fig:deep_autoencoder}
\end{figure}

The breakthrough results were published by Hinton et al. in 2006~\cite{hinton2006reducing}.
The authors noticed that the error backpropagation can be successfully employed to train multilayer neural networks
as long the weights are not randomly selected but already carry some knowledge about the training data.
To this end, the authors used a so-called deep belief network~(DBN),
which is a stack of restricted Boltzman machines~(RBMs), to \emph{pre-train} the network.
This pre-training phase is unsupervised, i.e. it does not require labeled data.
In practice, Contrastive Divergence
or Persistent Contrastive Divergence~\cite{tieleman2008training} learning algorithms are often used to train RBMs layer-by-layer.
After pre-training, the network can be \emph{fine-tuned} in either supervised or unsupervised manner.
In the first case, the network is treated as a standard multilayer perceptron (MLP)
and is fine-tuned with backpropagation algorithm, where the model loss is calculated based on the labels.
In the second case, the pre-trained network is \emph{unfolded} to form a deep autoencoder, as the one presented in~\figref{deep_autoencoder}.
In an unfolded network the second half of layers is an transposed copy of the first half.
After unfolding, the deep autoencoder is fine-tuned using error backpropagation.
The goal of the training of a deep autoencoder is to be able to recreate the input at the output of the network.
As a result of the training, a fixed-size (often low-dimensional) representation of the input data
is created in the center layer of the network.
As reported in~\cite{hinton2006reducing}, both described deep architectures yielded state-of-the-art results
on multiple tasks, including image classification and information retrieval.

Hinton's work brought about a revival of interest in neural networks. Many important works were published in the years to come.
Nair et al. showed~\cite{nair2010rectified} that using a rectified linear function instead of a sigmoid function
as an activation of hidden layers of the network noticeably improves the performance and stabilizes the training process.
Srivastava at al. proposed~\cite{srivastava2014dropout} a new regularization method called \emph{dropout}.
Their idea is to randomly disable some fraction of units in hidden layers.
More specifically, neurons' activations are randomly set to zero with some specific probability ($ 0.5 $ is a common value).
Dropout quickly became an integral part of neural network researcher's toolbox.
Wan et al. extended dropout to randomly disable some neurons' inputs, independently for each neuron in the layer.
In order to achieve this, weights are randomly set to zero. Their method is called \emph{DropConnect} \cite{wan2013regularization}
and they report results superior to dropout.

More recent deep learning research showed that it is possible to efficiently train deep neural networks without the pre-training phase
(e.g.~\cite{martens2010deep}).
However, it is believed that unsupervised, generative training is not less important than the supervised one
and, therefore, researchers are trying to improve the performance of pre-trained networks as well.
An example of work that goes in this direction is~\cite{sparseInit2015},
where authors adapted the \emph{Sparse Initialization} technique,
originally proposed for the network trained without pre-training~\cite{martens2010deep}, to initialize deep belief network.
Other modification to DBN training was proposed in~\cite{orthogonality2016},
where the authors encourage weight vectors to be orthogonal to each other during Contrastive Divergence training.
The idea is not novel in itself, but it was previously applied only to the networks trained in an supervised manner.

Other important deep learning research include batch normalization~\cite{ioffe2015batch}
and deep residual learning~\cite{he2015deep}, which enables training of neural networks with hundreds or even thousands layers.
A comprehensive introduction to deep learning can be found in~\cite{goodfellow2016dl, lecun2015deep}.

\subsection{Deep architectures}

There are many architectures of deep neural networks. One of the most popular is a \emph{convolutional neural network}~(CNN).
Originally CNNs were designed to process images.
To this end, they process input data sequentially, using small sliding window.
At each position of the window the same network weights are used.
Due to this parameter sharing, input images can have high resolution
, while weight matrices can be kept relatively small.
One of the consequences of this architecture is that not all input neurons have connections to all output neurons.
Because of this, they are not \emph{fully-connected} layers.
To combine results from outputs of convolutional layers, \emph{pooling} layers are used.
CNNs can have many alternated convolutional and pooling layers.
CNNs obtain state-of-the-art results in object detection
and scene classification tasks~\cite{lecun1998gradient, krizhevsky2012imagenet, szegedy2015going}.
Recently, they are used not only for image processing but for other content types as well, including text (e.g.~\cite{kim2014convolutional}).

Another popular architecture is a \emph{recurrent neural networks}~(RNN).
RNNs dates back to the down of artificial neural networks.
While CNNs use shared parameters in space, RNNs use shared parameters in time.
Prior to discovery of backpropagation algorithm,
RNNs remained rather conceptual models without practical applications.
A few years after backpropagation was proposed, its recurrent variant,
namely backpropagation through time~\cite{werbos1990backpropagation} (BPTT), was derived.
However, since BPTT suffers from vanishing and exploding gradinet problem no less than the classic error backpropagation,
training RNN was challenging.

One of the successful attempts to tackle this problem is to use memory cells with gated units, instead of simple neurons.
This approach was introduced in the late 1990s in the form of long short-term memory~(LSTM)~\cite{hochreiter1997long}.
Many variants of LSTM have been developed.
One of the extensions to the LSTM, which is gaining popularity, is Gated Recurrent Unit~\cite{cho2014learning} (GRU).
GRU is conceptually simpler than LSTM and perform no worse than it.

In recent years we observe a resurgence in RNNs.
For example, Sutskever et al. demonstrated~\cite{sutskever2011generating} how one can build character-level language model
by feeding big text corpus, character-by-character, to an RNN.
After training, the model can predict with high accuracy the next character in a stream of text.
This prediction can be used to automatically correct spelling mistakes or to implement an automatic language translation system.
Graves~\cite{graves2013generating} trained an RNN on handwritten characters.
Trained model was able to generate sequences of images which, when combined, looked like handwritten text.
In~\cite{yuan2016word} the authors demonstrate how to use LSTM networks to perform word sense disambiguation.
RNNs are used not only in academia but are already becoming popular in industry in form of production-ready software.
For example \emph{Google Allo} instant messaging mobile application uses LSTM-based RNN to suggest answers based on the conversation
history\footnote{\url{https://research.googleblog.com/2016/05/chat-smarter-with-allo.html}}.
LSTMs are also used for \emph{Smart Reply} feature recently added to
Gmail\footnote{\url{https://research.googleblog.com/2017/05/efficient-smart-reply-now-for-gmail.html}}.
Another popular application of LSTM is an Apple's \emph{QuickType}
keyboard\footnote{\url{https://www.techleer.com/articles/161-unfolding-of-rnn-popular-deep-learning-model}},
which suggest the next word to be typed.

Finally, recurrent neural networks should not be confused with recursive neural networks.
The later are designed to learn tree-like structures. One of the common applications is part-of-speech tagging.
For example,~\cite{zhu2015long} combine RNNs with recursive networks to obtain state-of-the-art result on sentiment analysis task.

Another deep architecture that has been gaining popularity in the recent years is a \emph{generative adversarial network} (GAN)
\cite{goodfellow2014generative}. GAN is a system of two networks working together. One of the networks is responsible
for stochastic generation of some type of data (e.g. images). The second network takes as an input either the output of the first network
or the real data provided by the user. The second network is optimized to be able to distinguish whether the input was real or confabulated.
However, the loss is also propagated to the first network, which causes this network to learn to generate data that
resemble real data. Consequently, the two networks compete with each other. GANs are used for, example, to generate highly realistic images,
e.g. for computer games' scenes.

Another neural architecture that has recently been successfully applied in deep settings is a \emph{siamese network}.
The siamese network is a neural model that consists of two or more identical subnetworks with shared parameters.
The subnetworks are connected at the top by some output layer, which measure the difference between outputs of the two subnetworks.
In general, supervised learning requires a lot of labeled data to perform well during prediction.
However, sometimes we want the network to be able to distinguish objects based on just one example from each of the target classes.
This is called \emph{one-shot learning} and it is one of the main applications of deep siamese networks.
Interestingly, the network is trained not with just one example from each class but multiple examples from each class.
However, those classes are different than classes used during prediction. During training, pairs of examples are sampled
from this multi-example-per-class training set. The pair consists of either two examples belonging to the same class
or to two different classes. Effectively, the network learns to distinguish whether
two examples belong to the same class or not. In the actual reference dataset, as we mention above, we have just one example
from each of the test classes. Prediction boils down to comparing a test example with all of examples from this
single-example-per-class dataset and selecting the class of the most similar example. Recently, convolutional siamese network proved
to yield state-of-the-art results in one-shot image recognition task~\cite{koch2015siamese}.

\subsection{Thought Vectors}

One could argue that Paragraph Vector models, described in \sectionref{document_embeddings},
can be used to generate dense representations of sentences,
by treating a sentence as a short document.
However, recently some works have been published which directly address a vectorized representation of sentences.
Those representations are often called \emph{though vectors}.
The name is based on an assumption that a single sentence carries just one thought.
The most important work that goes in this direction is \emph{skip-thought vectors}~\cite{kiros2015skip}.
The authors propose a model which is inspired by the skip-gram model from~\cite{mikolov2013efficient}.
Instead of predicting context words based on a given center word,
they predict two surrounding sentences (one to the left and one to the right) based on a given center sentence.
Prediction is done based on vectorized representations of sentences.
The center sentence needs to be encoded to form a vector. Consequently, surrounding sentences need to be decoded.
Skip-thought vectors is a generic approach. Any encoding and decoding model can be used.
The authors of~\cite{kiros2015skip} experimented with convolutional networks, LSTM-based recurrent networks and GRU-based networks.
A single cost function is defined based on encoder and two decoders (for sentences to the left and right).
The training leads to the optimal encoder and decoders parameters.
After training, any sentence can be encoded to obtain its though vector.
One advantage of skip-thought vectors over Paragraph Vector models is that the former does not require optimization at the test time,
i.e. inference phase is not needed.

Kiros et al. reported state-of-the-art results, outperforming known methods on semantic relatedness, paraphrase detection
and text classification tasks. However, more recently, one simple baseline method was proposed, which outperforms skip-thought vectors.
Specifically, Arora et al.~\cite{arora2016simple} presented dense vectorized representation of sentences in form of weighted average
of word embeddings, where the weights are inversely proportional to word frequencies. What distinguishes Arora et al. approach from
earlier similar ones is that, in addition to weighting, they remove from the weighted average its projection on its first principal component.
Arora et al. provide a neat probabilistic interpretation of their method, which is based on the assumption that each sentence is
backed by one \emph{discourse} and computing the first principal component is used to estimate a discourse vector.
\section{Cluster analysis}

The goal of clustering is to group similar objects together or partition different objects into separate groups.
Since there are many ways of defining similarity, there is multitude of clustering algorithms as well.
Probably the simplest one is \emph{k-means} algorithm.
K-means partition observations represented by numerical features into k distinct groups.
The algorithm starts with randomly assigning all observation to one of the k clusters.
Then, we iteratively repeat two steps:
\begin{enumerate}
\item for each cluster compute its \emph{centroid} by averaging observations' vectors feature-wise,
\item reassign each observation to a cluster with a centroid closest to the observation.
\end{enumerate}
Euclidean distance is used to find the closest cluster. 
After convergence, we get data separated into k distinct groups with small within-cluster variance.
Simplicity and low computational requirements leads to high popularity of k-means.
There are many variations of this algorithm.
One modification often used for clustering text documents is \emph{spherical k-means}~\cite{dhillon2001concept}.
One serious limitation of k-means algorithms is that we have to specify the number of expected clusters.
Sometimes we do not know in advance how many distinct groups there are in our data.

For experiments described later on in this dissertation we use hierarchical clustering algorithms.
These algorithms do not require specifying an expected number of groups to which data should be partitioned.
Hierarchical clustering can by either agglomerative or divisive.
The former is a \emph{bottom-up} approach, where we start with a separate cluster for each example in a dataset,
and then we recursively group together the most similar clusters.
The latter is a \emph{top-down} approach, where we start with one big cluster containing all data examples,
and then we recursively split it into more compact parts.

There are many more clustering algorithms.
Good overview of cluster analysis techniques is presented in~\cite{aggarwal2013data}.
A comprehensive survey of text document clustering methods can be found in~\cite{anastasiu2013document},
which has been published as a chapter in~\cite{aggarwal2013data}.

\section{Novel vector representation of text data}

\begin{figure}[htb!]
  \centering
  \includegraphics[width=0.4\linewidth]{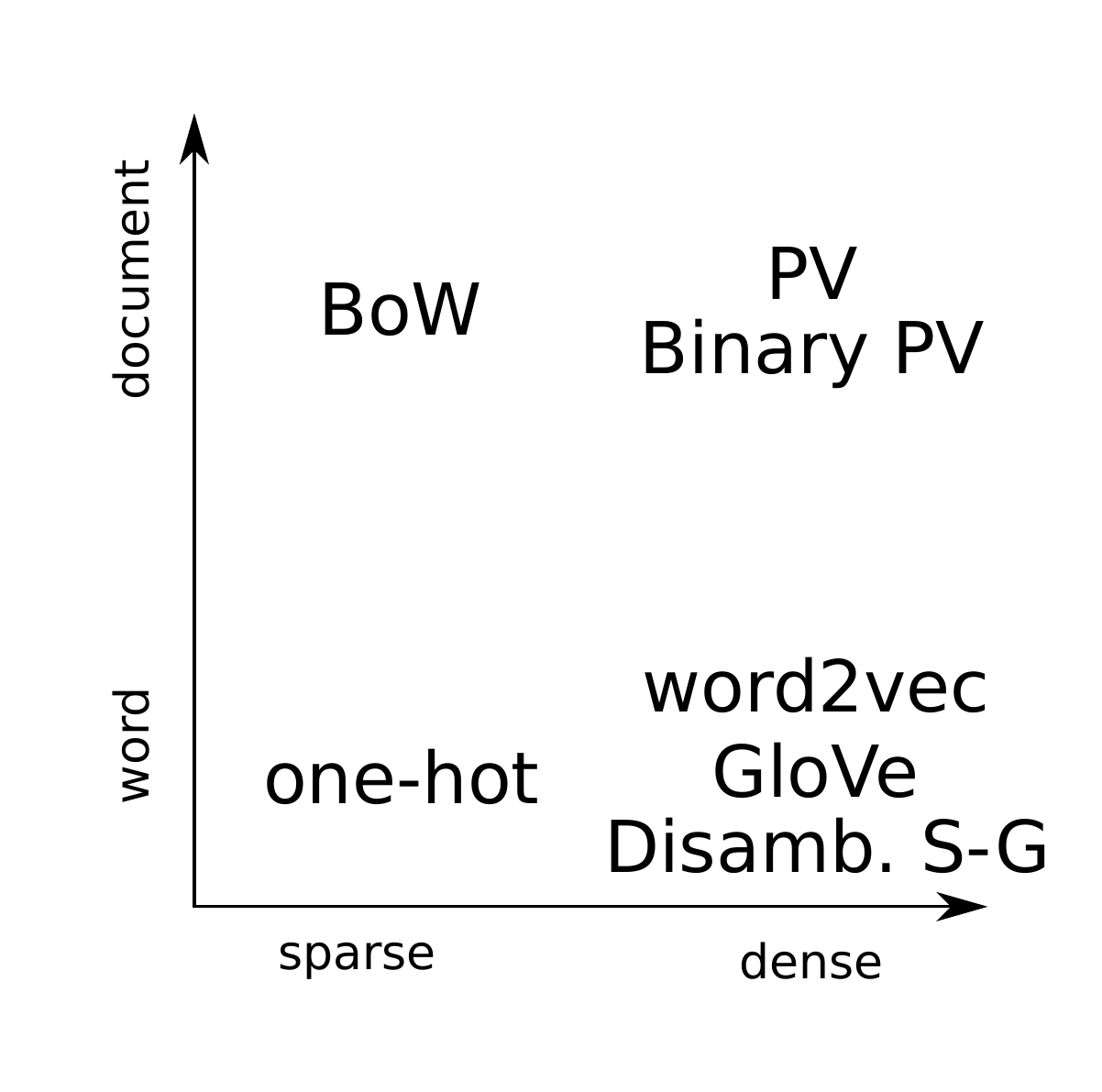}
  \caption{Vector representations of text data space with example models.}
  \label{fig:vector_rep_of_text_data}
\end{figure}

In~\figref{vector_rep_of_text_data} we summarize various ways of representing text data in a vector space.
In the following chapters we propose solutions that fit to one of the four areas of this chart.
Specifically, in \chapterref{dsh} we introduce a novel binary dense representation of text documents and demonstrate its
applicability to information retrieval task. In~\chapterref{disgram} we propose a simple neural network model for learning
multi-sense word embeddings. In both cases we compare proposed models with competing state-of-the-art solutions and demonstrate
their superiority.

\chapter{Binary Paragraph Vector models} \label{ch:dsh}

We live in the age of big data. Every day all around the world vast amount of data is collected by a variety of sensors, logging systems,
CCTV cameras and other devices. Moreover, due to social media popularity user generated content is also growing at a very high rate.
Data growth is considered to be exponential. According to~\cite{gantz2012digital} amount of digital data double every two years.
Therefore, being able to search and retrieve relevant information from huge datasets in relatively short time is crucial.
Basic information retrieval models, described in \sectionref{text_vector}, are not sufficient when the search space is large.

In many cases users are willing to compromise on quality of search results in favor of fast retrieval.
They prefer to have good but not ideal results immediately, rather than the best possible results after some delay.
One of the approaches that goes in this direction is approximate nearest neighbor~(ANN) search, pioneered by~\cite{arya1998optimal}.
There are many ANN algorithms available. Many of them employ hashing.
Hashing groups documents into \emph{buckets}. Each bucket has a short unique address.
A~fast \emph{hashing function} computes the address based on the content of the document,
enabling access in a constant time to other documents from the same bucket.
If the hashing function returns \emph{similar} addresses (i.e. addresses that differ in few bits only) for similar documents,
then in order to perform search we could compute an address for the query,
and then retrieve documents in the query's bucket and nearby buckets.
This idea was first implemented as Locality Sensitive Hashing~(LSH)~\cite{indyk1998approximate,broder1997resemblance}
in the late 1990s, and since then was extended in a variety of ways~\cite{charikar2002similarity,gong2011iterative,wang2014hashing}.
Most of the algorithms from this family are data-oblivious, i.e. can generate hashes for any type of data. Nevertheless,
some methods target specific kind of input data, like text, image or speech.
Thanks to deep learning a family of locality-preserving hashing methods is growing rapidly.
For example, Deng et al.~\cite{deng2010binary} demonstrated how to generate short binary codes of speech spectrograms.
More recently, Lin et al.~\cite{lin2015deep} generated binary hash codes of images
using convolutional neural network trained in supervised manner.
Binary codes have also been applied to fast cross-modal retrieval~\cite{wang2013semantic,masci2014multimodal}.

Fast retrieval of text documents was the main motivation for \emph{semantic hashing} model
proposed by Salakhutdinov \& Hinton~\cite{salakhutdinov2009semantic}.
The authors employ a deep autoencoder with a narrow coding layer to generate memory addresses of documents.
As described in \sectionref{deep_learning}, the deep autoencoder weights are pre-trained using deep belief network,
i.e. stacked restricted Boltzman machines, in a generative fashion.
Then, decoder layers are created as transposed copy of encoder layers and the whole network is fine-tuned using error backpropagation.
To make codes binary the authors used a sigmoid coding layer in the autoencoder.
However, that in itself would not be sufficient,
since sigmoid values could get stuck in a very narrow range and, therefore, not generalize well.
To force sigmoid activations to be distributed, the authors proposed to add Gaussian noise to the inputs of logistic units.
To counterbalance the noise, the network will learn to have activations either close to zero or close to one.
A~standard deviation of the Gaussian noise was selected based on the validation set.
After training, test data was feed through the encoder and its output was rounded to obtain binary codes.

Semantic hashing is reported to yield slightly better results than LSH and to be up to 50 times faster.
However, it still has one limitation.
The approach proposed by Salakhutdinov \& Hinton assumes that documents passed to the autoencoder network
are presented in a form of relatively low-dimensional bag-of-words vectors.
As discussed throughout this dissertation, there are good representations that can capture a lot more of the semantic content of documents.
Therefore, it would be desirable to be able to generate hashes from raw text documents, instead of their BoW representations.

In this chapter we propose a shallow neural network which does exactly that.
We combine semantic hashing with the Paragraph Vector model~\cite{le2014distributed}.
As the former, our model enables generation of relatively short binary codes of documents that compress or summarize their content.
However, in contrast to the original semantic hashing network, our model takes raw text documents as an input.
Therefore, our model combines two formerly separate approaches, namely document embedding and semantic hashing, in one.

\section{Model architecture} \label{sec:bpv_architecture}

\begin{figure}[b!]
  \centering
  \includegraphics[width=0.7\linewidth]{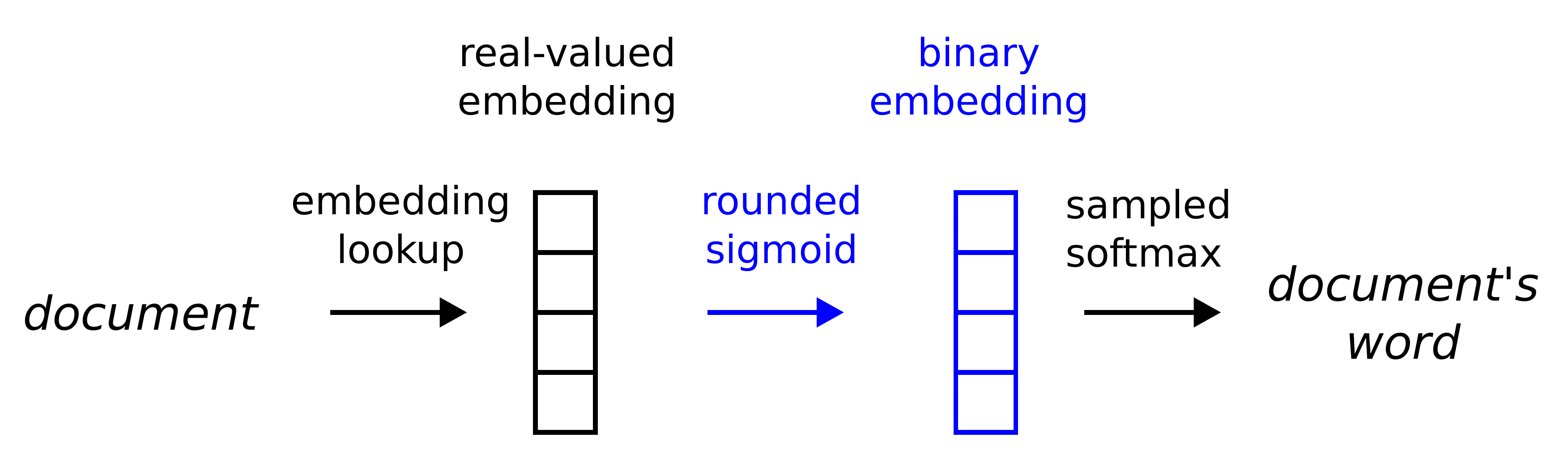}
  \caption{The Binary PV-DBOW model. Modifications to the original PV-DBOW model are highlighted in blue.}
  \label{fig:pv-dbow-bin}
\end{figure}

\begin{figure}[htb!]
  \centering
  \includegraphics[width=0.8\linewidth]{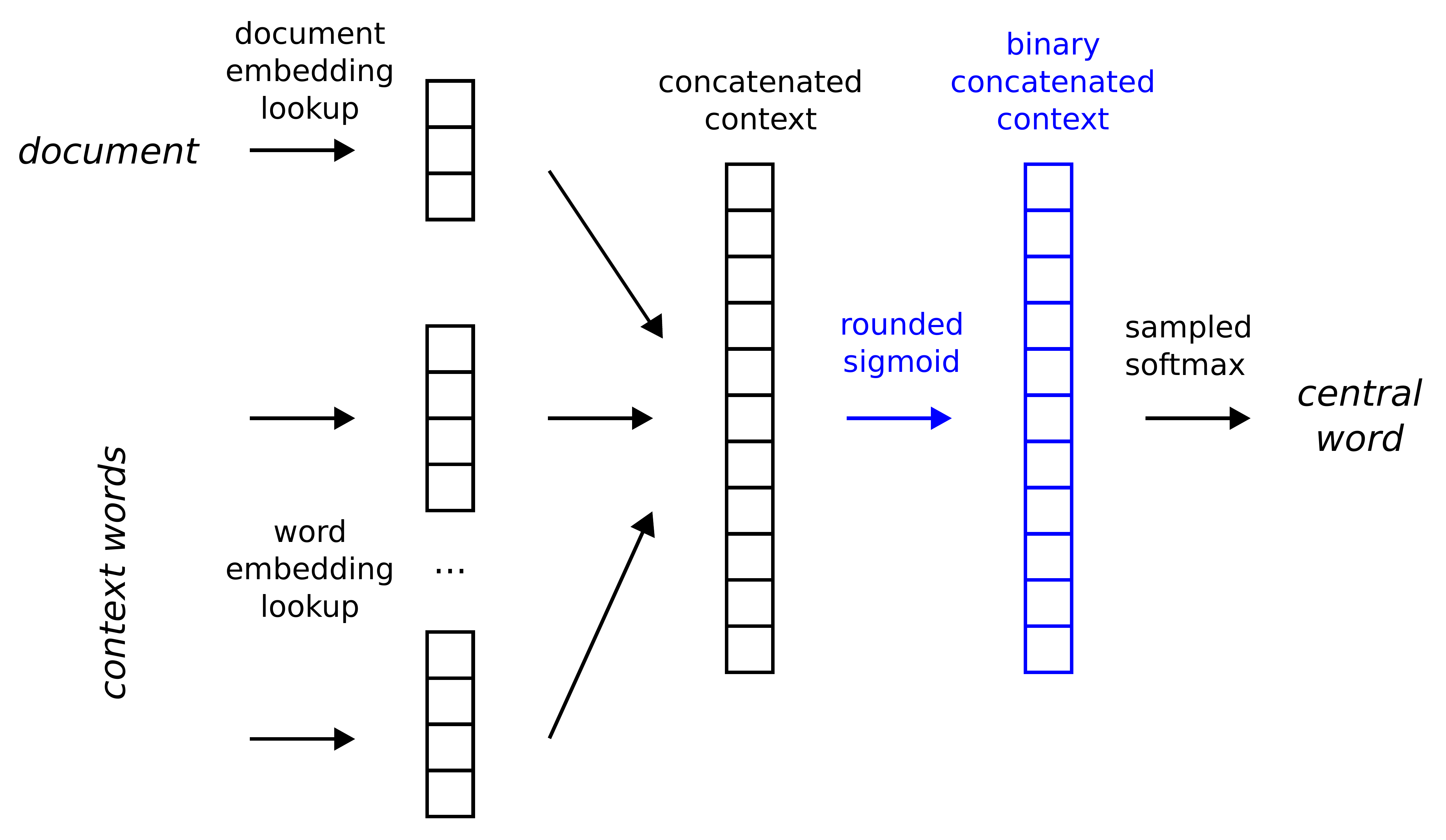}
  \caption{The Binary PV-DM model. Modifications to the original PV-DM model are highlighted in blue.}
  \label{fig:pv-dm-bin}
\end{figure}

As described in \sectionref{document_embeddings}, in their 2014 paper Le \& Mikolov proposed not a single machine learning model, but two
separate ones: Paragraph Vector Distributed Memory (PV-DM) and Paragraph Vector Distributed Bags of Words (PV-DBOW).
To learn binary codes we added a rounded sigmoid function to these two models. The resultant Binary Paragraph Vector models are depicted
in~\figref{pv-dbow-bin}
and~\figref{pv-dm-bin}.
Modifications to the original Paragraph Vector models are highlighted in blue.

As mentioned above, Salakhutdinov \& Hinton added Gaussian noise to the inputs of logistic units.
Later, Krizhevsky \& Hinton demonstrated~\cite{krizhevsky2011using} that binarization can be done in a simpler way.
In particular, they rounded the output of the logistic function during training to obtain binary activations.
This approach resemble the initial McCulloch \& Pitts neuron model with the Heaviside step function, which makes hard decisions.
However, since the rounded logistic function is not smooth, its derivative cannot be calculated everywhere
(it is zero for every value except 0.5 and positive infinity at 0.5). To overcome this limitation, Krizhevsky \& Hinton propose
to compute gradients for binarized layer like for a standard smooth logistic layer.

An alternative way for learning binary codes is to use binary stochastic neurons~(BSN) instead of deterministic ones.
BSN yield 0 or 1 at its output with a probability specified by the pre-acitvation.
There are many variants of BSNs. We evaluate some of them in \sectionref{bin_units_comparison}.
Nevertheless, of all tested approaches, the Krizhevsky's method yielded the best results
in our preliminary experiments. Therefore, we chose the Krizhevsky's approach as a binarization method in our models.

When seen as neural networks, both original Paragraph Vector models have just one hidden layer with a linear activation function.
The modification introduced in this chapter can be seen as a non-linear activation function for the hidden layer of those networks.

\subsection{Distributed bag of n-grams}

The paragraph vector distributed bag of words model learns paragraph embedding by predicting all words in a document.
Unlike the PV-DM model, PV-DBOW does not take a word context as an input, and therefore could be considered a weaker model.
We attempted to alleviate this weakness.

Instead of predicting document's words we tried to predict document's n-gram, i.e. sequences of consecutive words in the document.
We discovered that when predicting all words and all bigrams we get up to 5\% improvement over the unigram model.
Adding trigrams to the dictionary does not further improve the performance.

It is beneficial to take all words from a dictionary and all bigrams.
However, for large corpora the set of all uni- and bi-grams will be big
and, therefore, the output layer weigh matrix could be too big to fit in the memory.
Therefore, in practice the dictionary should be as big as possible, depending on available memory.
The dictionary can be shrunk by taking only a predefined number of globally most popular n-gram, or n-grams
that occur no less than a specified number of times in the corpus.
Finally, this extension is applicable only to the PV-DBOW model,
since in the case of PV-DM we explicitly predict center word (unigram) based on its context.

After conducting experiments with n-grams in binary PV-DBOW model we discovered that similar observation was previously made by
Li et al.~\cite{li2015learning}. However, they demonstrate the use of n-grams in a sentiment analysis task only and on one dataset only.
We show that inclusion of n-grams improves information retrieval results as well. One more observation by Li et al. that coincides with
ours is that, contrary to the Le and Mikolov's claims, PV-DBOW outperforms PV-DM, in spite of having simpler architecture.
This observation is also confirmed by~\cite{lau2016empirical}.

\subsection{Implementation}

The proposed models have been implemented using the TensorFlow library.
In addition to the models proposed above, we also implemented the original PV-DBOW and PV-DM models,
to generate 32-bit real-valued codes.
In the case of the PV-DBOW model, training data consists of pairs of document and n-gram identifiers.
In the case of PV-DM, a single training example consists of a document identifier, context words identifiers and a center word identifier.
Training examples are globally shuffled and split into mini-batches. It is important to shuffle examples globally, not only within documents.
When training with examples shuffled only within documents, the networks fit to the leading documents in a dataset,
which consequently leads to poor overall performance. This requirement poses a technical challenge, since training set size often exceeds
available memory. Resorting to distributed computing\footnote{Global shuffling can be easily done using Apache Hadoop
or Apache Spark distributed processing frameworks.}
is often the only solution. Shuffled training examples can be written to a file and then the network can be trained in an online fashion
with low memory requirements, by reading examples from the file in mini-batches.

At the input to the network, both word and document identifiers have to be presented using one-hot representation.
Luckily, we do not need to convert identifiers into one-hot codes.
TensorFlow provides an \emph{embedding\_lookup} function, which takes integer identifiers
and the embedding matrix as inputs and returns embeddings associated with provided identifiers.
In the case of PV-DBOW this function effectively implements the whole first layer of the network
depicted in~\figref{neural_network}.

As explained in \sectionref{tensorflow}, TensorFlow is multi-threaded by nature
and uses as many CPU or GPU cores as available.
However, sometimes computations are slowed down by I/O operations.
To speed up training, we feed mini-batches to TensorFlow session from multiple threads.
This way the overall training time is shortened approximately 6 times.
However, since we use Python API, all threads from our thread pool use just one CPU core.
This is due to the global interpreter lock, a standard Python mechanism that prevents multiple threads from running in parallel.
To overcome this limitation, we could use multiple processes to feed the data. However, in order to share a TensorFlow session
between processes, we would have to set up a distributed TensorFlow cluster.

There are at least two ways of implementing the Krizhevsky's approach to binarization. One is based on the \emph{stop\_gradient}
TensorFlow function. This function prevents gradients from being computed for all tensors that constitute an input to it.
The Krizhevsky's binarization can be implemented in the following way:
\begin{equation}  \label{eq:kriz_bin}
f( x ) = \sigma ( x ) + \stopgradient ( \roundfunction ( \sigma ( x ) ) - \sigma ( x ) ),
\end{equation}
where $ x $ are inputs to the binarized sigmoid corresponding to the current mini-batch embeddings and $ f ( x ) $ is a hidden layer
activation function. Effectively, in the forward pass $ f ( x) $ is an identity function and, therefore, \equationref{kriz_bin}
is reduced to:
\begin{equation}
f_{forward} ( x ) = \roundfunction ( \sigma ( x ) ).
\end{equation}
In the backward pass, the graph vertices that rely on \emph{stop\_gradient} are blocked and consequently gradients are computed for:
\begin{equation}
f_{backward} ( x ) = \sigma ( x ).
\end{equation}

As an alternative implementation, we can override gradient function for the round operation with identity function.
We can do this easily since TensorFlow allows to override gradient functions for selected operations.

\section{Experiments} \label{sec:dsh_results}

We tested our implementation in an information retrieval task using three popular text datasets: 20 Newsgroups,
RCV1-v2 and English Wikipedia. The datasets are described in \sectionref{datasets}.
We removed stopwords, words shorter than two characters and words longer than 15 characters.
We also experimented with stemming the corpora, but ultimately we discovered that stemming does not improve performance,
and decided not to apply it.

Our model has a lot of hyperparameters. Performing a full grid search over all the hyperparameters would be infeasible.
Therefore, we decided to make some reasonable choices based on common sense.
We decided to use AdaGrad~\cite{duchi2011adaptive} optimization method,
since it performed better than other methods we used in preliminary experiments
(in addition to AdaGrad we experimented with stochastic gradient descent with momentum,
Adam~\cite{kingma2014adam}, Adadelta~\cite{zeiler2012adadelta} and FTRL~\cite{mcmahan2013ad} optimizers).
Mini batch size was set to 128. As explained in \sectionref{word_embeddings}, in the Paragraph Vector model, as well as in word2vec,
gradient of the softmax function need to be approximated.
To approximate this gradient we used sampled softmax method~\cite{cho2015using}
with 64 classes (words or n-grams from the dictionary) sampled for each mini-batch.
Embeddings were initialized with random numbers drawn from an uniform distribution
in the range $ [-\frac{0.5}{d}, \frac{0.5}{d}], $ where $ d $ is an embedding size.
Prediction layer weights and biases were initialized with zeros.

In the case of the PV-DM model we decided to concatenate a document embedding with context words embeddings, instead of summing or averaging.
This way order of words within a context is not lost and also the dimensionality of the document embeddings is not tied to
the dimensionality of the word embeddings. We experimented with different windows sizes.
To our surprise, the best result was obtained for a minimal one-word one-sided window.
To verify correctness of our implementation,
we trained the PV-DM model implementation available in a popular \emph{gensim}~\cite{rehurek2010software} library
on both the 20 Newsgroups and RCV1-v2 datasets for different window sizes.
As in the case of our implementation, the best results were obtained for a context window of size just 1.
One explanation for this observation could be that 20 Newsgroups and RCV1 are too small datasets to learn high-quality word embeddings.
If word embeddings, which are learnt alongside document embeddings, are of low quality,
then instead of helping to predict the center word they could make the prediction harder.

To select remaining hyperparameters we created a validation subset by randomly extracting 25\% documents from the training set.
To improve quality of the embeddings we applied dropout~\cite{srivastava2014dropout} during training.
The probability of keeping the activations was selected on the validation sets separately for the PV-DBOW,
Binary PV-DBOW, PV-DM and Binary PV-DM models and separately for each dataset.
Two others hyperparameters that needs to be selected based on the validation sets are the epoch number and the learning rate.
For simplicity, we decided to use the same learning rate in the training and the inference phase.

All the experiment were conducted on the HP Apollo XL750f Gen9 liquid cooled HPC machines
equipped with two Intel Xeon E5-2680v3 processors, 128 GB RAM and two Nvidia Tesla K40 GPUs.
We used Python version 2.7.5 and TensorFlow version 0.11. As explained in \sectionref{tensorflow},
TensorFlow is able to run the experiment on either CPU or GPU, and, when GPU is available, it is advised to run on it.
However, at the time of conducting the experiments not all the TensorFlow kernels used in our model (e.g. AdaGrad optimizer)
had GPU implementations.
Therefore, some parts of the computation graph were evaluated on the GPU and some parts on the CPU.

\subsection{Information retrieval metrics} \label{sec:dsh_ir}

Probably the most popular metrics used to evaluate results of information retrieval (IR) systems are precision and recall.
The precision tells us the fraction of relevant results among all returned results.
The recall tells us the fraction of returned results among all relevant results. IR system rarely returns a single result.
More often a list of ranked results, i.e. sorted by relevance, is returned.
In practice, for ranking we often use cosine similarity in case of documents represented by real-valued vectors and the Hamming distance
in case of documents represented by binary vectors.
When we ask for a very limited number of results, e.g. 10, then most likely most of the results are relevant
and, therefore, an average precision among them is high.
When we increase the number of requested results, the precision is expected to decline but the recall will grow.
If we query the system for all stored documents, then the recall will be 1 but the precision most likely be very low.

The datasets we use do not come with predefined sets of test queries.
Therefore, in order to evaluate the performance of the model we sample a document from a test set
and treat it as a query to retrieve relevant documents.
We repeat this operation many times and average the results.
To speed-up the computations we do this in mini-batches. First, we select some number of query documents.
Then we compute similarity between those query documents and remaining test documents.
We can do this in a batched way regardless whether we are using cosine similarity or the Hamming distance.
Finally, we calculate relevancy by comparing with actual topic assignments. This comparison also is batched.
To this end, we store labels as one-hot vectors (or a-few-hot vectors in the case of multi-label datasets)
and we perform matrix multiplication between query documents labels and all test documents labels.

Precision and recall are defined at specified cut-off level of ranked result list.
To measure the performance of IR system precisions are often computed for all possible recall values.
We can then plot a precision-recall curve. Systems having bigger area under the curve (AUC) perform better than those having lower area.
One of the approximation to AUC is mean average precision (MAP).
In order to compute average precision, we sum precisions at different recall levels
and then divide it by the total number of relevant documents for a given query.
MAP is an average precision averaged over all queries.

One of the limitations of precision and recall metrics is that they work only for a binary relevance measure\footnote{In practice,
we can relax the definitions of precision and recall to deal with multi-level relevancy, but it is not a standard precision-recall definition}.
Sometimes, instead of telling whether the result is relevant to the query or not, it is better to specify the degree of relevancy.
Some documents can be partially relevant to the query.
The most popular IR metric used to evaluate multi-level relevance systems
is normalized discounted cumulative gain at a $k$-th position ($ NDCG@k $)~\cite{jarvelin2002cumulated}.
The gain is just a different name for how a given document is relevant to the query.
Cumulative gain at a specific position on the result list, e.g. on the $ 10 $-th position,
is a sum of all the gains from the top to this position on the list.
Discounted cumulative gain is a cumulative gain where gains further on the list are penalized in the following way:
\begin{equation}
DCG @ k = r_1 + \sum_{i=2}^k \frac{r_i}{\log_2 i},
\end{equation}
where $ r_i $ is a relevance of a document at the $ i $-th position on the result list to the query.
Finally, normalized $ DCG@k $ ($ NDCG@k $) is a $ DCG $ at a position $ k $ divided by the `ideal' $ DCG $ at this position,
i.e. $ DCG $ for the case where most relevant documents are placed on the top of the result list.

\subsection{20 Newsgroups}

Since 20 Newsgroups is a relatively small dataset, we decided to predict all words (and, in the case of PV-DBOW, all bigrams) during training.
The total number of classes to be predicted for PV-DBOW (unique words and bigrams) was slightly over $ 10^6 $.
Using the validation set we selected following hyperparameters: 10 training epochs, 50\% dropout,
learning rate 1.3 for PV-DBOW, 0.3 for Binary PV-DBOW,
1.2 for PV-DM and 0.2 for Binary PV-DM.
In the case of the 20 Newsgroups dataset a relevancy measure is straightforward.
If a retrieved document has the same label as a query document, then it is relevant. Otherwise it it not.

Both Paragraph Vector and Binary Paragraph Vector results for different code sizes are reported in~\tabref{dsh_tng_results}.
\begin{table}[htb]
  \centering
    \begin{tabular}{|c|c|c|c|c|}
      \hline
      Code                 & \multirow{2}{*}{Model}          & Include              & \multirow{2}{*}{MAP} & \multirow{2}{*}{NDCG@10} \\
      dimensionality       &                                 & bigrams              &                      &                          \\ \hline
      \multirow{6}{*}{300} & \multirow{2}{*}{PV-DBOW}        & no                   & 0.37                 & 0.75                     \\ \cline{3-5}
                           &                                 & yes                  & 0.43                 & 0.76                     \\ \cline{2-5}
                           & \multirow{2}{*}{Binary PV-DBOW} & no                   & 0.28                 & 0.68                     \\ \cline{3-5}
                           &                                 & yes                  & \textbf{0.32}        & \textbf{0.71}            \\ \cline{2-5}
                           & PV-DM                           & \multirow{2}{*}{N/A} & 0.38                 & 0.73                     \\ \cline{2-2} \cline{4-5}
                           & Binary PV-DM                    &                      & 0.29                 & 0.63                     \\ \hline
      \multirow{6}{*}{128} & \multirow{2}{*}{PV-DBOW}        & no                   & 0.4                  & 0.75                     \\ \cline{3-5}
                           &                                 & yes                  & 0.45                 & 0.75                     \\ \cline{2-5}
                           & \multirow{2}{*}{Binary PV-DBOW} & no                   & 0.34                 & 0.69                     \\ \cline{3-5}
                           &                                 & yes                  & \textbf{0.35}        & \textbf{0.69}            \\ \cline{2-5}
                           & PV-DM                           & \multirow{2}{*}{N/A} & 0.41                 & 0.73                     \\ \cline{2-2} \cline{4-5}
                           & Binary PV-DM                    &                      & 0.34                 & 0.65                     \\ \hline
      \multirow{6}{*}{64}  & \multirow{2}{*}{PV-DBOW}        & no                   & 0.42                 & 0.74                     \\ \cline{3-5}
                           &                                 & yes                  & 0.46                 & 0.74                     \\ \cline{2-5}
                           & \multirow{2}{*}{Binary PV-DBOW} & no                   & 0.36                 & 0.65                     \\ \cline{3-5}
                           &                                 & yes                  & \textbf{0.36}        & \textbf{0.65}            \\ \cline{2-5}
                           & PV-DM                           & \multirow{2}{*}{N/A} & 0.43                 & 0.72                     \\ \cline{2-2} \cline{4-5}
                           & Binary PV-DM                    &                      & 0.33                 & 0.6                      \\ \hline
      \multirow{6}{*}{32}  & \multirow{2}{*}{PV-DBOW}        & no                   & 0.43                 & 0.71                     \\ \cline{3-5}
                           &                                 & yes                  & 0.46                 & 0.72                     \\ \cline{2-5}
                           & \multirow{2}{*}{Binary PV-DBOW} & no                   & 0.32                 & 0.53                     \\ \cline{3-5}
                           &                                 & yes                  & \textbf{0.32}        & \textbf{0.54}            \\ \cline{2-5}
                           & PV-DM                           & \multirow{2}{*}{N/A} & 0.43                 & 0.7                      \\ \cline{2-2} \cline{4-5}
                           & Binary PV-DM                    &                      & 0.29                 & 0.49                     \\ \hline
    \end{tabular}
  \longcaption{Information retrieval 20 Newsgroups results.}{The best binary results for each code dimensionality are highlighted.}
  \label{tab:dsh_tng_results}
\end{table}
The results were generated in the following way.
We took 200 recall values evenly distributed between 0 and 1.
For each of those recall values we took the first document from the dataset and treated it as a query.
All remaining documents were sorted according to relevance.
The list was cut off when the given recall level was reached.
For small recall values, cut-off level was close to the beginning of the list.
For higher recall values, more documents had to be taken to reach the expected recall.
Afterwards we evaluated the precision among the documents remaining in the list.
The whole procedure was repeated for all test documents in the dataset, i.e. each document in turn was treated as a query.
Obtained precision values were averaged. 

\begin{figure}[htb!]
  \centering
  \begin{subfigure}[b]{0.495\linewidth}
    \centering
    \includegraphics[width=\textwidth]{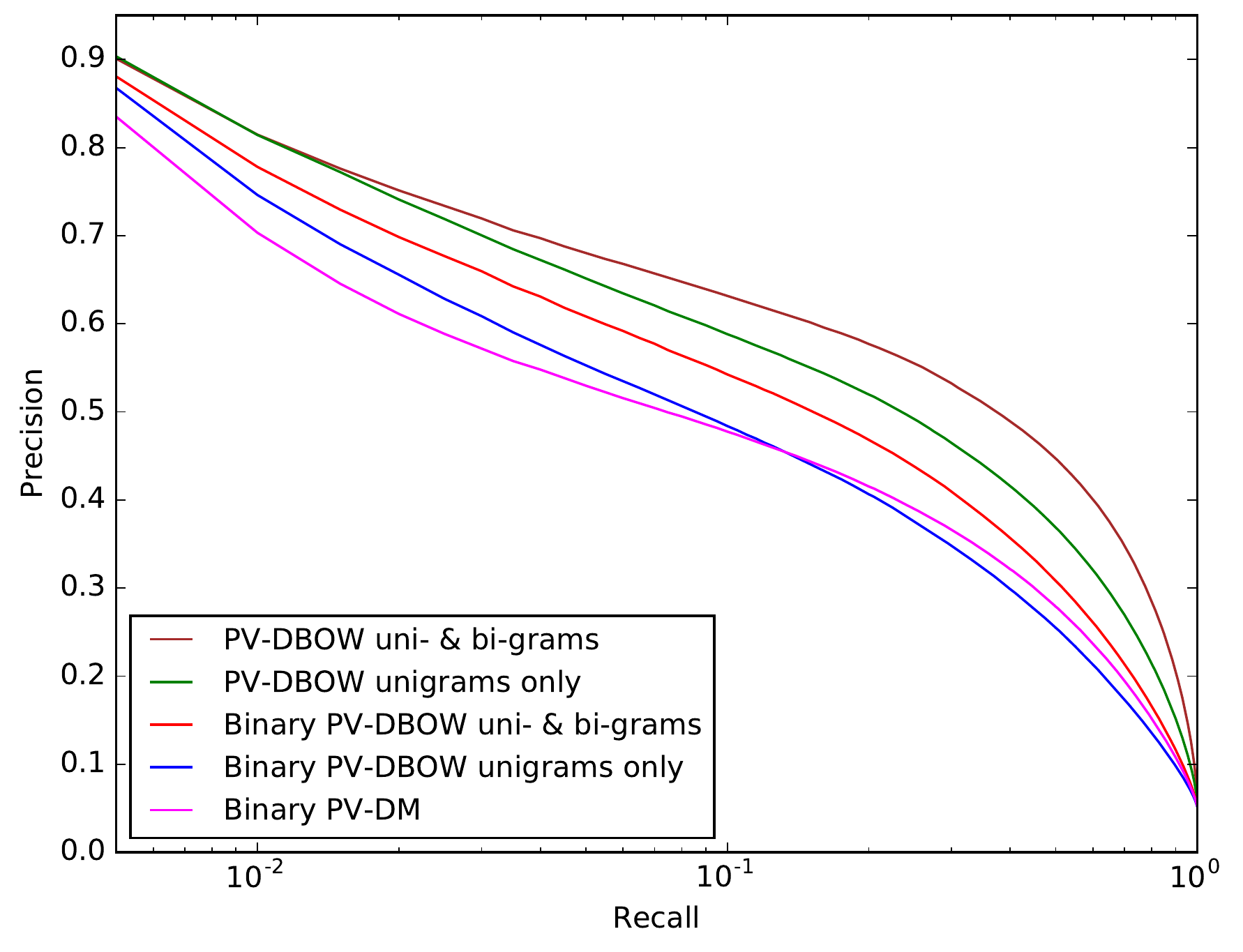}
    \caption{300 dimensions}
  \end{subfigure}
  \hfill
  \begin{subfigure}[b]{0.495\linewidth}
    \centering
    \includegraphics[width=\textwidth]{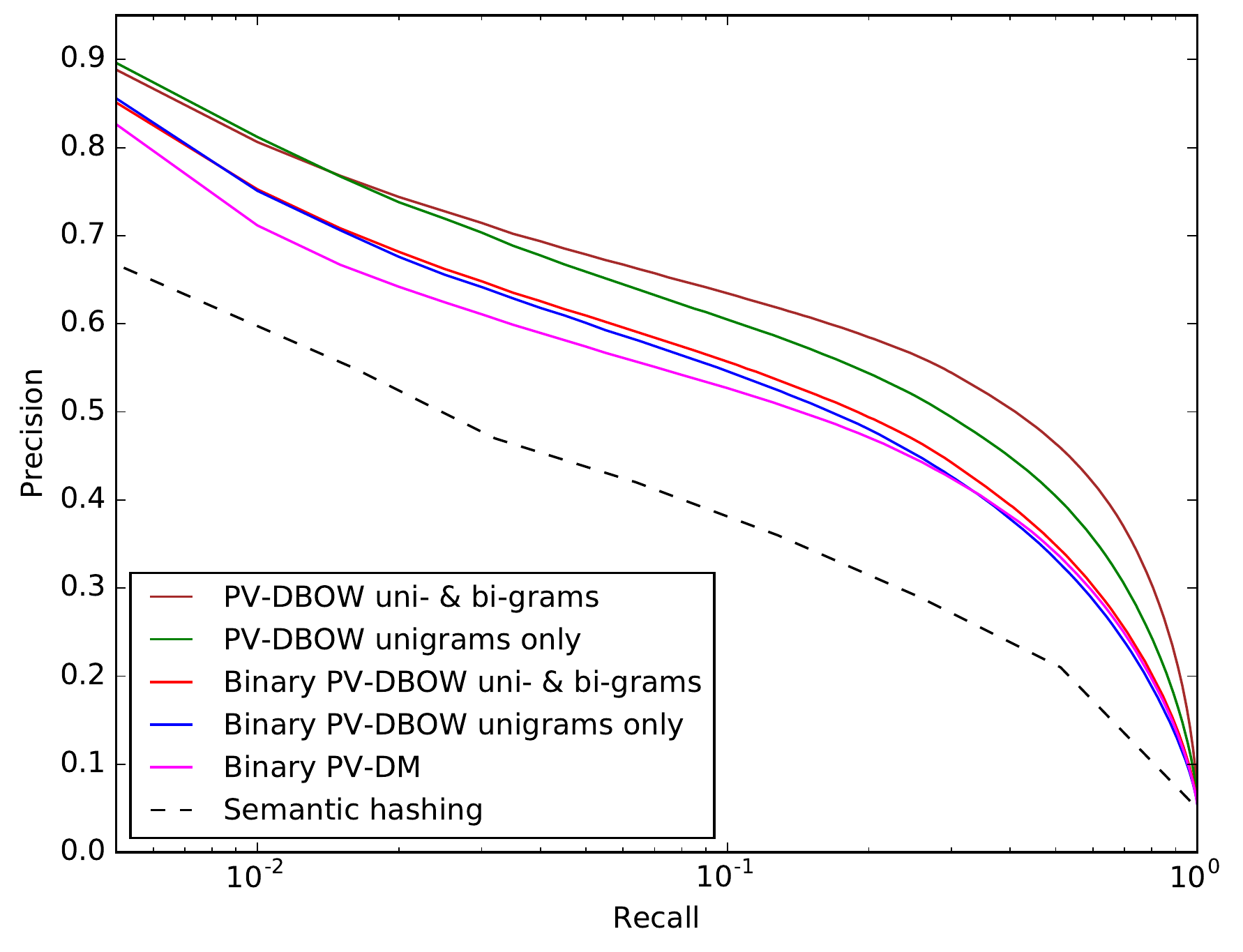}
    \caption{128 dimensions}
  \end{subfigure}
  \hfill
  \begin{subfigure}[b]{0.495\linewidth}
    \centering
    \includegraphics[width=\textwidth]{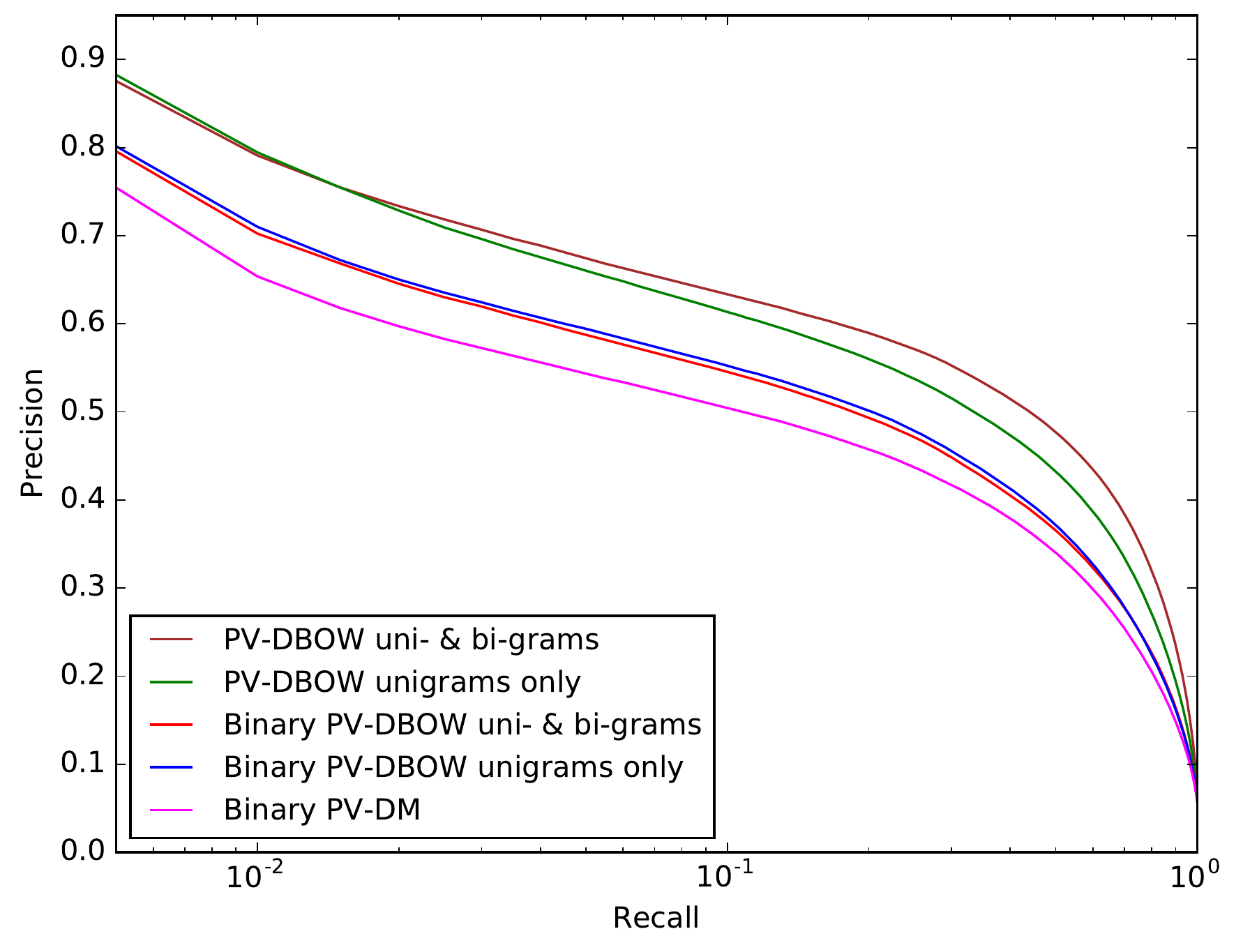}
    \caption{64 dimensions}
  \end{subfigure}
  \hfill
  \begin{subfigure}[b]{0.495\linewidth}
    \centering
    \includegraphics[width=\textwidth]{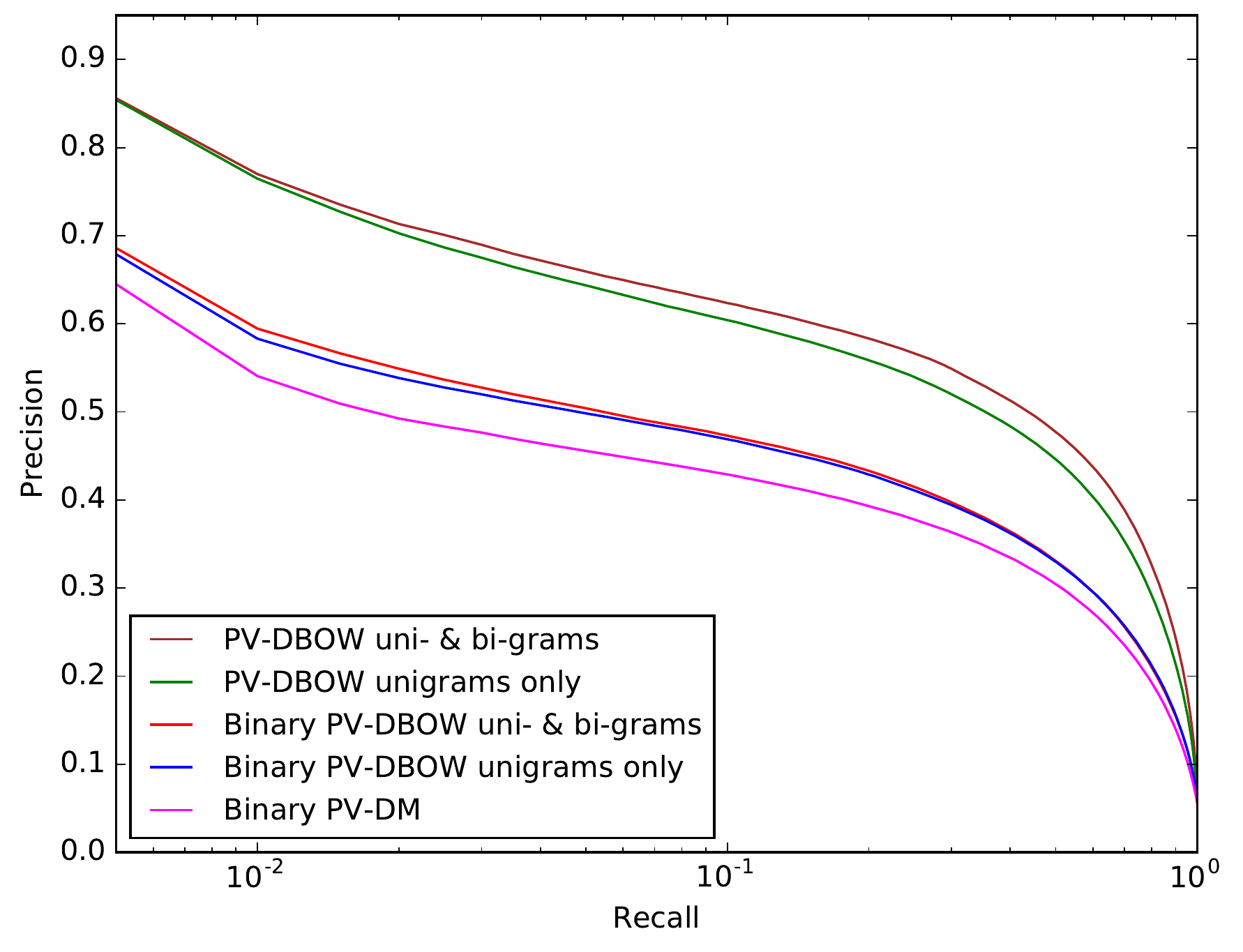}
    \caption{32 dimensions}
  \end{subfigure}
  \longcaption{The 20 Newsgroups dataset precision-recall curves for different code dimensionalities and different model variants.}
              {For real-valued codes cosine distance was used as a similarity measure.
               For binary codes the Hamming distance was used as a similarity measure.
               For comparison, on plot (b) we also report results from~\cite[Fig. 6]{salakhutdinov2009semantic}.}
  \label{fig:dsh_tng_precision_recall}
\end{figure}

To shed more light on the information retrieval results we also drew the precision-recall curves~\figref{dsh_tng_precision_recall}.
The curves are generated by connecting averaged precision values for 200 different recall values evenly distributed between 0 and 1.
As expected, binary codes give worse results than real-valued ones, but the difference is very small.
Taking into account that binary codes use 32 times less memory than floating point ones,
the closeness of the results seem to be impressive. The results are also much better than the seminal semantic hashing results.
For example, 128-dimensional binary codes results are almost 20\% higher than those reported in~\cite{salakhutdinov2009semantic}.

The precision-recall curves reveal that PV-DBOW gives slightly better results than PV-DM for both real-valued and binary codes.
Since this behavior is contrary to the results reported in~\cite{le2014distributed},
we verified experiments for floating point codes using the gensim library.
Surprisingly, we identified a similar trend.
As discussed earlier, this behavior can be caused by relatively small dataset size, not sufficient for PV-DM to perform well.

\begin{figure}[htb!]
  \centering
  \begin{subfigure}[b]{0.495\linewidth}
    \centering
    \includegraphics[width=\textwidth]{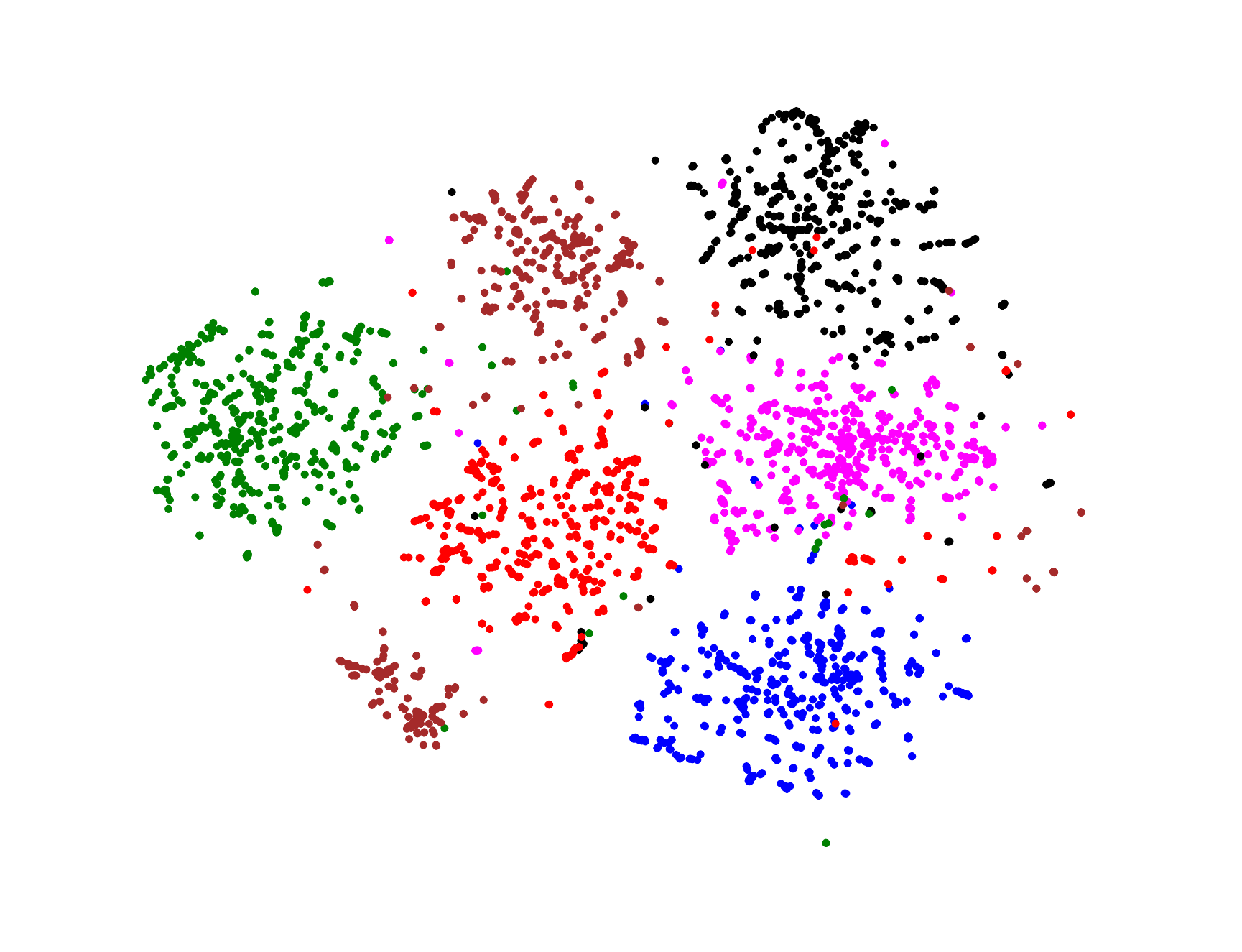}
    \caption{300 dimensions}
  \end{subfigure}
  \hfill
  \begin{subfigure}[b]{0.495\linewidth}
    \centering
    \includegraphics[width=\textwidth]{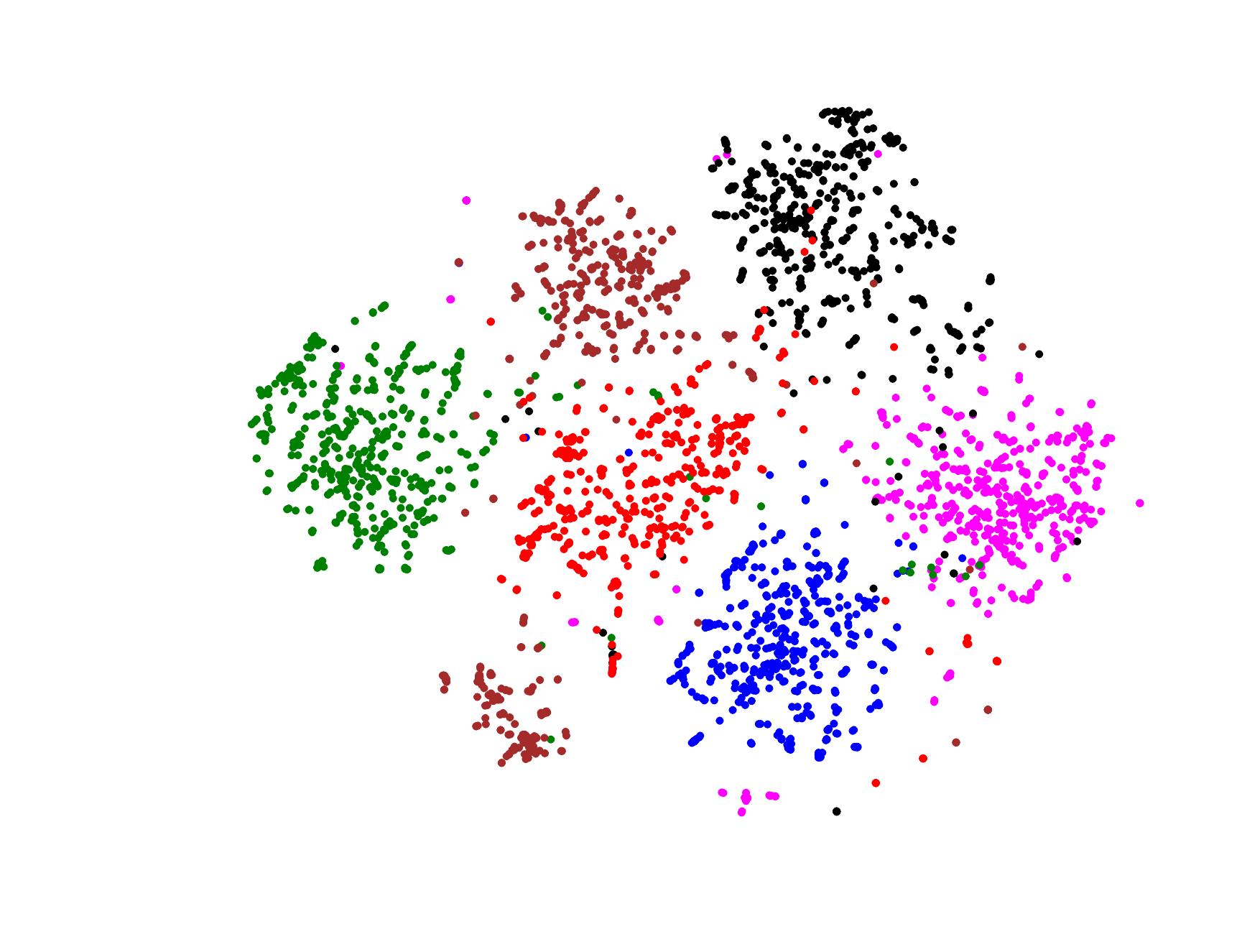}
    \caption{128 dimensions}
  \end{subfigure}
  \hfill
  \begin{subfigure}[b]{0.495\linewidth}
    \centering
    \includegraphics[width=\textwidth]{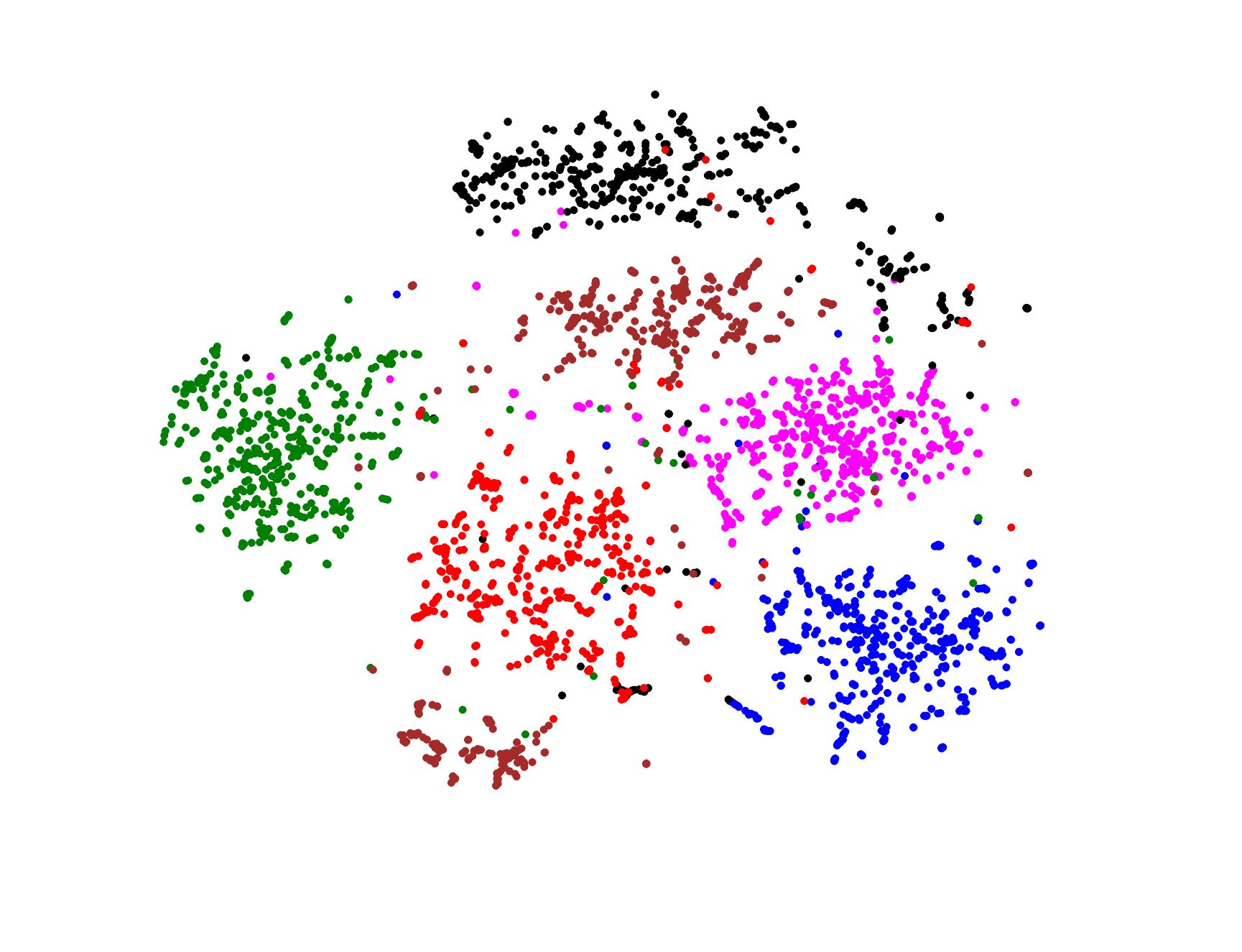}
    \caption{64 dimensions}
  \end{subfigure}
  \hfill
  \begin{subfigure}[b]{0.495\linewidth}
    \centering
    \includegraphics[width=\textwidth]{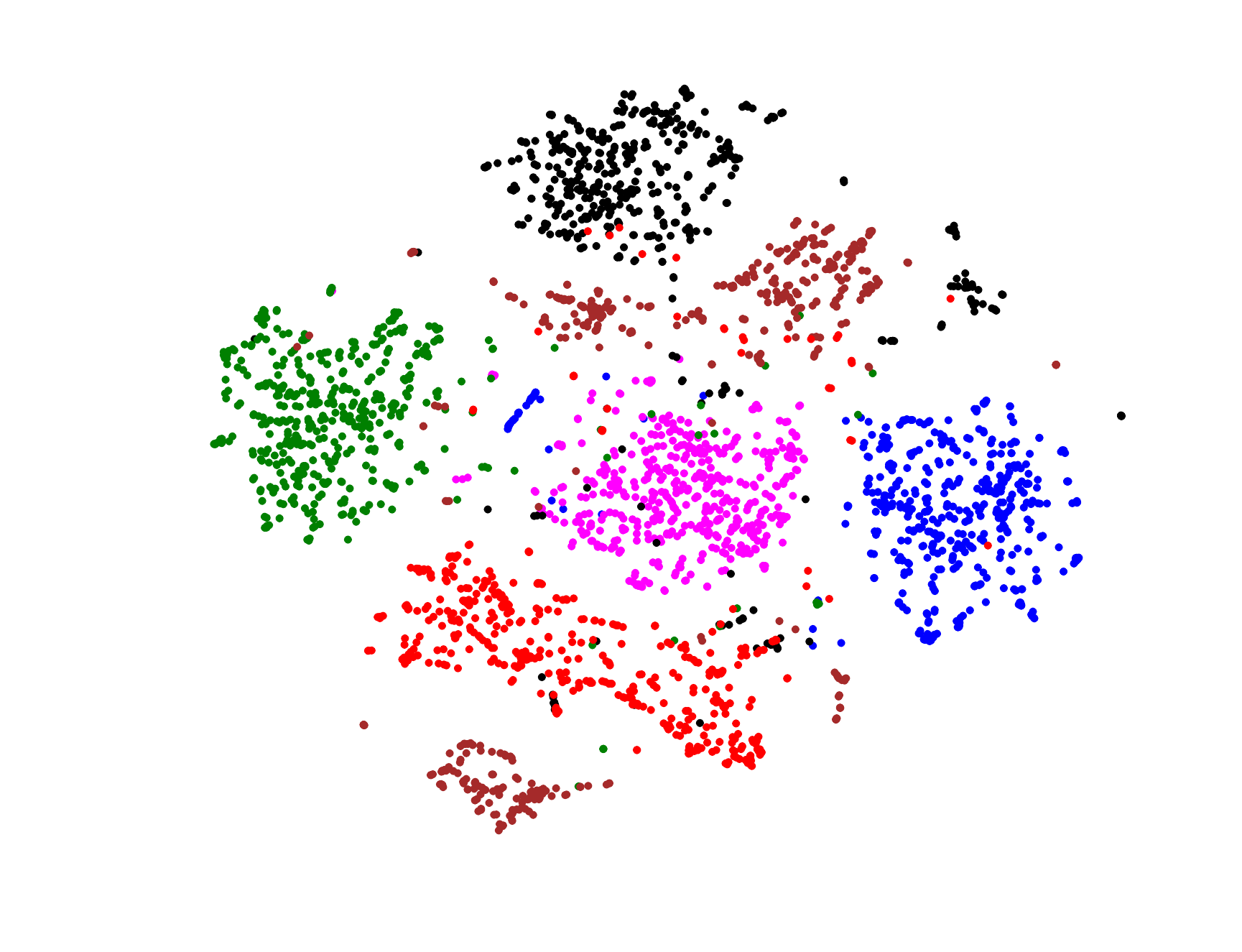}
    \caption{32 dimensions}
  \end{subfigure}
  \longcaption{t-SNE visualizations of real-valued point codes of seven selected newsgroups
               from the 20 Newsgroups dataset for different code dimensionalities.}
              {Cosine distance was used as a similarity measure.
               Selected groups: green - soc.religion.christian, red - talk.politics.guns, blue - rec.sport.hockey,
               brown - talk.politics.mideast, magenta - comp.graphics, black - sci.crypt.}
  \label{fig:tng_fp_visualise}
\end{figure}

\begin{figure}[htb!]
  \centering
  \begin{subfigure}[b]{0.495\linewidth}
    \centering
    \includegraphics[width=\textwidth]{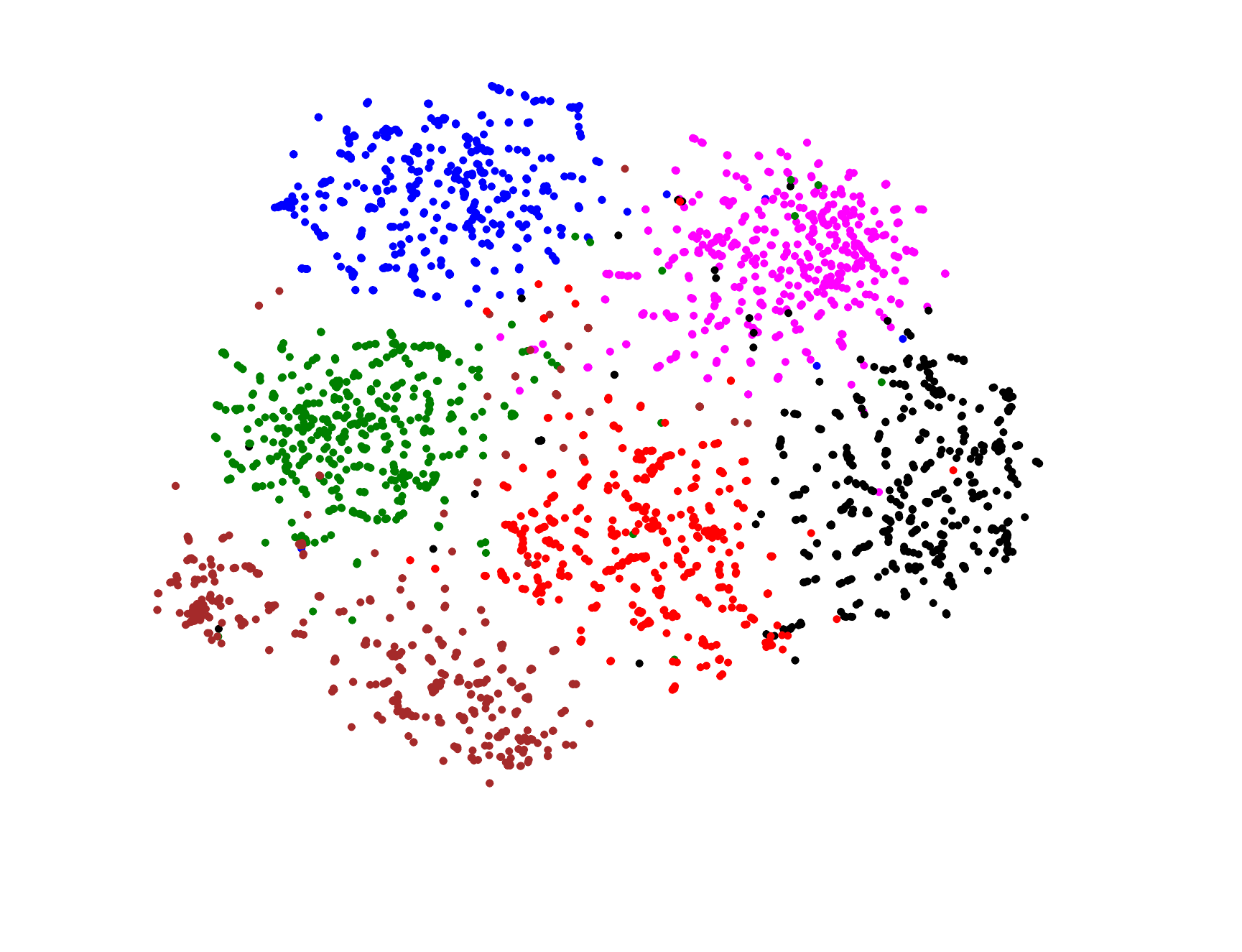}
    \caption{300 dimensions}
  \end{subfigure}
  \hfill
  \begin{subfigure}[b]{0.495\linewidth}
    \centering
    \includegraphics[width=\textwidth]{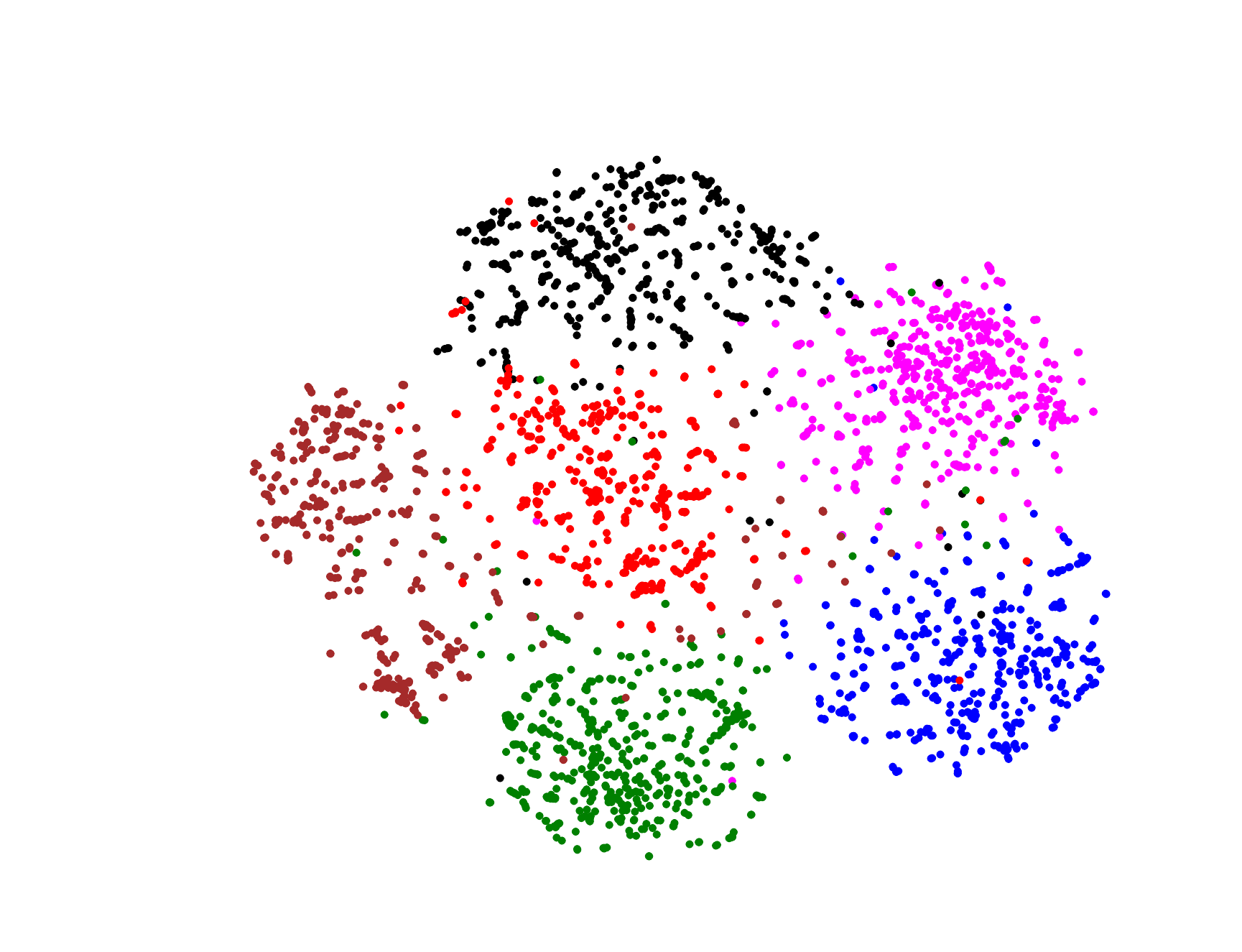}
    \caption{128 dimensions}
  \end{subfigure}
  \hfill
  \begin{subfigure}[b]{0.495\linewidth}
    \centering
    \includegraphics[width=\textwidth]{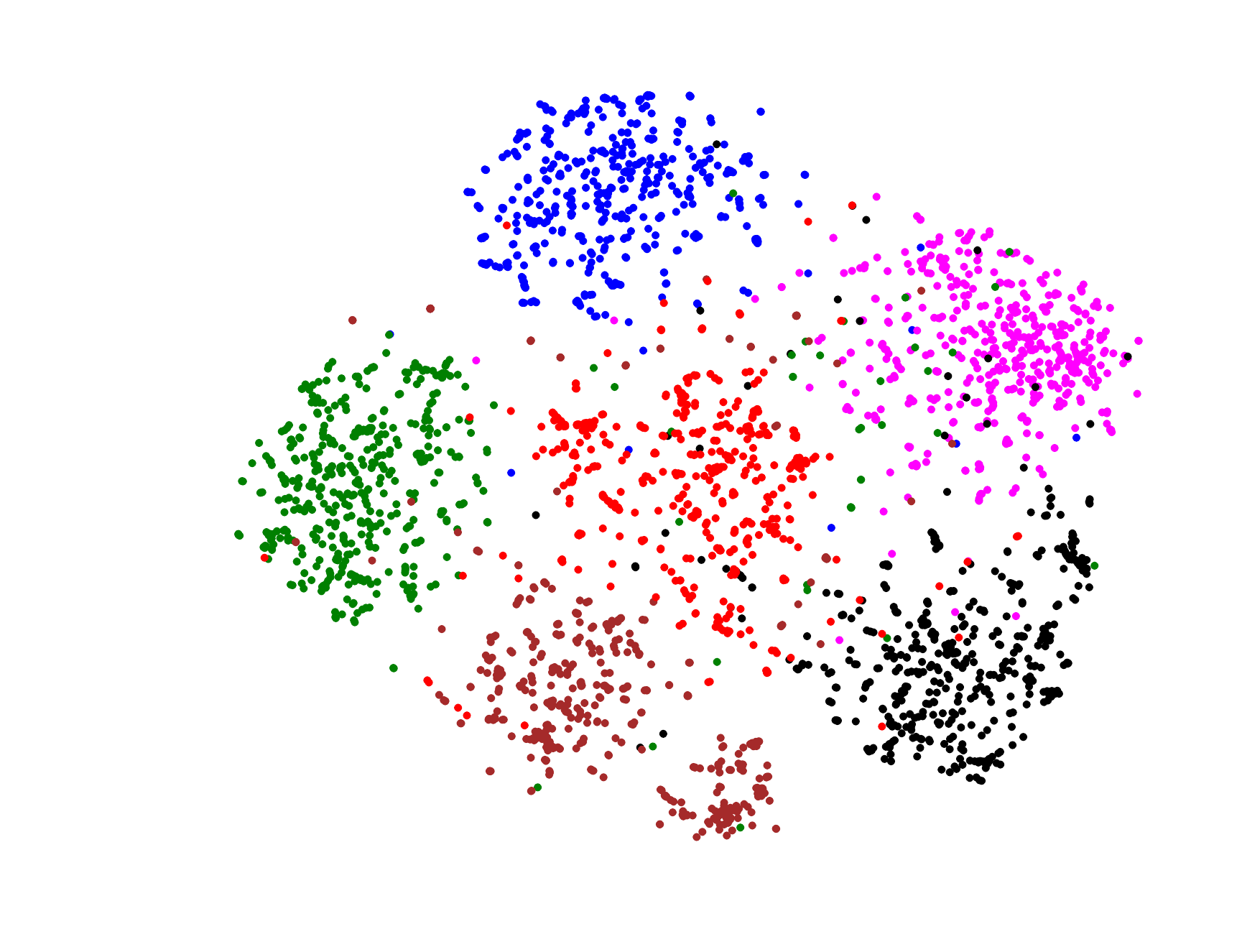}
    \caption{64 dimensions}
  \end{subfigure}
  \hfill
  \begin{subfigure}[b]{0.495\linewidth}
    \centering
    \includegraphics[width=\textwidth]{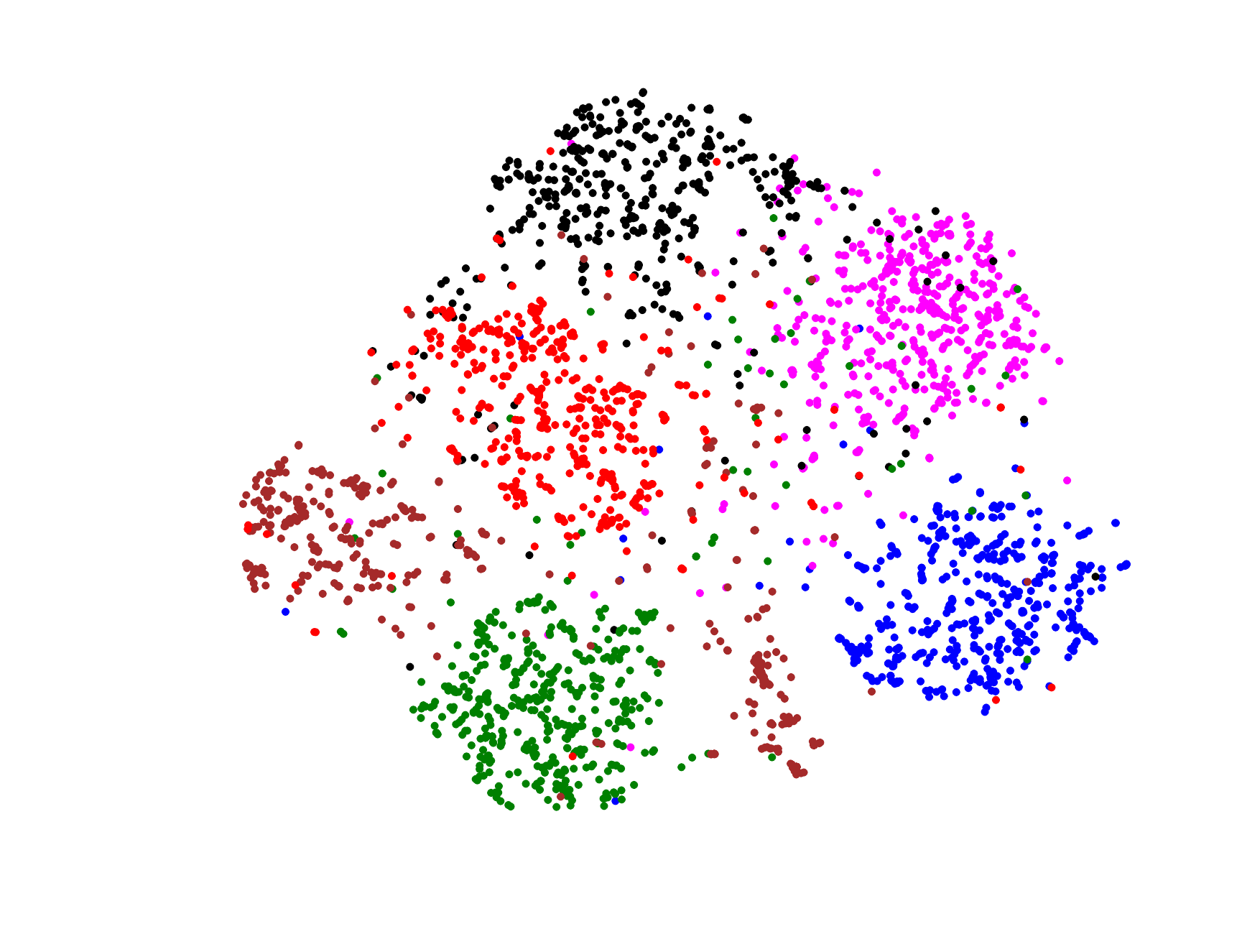}
    \caption{32 dimensions}
  \end{subfigure}
  \longcaption{t-SNE visualizations of binary codes of seven selected newsgroups
               from the 20 Newsgroups dataset for different code dimensionalities.}
              {The Hamming distance was used as a similarity measure.
               The color-group mappings are the same as in~\figref{tng_fp_visualise}.}
  \label{fig:tng_bin_visualise}
\end{figure}

In addition to the information retrieval experiment we also generated two-dimensional visualizations of coded for documents from selected
newsgroups. To reduce dimensionality from that of our model (respectively 32, 64, 128 and 300 dimensions) to two
we used the stat-of-the-art t-distributed Stochastic Neighbor Embedding~(t-SNE)~\cite{maaten2008visualizing} method.
The visualizations of real-valued codes are presented in~\figref{tng_fp_visualise}
and of binary codes in~\figref{tng_bin_visualise}.
The selection of newsgroups and colors are the same as in~\cite[Fig. 5]{salakhutdinov2009semantic}.
As we can see, binary codes provide almost as good visual separation of topics as real-valued ones.
Moreover, the separation appears to be better than the one presented in~\cite{salakhutdinov2009semantic}.

\subsection{RCV1}

Since the RCV1-v2 corpus is approximately 35 times bigger than 20 Newsgroups its set of unique uni- and bigrams is much bigger as well.
Specifically, it has approximately $ 10^7 $ elements. Using such a big dictionary would require huge softmax weight matrix.
Therefore, we shrank the dictionary by taking uni- and bigrams occurring at least 10 times in the training set.
Resultant dictionary has approximately $ 8 \times 10^5 $ elements.
Following~\cite{salakhutdinov2009semantic}, to form a test set we split the dataset into two halves.
Experiments on the validation set revealed that the optimal epoch number for RCV1 is just one,
learning rate 1.1 for PV-DBOW, 0.4 for Binary PV-DBOW, 1.1 for PV-DM and 0.5 for Binary PV-DM. Validation experiments also established
50\% dropout rate for PV-DBOW, 30\% rate for Binary PV-DBOW and 10\% dropout rate for PV-DM and Binary PV-DM.

In the case of the RCV1 dataset, deciding whether the document is relevant to the query or not
is not as straightforward as in the case of the 20 Newsgroups dataset.
As explained in \sectionref{datasets_rcv1}, each document is not assigned to a single topic
but is described by a hierarchy of topics.
Therefore, we cannot apply a binary relevance measure.
Instead, we calculate the relevance as a fraction of overlapping labels in a retrieved document and a query document.
The same strategy was adopted by Salakhutdinov and Hinton.

\begin{figure}[htb!]
  \centering
  \begin{subfigure}[b]{0.495\linewidth}
    \centering
    \includegraphics[width=\textwidth]{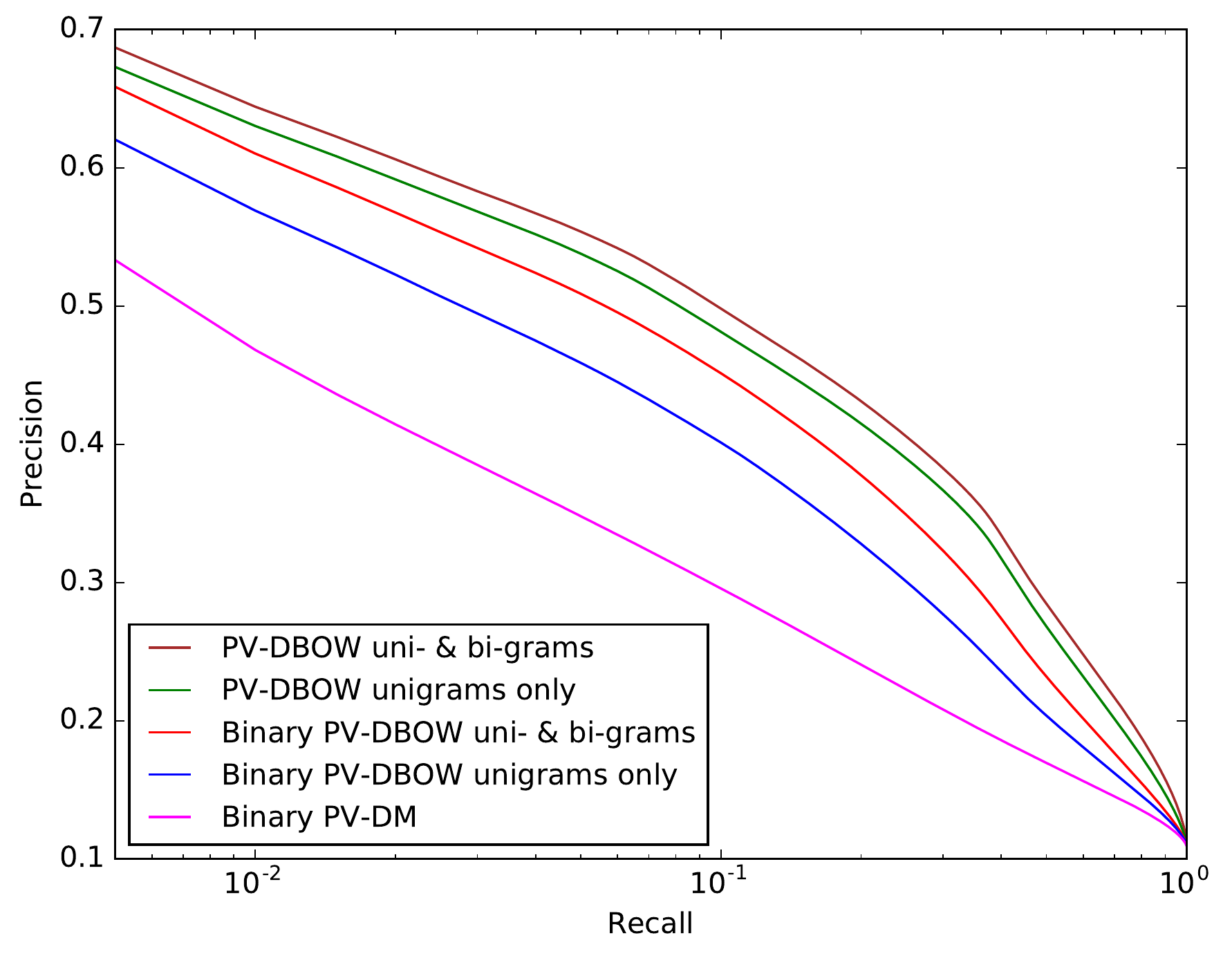}
    \caption{300 dimensions}
  \end{subfigure}
  \hfill
  \begin{subfigure}[b]{0.495\linewidth}
    \centering
    \includegraphics[width=\textwidth]{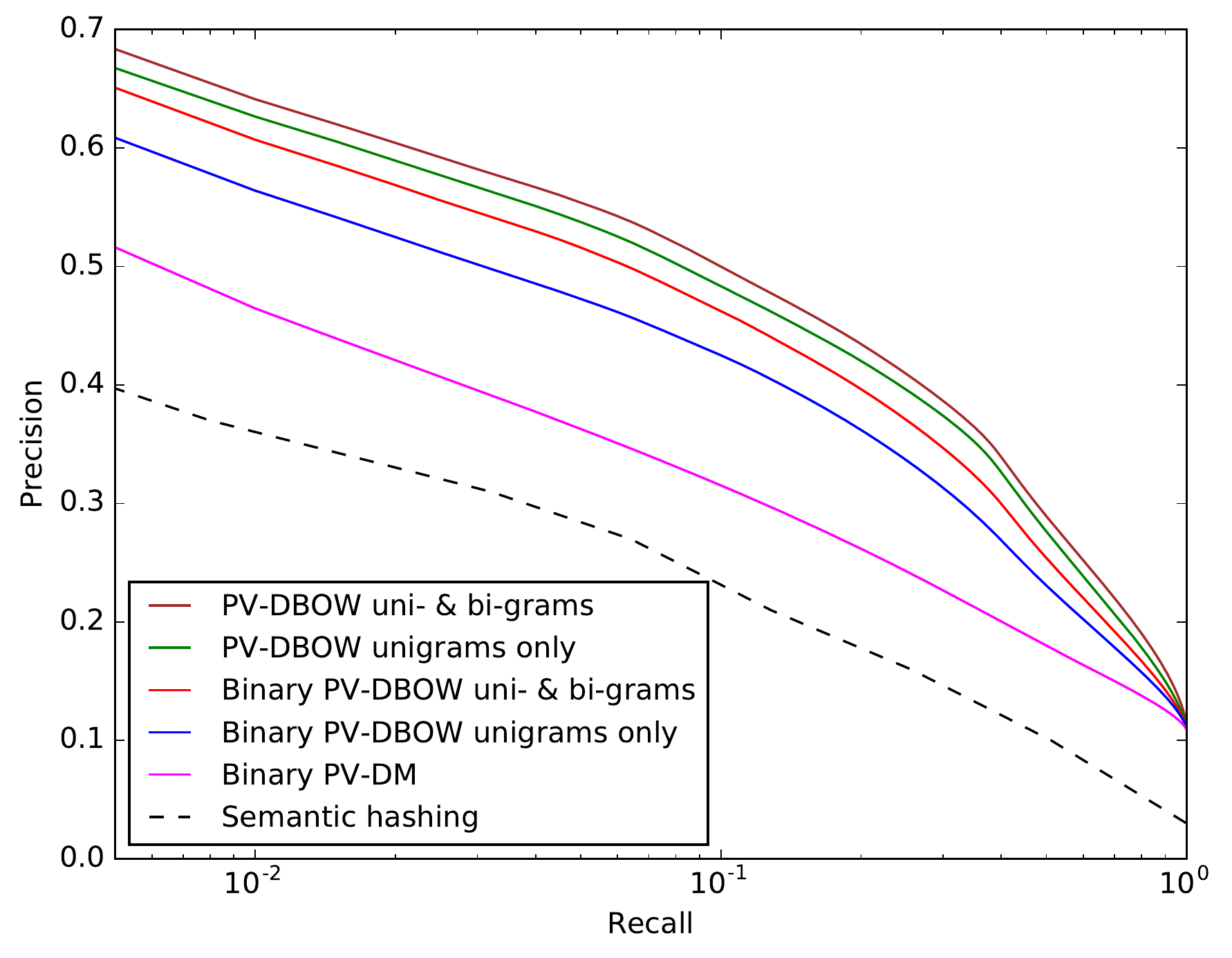}
    \caption{128 dimensions}
  \end{subfigure}
  \hfill
  \begin{subfigure}[b]{0.495\linewidth}
    \centering
    \includegraphics[width=\textwidth]{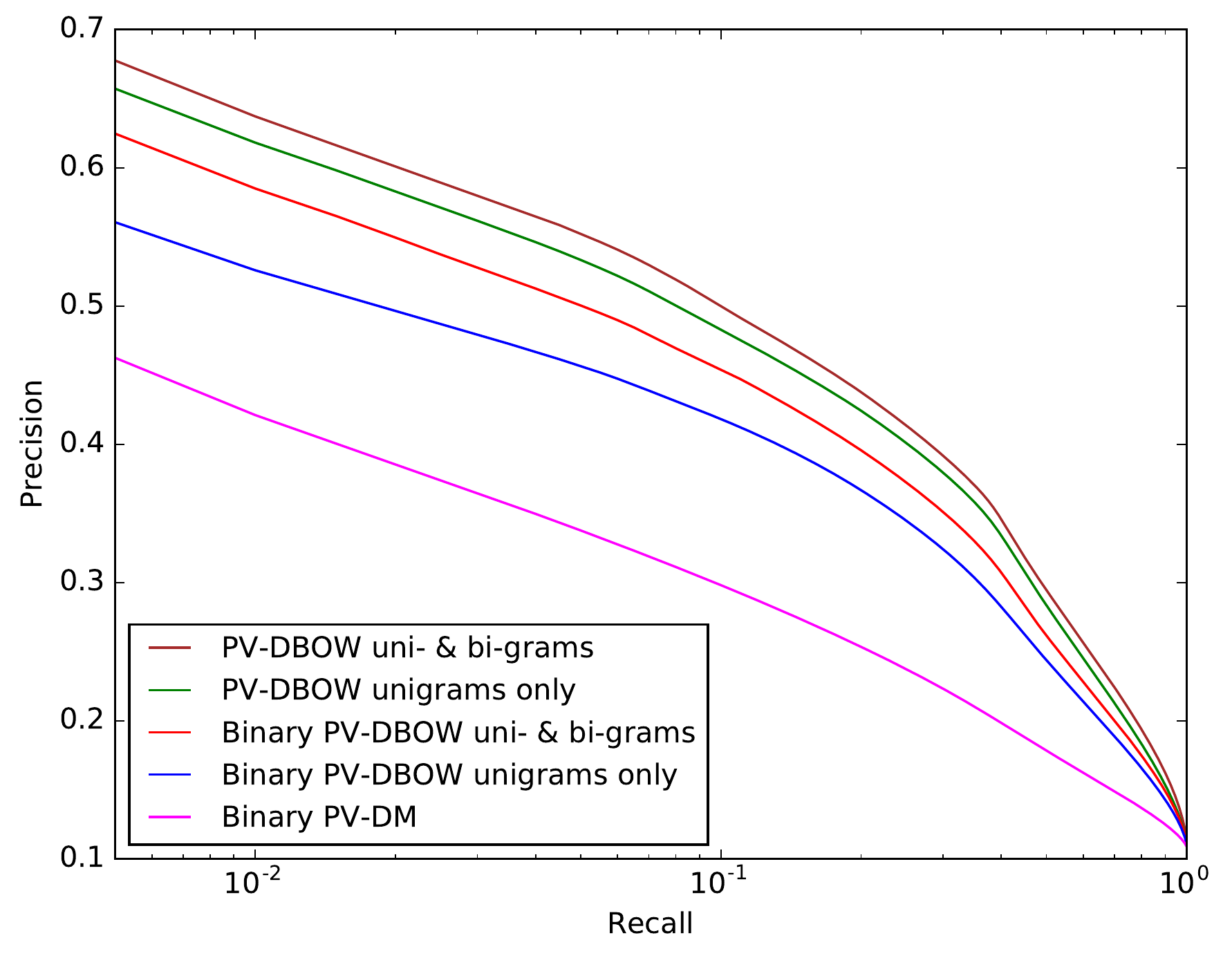}
    \caption{64 dimensions}
  \end{subfigure}
  \hfill
  \begin{subfigure}[b]{0.495\linewidth}
    \centering
    \includegraphics[width=\textwidth]{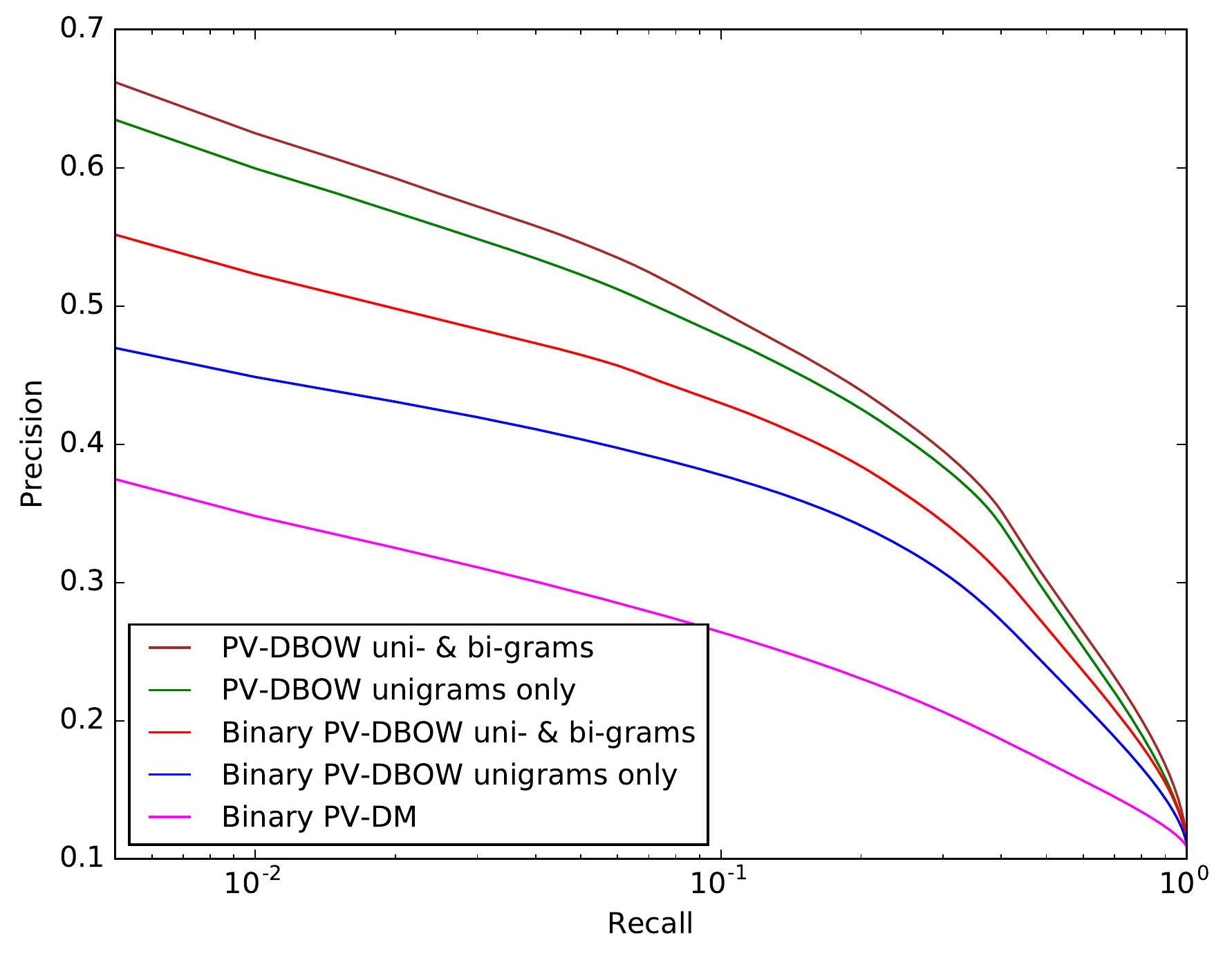}
    \caption{32 dimensions}
  \end{subfigure}
  \longcaption{The precision-recall curves for the RCV1 dataset for different code dimensionalities and model variants.}
              {For real-valued codes cosine distance was used as a similarity measure.
               For binary codes the Hamming distance was used as a similarity measure.
               For comparison, in plot (b) we also report results from~\cite[Fig. 7]{salakhutdinov2009semantic}.}
  \label{fig:dsh_rcv1_precision_recall}
\end{figure}

Results for both Paragraph Vector and Binary Paragraph Vector for different code sizes are reported in~\tabref{dsh_rcv1_results}.
\begin{table}[htb]
  \centering
    \begin{tabular}{|c|c|c|c|c|}
      \hline
      Code                 & \multirow{2}{*}{Model}          & Include              & \multirow{2}{*}{MAP} & \multirow{2}{*}{NDCG@10} \\
      dimensionality       &                                 & bigrams              &                      &                          \\ \hline
      \multirow{6}{*}{300} & \multirow{2}{*}{PV-DBOW}        & no                   & 0.25                 & 0.79                     \\ \cline{3-5}
                           &                                 & yes                  & 0.26                 & 0.8                      \\ \cline{2-5}
                           & \multirow{2}{*}{Binary PV-DBOW} & no                   & 0.21                 & 0.76                     \\ \cline{3-5}
                           &                                 & yes                  & \textbf{0.23}        & \textbf{0.78}            \\ \cline{2-5}
                           & PV-DM                           & \multirow{2}{*}{N/A} & 0.22                 & 0.78                     \\ \cline{2-2} \cline{4-5}
                           & Binary PV-DM                    &                      & 0.18                 & 0.72                     \\ \hline
      \multirow{6}{*}{128} & \multirow{2}{*}{PV-DBOW}        & no                   & 0.25                 & 0.79                     \\ \cline{3-5}
                           &                                 & yes                  & 0.27                 & 0.8                      \\ \cline{2-5}
                           & \multirow{2}{*}{Binary PV-DBOW} & no                   & 0.22                 & 0.74                     \\ \cline{3-5}
                           &                                 & yes                  & \textbf{0.24}        & \textbf{0.77}            \\ \cline{2-5}
                           & PV-DM                           & \multirow{2}{*}{N/A} & 0.23                 & 0.78                     \\ \cline{2-2} \cline{4-5}
                           & Binary PV-DM                    &                      & 0.18                 & 0.69                     \\ \hline
      \multirow{6}{*}{64}  & \multirow{2}{*}{PV-DBOW}        & no                   & 0.26                 & 0.77                     \\ \cline{3-5}
                           &                                 & yes                  & 0.27                 & 0.79                     \\ \cline{2-5}
                           & \multirow{2}{*}{Binary PV-DBOW} & no                   & 0.23                 & 0.69                     \\ \cline{3-5}
                           &                                 & yes                  & \textbf{0.25}        & \textbf{0.74}            \\ \cline{2-5}
                           & PV-DM                           & \multirow{2}{*}{N/A} & 0.23                 & 0.78                     \\ \cline{2-2} \cline{4-5}
                           & Binary PV-DM                    &                      & 0.18                 & 0.63                     \\ \hline
      \multirow{6}{*}{32}  & \multirow{2}{*}{PV-DBOW}        & no                   & 0.26                 & 0.75                     \\ \cline{3-5}
                           &                                 & yes                  & 0.27                 & 0.77                     \\ \cline{2-5}
                           & \multirow{2}{*}{Binary PV-DBOW} & no                   & 0.22                 & 0.6                      \\ \cline{3-5}
                           &                                 & yes                  & \textbf{0.25}        & \textbf{0.66}            \\ \cline{2-5}
                           & PV-DM                           & \multirow{2}{*}{N/A} & 0.23                 & 0.77                     \\ \cline{2-2} \cline{4-5}
                           & Binary PV-DM                    &                      & 0.17                 & 0.53                     \\ \hline
    \end{tabular}

  \longcaption{Information retrieval RCV1 results.}{The best binary results for each code dimensionality are highlighted.}
  \label{tab:dsh_rcv1_results}
\end{table}
The results were calculated similarly to the 20 Newsgroups case.
However, since RCV1 is a much bigger dataset, treating each document as a query would be computationally prohibitive.
Therefore, we randomly selected 10 percent of test documents as queries.
Comparison of precision-recall curves is depicted in~\figref{dsh_rcv1_precision_recall}.
As in the case of the 20 Newsgroups dataset, PV-DBOW gives slightly better results than PV-DM.
Results for 128-dimensional binary codes are approximately 20\% higher than those reported in~\cite{salakhutdinov2009semantic}.

\begin{figure}[htb!]
  \centering
  \begin{subfigure}[b]{0.495\linewidth}
    \centering
    \includegraphics[width=\textwidth]{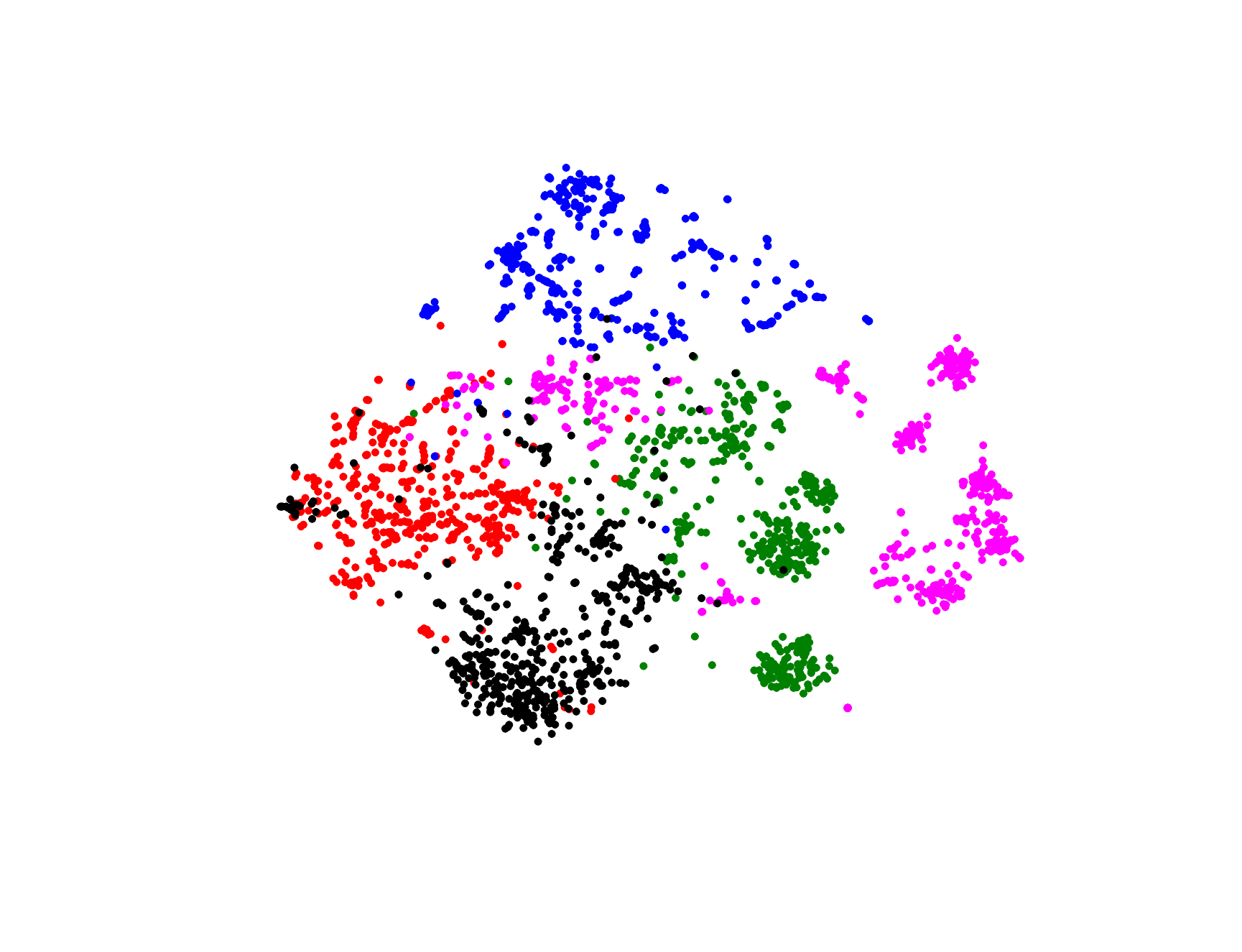}
    \caption{300 dimensions}
  \end{subfigure}
  \hfill
  \begin{subfigure}[b]{0.495\linewidth}
    \centering
    \includegraphics[width=\textwidth]{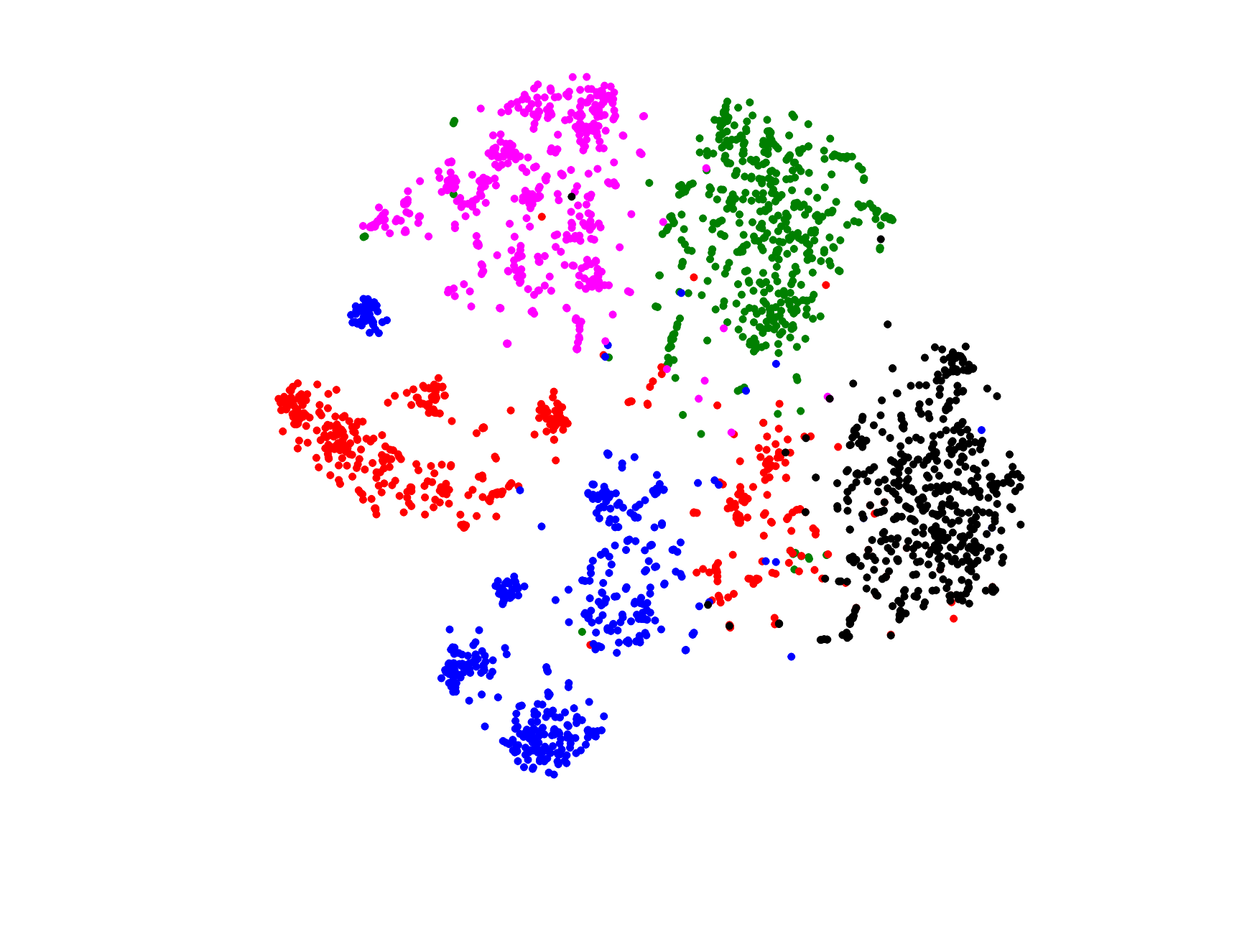}
    \caption{128 dimensions}
  \end{subfigure}
  \hfill
  \begin{subfigure}[b]{0.495\linewidth}
    \centering
    \includegraphics[width=\textwidth]{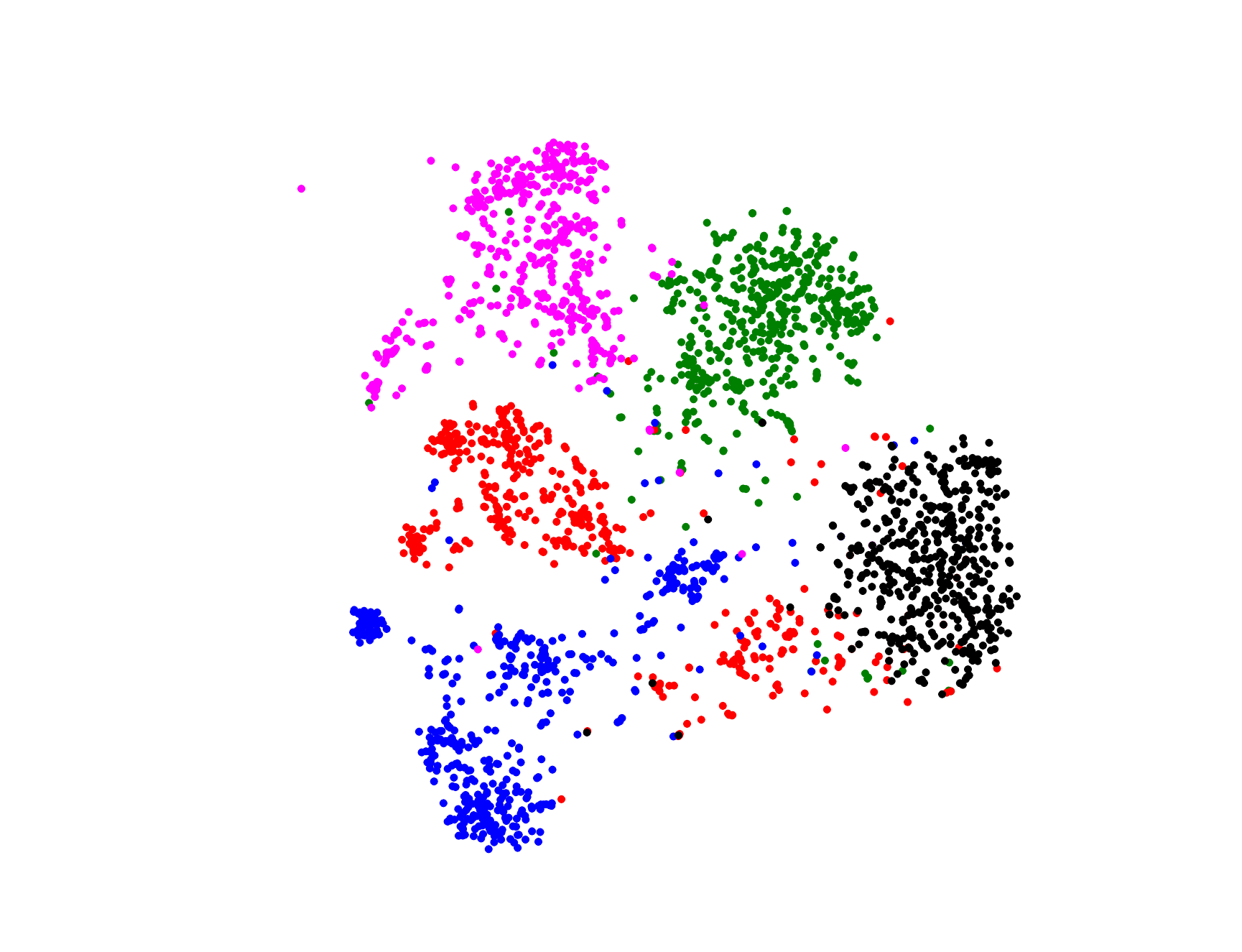}
    \caption{64 dimensions}
  \end{subfigure}
  \hfill
  \begin{subfigure}[b]{0.495\linewidth}
    \centering
    \includegraphics[width=\textwidth]{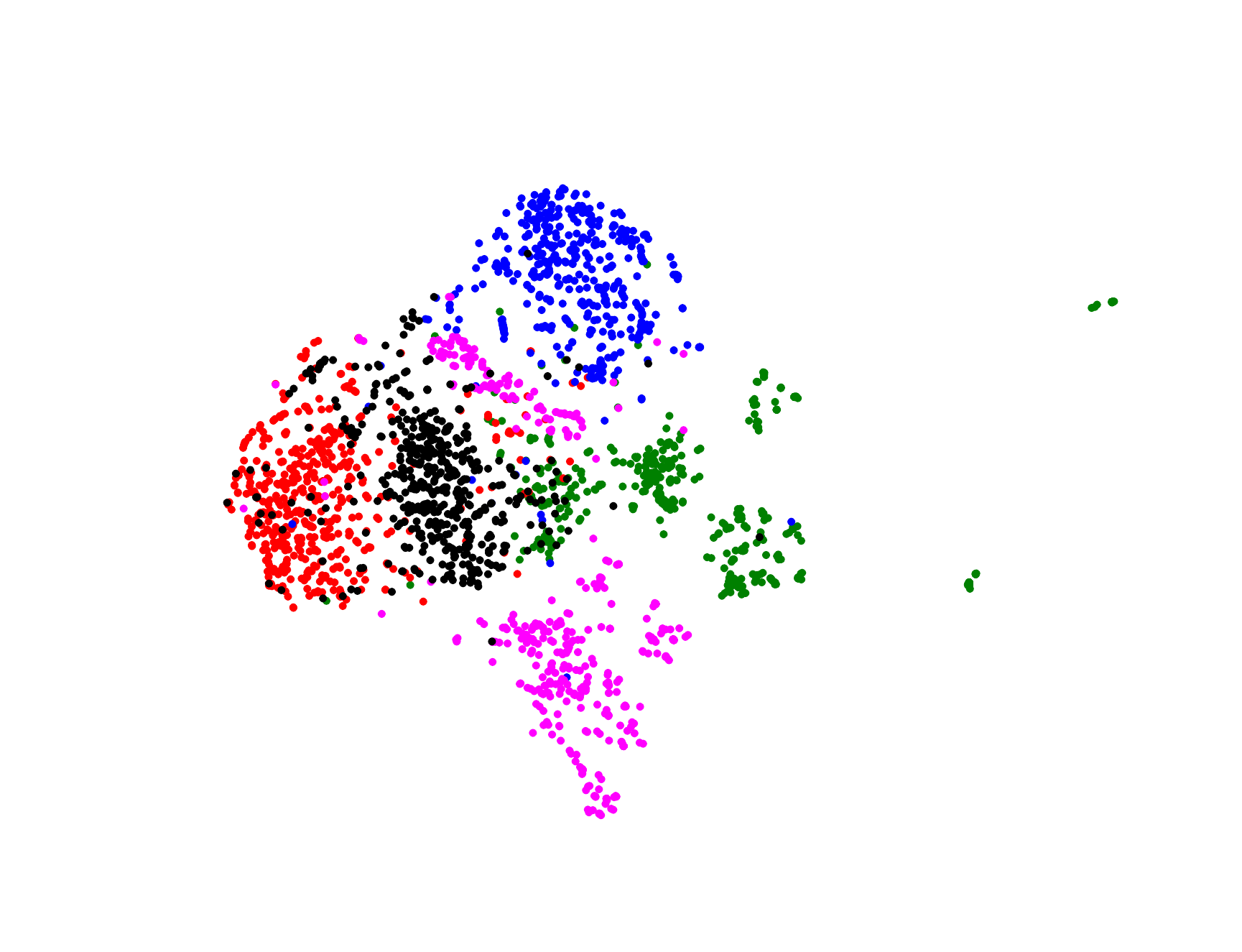}
    \caption{32 dimensions}
  \end{subfigure}
  \longcaption{t-SNE visualizations of binary codes of six selected topics from the RCV1 dataset for different code dimensionalities.}
              {The Hamming distance was used as a similarity measure.
               Selected topics: green - disasters and accidents, red - government borrowing, blue - accounts/earnings,
               magenta - energy markets, black - EC monetary/economic.}
  \label{fig:rcv1_bin_visualise}
\end{figure}

We also generated two-dimensional visualizations of codes for selected topics from the RCV1 dataset.
They are presented in~\figref{rcv1_bin_visualise}.
Topic selection and color assignment are the same as in~\cite[Fig. 5]{salakhutdinov2009semantic}.
Three of five topics are clearly separated. Within two topics, i.e. \emph{government borrowing} and \emph{accounts/earnings},
a few subgroups can be discerned.

\subsection{English Wikipedia}

In order to test Binary PV models with dataset even bigger than RCV1 we trained them on the English Wikipedia snapshot\footnote{From April 5th, 2016},
which has almost $ 4 \times 10^6 $ articles. We used words and bigrams with at least 100 occurrences, which gives a vocabulary with approximately $ 1.5 \times 10^6 $ elements.
To test our models we held out randomly selected 10\% of all articles. From the remaining documents we took another 10\% to form a validation set,
which was used to select model hyperparameters. The following hyperparameters were selected with validation experiments:
learning rate 1.4 for PV-DBOW, 0.9 for Binary PV-DBOW, 0.5 for PV-DM and 0.6 for Binary PV-DM.
A~dropout rate of 10\% for PV-DBOW and Binary PV-DBOW, and no dropout for PV-DM and Binary PV-DM.

To assess the relevancy of articles from English Wikipedia we employ categories assigned to them.
Since Wikipedia categorization is complex (see \sectionref{datasets_enwiki} for details), we adopted a simplified relevancy measure:
two articles are relevant if they share at least one category.
We also removed from the test set categories with less than 20 documents as well as documents that were
left with no categories. Overall, the relevancy is measured over almost $ 10^6 $ categories, making English Wikipedia harder than the
previous two benchmarks.

\begin{figure}[htb!]
  \centering
  \begin{subfigure}[b]{0.495\linewidth}
    \centering
    \includegraphics[width=\textwidth]{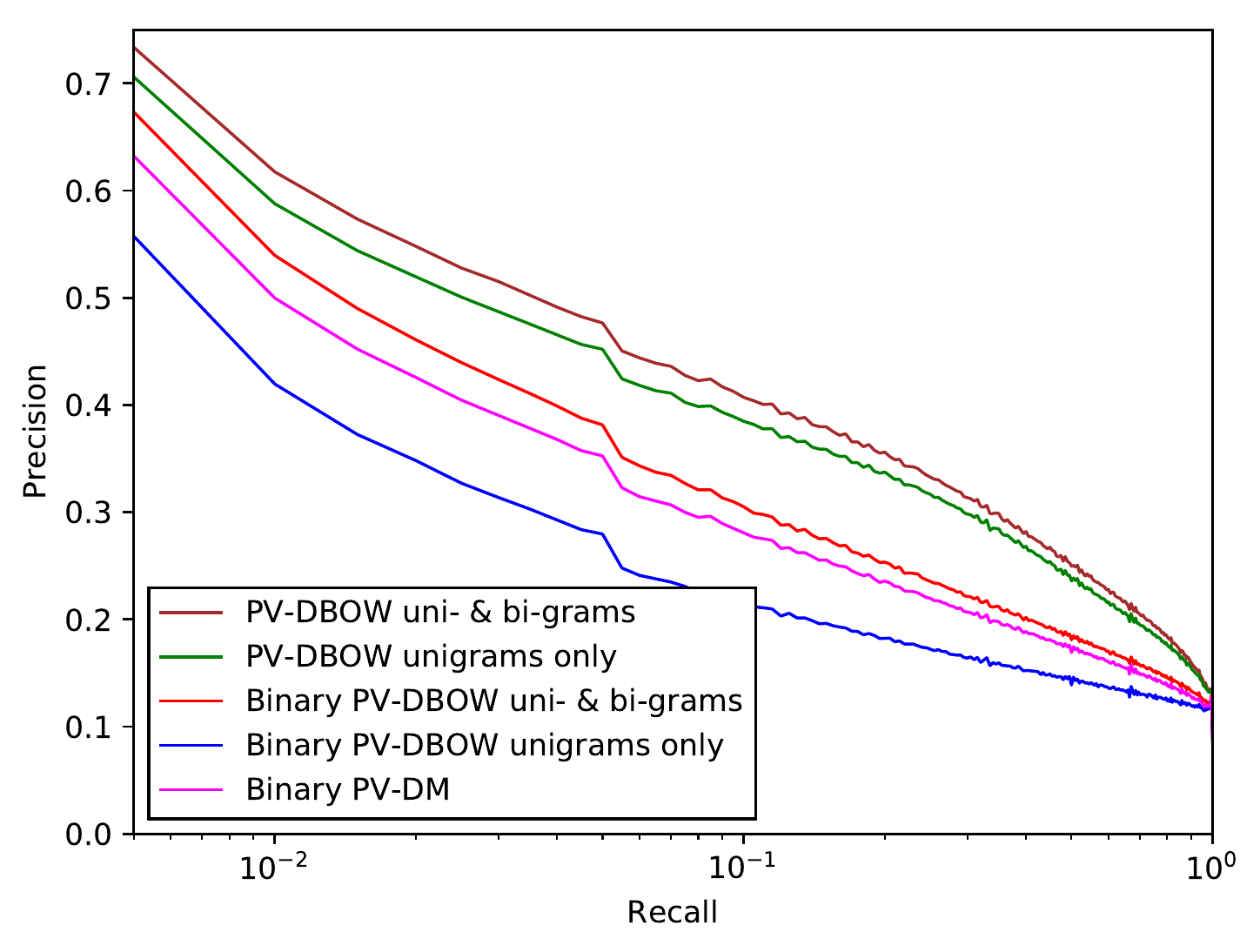}
    \caption{300 dimensions}
  \end{subfigure}
  \hfill
  \begin{subfigure}[b]{0.495\linewidth}
    \centering
    \includegraphics[width=\textwidth]{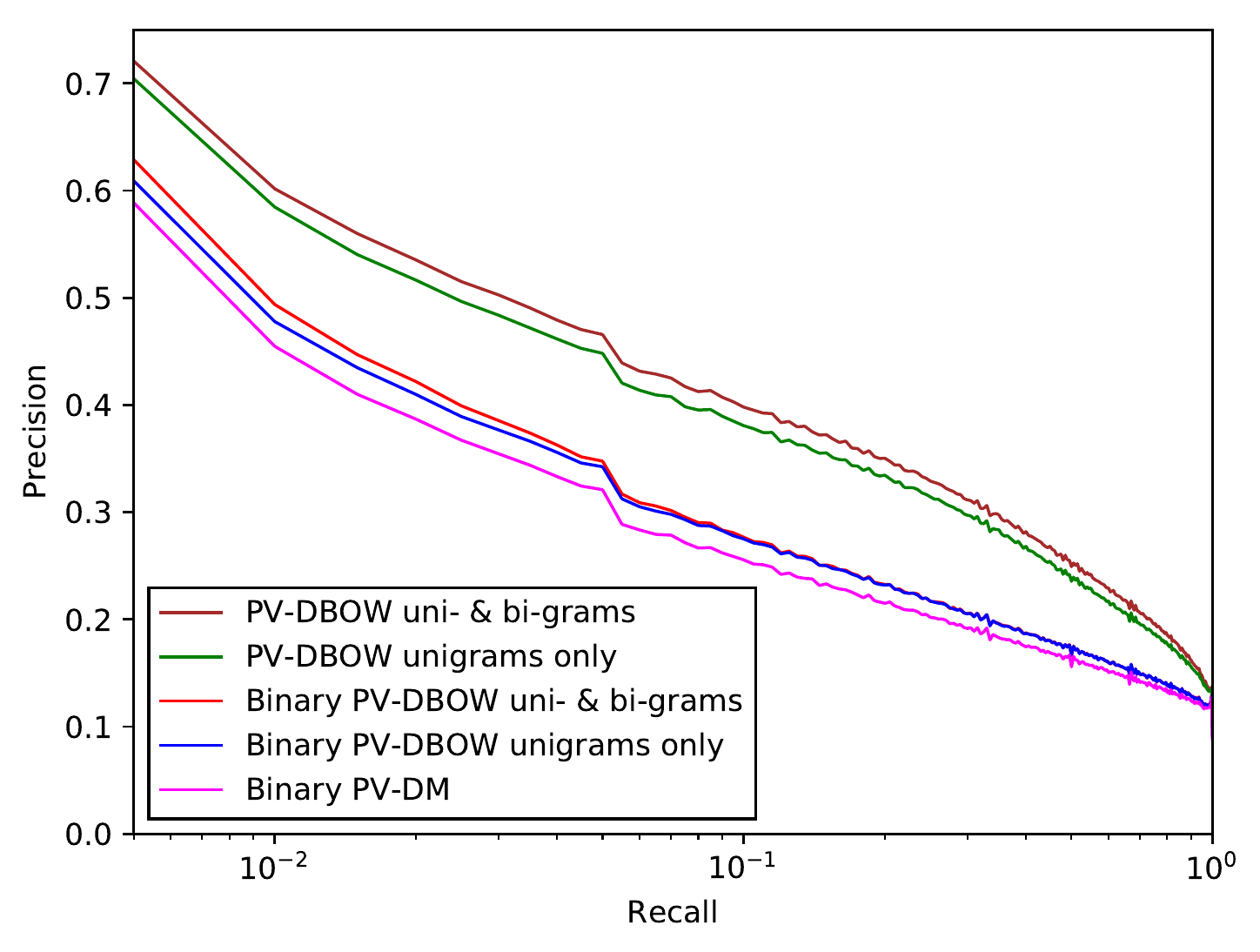}
    \caption{128 dimensions}
  \end{subfigure}
  \hfill
  \begin{subfigure}[b]{0.495\linewidth}
    \centering
    \includegraphics[width=\textwidth]{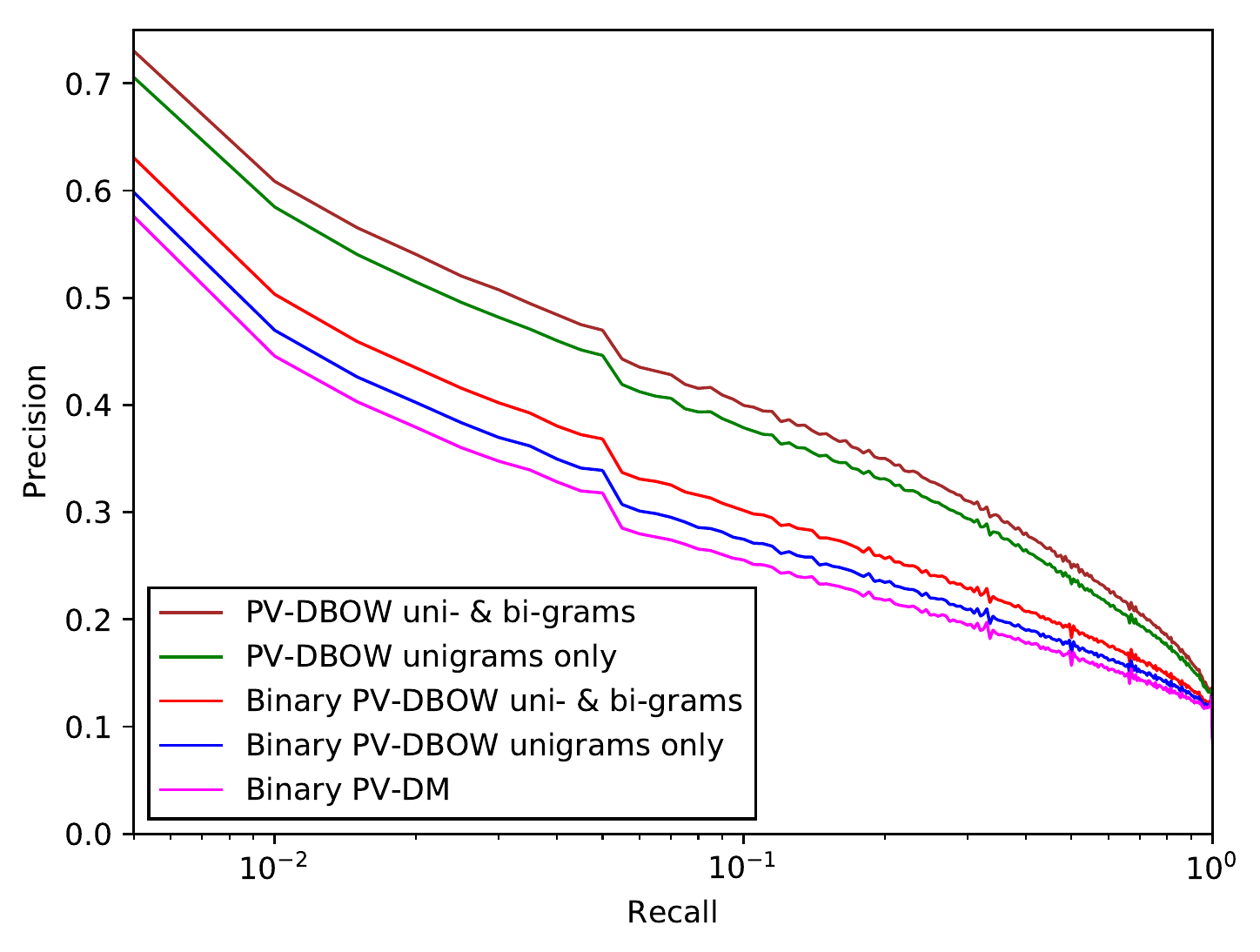}
    \caption{64 dimensions}
  \end{subfigure}
  \hfill
  \begin{subfigure}[b]{0.495\linewidth}
    \centering
    \includegraphics[width=\textwidth]{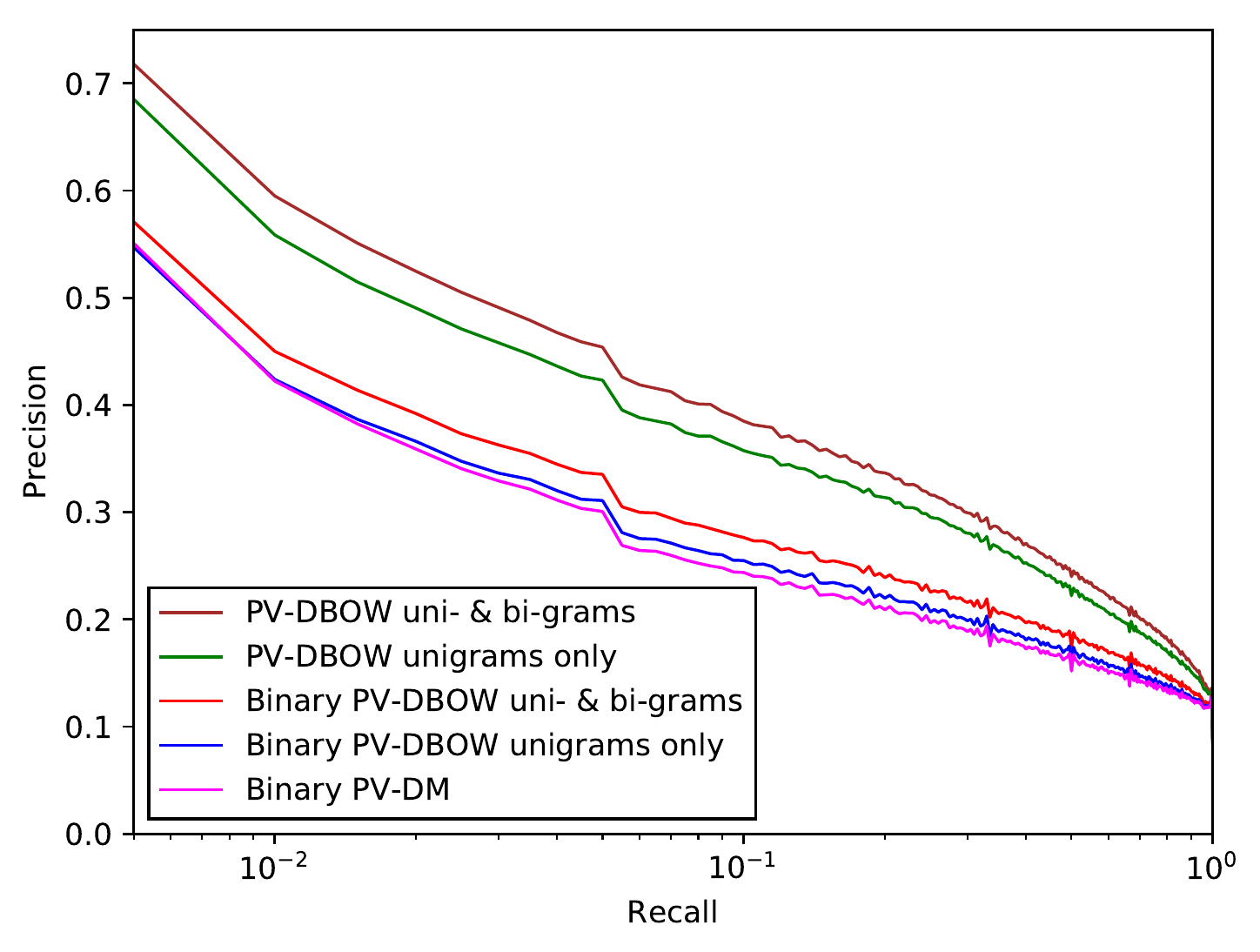}
    \caption{32 dimensions}
  \end{subfigure}
  \longcaption{The English Wikipedia precision-recall curves for different code dimensionalities and different model variants.}
              {For real-valued codes cosine distance was used as a similarity measure.
               For binary codes the Hamming distance was used as a similarity measure.}
  \label{fig:dsh_enwiki_precision_recall}
\end{figure}

Results for Paragraph Vector and Binary Paragraph Vector for different code sizes are reported in~\tabref{dsh_enwiki_results}.
\begin{table}[htb]
  \centering
    \begin{tabular}{|c|c|c|c|c|}
      \hline
      Code                 & \multirow{2}{*}{Model}          & Include              & \multirow{2}{*}{MAP} & \multirow{2}{*}{NDCG@10} \\
      dimensionality       &                                 & bigrams              &                      &                          \\ \hline
      \multirow{6}{*}{300} & \multirow{2}{*}{PV-DBOW}        & no                   & 0.25                 & 0.59                     \\ \cline{3-5}
                           &                                 & yes                  & 0.26                 & 0.61                     \\ \cline{2-5}
                           & \multirow{2}{*}{Binary PV-DBOW} & no                   & 0.15                 & 0.46                     \\ \cline{3-5}
                           &                                 & yes                  & \textbf{0.2}         & \textbf{0.54}            \\ \cline{2-5}
                           & PV-DM                           & \multirow{2}{*}{N/A} & 0.25                 & 0.59                     \\ \cline{2-2} \cline{4-5}
                           & Binary PV-DM                    &                      & 0.18                 & 0.51                     \\ \hline
      \multirow{6}{*}{128} & \multirow{2}{*}{PV-DBOW}        & no                   & 0.25                 & 0.59                     \\ \cline{3-5}
                           &                                 & yes                  & 0.26                 & 0.6                      \\ \cline{2-5}
                           & \multirow{2}{*}{Binary PV-DBOW} & no                   & 0.18                 & 0.48                     \\ \cline{3-5}
                           &                                 & yes                  & \textbf{0.18}        & \textbf{0.49}            \\ \cline{2-5}
                           & PV-DM                           & \multirow{2}{*}{N/A} & 0.24                 & 0.59                     \\ \cline{2-2} \cline{4-5}
                           & Binary PV-DM                    &                      & 0.16                 & 0.46                     \\ \hline
      \multirow{6}{*}{64}  & \multirow{2}{*}{PV-DBOW}        & no                   & 0.24                 & 0.58                     \\ \cline{3-5}
                           &                                 & yes                  & 0.26                 & 0.6                      \\ \cline{2-5}
                           & \multirow{2}{*}{Binary PV-DBOW} & no                   & 0.17                 & 0.46                     \\ \cline{3-5}
                           &                                 & yes                  & \textbf{0.19}        & \textbf{0.49}            \\ \cline{2-5}
                           & PV-DM                           & \multirow{2}{*}{N/A} & 0.24                 & 0.57                     \\ \cline{2-2} \cline{4-5}
                           & Binary PV-DM                    &                      & 0.16                 & 0.44                     \\ \hline
      \multirow{6}{*}{32}  & \multirow{2}{*}{PV-DBOW}        & no                   & 0.23                 & 0.55                     \\ \cline{3-5}
                           &                                 & yes                  & 0.25                 & 0.58                     \\ \cline{2-5}
                           & \multirow{2}{*}{Binary PV-DBOW} & no                   & 0.16                 & 0.41                     \\ \cline{3-5}
                           &                                 & yes                  & \textbf{0.17}        & \textbf{0.44}            \\ \cline{2-5}
                           & PV-DM                           & \multirow{2}{*}{N/A} & 0.23                 & 0.55                     \\ \cline{2-2} \cline{4-5}
                           & Binary PV-DM                    &                      & 0.15                 & 0.41                     \\ \hline
    \end{tabular}

  \longcaption{Information retrieval results for English Wikipedia.}{The best binary results for each code dimensionality are highlighted.}
  \label{tab:dsh_enwiki_results}
\end{table}
As in the case of RCV1, when calculating these numbers we randomly selected 10 percent of test documents and used only those documents as
queries. Results for this benchmark are weaker than in the case of the two previous datasets. This is not a surprise,
since Wikipedia has a much wider range of topics than both 20 Newsgroups and RCV1. Comparison of precision-recall curves is depicted in~\figref{dsh_enwiki_precision_recall}.
To obtain those results we needed to overcome some technical difficulties stemmed from dataset size. As explained in~\sectionref{dsh_ir},
in order to compute relevancy between query documents and test documents in a batched way, we perform matrix multiplication between
a mini-batch of query document labels and all test document labels. To this end, we need to store labels in a one-hot (or a few-hot) format.
Since there are approximately $ 10^6 $ labels and $ 10^5 $ test documents, storing labels as dense matrices would be very memory demanding.
Therefore, we store them in a sparse, compressed row matrix format.


\subsection{Comparison of binarization methods} \label{sec:bin_units_comparison}

As stated in \sectionref{bpv_architecture}, instead of the deterministic binarization method, proposed by Salakhutdinov and Krizhevsky,
one can also employ binary stochastic neurons. Binary stochastic neurons yield either 0 or 1 at their output. In order to train the network
with BSNs using error backpropagation one needs to estimate the gradient of the expected loss under stochastic activations.

Probably the simplest gradient estimator for BSNs is the \emph{straight-through} estimator proposed by Bengio et al.~\cite[section 4]{bengio2013estimating}.
In this approach it is assumed that errors are backpropagated as if BSN was the identity function. Therefore, the gradient is simply:
\begin{equation} \label{eq:dsh_bsn_st_1}
\frac{d BSN ( a )}{d a} = 1,
\end{equation}
where $ a $ is neuron pre-activation. Bengio et al. also consider a modification to the straight-through estimator, where the gradient
equals the derivative of the sigmoid function used to calculate the probability of activation of the BSN:
\begin{equation} \label{eq:dsh_bsn_st_2}
\frac{d BSN ( a )}{d a} = \frac{d \sigma ( a )}{d a} = \sigma ( a ) (1 - \sigma(a)).
\end{equation}
Bengio et al. assert that \equationref{dsh_bsn_st_2} produces better results than \equationref{dsh_bsn_st_1}.
However, Raiko et al. claim the opposite~\cite[section 3]{raiko2014techniques}.
We therefore consider both variants in our evaluation.

Performance of BSNs can be improved by using the \emph{slope-annealing} trick proposed by Chung et al.~\cite[section 3.3]{chung2016hierarchical}.
The idea is to gradually increase the slope $ s $ of the sigmoid function during training to increase the probability of activation $ P $:
\begin{equation}
P = \sigma ( s * a ).
\end{equation}
At the beginning of the training the network weights are not yet fitted to data, and, consequently, decision made by BSN can be random.
Therefore, we start the training with a slope $ s = 1.0 $. Later on, when weights are more fitted to the data, we increase the slope
and consequently make BSNs more `confident'.

In~\tabref{bpv_bin_units} comparison of both deterministic and stochastic binary units is presented.
\begin{table}[htb]
  \centering
    \begin{tabular}{|l|l|l|}
      \hline
      Binarization approach                                                   & MAP                   & NDCG@10               \\ \hline
      Salakhutdinov \& Hinton binarization                                    & 0.19                  & 0.37                  \\ \hline
      Krizhevsky \& Hinton binarization                                       & \textbf{0.32}         & \textbf{0.54}         \\ \hline
      BSN with \equationref{dsh_bsn_st_1} straight-through gradient estimator & 0.25                  & 0.42                  \\ \hline
      BSN with \equationref{dsh_bsn_st_1} straight-through gradient estimator & \multirow{2}{*}{0.3}  & \multirow{2}{*}{0.5}  \\
      with slope-annealing                                                    &                       &                       \\ \hline
      BSN with \equationref{dsh_bsn_st_2} straight-through gradient estimator & 0.14                  & 0.33                  \\ \hline
      BSN with \equationref{dsh_bsn_st_2} straight-through gradient estimator & \multirow{2}{*}{0.22} & \multirow{2}{*}{0.48} \\
      with slope-annealing                                                    &                       &                       \\ \hline
     \end{tabular}
    \caption{Comparison of performance of different binary units for 32 bit model trained on the 20 Newsgroups dataset.}
  \label{tab:bpv_bin_units}
\end{table}
In case of the former, we tried the approaches proposed by Salakhutdinov and Krizhevsky.
In case of the latter, we examined straight-through gradient estimator in both described variants (\equationref{dsh_bsn_st_1} and \equationref{dsh_bsn_st_2}).
Model hyperparameters were selected separately for each binarization approach using hold-out validation set. Standard deviation of the Gaussian noise
added to the pre-activation signal in the Salakhutdinov's binarization is 0.4. Learning rate for that case is 1.0. Learning rates for BSN have higher values.
They are 1.5 for \equationref{dsh_bsn_st_1} and as high as 11.0 for \equationref{dsh_bsn_st_2}. In contrast to the Krizhevsky's binarization,
dropout does not improve the results of the Salakhutdinov's binarization. Neither in case of BSNs. In variants that use slope annealing we increase
the slope hiperparameter by 0.1 in each epoch. As can be concluded from the table~\tabref{bpv_bin_units}, BSNs give relatively good results,
but not as good as a simple deterministic approach proposed by Krizhevsky \& Hinton.

\subsection{Comparison against indirect hashing approaches}

\begin{figure}[htb!]
  \centering
  \begin{subfigure}[b]{0.495\linewidth}
    \centering
    \includegraphics[width=\textwidth]{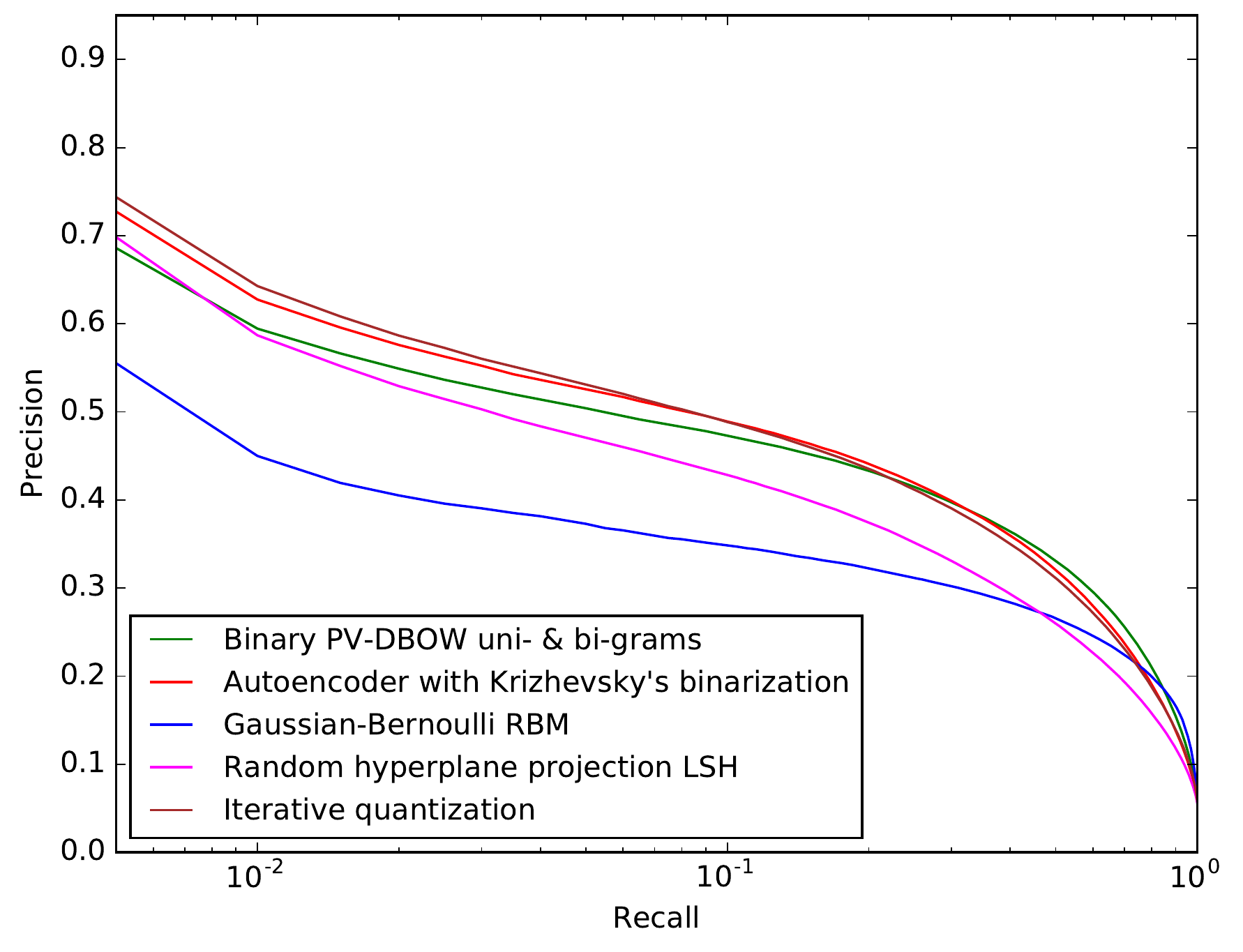}
    \caption{20 Newsgroups}
  \end{subfigure}
  \hfill
  \begin{subfigure}[b]{0.495\linewidth}
    \centering
    \includegraphics[width=\textwidth]{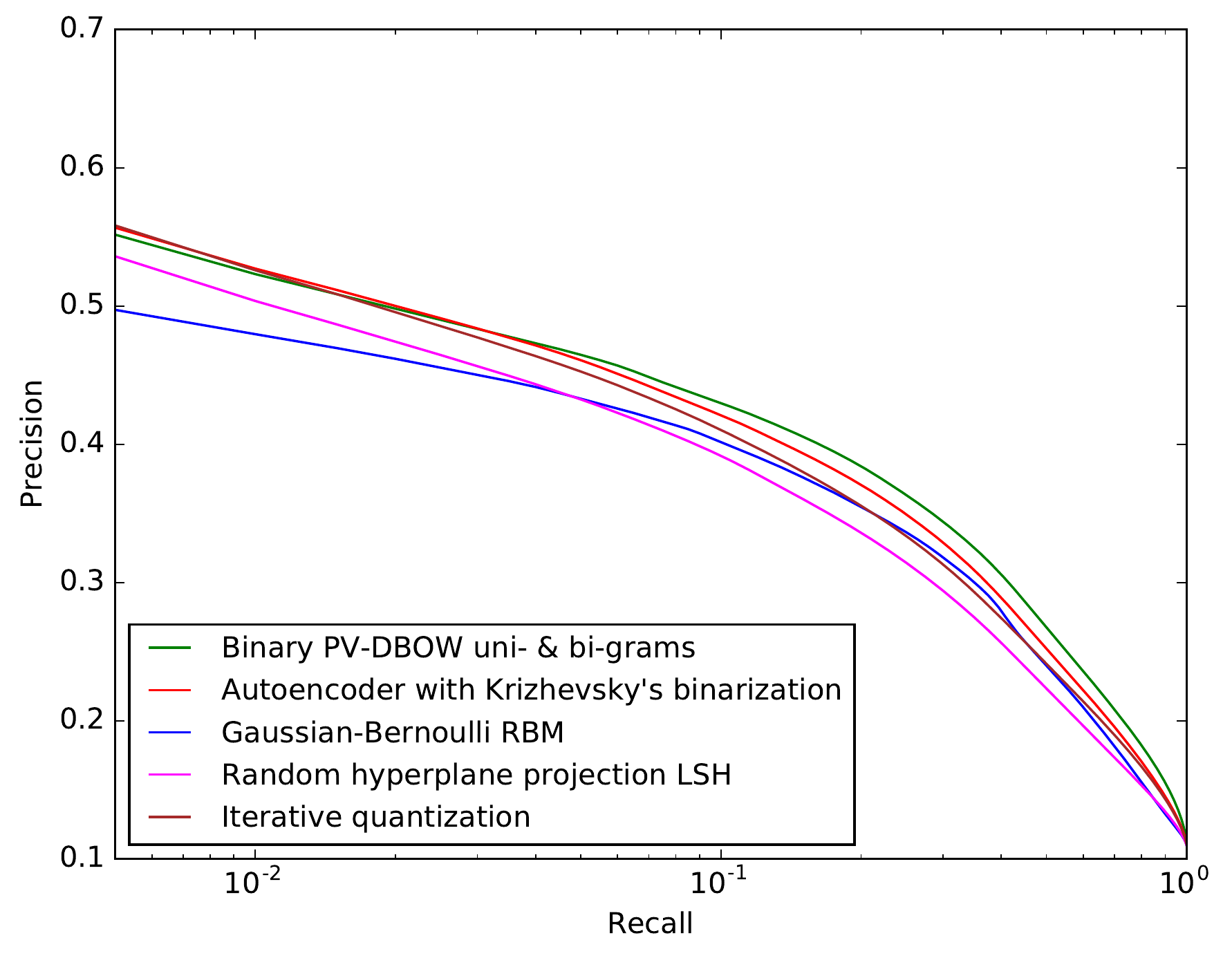}
    \caption{RCV1}
  \end{subfigure}
  \hfill
  \begin{subfigure}[b]{0.495\linewidth}
    \centering
    \includegraphics[width=\textwidth]{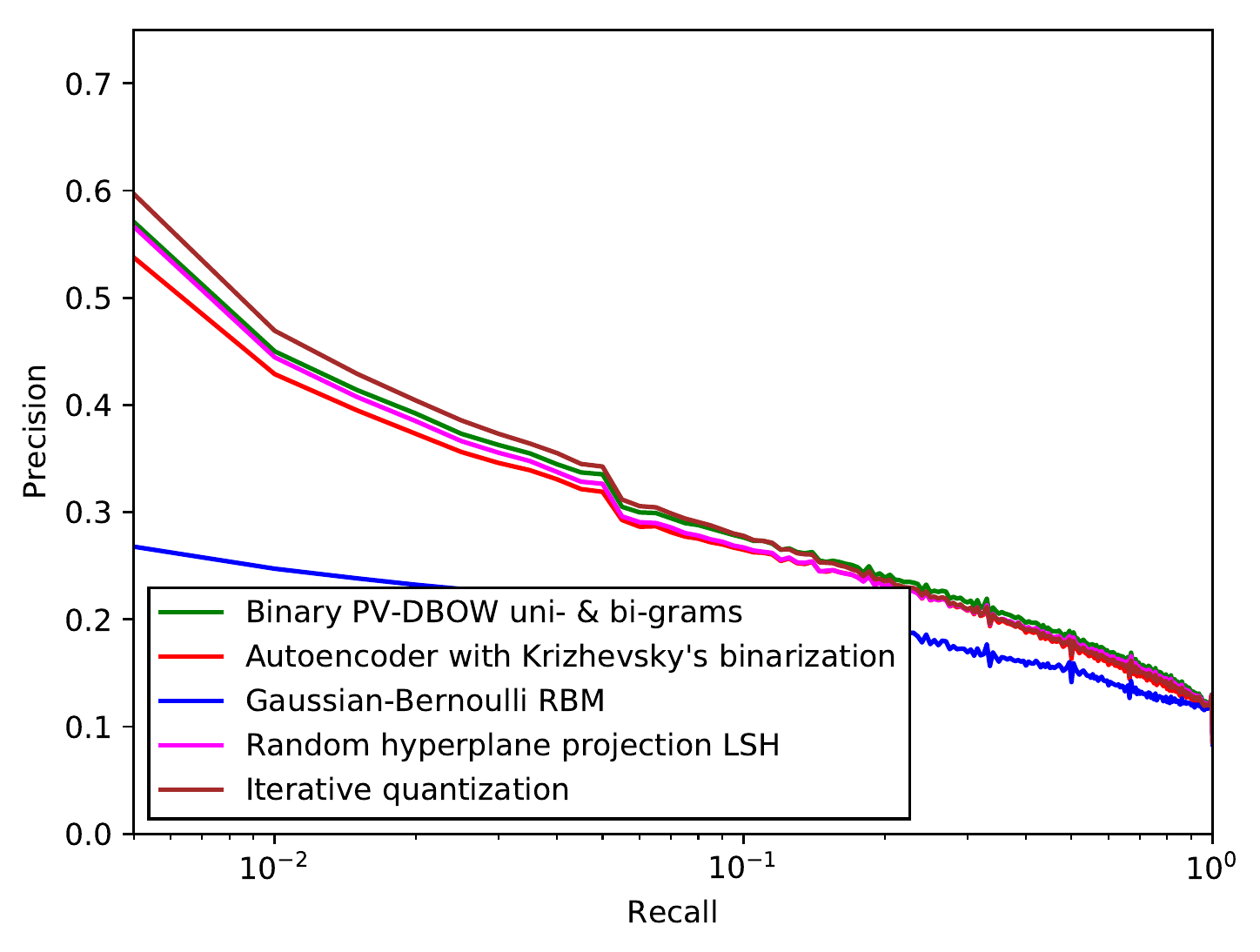}
    \caption{English Wikipedia}
  \end{subfigure}
  \longcaption{Information retrieval results for $32$-bit binary codes}
              {constructed by first inferring $32$d real-valued paragraph vectors and then employing another unsupervised model
               or hashing algorithm for binarization. Paragraph vectors were inferred using PV-DBOW with bigrams.}
  \label{fig:bpv_ae_and_rbm}
\end{figure}

The Binary Paragraph Vector models enable simple generation of document hashes. Alternatively, similar codes can be obtained in two stages,
by first learning real-valued document embeddings and then applying to these embeddings an off-the-shelf state-of-the-art locality-preserving
hashing technique. To validate this approach we generated real-valued codes using PV-DBOW model and then applied to them one of the four hashing techniques: an autoencoder
with sigmoid coding layer and the Krizhevsky's binarization, a Gaussian-Bernoulli Restricted Boltzmann Machine (RBM)~\cite{welling2004exponential},
random hyperplane projection (a.k.a. SimHash)~\cite{charikar2002similarity} and iterative quantization (ITQ)~\cite{gong2011iterative}.
We considered two variants, where hashes are generated either from low-dimensional or high-dimensional document embeddings.
Results are presented in~\tabref{bpv_baseline} and~\figref{bpv_ae_and_rbm}.
\begin{table}[htb]
  \small
  \centering
    \begin{tabular}{|c|c|c|c|c|c|c|c|}
      \hline
      \multirow{2}{*}{Binarization model} & \multirow{2}{*}{\rot{Embed. size~}} & \multicolumn{2}{|c|}{20 Newsgroups} & \multicolumn{2}{|c|}{RCV1} & \multicolumn{2}{|c|}{EN Wikipedia} \\ \cline{3-8}
                                          &                                     & \rot{MAP} & \rot{NDCG@10~}          & \rot{MAP} & \rot{NDCG@10}  & \rot{MAP} & \rot{NDCG@10}          \\ \hline
      Autoencoder with                    & 32                                  & 0.32      & 0.57                    & 0.24      & 0.67           & 0.16      & 0.42                   \\ \cline{2-8}
      Krizhevsky's binarization           & 300                                 & 0.29      & 0.56                    & 0.22      & 0.71           & 0.17      & 0.44                   \\ \hline
      Gaussian-Bernoulli                  & 32                                  & 0.26      & 0.39                    & 0.23      & 0.52           & 0.12      & 0.19                   \\ \cline{2-8}
      RBM                                 & 300                                 & 0.36      & 0.55                    & 0.26      & 0.71           & 0.16      & 0.33                   \\ \hline
      Random hyperplane                   & 32                                  & 0.27      & 0.53                    & 0.21      & 0.66           & 0.16      & 0.44                   \\ \cline{2-8}
      projection (SimHash)                & 300                                 & 0.13      & 0.31                    & 0.17      & 0.56           & 0.12      & 0.36                   \\ \hline
      Iterative                           & 32                                  & 0.31      & 0.58                    & 0.23      & 0.68           & 0.17      & 0.46                   \\ \cline{2-8}
      quantization                        & 300                                 & 0.31      & 0.58                    & 0.22      & 0.71           & 0.17      & 0.44                   \\ \hline
    \end{tabular}
    \longcaption{Information retrieval results for $32$-bit binary codes}{constructed by first inferring $32$d or $300$d real-valued paragraph vectors and
                 then employing another unsupervised model or hashing algorithm for binarization. Paragraph vectors were inferred using PV-DBOW
                 with bigrams.}
  \label{tab:bpv_baseline}
\end{table}

As can be seen in~\figref{bpv_ae_and_rbm},
two of the tested methods, namely an autoencoder and ITQ, yielded precisions for low recall values slightly higher than Binary Paragraph Vector.
However, the area under the curve is still smaller than in the case of Binary PV.
Given the simplicity of end-to-end training of Binary Paragraph Vector models there is no clear benefit from using two-stage approach.
The training and inference time in the Binary Paragraph Vector models is almost the same as that of the original PV.

Autoencoder used in this comparative experiment was implemented using TensorFlow library.
We trained it for 10 epochs, splitting data into mini-batches of 100 examples. AdaGrad with learning rate of 1.0 was used for optimization.
A~dropout rate of 10\% was used. RBM was implemented using AGH deep learning library described in \sectionref{dlcuda}.
We trained it for 10 epochs with learning rate of 0.01. The cost function was optimized by gradient descent optimizer with momentum equal to 0.5,
This model also used a weight cost of 0.0002. The training was carried out in mini-batches of 128 examples.
For SimHash and ITQ we used simple implementations provided by Dong Guosheng\footnote{Available at \url{https://github.com/dongguosheng/lsh}}
with default hyperparameters.

\subsection{Transfer learning}

\begin{figure}[htb!]
  \centering
  \begin{subfigure}[b]{0.495\linewidth}
    \centering
    \includegraphics[width=\textwidth]{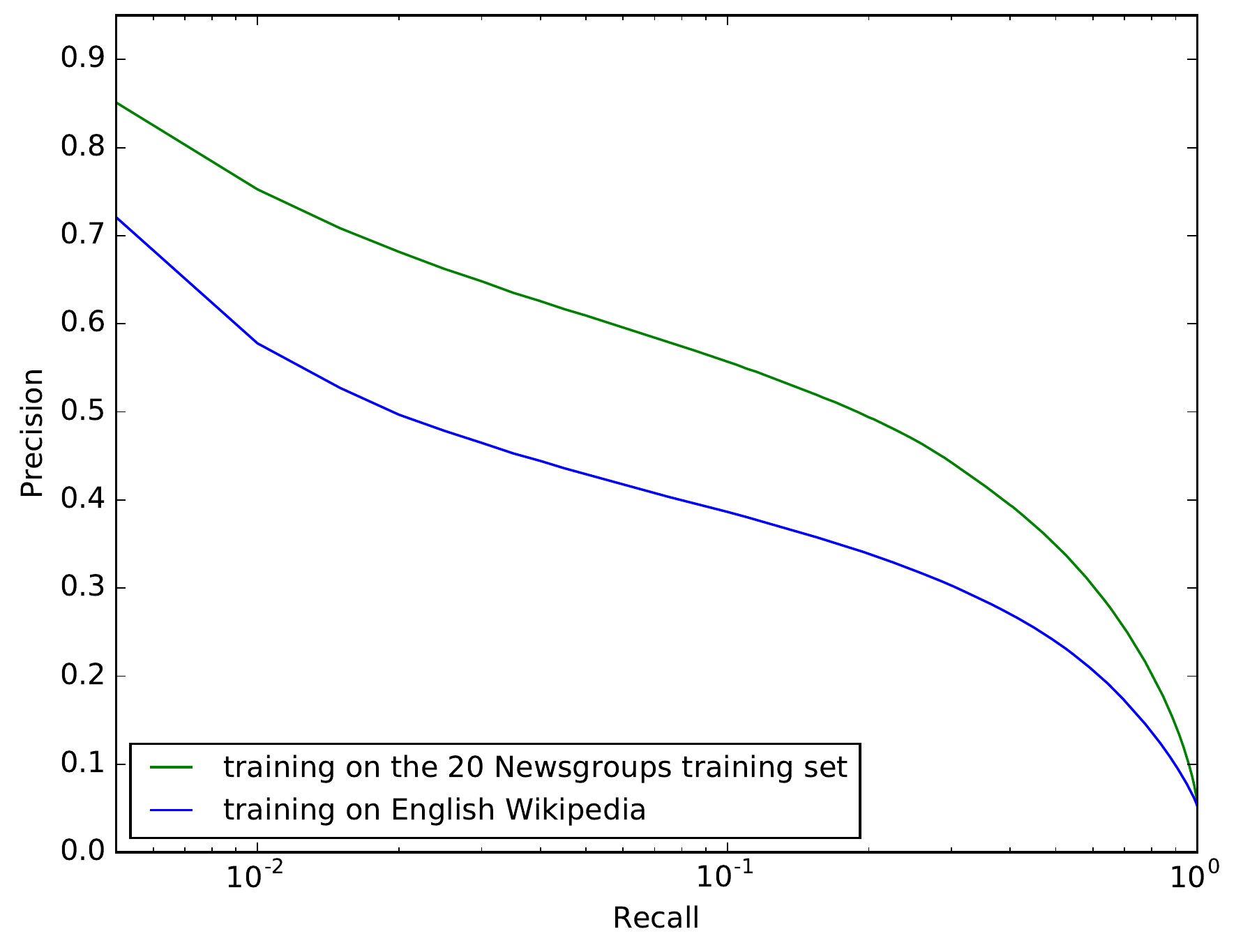}
    \caption{20 Newsgroups}
  \end{subfigure}
  \hfill
  \begin{subfigure}[b]{0.495\linewidth}
    \centering
    \includegraphics[width=\textwidth]{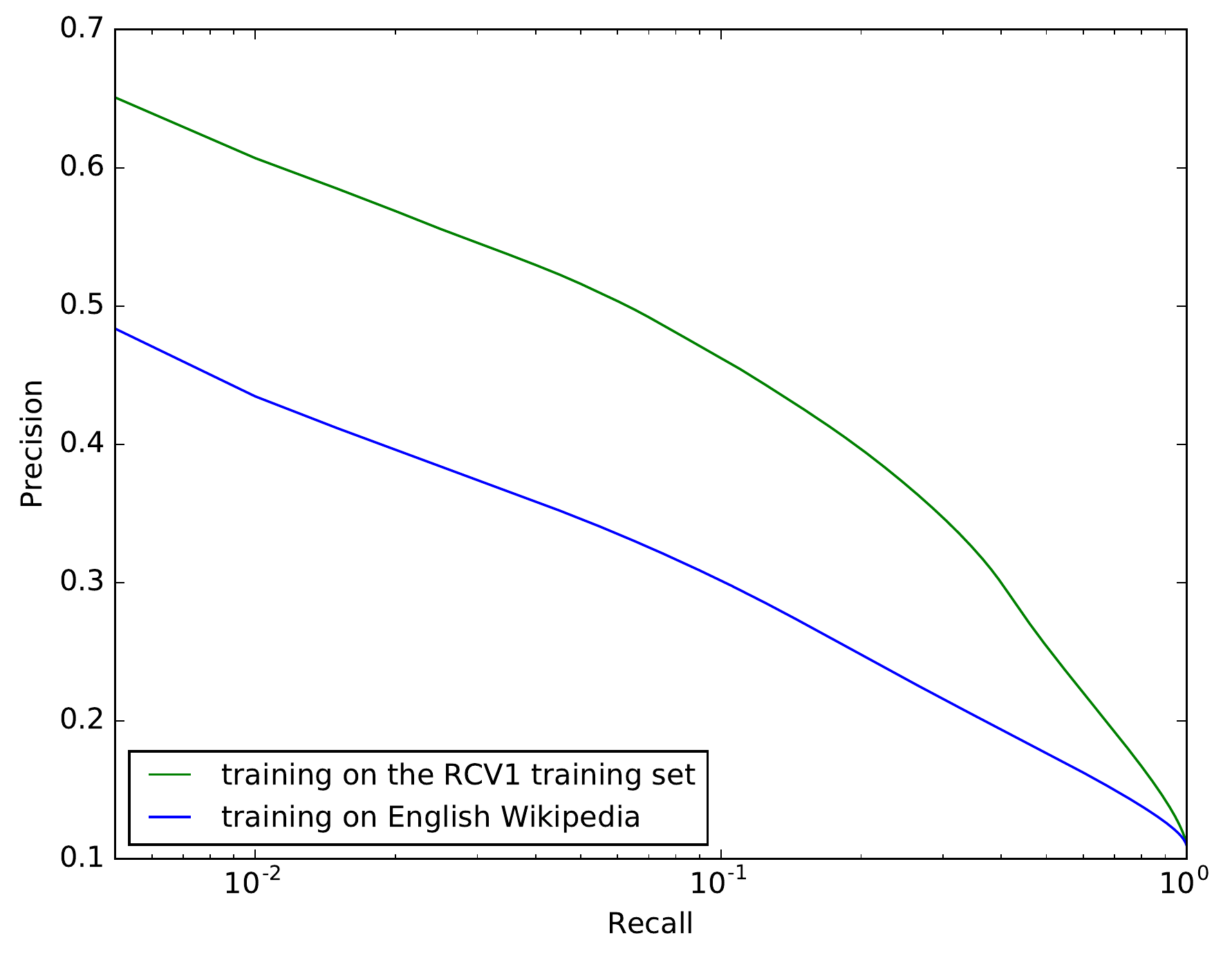}
    \caption{RCV1}
  \end{subfigure}
  \caption{128-dimensional binarized PV-DBOW model.}
  \label{fig:dsh_transfer_learning}
\end{figure}

As shown in the previous sections, our model enables learning of short, high-quality binary codes for text documents.
However, in all cases presented above (20 Newsgroups, RCV1 and Wikipedia) the models were evaluated on test examples held out from the datasets
used to fit the model's parameters. In other words, test examples were from the same domain as training examples. One could pose
a question: what if we want to generate binary codes for documents that are not associated with any domain-specific corpus?
One of the solution in this case could be to train the model based on a big generic text corpus covering wide variety of domains.
To validate this approach, we trained the Binary PV-DBOW model on English Wikipedia snapshot
and then we inferred binary codes for the test parts of the 20 Newsgroups and RCV1 datasets.
The results are presented in~\tabref{dsh_transfer_learning} and in~\figref{dsh_transfer_learning}.
\begin{table}[htb]
\begin{minipage}{0.95\textwidth}
  \centering
    \begin{tabular}{|c|c|c|}
      \hline
      Dataset       & MAP  & NDCG@10 \\ \hline
      20 Newsgroups & 0.19 & 0.51    \\ \hline
      RCV1          & 0.18 & 0.66    \\ \hline
    \end{tabular}
  \caption{Information retrieval results for transfer learning for 128-dimensional binary codes.}
  \label{tab:dsh_transfer_learning}
\end{minipage}
\end{table}

As expected, the results are worse than those for the cases when the model was trained on domain-specific training sets.
However, the precision-recall curves are, approximately, only $ 10\% $ lower than in case of training with domains-specific data.
It can, therefore, be concluded that the codes learned in the transfer learning settings preserve a lot of the semantic similarity
of documents.

\section{Real-Binary Paragraph Vector model}

\begin{figure}[htb!]
  \centering
  \includegraphics[width=0.8\linewidth]{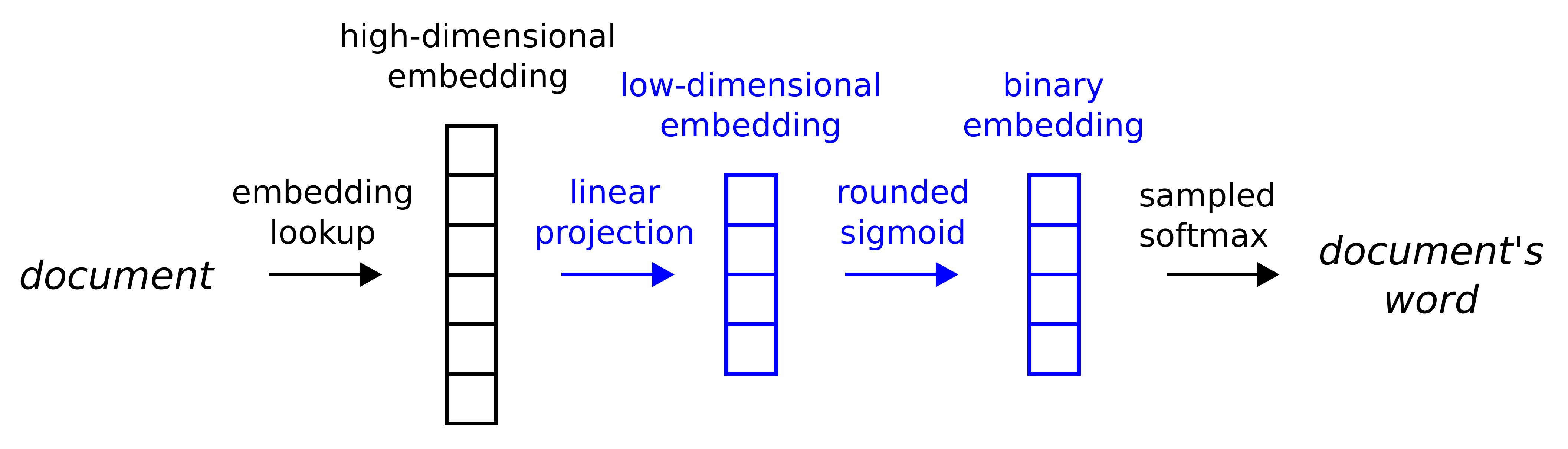}
  \longcaption{The Real-Binary PV-DBOW model.}{Modifications to the original PV-DBOW model are highlighted in blue.}
  \label{fig:pv-dbow-bin-projections}
\end{figure}

As described in~\cite{salakhutdinov2009semantic}, when a document collection is huge, it is beneficial to perform a two-stage search.
First, a shortlist of candidate documents is selected. The shortlist is constructed by filtering documents using short binary codes
and a Hamming ball of some small radius, e.g. 5. Then, by using longer, possibly real-valued codes, precise ranking is performed
for documents within the Hamming ball. In this section we propose a neural model that can jointly learn short binary codes and longer,
real-valued representations. To this end, we added additional linear projection layer in the middle of the Binary PV-DBOW network,
as depicted in~\figref{pv-dbow-bin-projections}.
During training, projection weights are optimized alongside softmax weights and document embeddings.
During the inference stage, the projection weights are fixed.
We call this model Real-Binary Paragraph Vector Distributed Bag of Words (Real-Binary PV-DBOW).
Projection weights are randomly initialized by sampling from a uniform distribution.
We also tried initializing the projection weights using the method described in~\cite{glorot2010understanding},
but this did not improve results.

\begin{figure}[htb!]
  \centering
  \begin{subfigure}[b]{0.495\linewidth}
    \centering
    \includegraphics[width=\textwidth]{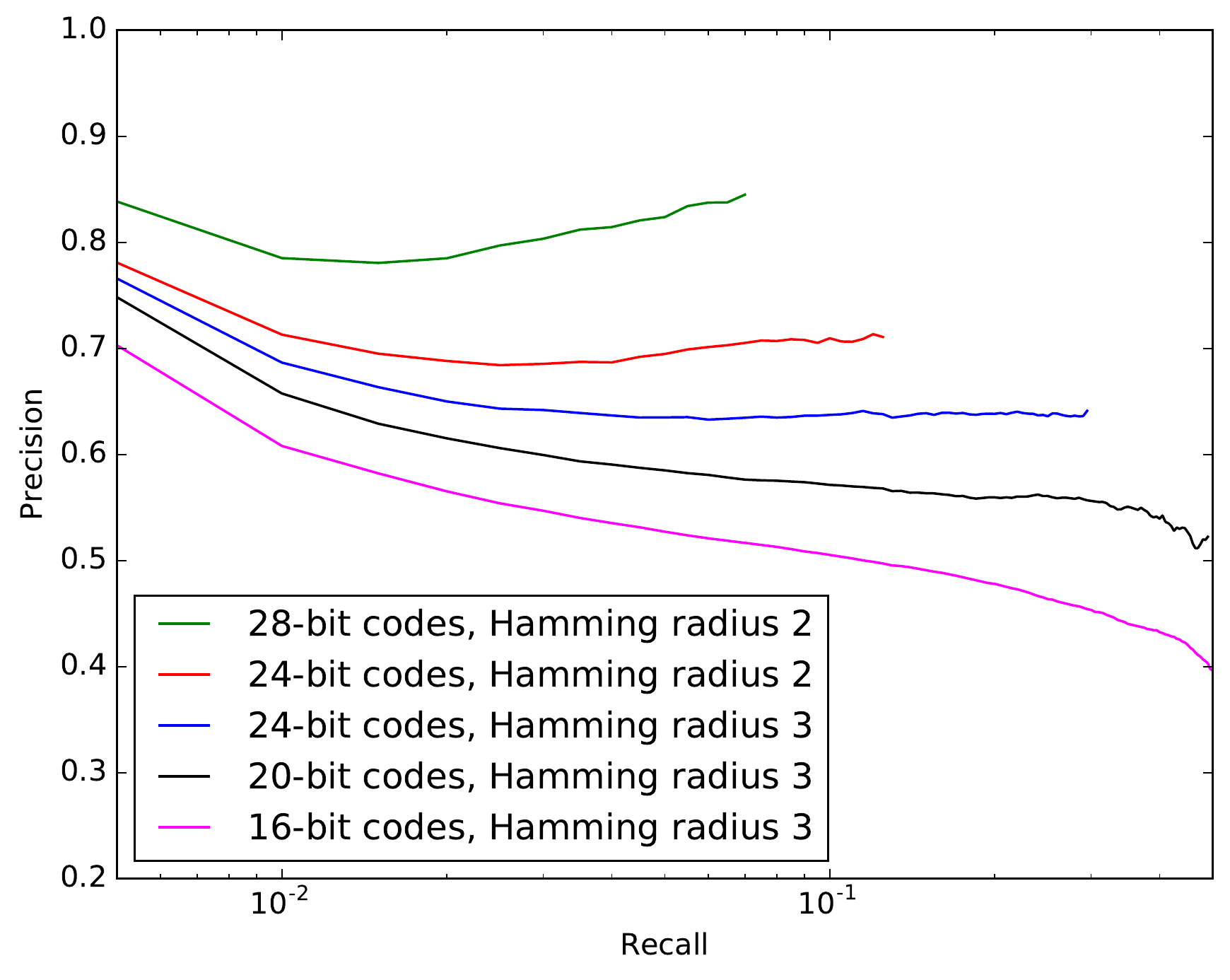}
    \caption{20 Newsgroups}
  \end{subfigure}
  \hfill
  \begin{subfigure}[b]{0.495\linewidth}
    \centering
    \includegraphics[width=\textwidth]{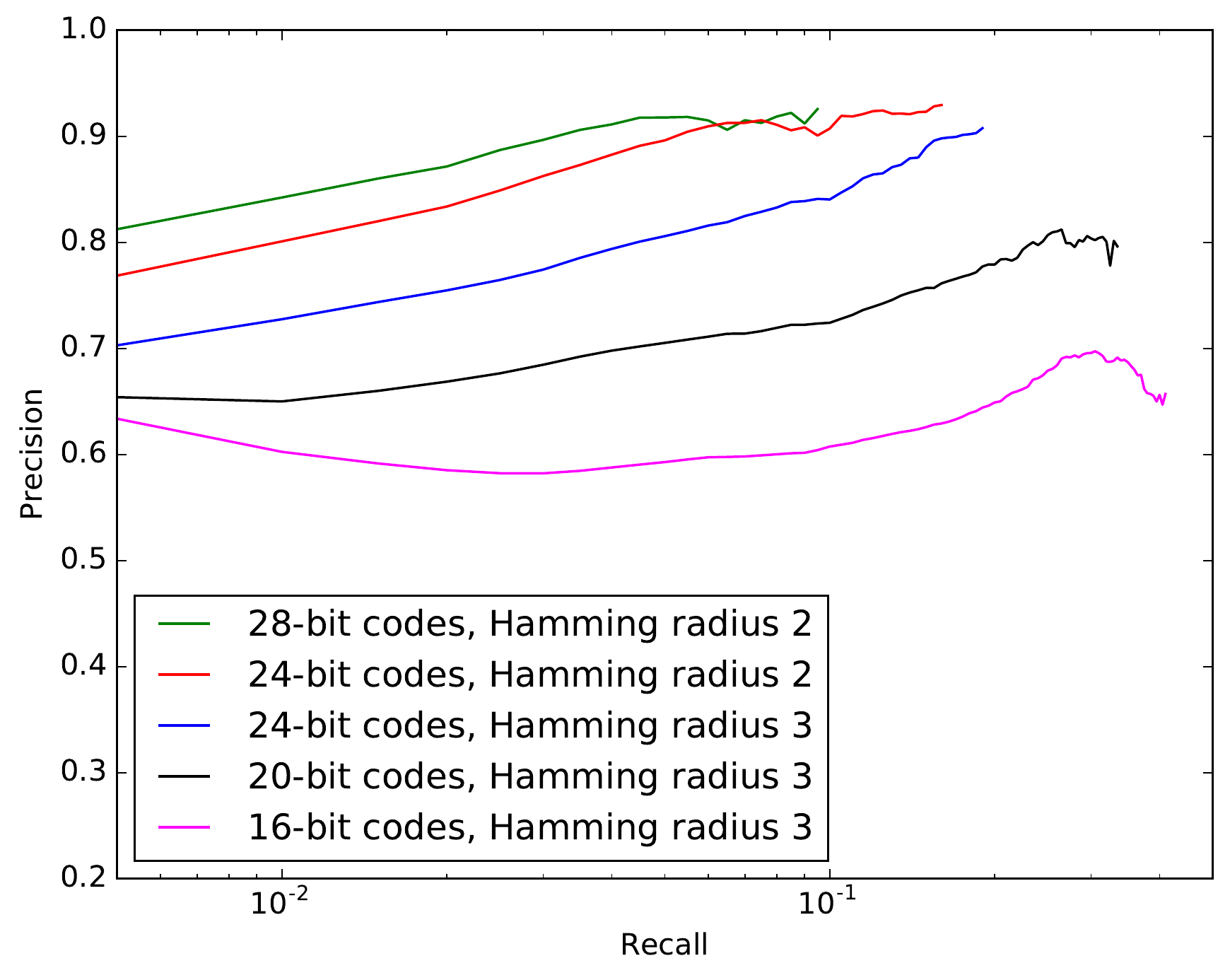}
    \caption{RCV1}
  \end{subfigure}
  \hfill
  \begin{subfigure}[b]{0.495\linewidth}
    \centering
    \includegraphics[width=\textwidth]{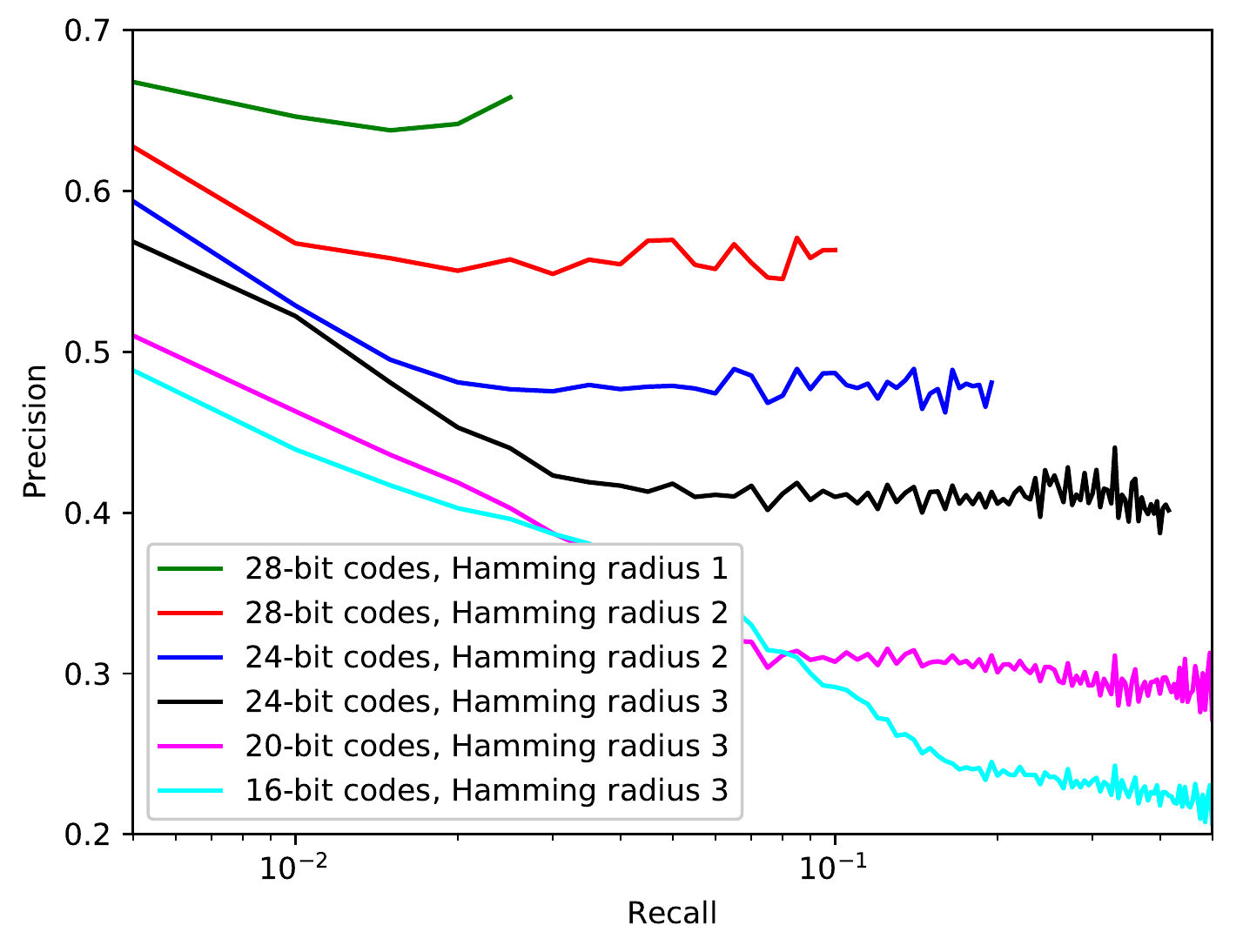}
    \caption{English Wikipedia}
  \end{subfigure}
  \longcaption{Information retrieval results for Real-Binary PV-DBOW model with 300-dimensional real-valued codes.}
              {To speed up the search, the search space is narrowed to documents with binary codes within small Hamming distance to the query.
               These documents are ranked accordingly to the cosine distance of real-valued representations.
               On the plot, precisions are drawn only for recall levels for which at least 10 documents were returned.}
  \label{fig:dsh_narrowing_search_space}
\end{figure}

We evaluated the Real-Binary PV-DBOW using the same text datasets as in the case of Binary PV-DBOW, namely 20 Newsgroups, RCV1 and English 
Wikipedia. The only metric that we report in this evaluation is NDCG@10. We do not report MAP, because MAP is estimated using the whole
recall range and the inherent trait of the filtering approach is that high recall levels may not be accessible.
In~\tabref{dsh_narrowing_search_space} and in~\figref{dsh_narrowing_search_space}
\begin{table}[htb]
  \centering
    \begin{tabular}{|c|c|c|c|c|c|c|c|}
      \hline
      \multirow{3}{*}{Code size} & \multirow{3}{*}{Radius} & \multicolumn{6}{c|}{NDCG@10}                                                                            \\ \cline{3-8}
                                 &                         & \multicolumn{2}{c|}{20 Newsgroups} & \multicolumn{2}{c|}{RCV1} & \multicolumn{2}{c|}{English Wikipedia} \\ \cline{3-8}
                                 &                         & A    & B                           & A    & B                  & A    & B                               \\ \hline
      \multirow{2}{*}{28}        & 1                       & 0.79 & 0.85                        & 0.77 & 0.85               & 0.66 & 0.7                             \\ \cline{2-1} \cline{3-8}
                                 & \multirow{2}{*}{2}      & 0.72 & 0.8                         & 0.73 & 0.81               & 0.62 & 0.65                            \\ \cline{1-1} \cline{3-8}
      \multirow{2}{*}{24}        &                         & 0.65 & 0.79                        & 0.7  & 0.76               & 0.56 & 0.59                            \\ \cline{2-2} \cline{3-8}
                                 & \multirow{3}{*}{3}      & 0.63 & 0.76                        & 0.69 & 0.74               & 0.5  & 0.55                            \\ \cline{1-1} \cline{3-8}
      20                         &                         & 0.6  & 0.73                        & 0.73 & 0.79               & 0.42 & 0.52                            \\ \cline{1-1} \cline{3-8}
      16                         &                         & 0.55 & 0.72                        & 0.72 & 0.79               & 0.39 & 0.52                            \\ \hline
    \end{tabular}
    \longcaption{Information retrieval results for the Real-Binary PV-DBOW model.}{Real valued representations have $300$ dimensions.
             (A) Binary codes are used for selecting documents within a given Hamming distance to the query and real-valued
             representations are used for ranking. (B) For comparison, variant~A was repeated with binary codes inferred using
             plain Binary PV-DBOW and a real-valued representation inferred using original PV-DBOW model.}
    \label{tab:dsh_narrowing_search_space}
\end{table}
we report performance of Real-Binary PV-DBOW for different Hamming ball radii. As can be seen, vast majority of documents that are left in
the Hamming ball after filtering are very relevant to the query. In particular, NDCG@10 values for Real-Binary PV-DBOW are higher than
results for plain Binary PV-DBOW. In addition to the main experiment, where both binary codes and real-valued representations are learned
by Real-Binary PV-DBOW (A), we also report results for variant (B), where binary codes are inferred using plain Binary PV-DBOW and
real-valued representations are inferred using original PV-DBOW model. The second variant gives slightly better results, but in the first
approach we get both real-valued and binary codes from the same model, thereby reducing the memory requirements.

\begin{figure}[hbt!]
  \centering
  \begin{subfigure}[b]{0.495\linewidth}
    \centering
    \includegraphics[width=\textwidth]{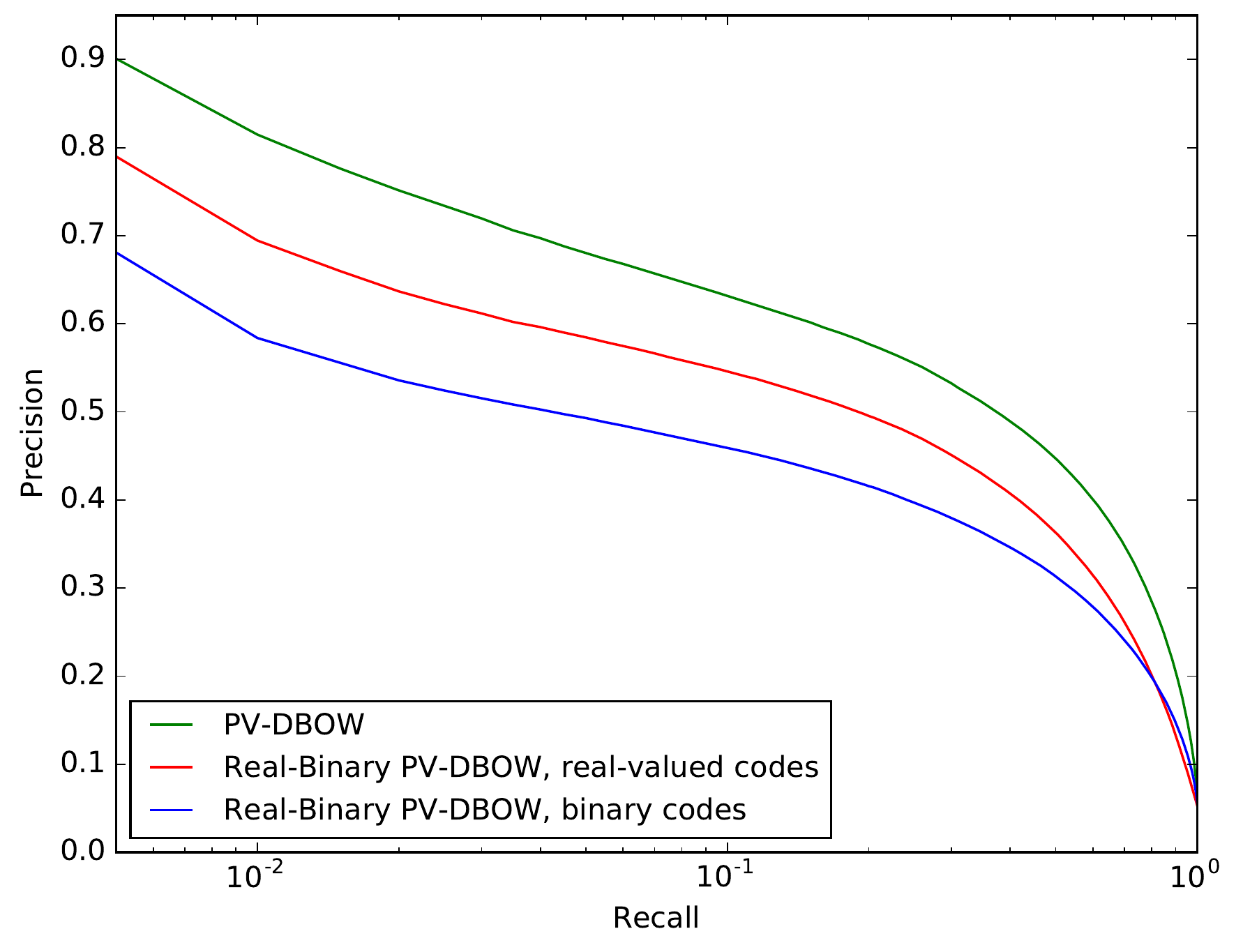}
    \caption{20 Newsgroups}
  \end{subfigure}
  \hfill
  \begin{subfigure}[b]{0.495\linewidth}
    \centering
    \includegraphics[width=\textwidth]{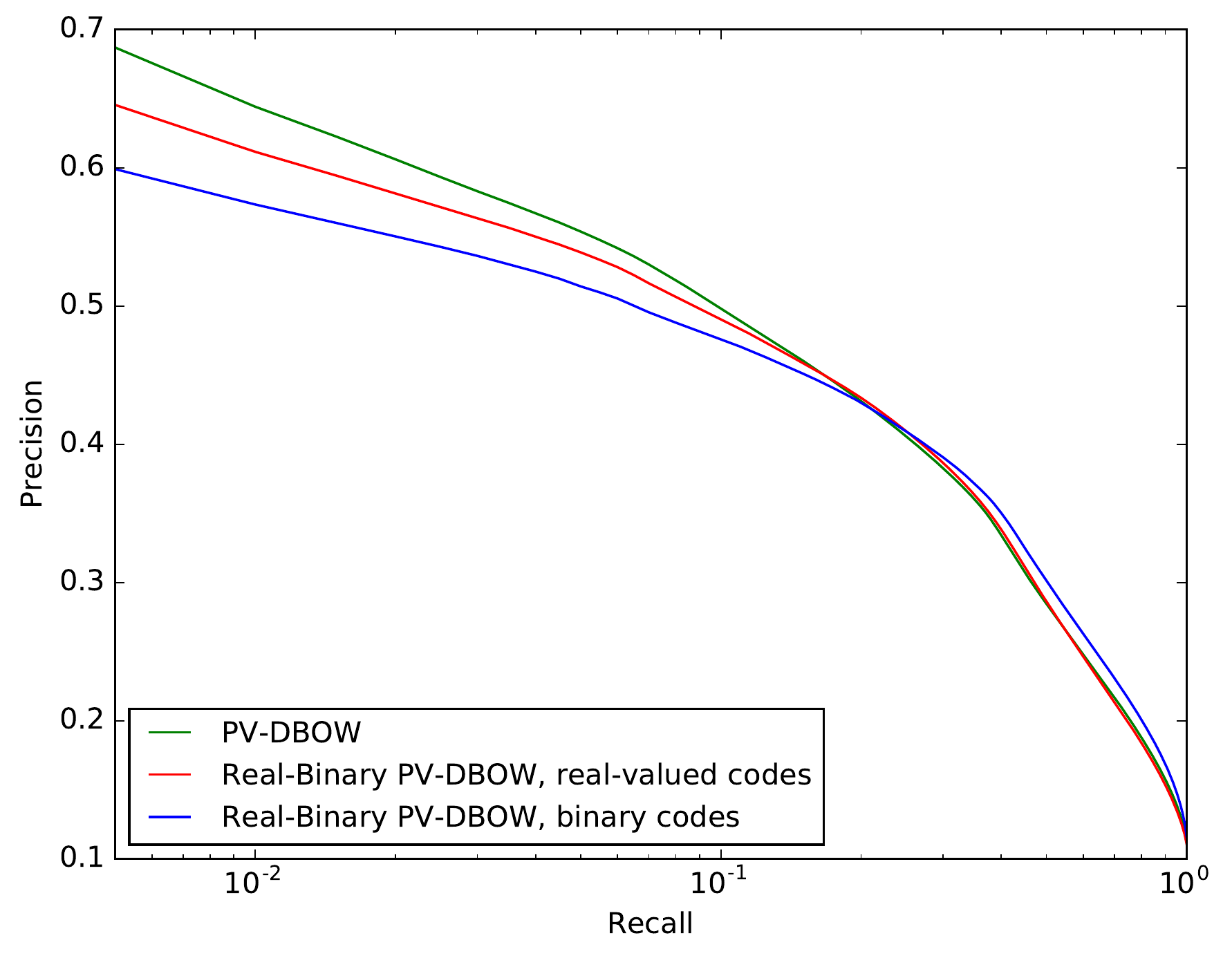}
    \caption{RCV1}
  \end{subfigure}
  \hfill
  \begin{subfigure}[b]{0.495\linewidth}
    \centering
    \includegraphics[width=\textwidth]{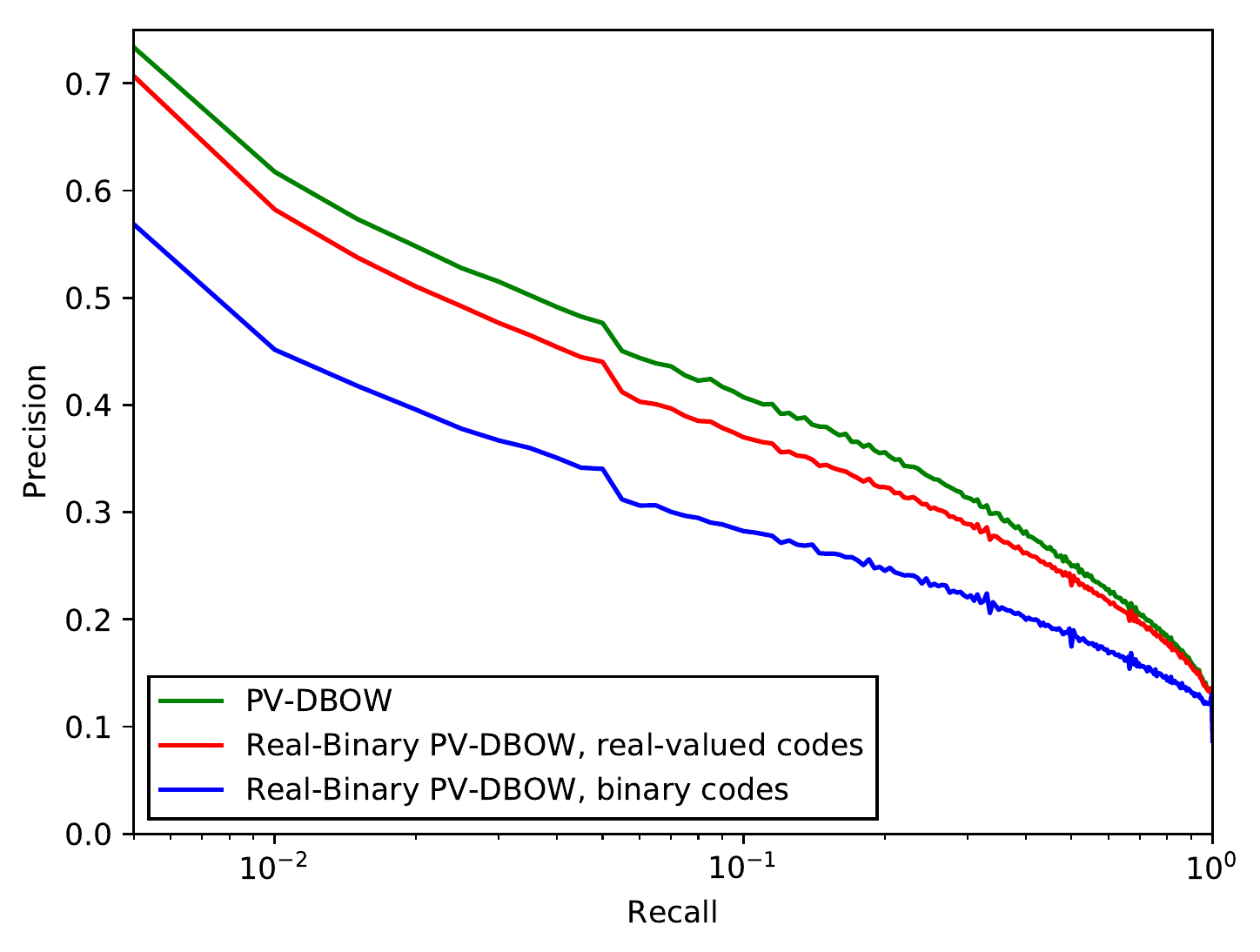}
    \caption{English Wikipedia}
  \end{subfigure}
  \longcaption{Performance comparison between real-valued codes learned by Real-Binary PV-DBOW (red curve) and original PV-DBOW (green curve).}
              {In both cases 300-dimensional real-valued codes were generated and cosine distance was used as a similarity measure.
               For reference, performance for binary codes learned by Real-Binary PV-DBOW (blue curve) is also presented. In this case
               the Hamming distance was used as a similarity measure.}
  \label{fig:dsh_projections_compare}
\end{figure}
\figref{dsh_projections_compare} compares accuracy of 300-dimensional real-valued codes learned by Real-Binary PV-DBOW with the codes
learned by the original PV-DBOW model. Since the two-tier model entails strong dimensionality reduction in the projection matrix,
codes generated by this model are slightly worse than those generated by the vanilla PV-DBOW.
However, the difference is relatively small, and we believe that it can be acceptable for searching with pre-filtering.

\section{Conclusions}

In this chapter we presented three novel neural network models for direct learning of low-dimensional binary codes of text documents.
Our models are simple extensions to the well-known Paragraph Vector Distributed Bag of Words and Paragraph Vector Distributed Memory neural
networks, collectively referred to as \emph{doc2vec}. A useful trait of the codes learned by our models is that, documents having the same,
or similar, topics end up having high probability of code collisions. This feature makes the models eligible for approximate nearest neighbor
search. In particular, we obtain state-of-the-art results in an information retrieval task on three benchmark datasets.

Binary PV-DM model accounts for the order of words in the documents. Nevertheless, it yields worse results than Binary PV-DBOW, which
disregards word ordering. This observation is consistent with previous studies on the original PV models. To improve results with PV-DM,
we tried to pre-train its word vectors on English Wikipedia. Unfortunately, this settings did not produce noticeably better results.


One of the use cases of approximate nearest neighbor search is document pre-filtering. The idea is to use binary codes to select some number
of candidates, and then apply a more precise search method, based on real-valued codes, for final search or ranking.
Since in this scenario we need both short binary codes and longer real-valued codes, we proposed a two-tier neural network model
that simultaneously learns both kinds of representations. Advantages of this model are shorter combined training
time and smaller memory requirements.

We demonstrated experimentally that the Binary Paragraph Vector models can be used for information retrieval. We believe that they can be used
to other tasks as well. One of the applications could be probabilistic data structures~(PDS). PDS are used to generate and store some
summarization of data. In contrast to deterministic data structures, like hash tables, they require significantly less memory
and, therefore, can be applied to much bigger datasets. They are often used for web analytics and big data mining, where they provide
approximations of some basic statistics, like cardinality. They are particularly useful for on-line and stream processing. Examples of PDS
are Bloom filter~\cite{bloom1970space}, HyperLogLog~\cite{durand2003loglog} or Count-Min Sketch~\cite{cormode2005improved}.
We believe that one promising research direction is to use Binary Paragraph Vector as a basis for probabilistic data structures applied to
stream or real-time processing.

\chapter{Probabilistic multi-sense word embeddings} \label{ch:disgram}

An inherent limitation of leading word embedding models~\cite{mikolov2013efficient, pennington2014glove, bojanowski2016enriching, wu2017starspace}
is that for each word, even ambiguous one, only one vector is learned. As we discuss in \sectionref{multi_sense_word_embeddings},
there are ongoing efforts to overcome this limitation. Currently these efforts focus on multi-sense word embedding models,
which are able to learn multiple vector representations per word.

In this chapter we propose a novel neural model for learning multi-sense word embeddings and we perform its thorough experimental evaluation.
On three out of four benchmark datasets our model gives better results in the word sense induction task than competing state-of-the-art solutions.
Also, in contrast to the previously proposed neural models, our solution is end-to-end differentiable and has an elegant probabilistic interpretation.

\section{Disambiguated Skip-gram model}

\begin{figure}[htb!]
  \centering
  \includegraphics[width=0.9\linewidth]{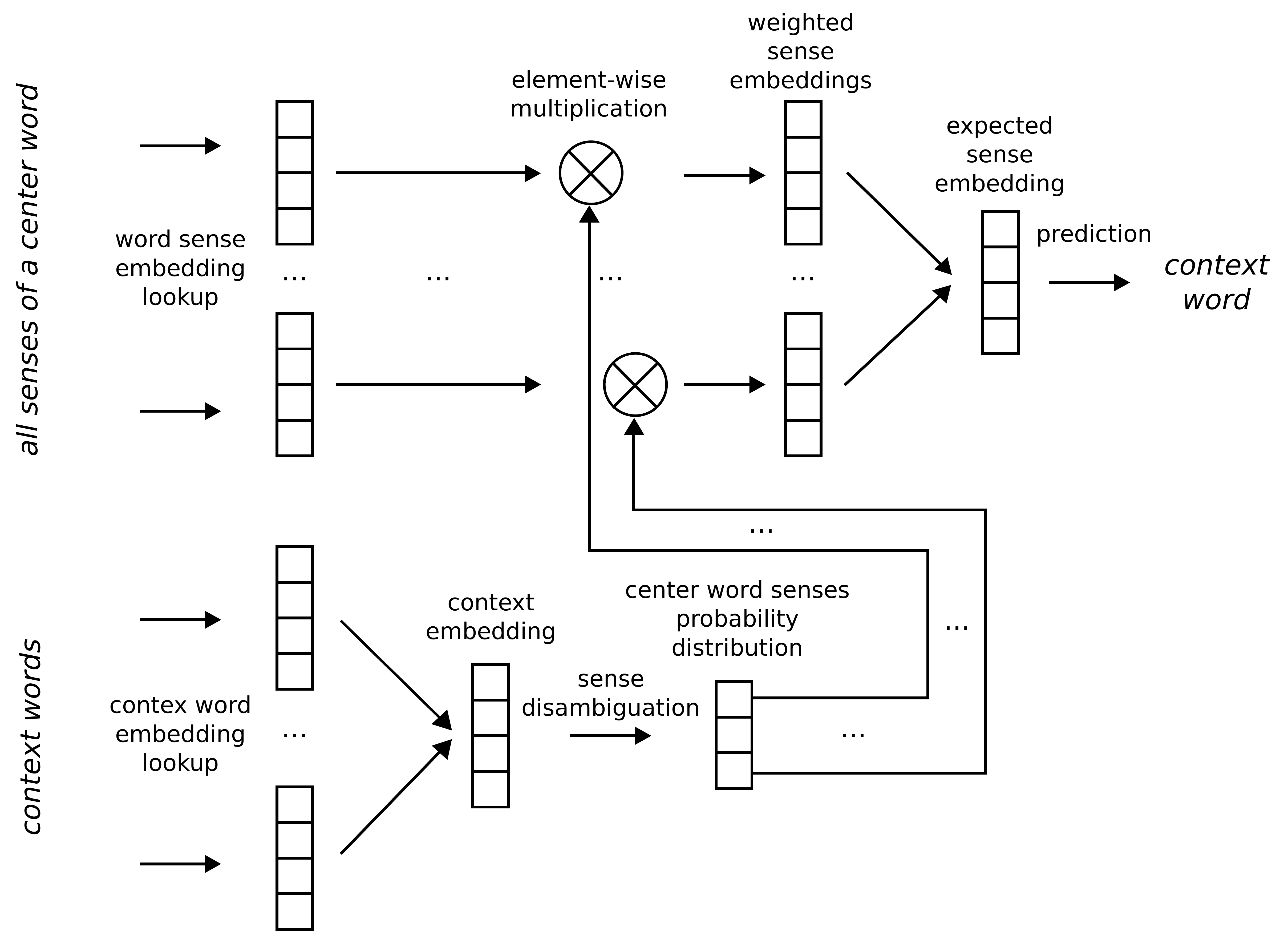}
  \caption{Disambiguated Skip-gram model.}
  \label{fig:disgram}
\end{figure}

We propose a neural model for learning multi-sense word embeddings that is a simple extension to the skip-gram model (\sectionref{word_embeddings}).
In skip-gram context words are conditioned on the center word (\equationref{skip_gram}).
Our model learns multi-sense word vectors by attempting to predict surrounding context words based on a given sense of the center word.
Therefore, for each vocabulary word $ w $ we define a fixed number~$ k $ of \emph{sense embedding vectors}:
$ \vect{v}_{w,s}, \ s = 1, \ldots, k $. We then define the probability of a context word $ c \in C_w $ given a sense $ s $ of the
center word $ w $ as:
\begin{equation} \label{eq:disgram_ctx}
P(c \mid w,s) = \frac{\me^{\vect{v}_{w,s}^\mathrm{T} \vect{u}_c}}{\sum_{c' \in V} \me^{\vect{v}_{w,s}^\mathrm{T} \vect{u}_{c'}}},
\end{equation}
where $ \vect{u}_c $ are output embedding vectors (defined as in the vanilla skip-gram). This parametrization has been used previously in
the MPSG model~\cite{tian2014probabilistic}. In contrast to previous works on multi-sense word embeddings, we also define a probabilistic
model for word senses. In particular, to predict the sense of the center word we condition this sense on the context. To this end, for each
vocabulary word~$ w $, we introduce $ k $ \emph{sense disambiguation vectors} $ \vect{q}_{w,s}, \ s = 1, \ldots, k $ and a \emph{context
embedding vector} $ \vect{r}_w $. We then define the probability that the center word $ w $ is in the sense $ s $ as:
\begin{equation} \label{eq:disgram_sense}
P(s \mid w,C_w) = \frac{\me^{\vect{q}_{w,s}^\mathrm{T} \bar{\vect{r}}_w}}{\sum_{j=1, \ldots, k} \me^{\vect{q}_{w,j}^\mathrm{T} \bar{\vect{r}}_w}},
\end{equation}
where $ \bar{\vect{r}}_w $ is a vector representation of the context of the word $ w $, defined as the average of the context embedding
vectors:
\begin{equation}
  \bar{\vect{r}}_w = \frac{1}{\#C_w}\sum_{c \in C_w} \vect{r}_c.
\end{equation}
Our approach can be seen as extending the skip-gram model by adding to it a sense disambiguation subnetwork (depicted in the bottom
of~\figref{disgram}).
Therefore, we call our model \emph{Disambiguated Skip-gram}.

Ideally, we would disambiguate center words during training by sampling senses from the categorical probability distribution defined
by~\equationref{disgram_sense}. Unfortunately, we would then not be able to backpropagate gradients through our model, because sampling
is not differentiable. Alternatively, we could simply ignore the sampling and predict the context words based on the expected sense embeddings,
i.e. maximize the log-probability:
\begin{equation} \label{eq:disgram_mean_objective}
\log P(C_w \mid \vect{e}_w) = \log \prod_{c \in C_w} \frac{\me^{\vect{e}_w^\mathrm{T} \vect{u}_c}}
                                                          {\sum_{c' \in V} \me^{\vect{e}_w^\mathrm{T} \vect{u}_{c'}}},
\end{equation}
where:
\begin{equation}
\vect{e}_w = \sum_{s=1}^{k} P(s \mid w,C_w) \vect{v}_{w,s}.
\end{equation}
However, modeling word senses with expected sense embedding vectors is not necessarily the best choice. In most cases a word in a given
context has one and only one meaning. For example, a word \emph{palm} in a given context is either a body part or a kind of tree.
A mean of the vector representations of these two senses does not represent a specific sense. This is the reason why samples form the
categorical distribution defined in~\equationref{disgram_sense} are more suitable for modeling senses of the center words.

As we mentioned above, we cannot backpropagate gradients through the samples from the categorical distribution defined
by~\equationref{disgram_sense}. Therefore we have to use some gradient estimator. There are many gradient estimators for stochastic
binary units (some of them are described in~\sectionref{bin_units_comparison}). Well-performing gradient estimators for backpropagation
through samples from a categorical distribution were proposed only recently, independently by two research teams, under names
\emph{Gumbel-Softmax} distribution~\cite{jang2016categorical} and \emph{concrete} distribution~\cite{maddison2016concrete}. The general
idea in these estimators is to express the samples via the \emph{Gumbel-Max trick} \cite{gumbel1954statistical} and then approximate the
$ \argmax $ operator in this trick with a softmax function with temperature hyperparameter. Concretely, given a categorical probability
distribution $ p_1,\ldots,p_k $ the Gumbel-Max trick expresses the samples as:
\begin{equation}
z = \argmax_{i=1,\ldots,k} (\log p_i + g_i),
\end{equation}
where $ g_i $ are i.i.d. samples form $ Gumbel (0 , 1) $ distribution. The Gumbel-Max distribution replaces the $ \argmax $ operator with
a continuous relaxation:
\begin{equation} \label{eq:softmax_gumbel}
\tilde{z}_i = \frac{e^\frac{\log(p_i) + g_i}{t}}{\sum_{j=1}^k e^\frac{\log(p_i) + g_i}{t}},~~\text{for}~~i=1..k,
\end{equation}
where $ t $ is the temperature hyperparameter. At the beginning of training the temperature is set to one.
During training the temperature decays toward zero, which consequently moves the softmax function towards the $ \argmax $ operator.
Note that gradients are not propagated to the Gumbel noise $ g_i $, as it is sampled from a fixed distribution. Overall, this gradient
estimator can be seen as a \emph{reparameterization trick} \cite{kingma2013auto} for an approximation to the categorical distribution.

To devise an effective training algorithm for Disambiguated Skip-gram we substitute samples from the categorical distribution over center
word senses with samples from the Gumbel-Max distribution over these senses. This means that we effectively apply softmax twice: first to
calculate the sense probabilities (\equationref{disgram_sense}) and then to approximate the $ \argmax $ operator in the Gumbel-Max trick.
We then use the resultant continuous samples $ \tilde{z}_{w,s} $ to calculate a \emph{relaxed sense embedding vector}:
\begin{equation}
\tilde{\vect{v}}_w = \sum_{s=1}^{k} \tilde{z}_{w,s} \vect{v}_{w,s}.
\end{equation}
Finally, we define the training objective for Disambiguated Skip-gram as the maximization of the log-probability:
\begin{equation} \label{eq:disgram_objective}
L = \log P(C_w \mid \tilde{\vect{v}}_w) = \log \prod_{c \in C_w} \frac{\me^{\tilde{\vect{v}}_w^\mathrm{T} \vect{u}_c}}
                                                                      {\sum_{c' \in V} \me^{\tilde{\vect{v}}_w^\mathrm{T} \vect{u}_{c'}}}.
\end{equation}
Note that when the temperature in the Gumbel-Max distribution $ t \to 0 $ the objective in~\equationref{disgram_objective} approaches
log-probability of the context words under true samples from the categorical distribution over the center word senses.

Our model could be considered similar to the multiple-sense skip-gram (MSSG) model proposed by Neelakantan et al.~\cite{neelakantan2015efficient}.
However, in contrast to our solution, they do not condition the word senses on the contexts in a probabilistic manner.
Instead, to select a proper sense for the center word they perform a hard cluster assignment.
This sense selection approach has no probabilistic interpretation.

\subsection{Regularization in Disambiguated Skip-gram}

In Disambiguated Skip-gram context embedding vectors $ \vect{r}_w $ perform a role similar to output embedding vectors $ \vect{u}_w $.
Specifically, they both represent context words, albeit in two different tasks: sense disambiguation and context word prediction, respectively.
In practice we simplify Disambiguated Skip-gram by tying these two set of vectors. This also gives us a performance benefit over
Neelakantan et al. MSSG algorithm. In particular, to build the context representation Neelakantan et al. use so-called \emph{global vectors},
which are single-sense word vectors learned in addition to multi-sense embeddings. In Disambiguated Skip-gram this role is fulfilled by
context embedding vectors. We get this vectors essentially `for free', because they are tied to output embeddings.
We also tried modeling contexts by averaging embeddings for all senses of all context words. However, this approach did not yielded
promising results.

In our model each word has its own set of sense disambiguation vectors $ \vect{q}_{w,s} $. In practical implementation we first look up
these vectors from a weights matrix and then we predict the sense of the center word using the looked-up vectors. One problem with this
approach is that the sense disambiguation vectors for rare words receive few updates during training and therefore are difficult to fit.
To overcome this limitation, we first pre-train Disambiguated Skip-gram with sense disambiguation vectors tied between all words from the
vocabulary. Furthermore, for simplicity we use the objective defined by~\equationref{disgram_mean_objective}
during this pre-training. After the pre-training phase we initialize separate sense disambiguation vectors for all vocabulary
words using the pre-trained values. This way we first learn multi-sense word embeddings based on a shared disambiguation model and then we
fine-tune those embeddings using a separate disambiguation model for each vocabulary word. We carry out this fine-tuning for multiple epochs.

Different words have different sense distributions and some senses occur more often then others. Therefore, it would be beneficial to have
a mechanism for controlling the granularity of the senses learned by Disambiguated Skip-gram. One way to achieve this is to control how
`confident' are the probability distributions predicted by the disambiguation subnetwork (\equationref{disgram_sense}).
A technique that can influence the confidence of the softmax activations was recently studied in~\cite{pereyra2017regularizing}.
The basic idea is to penalize distributions with high or low entropy (\equationref{entropy}).
We employ this idea by adding to the training objective an entropy $ S $ of the distribution defined in~\equationref{disgram_sense}
multiplied by a hyperparameter $ \gamma $. This hyperparameter, which we further call an \emph{entropy cost}, controls the strength of the
regularization. One caveat is that when softmax activations are close to zero, the entropy term can be undefined, since
$ \lim_{ x \to 0 } \log ( x ) = - \infty $. To prevent this, we add a small constant $ \epsilon $ to the softmax activations under
the logarithm, resulting in numerically stable expression:
\begin{equation} \label{eq:save_entropy}
L_e = - \gamma \sum_{s=1}^k P(s \mid w,C_w) \log( P(s \mid w,C_w) + \epsilon ),
\end{equation}
We employ a positive entropy cost in order to learn more fine-grained senses and a negative entropy cost in order to learn more balanced
distributions. In practice, we add a small negative entropy cost during pre-training and often use a positive entropy cost during fine-tuning.

Another regularization method that we use in Disambiguated Skip-gram is inspired by our earlier work~\cite{orthogonality2016}.
Therein we investigated the impact of regularization term that
encourages orthogonality between weight vectors on the performance of pre-trained deep neural networks. The goal of our research was to force
the network to learn more diverse sets of latent features in hidden layers, and consequently to obtain better results in common machine
learning tasks. In order to learn a broad set of features, weight vectors in hidden layers should point in different directions. We encouraged
orthogonality between latent features by introducing an additional term to the weight update rule, which penalizes parallel components
of the weight vectors:
\begin{equation}
\vect{w}_k \leftarrow \vect{w}_k - \frac{1}{n-1} \sum_{j \neq k} o_{kj} \mathbf{w}_j, ~~~ k = 1, \ldots, n,
\end{equation}
where $ n $ is a number of hidden units, $ o_{kj} $ is a non-orthogonality coefficient (defined below) and $ \mathbf{w}_j $ is the
$ j $-th weight vector. In the summation we skip the case where $ j = k $, because obviously we do not want to penalize the network for
having weight vectors parallel to themselves. We investigated three variants of the non-orthogonality coefficients, namely cosine of the
angle between the weight vectors:
\begin{equation} \label{eq:o_cosine}
o_{ij} = \frac{\mathbf{w}_i^\mathrm{T} \mathbf{w}_j}{\left \| \mathbf{w}_i \right \| \left \| \mathbf{w}_j \right \|} , ~~~ i \neq j,
\end{equation}
the dot product between the weight vectors:
\begin{equation}\label{eq:o_dot_prod}
o_{ij} = \mathbf{w}_i^\mathrm{T} \mathbf{w}_j , ~~~ i \neq j,
\end{equation}
and the Gram-Schmidt orthogonalization:
\begin{equation} \label{eq:o_gram_schmidt}
o_{ij} = \frac{\mathbf{w}_i^\mathrm{T} \mathbf{w}_j}{\mathbf{w}_i^\mathrm{T} \mathbf{w}_i} , ~~~ i \neq j.
\end{equation}
In~\cite{orthogonality2016}, we have chosen the first of those variants as our orthogonalization strategy and we
have shown that the resulting regularization term improves performance of pre-trained deep networks on image recognition and document
retrieval tasks.

Our idea in this work is to encourage different sense embedding vectors to point in different directions,
similar to the feature vectors in~\cite{orthogonality2016}. To this end, we penalize the \emph{Huber loss}~\cite{huber1964robust} over
dot products between sense embedding vectors:
\begin{equation} \label{eq:huber_loss}
L_H(x_w)=\left\{
\begin{matrix}
  \frac{1}{2} x_w^2   & \text{for}~~|x_w|\leq 1.0,  \\
  |x_w| - \frac{1}{2} & \text{otherwise,}
\end{matrix}\right.
\end{equation}
where:
\begin{equation}
x_w = \sum_{\substack{i, j = 1,\ldots,k \\ i \neq j}} \vect{v}_{w,i}^\mathrm{T} \vect{v}_{w,j}, ~~~ w \in V.
\end{equation}
We multiply the Hubber loss by a hyperparameter $ \delta $, which we further call \emph{parallel penalty}, and add it to the training
objective (\equationref{disgram_objective}). The advantage of the Huber loss over the classic squared error cost is that it grows less
rapidly and, therefore, is easier to optimize during training. Specifically, for inputs $ x_w \leq 1.0 $ the Huber loss is equal to the
squared loss, while for bigger inputs it becomes linear.

In addition to the regularizers introduced above, we tried regularizing the model by penalizing the weights in the disambiguation subnetwork
using standard $ L_1 $ and $ L_2 $ norms, but in our preliminary experiments we did not observed any benefit from doing so. Also, to further
improve the training we tried applying batch normalization ~\cite{ioffe2015batch} in the context prediction subnetwork, as well as in the
sense disambiguation subnetwork. We did not observe any tangible benefit from that.

\section{Experiments}

To assess the performance of Disambiguated Skip-gram we performed experiments similar to those reported in~\cite[section 5]{bartunov2016breaking}
and~\cite[section 6]{neelakantan2015efficient}. We trained Disambiguated Skip-gram with the Westbury Lab Wikipedia corpus, described in \sectionref{datasets_enwiki}.
The corpus was lowercased, stopwords were removed and numbers were converted to a unique token. We did not apply any stemming or lemmatization.
However, we include in the vocabulary only words that occurs at least 100 times in the corpus. This gives a vocabulary of $ 1.32 \times 10^5 $ words.

We trained Disambiguated Skip-gram in four basic variants: for 3 and 5 senses and for 50 and 300 dimensions.
Additionally, we repeated those tests with the entropy cost and the parallel penalty.
We apply one epoch of pre-training with a learning rate decaying from $ 1.0 $ to zero.
We then train the network for three fine-tuning epochs with learning rate equals to $ 0.1 $, $ 0.05 $ and $ 0.01 $, respectively.
As we mention above, during fine-tuning we replace vanilla softmax in the subnetwork with the Gumbel-Softmax distribution.
In the first epoch of fine-tuning we decay the Gumbel-Softmax temperature from $ 1.0 $ to $ 0.5 $.
In the following epochs we keep the temperature fixed at a $ 0.5 $.
We also tried letting the network to \emph{learn} the Gumbel-Softmax temperature by itself, i.e. considered it a model parameter,
effectively reducing the number of hyperparameters. Unfortunately, this did not lead to good performance.

After fine-tuning, we feed the corpus once more through our model and average the sense probability distributions separately for each word
from the vocabulary. This way, we estimate marginal sense probabilities for all words in the vocabulary. This allows us to prune sense
representation, i.e. senses with low marginal probability. This technique is especially useful when combined with high entropy cost $ \gamma $.

We implemented Disambiguated Skip-gram using TensorFlow~\cite{abadi2016tensorflow}.
All experiments were conducted on the HP Apollo XL730f Gen9 liquid cooled HPC machines
equipped with two Intel Xeon E5-2680v3 processors (24 cores) and 128 GB RAM.
We used Python version 2.7.5 and TensorFlow version 1.3.
It takes from 9 to 12 hours to pre-train the model, depending on dimensionality and the number of senses.
Duration of a single fine-tuning epoch vary from 10 to 18 hours.

\subsection{Qualitative evaluation}

In order to qualitatively evaluate our model, we selected 5 nearest neighbors (word senses) for 10 popular ambiguous words.
We used cosine similarity to build the nearest neighbors lists. If multiple senses of some word appear on the nearest neighbors list,
we merge them into one neighbor. We also merge simple variants of words (e.g. plurals).
Results for four different values of the entropy cost $ \gamma $ are presented in~\tabref{disgram_neighbors_1}
\newcommand{\disgramNNcaption}[1]{
\longcaption{Marginal probabilities and nearest neighbors for selected words from the vocabulary.}
            {Codes learned with a 3-sense, 300-dimensional Disambiguated Skip-gram model trained with different values of the entropy cost $ \gamma $.
             The neighbors are selected are presented only for senses with a marginal probability higher than $ 0.05 $. (Part #1)}
}
\newcommand{\disgramNNheader}{
\hline
\multirow{2}{*}{Word}  & \multirow{2}{*}{$ \gamma $} & \multicolumn{2}{|c|}{Sense 1}                     & \multicolumn{2}{|c|}{Sense 2}                        & \multicolumn{2}{|c|}{Sense 3}                      \\ \cline{3-8}
                       &                             & $ P $                 & Nearest neighbors         & $ P $                 & Nearest neighbors            & $ P $                 & Nearest neighbors          \\ \hline \hline
}
\begin{table}[!p]
  \scriptsize
  \centering
  \setlength{\tabcolsep}{3pt}
    \begin{tabular}{|c|c|c|c|c|c|c|c|}
      \disgramNNheader
      \multirow{8}{*}{apple} & \multirow{2}{*}{0.0}  & \multirow{2}{*}{0.28} & wozniak macworld          & \multirow{2}{*}{0.46} & macintosh imac iigs          & \multirow{2}{*}{0.26} & strawberry peach           \\
                             &                       &                       & macintosh ipod sculley    &                       & iie iic                      &                       & raspberry blueberry plum   \\ \cline{2-8}
                             & \multirow{2}{*}{0.25} & \multirow{2}{*}{0.25} & wozniak blackberry        & \multirow{2}{*}{0.64} & macintosh iigs imac          & \multirow{2}{*}{0.11} & peach pecan persimmon      \\
                             &                       &                       & tomato potato popcorn     &                       & iie iic                      &                       & prune blueberry            \\ \cline{2-8}
                             & \multirow{2}{*}{0.5}  & \multirow{2}{*}{0.01} &                           & \multirow{2}{*}{0.95} & macintosh blackberry         & \multirow{2}{*}{0.04} &                            \\
                             &                       &                       &                           &                       & iigs imac apricot            &                       &                            \\ \cline{2-8}
                             & \multirow{2}{*}{0.75} & \multirow{2}{*}{0}    &                           & \multirow{2}{*}{1}    & macintosh blackberry         & \multirow{2}{*}{0}    &                            \\
                             &                       &                       &                           &                       & iigs imac apricot            &                       &                            \\ \hline \hline
      \multirow{8}{*}{fox}   & \multirow{2}{*}{0.0}  & \multirow{2}{*}{0.52} & nbc cbs network           & \multirow{2}{*}{0.24} & badger wolf coyote           & \multirow{2}{*}{0.25} & miller allen plummer       \\
                             &                       &                       & syndication espn          &                       & weasel marten                &                       & crowe buck                 \\ \cline{2-8}
                             & \multirow{2}{*}{0.25} & \multirow{2}{*}{0.6}  & nbc cbs abc               & \multirow{2}{*}{0.18} & badger squirrel weasel       & \multirow{2}{*}{0.22} & miller allen terry         \\
                             &                       &                       & syndication network       &                       & raccoon marten               &                       & russell soper              \\ \cline{2-8}
                             & \multirow{2}{*}{0.5}  & \multirow{2}{*}{0.68} & cbs nbc abc               & \multirow{2}{*}{0.14} & badger marten raccoon        & \multirow{2}{*}{0.18} & allen russell miller       \\
                             &                       &                       & cable colmes              &                       & beaver mink                  &                       & turner berry               \\ \cline{2-8}
                             & \multirow{2}{*}{0.75} & \multirow{2}{*}{0.86} & cbs nbc abc               & \multirow{2}{*}{0.09} & vulpes porcupine raccoon     & \multirow{2}{*}{0.06} & raccoon sauk hammond       \\
                             &                       &                       & colmes wttg               &                       & marten mink                  &                       & mendota meskwaki           \\ \hline \hline
      \multirow{8}{*}{net}   & \multirow{2}{*}{0.0}  & \multirow{2}{*}{0.26} & crossbar puck lob         & \multirow{2}{*}{0.32} & trawl streamline maximising  & \multirow{2}{*}{0.41} & ebitda earnings annualized \\
                             &                       &                       & sliothar offside          &                       & minimises counteracts        &                       & taxable depreciation       \\ \cline{2-8}
                             & \multirow{2}{*}{0.25} & \multirow{2}{*}{0.24} & crossbar puck lob         & \multirow{2}{*}{0.33} & trawl minimises maximising   & \multirow{2}{*}{0.43} & ebitda annualized jpy      \\
                             &                       &                       & offside dribbled          &                       & streamlines stickiness       &                       & deadweight gni             \\ \cline{2-8}
                             & \multirow{2}{*}{0.5}  & \multirow{2}{*}{0.23} & crossbar puck lob         & \multirow{2}{*}{0.77} & ebitda deadweight isk        & \multirow{2}{*}{0}    &                            \\
                             &                       &                       & header dribbled           &                       & annualized deducting         &                       &                            \\ \cline{2-8}
                             & \multirow{2}{*}{0.75} & \multirow{2}{*}{0}    &                           & \multirow{2}{*}{1}    & ebitda deducting deadweight  & \multirow{2}{*}{0}    &                            \\
                             &                       &                       &                           &                       & offsetting isk               &                       &                            \\ \hline \hline
      \multirow{8}{*}{rock}  & \multirow{2}{*}{0.0}  & \multirow{2}{*}{0.41} & band indie punk           & \multirow{2}{*}{0.34} & punk rockabilly pop          & \multirow{2}{*}{0.26} & boulder quartzite          \\
                             &                       &                       & alternative supergroup    &                       & psychedelia funk             &                       & granite sandstone basalt   \\ \cline{2-8}
                             & \multirow{2}{*}{0.25} & \multirow{2}{*}{0.7}  & alternative glam progre-  & \multirow{2}{*}{0.17} & boulder basalt outcrop       & \multirow{2}{*}{0.13} & granite bluff pine         \\
                             &                       &                       & ssive indie psychedelic   &                       & quartzite cliffs             &                       & pigeon ledge               \\ \cline{2-8}
                             & \multirow{2}{*}{0.5}  & \multirow{2}{*}{0.72} & alternative punk indie    & \multirow{2}{*}{0.17} & basalt boulders quartzite    & \multirow{2}{*}{0.11} & pine bluff eagle           \\
                             &                       &                       & glam progressive          &                       & cliff outcropping            &                       & pigeon turtle              \\ \cline{2-8}
                             & \multirow{2}{*}{0.75} & \multirow{2}{*}{0.73} & alternative punk glam     & \multirow{2}{*}{0.18} & basalt boulders quartzite    & \multirow{2}{*}{0.1}  & big little sandy           \\
                             &                       &                       & indie psychobilly         &                       & outcrop cliffs               &                       & alum butte                 \\ \hline \hline
      \multirow{8}{*}{plant} & \multirow{2}{*}{0.0}  & \multirow{2}{*}{0.18} & factory botanical labo-   & \multirow{2}{*}{0.45} & flowering perennial          & \multirow{2}{*}{0.38} & refinery smelter petroche- \\
                             &                       &                       & ratory farm nurseryman    &                       & shrub grass fungus           &                       & mical processing factory   \\ \cline{2-8}
                             & \multirow{2}{*}{0.25} & \multirow{2}{*}{0.08} & weed planted shed         & \multirow{2}{*}{0.46} & flowering grasses shrub      & \multirow{2}{*}{0.45} & refinery factory megawatt  \\
                             &                       &                       & grinder laboratory        &                       & fungus herbaceous            &                       & smelter cogeneration       \\ \cline{2-8}
                             & \multirow{2}{*}{0.5}  & \multirow{2}{*}{0.04} &                           & \multirow{2}{*}{0.48} & flowering shrub grasses      & \multirow{2}{*}{0.48} & refinery factory smelter   \\
                             &                       &                       &                           &                       & herbaceous fungus            &                       & megawatt sellafield        \\ \cline{2-8}
                             & \multirow{2}{*}{0.75} & \multirow{2}{*}{0.02} &                           & \multirow{2}{*}{0.98} & flowering woody              & \multirow{2}{*}{0}    &                            \\
                             &                       &                       &                           &                       & herbaceous shrub aster       &                       &                            \\ \hline
    \end{tabular}
  \disgramNNcaption{I}
  \label{tab:disgram_neighbors_1}
\end{table}
and~\tabref{disgram_neighbors_2}.
\begin{table}[!p]
  \scriptsize
  \centering
  \setlength{\tabcolsep}{3pt}
    \begin{tabular}{|c|c|c|c|c|c|c|c|}
      \disgramNNheader
      \multirow{8}{*}{bank}  & \multirow{2}{*}{0.0}  & \multirow{2}{*}{0.40} & savings hsbc citibank     & \multirow{2}{*}{0.35} & credit loans depositors      & \multirow{2}{*}{0.25} & mouth confluence           \\
                             &                       &                       & barclays lloyds           &                       & fdic lending                 &                       & opposite upstream side     \\ \cline{2-8}
                             & \multirow{2}{*}{0.25} & \multirow{2}{*}{0.41} & savings hsbc citibank     & \multirow{2}{*}{0.37} & depositors fdic credit       & \multirow{2}{*}{0.22} & confluence kolpa river     \\
                             &                       &                       & lloyds barclays           &                       & lending landsbanki           &                       & opposite mouth             \\ \cline{2-8}
                             & \multirow{2}{*}{0.5}  & \multirow{2}{*}{0.1}  & cashier savings citibank  & \multirow{2}{*}{0.3}  & depositor fdic lenders       & \multirow{2}{*}{0.6}  & savings hsbc citibank      \\
                             &                       &                       & robbers jpmorgan          &                       & liquidity unsecured          &                       & jpmorgan bancorp           \\ \cline{2-8}
                             & \multirow{2}{*}{0.75} & \multirow{2}{*}{0.07} & bookie bookmaker heist    & \multirow{2}{*}{0.19} & depositors overdraft un-     & \multirow{2}{*}{0.75} & hsbc citibank jpmorgan     \\
                             &                       &                       & holdup cashier            &                       & secured issuer borrowers     &                       & lloyds icici               \\ \hline \hline
      \multirow{8}{*}{mouse} & \multirow{2}{*}{0.0}  & \multirow{2}{*}{0.47} & mickey rabbit goofy       & \multirow{2}{*}{0.35} & cursor joystick trackball    & \multirow{2}{*}{0.19} & rodent vole shrew          \\
                             &                       &                       & cat porky                 &                       & touchpad touchscreen         &                       & pygmy rat                  \\ \cline{2-8}
                             & \multirow{2}{*}{0.25} & \multirow{2}{*}{0.49} & rabbit goofy cat          & \multirow{2}{*}{0.5}  & rat mice rodent              & \multirow{2}{*}{0.01} &                            \\
                             &                       &                       & porky tigger              &                       & mus elegans                  &                       &                            \\ \cline{2-8}
                             & \multirow{2}{*}{0.5}  & \multirow{2}{*}{0.49} & rabbit goofy porky        & \multirow{2}{*}{0.51} & mice rodent rat              & \multirow{2}{*}{0}    &                            \\
                             &                       &                       & tigger tweety             &                       & mus elegans                  &                       &                            \\ \cline{2-8}
                             & \multirow{2}{*}{0.75} & \multirow{2}{*}{0.48} & goofy porky tweety        & \multirow{2}{*}{0.52} & rodent mice mus              & \multirow{2}{*}{0}    &                            \\
                             &                       &                       & mickey tigger             &                       & rat musculus                 &                       &                            \\ \hline \hline
      \multirow{8}{*}{table} & \multirow{2}{*}{0.0}  & \multirow{2}{*}{0.38} & foosball carom lang-      & \multirow{2}{*}{0.4}  & sortable column lookup       & \multirow{2}{*}{0.22} & sortable list alphabe-     \\
                             &                       &                       & uishing pool slipping     &                       & hashed tray                  &                       & tical descending brackets  \\ \cline{2-8}
                             & \multirow{2}{*}{0.25} & \multirow{2}{*}{0.36} & foosball ept languishing  & \multirow{2}{*}{0.48} & sortable column lookup       & \multirow{2}{*}{0.16} & sortable descending list   \\
                             &                       &                       & pool leaderboard          &                       & tray hashed                  &                       & alphabetically please      \\ \cline{2-8}
                             & \multirow{2}{*}{0.5}  & \multirow{2}{*}{0.31} & standings ept foosball    & \multirow{2}{*}{0.66} & sortable tray column         & \multirow{2}{*}{0.03} &                            \\
                             &                       &                       & leaderboard ittf          &                       & chairs buckets               &                       &                            \\ \cline{2-8}
                             & \multirow{2}{*}{0.75} & \multirow{2}{*}{0.26} & standings relegation      & \multirow{2}{*}{0.74} & sortable tray tallies        & \multirow{2}{*}{0}    &                            \\
                             &                       &                       & ittf ept creditable       &                       & lists chairs                 &                       &                            \\ \hline \hline
      \multirow{8}{*}{light} & \multirow{2}{*}{0.0}  & \multirow{2}{*}{0.3}  & shade reflection ibo      & \multirow{2}{*}{0.42} & sunlight illumination        & \multirow{2}{*}{0.28} & heavy lvt lrv lynx         \\
                             &                       &                       & radiant shines            &                       & bright refracted luminous    &                       & wheeled                    \\ \cline{2-8}
                             & \multirow{2}{*}{0.25} & \multirow{2}{*}{0.28} & lighter ibo illuminates   & \multirow{2}{*}{0.44} & sunlight bright dichroic     & \multirow{2}{*}{0.28} & heavy lvt lynx             \\
                             &                       &                       & shade wbu                 &                       & refracted phosphors          &                       & lrv searchlights           \\ \cline{2-8}
                             & \multirow{2}{*}{0.5}  & \multirow{2}{*}{0.24} & flyweight ibo welterwei-  & \multirow{2}{*}{0.48} & refracted sunlight dichroic  & \multirow{2}{*}{0.29} & heavy lvt lrv              \\
                             &                       &                       & ght wbu bantamweight      &                       & luminous phosphors           &                       & shermans unarmoured        \\ \cline{2-8}
                             & \multirow{2}{*}{0.75} & \multirow{2}{*}{0.16} & flyweight featherweight   & \multirow{2}{*}{0.55} & refracted illuminates        & \multirow{2}{*}{0.29} & lvt heavy shermans         \\
                             &                       &                       & welterweight ibo          &                       & sunlight glows shine         &                       & turreted lrv               \\ \hline \hline
      \multirow{8}{*}{core}  & \multirow{2}{*}{0.0}  & \multirow{2}{*}{0.21} & backbone nucleus co-      & \multirow{2}{*}{0.4}  & microarchitecture threading  & \multirow{2}{*}{0.39} & integrates competencies    \\
                             &                       &                       & hesive interplay nexus    &                       & opteron ultrasparc merom     &                       & focus components combines  \\ \cline{2-8}
                             & \multirow{2}{*}{0.25} & \multirow{2}{*}{0.17} & backbone integral exp-    & \multirow{2}{*}{0.42} & microarchitecture threading  & \multirow{2}{*}{0.41} & combine competencies com-  \\
                             &                       &                       & anding interplay newer    &                       & opteron xeon merom           &                       & ponent integral structured \\ \cline{2-8}
                             & \multirow{2}{*}{0.5}  & \multirow{2}{*}{0.13} & assembled integral foun-  & \multirow{2}{*}{0.46} & microarchitecture threading  & \multirow{2}{*}{0.4}  & competencies component     \\
                             &                       &                       & ding dynamic comprised    &                       & opteron merom penryn         &                       & integral envisions         \\ \cline{2-8}
                             & \multirow{2}{*}{0.75} & \multirow{2}{*}{0.09} & comprised founding proto  & \multirow{2}{*}{0.77} & microarchitecture merom      & \multirow{2}{*}{0.14} & incorporating fringe       \\
                             &                       &                       & sugarcubes replacements   &                       & penryn conroe opteron        &                       & absorbing gastown nucleus  \\ \hline
    \end{tabular}
  \disgramNNcaption{II}
  \label{tab:disgram_neighbors_2}
\end{table}
In the tables we also include marginal probabilities of senses. As we can see, for higher values of the entropy cost $ \gamma $, marginal probabilities
are less balanced. In those cases, there is one or two dominating senses, while other senses are pruned.

To shed light on relations between senses of the test words we also projected embedding vectors to $ \mathbb{R}^2 $
and pictured placed nearest neighbors on a plane. These results are presented in~\figref{disgram_neighbors_pca_1}
\newcommand{\disgramVisCaption}[1]{
\longcaption{Visualization of the nearest neighbors for selected words from the vocabulary.}
            {Codes learned with a 3-sense, 300-dimensional Disambiguated Skip-gram model. Dimensionality reduced with Principal Components
             Analysis. The neighbors were selected in the original space using the cosine distance. (Part #1)}
}
\begin{figure}[!p]
  \centering
  \begin{subfigure}[b]{0.495\linewidth}
    \centering
    \includegraphics[width=\textwidth]{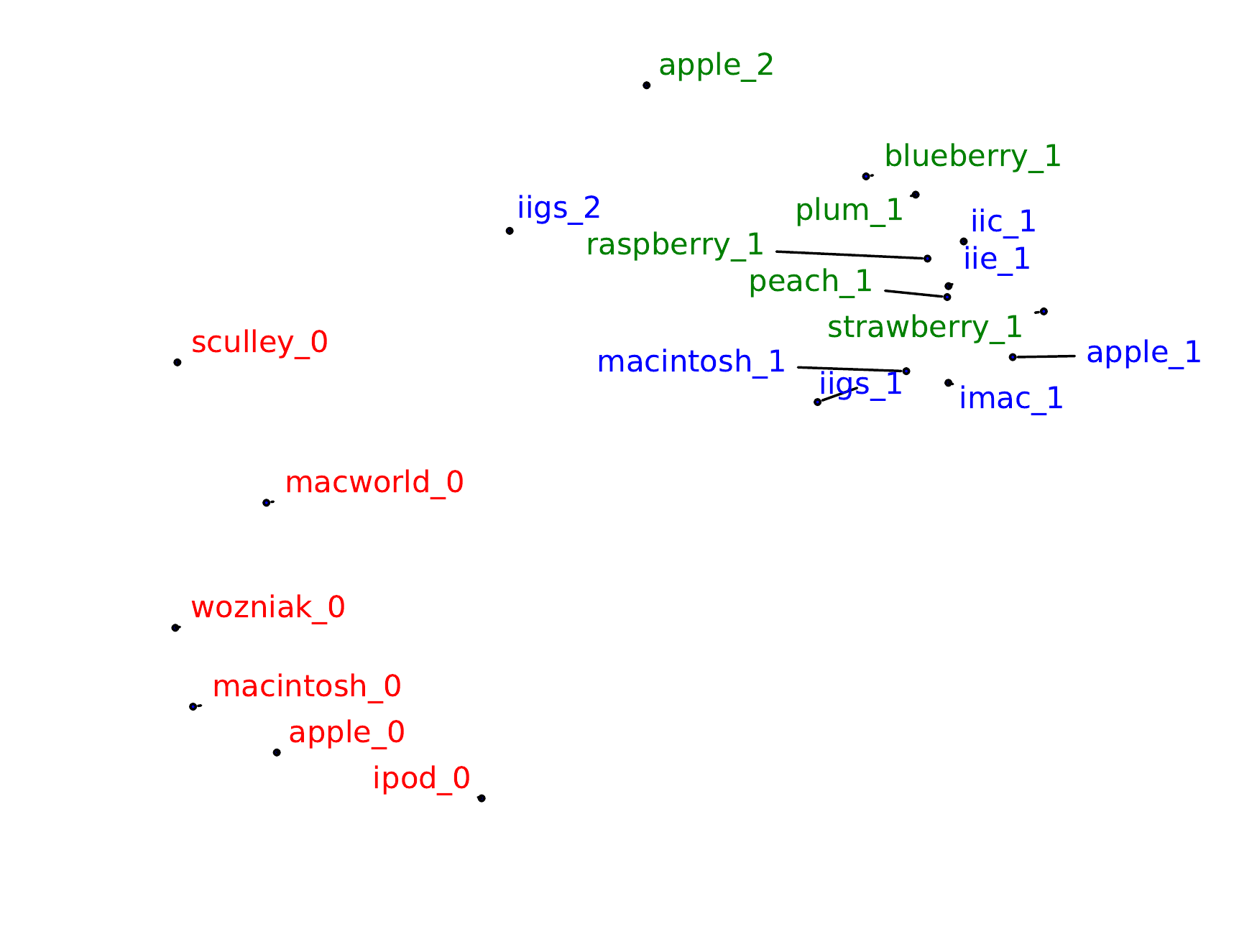}
    \caption{apple}
  \end{subfigure}
  \hfill
  \begin{subfigure}[b]{0.495\linewidth}
    \centering
    \includegraphics[width=\textwidth]{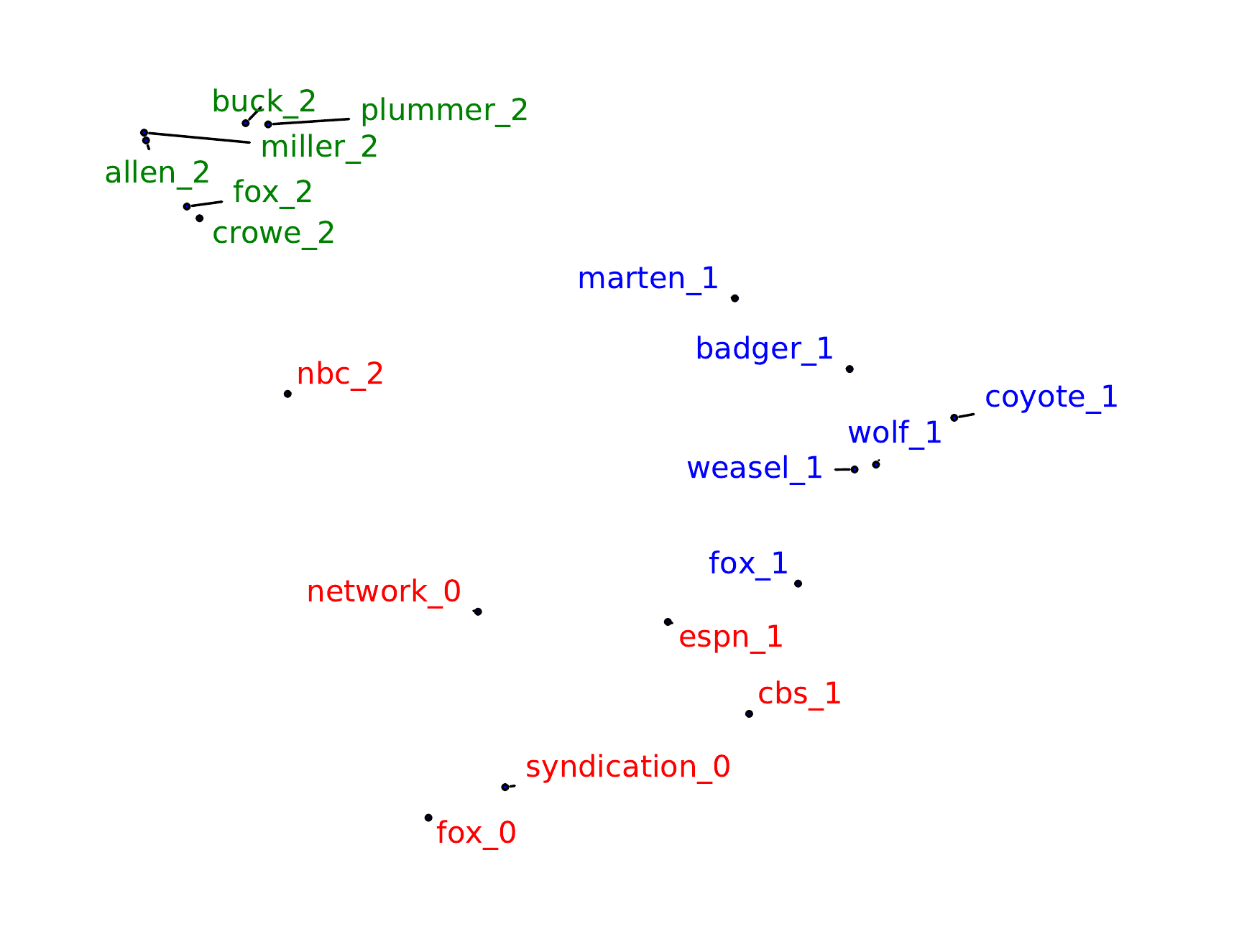}
    \caption{fox}
  \end{subfigure}
  \hfill
  \begin{subfigure}[b]{0.495\linewidth}
    \centering
    \includegraphics[width=\textwidth]{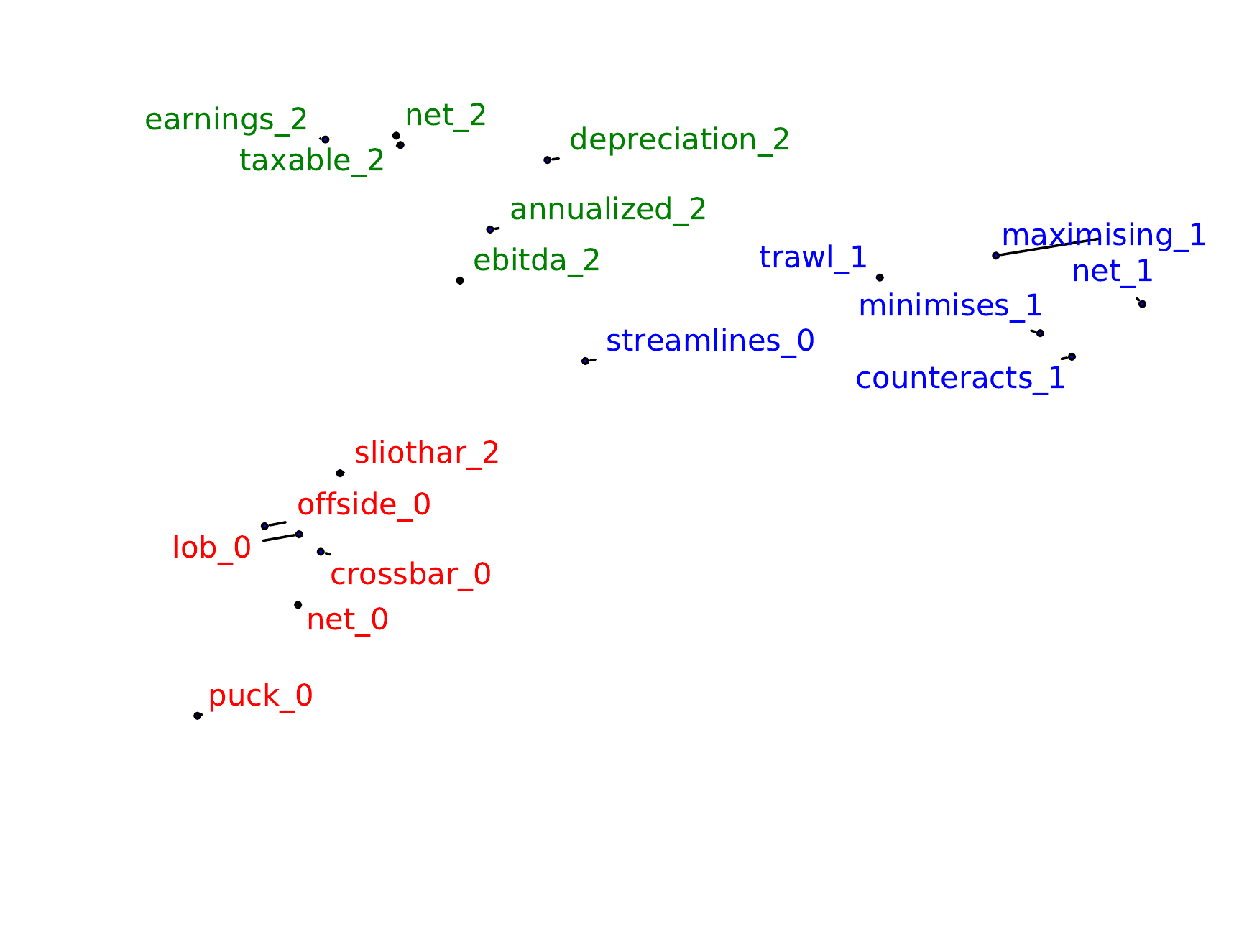}
    \caption{net}
  \end{subfigure}
  \hfill
  \begin{subfigure}[b]{0.495\linewidth}
    \centering
    \includegraphics[width=\textwidth]{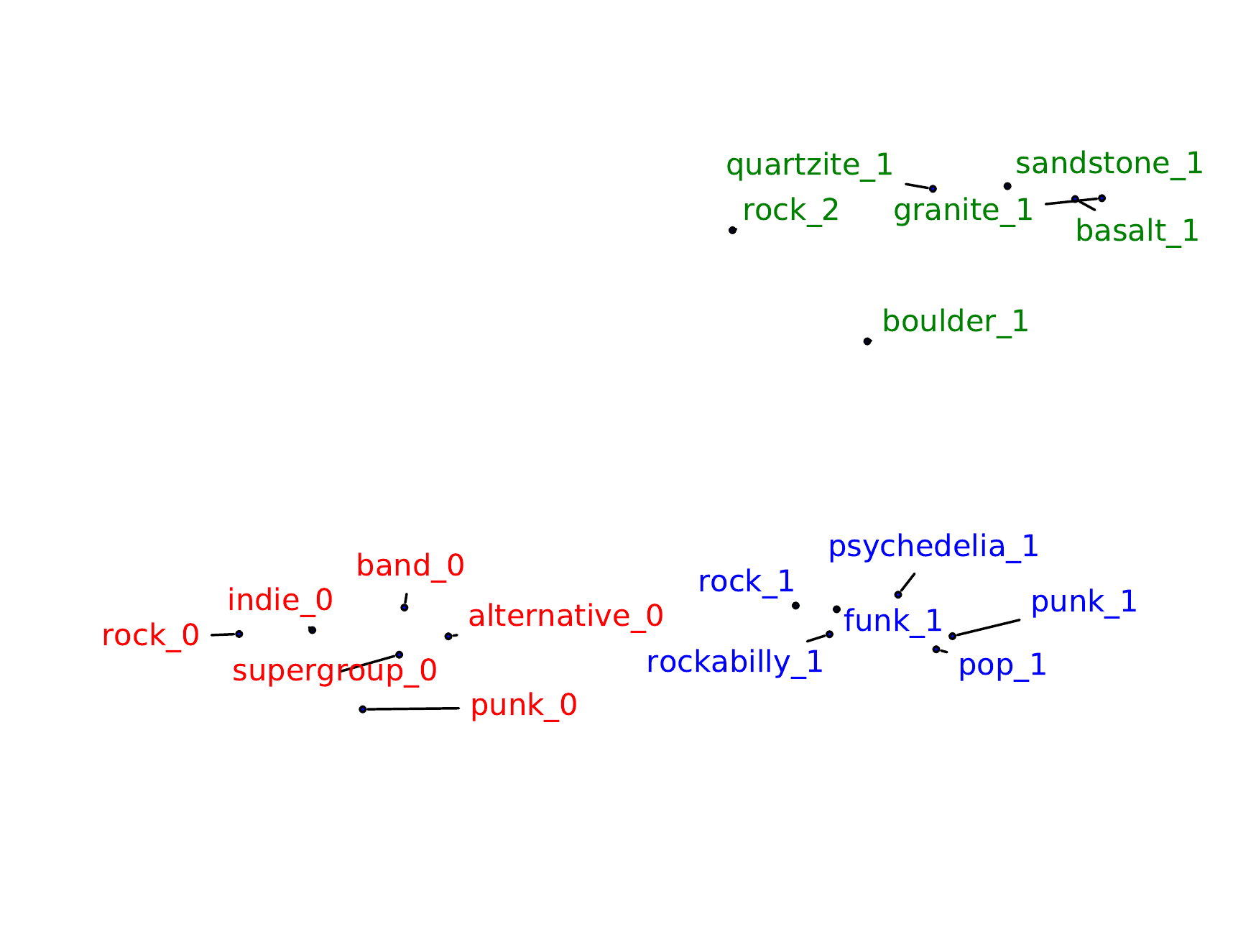}
    \caption{rock}
  \end{subfigure}
  \hfill
  \begin{subfigure}[b]{0.495\linewidth}
    \centering
    \includegraphics[width=\textwidth]{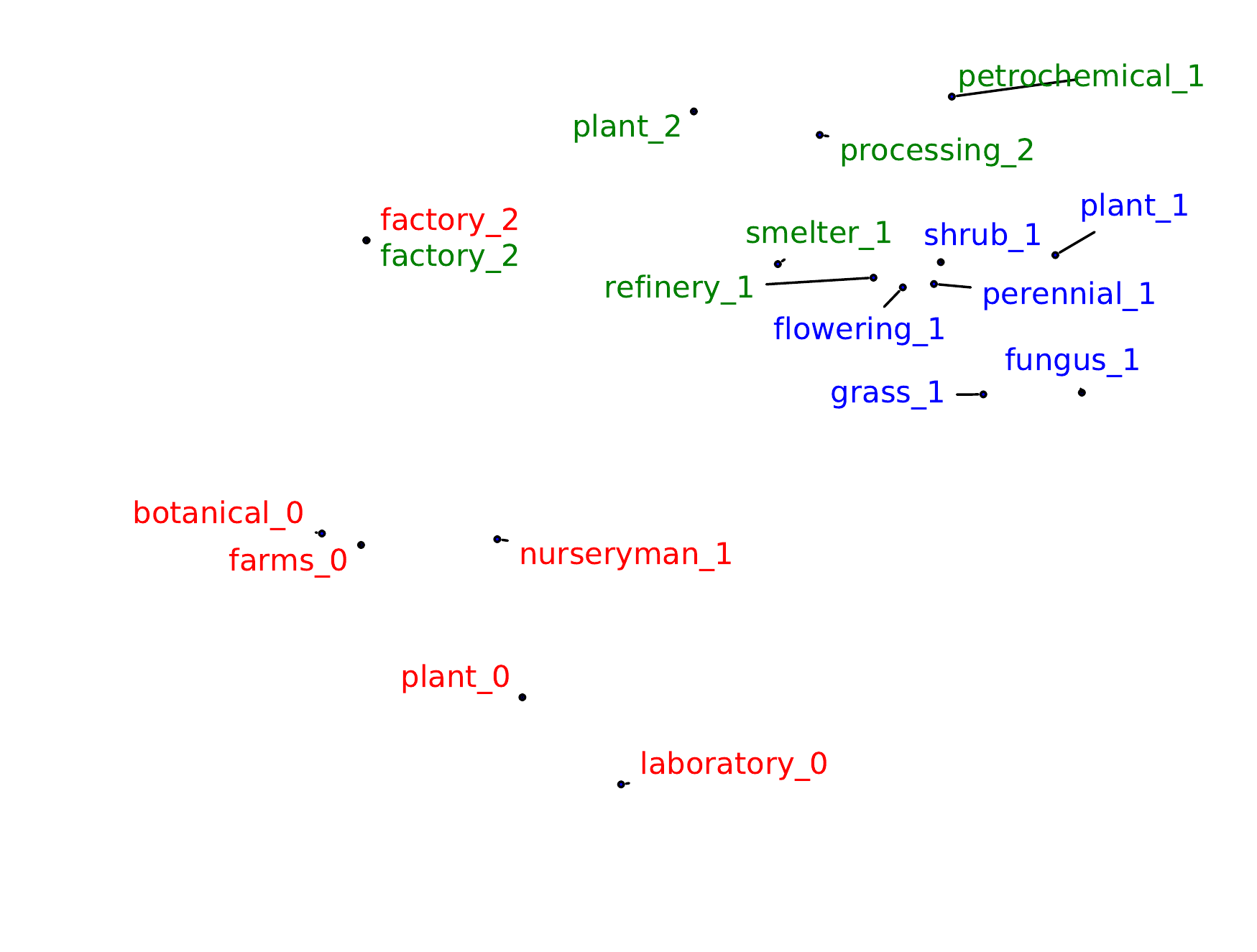}
    \caption{plant}
  \end{subfigure}
  \disgramVisCaption{I}
  \label{fig:disgram_neighbors_pca_1}
\end{figure}
and~\figref{disgram_neighbors_pca_2}.
\begin{figure}[!p]
  \centering
  \begin{subfigure}[b]{0.495\linewidth}
    \centering
    \includegraphics[width=\textwidth]{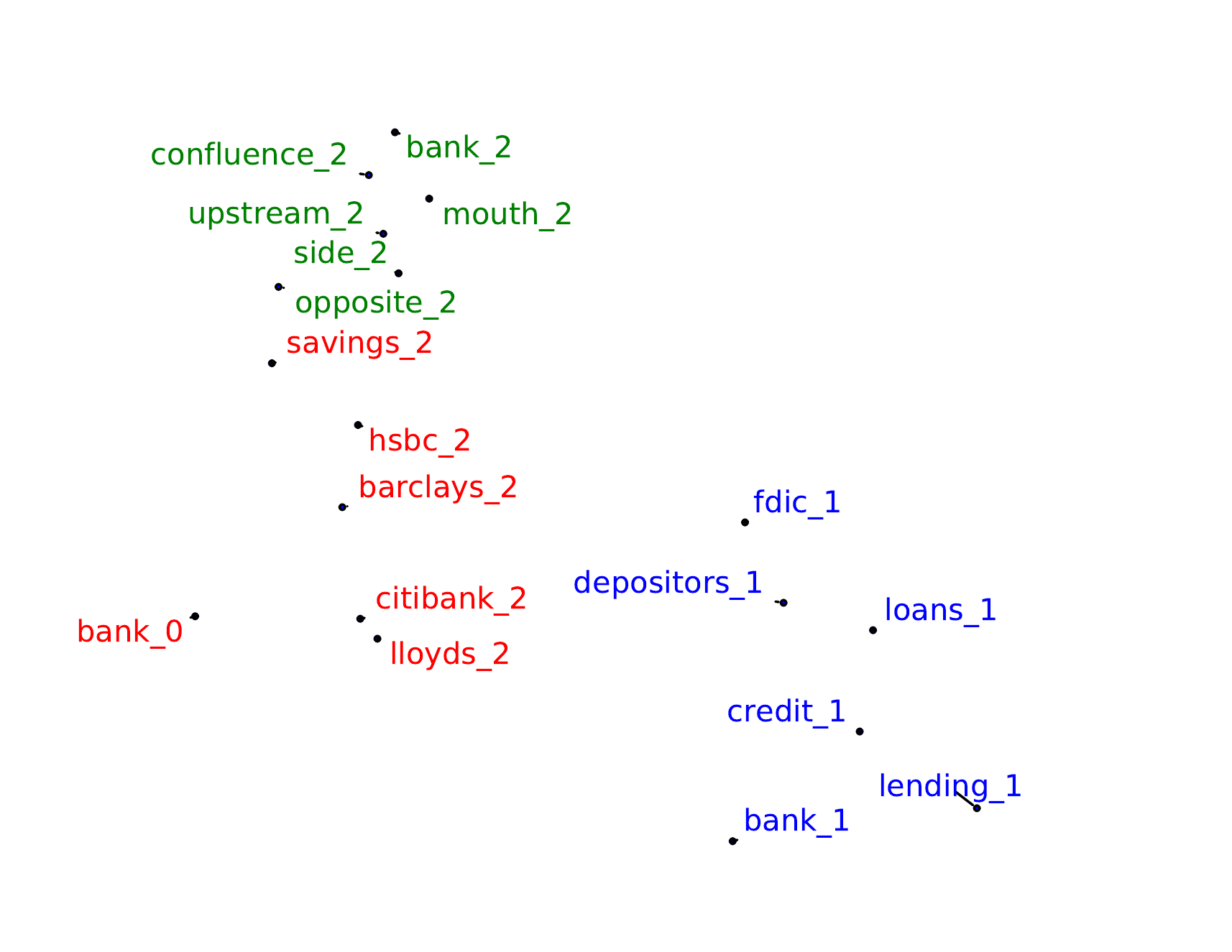}
    \caption{bank}
  \end{subfigure}
  \hfill
  \begin{subfigure}[b]{0.495\linewidth}
    \centering
    \includegraphics[width=\textwidth]{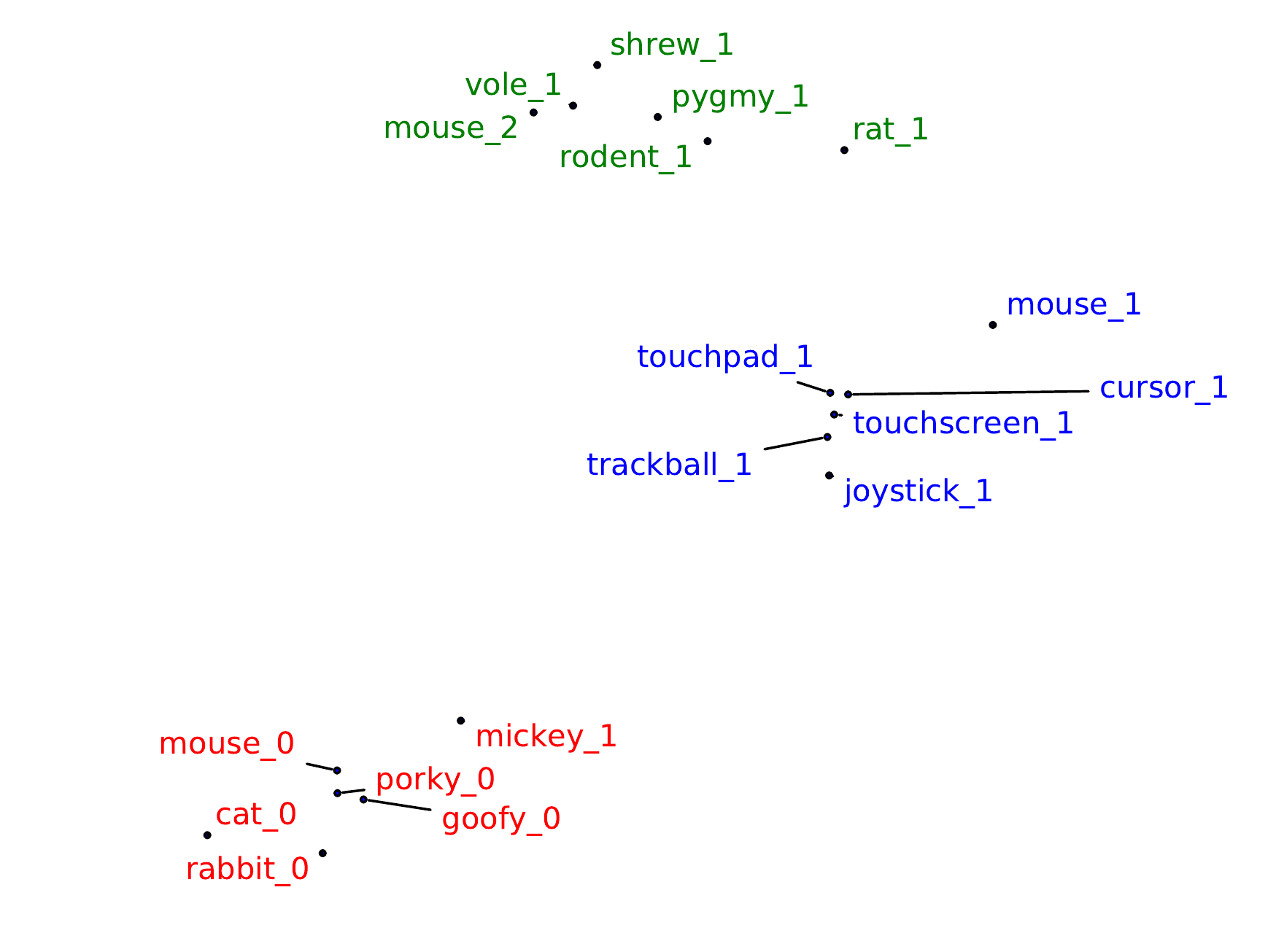}
    \caption{mouse}
  \end{subfigure}
  \hfill
  \begin{subfigure}[b]{0.495\linewidth}
    \centering
    \includegraphics[width=\textwidth]{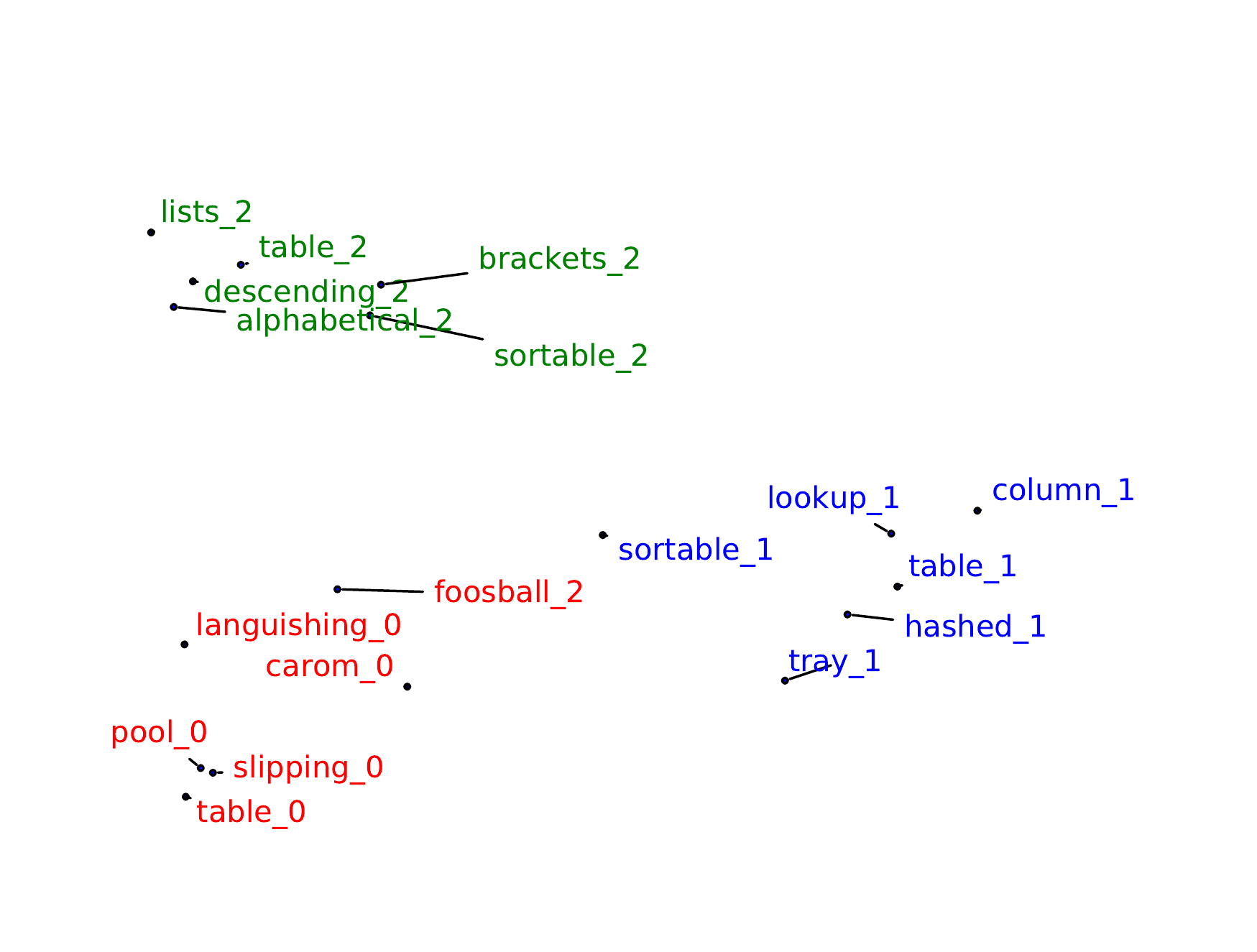}
    \caption{table}
  \end{subfigure}
  \hfill
  \begin{subfigure}[b]{0.495\linewidth}
    \centering
    \includegraphics[width=\textwidth]{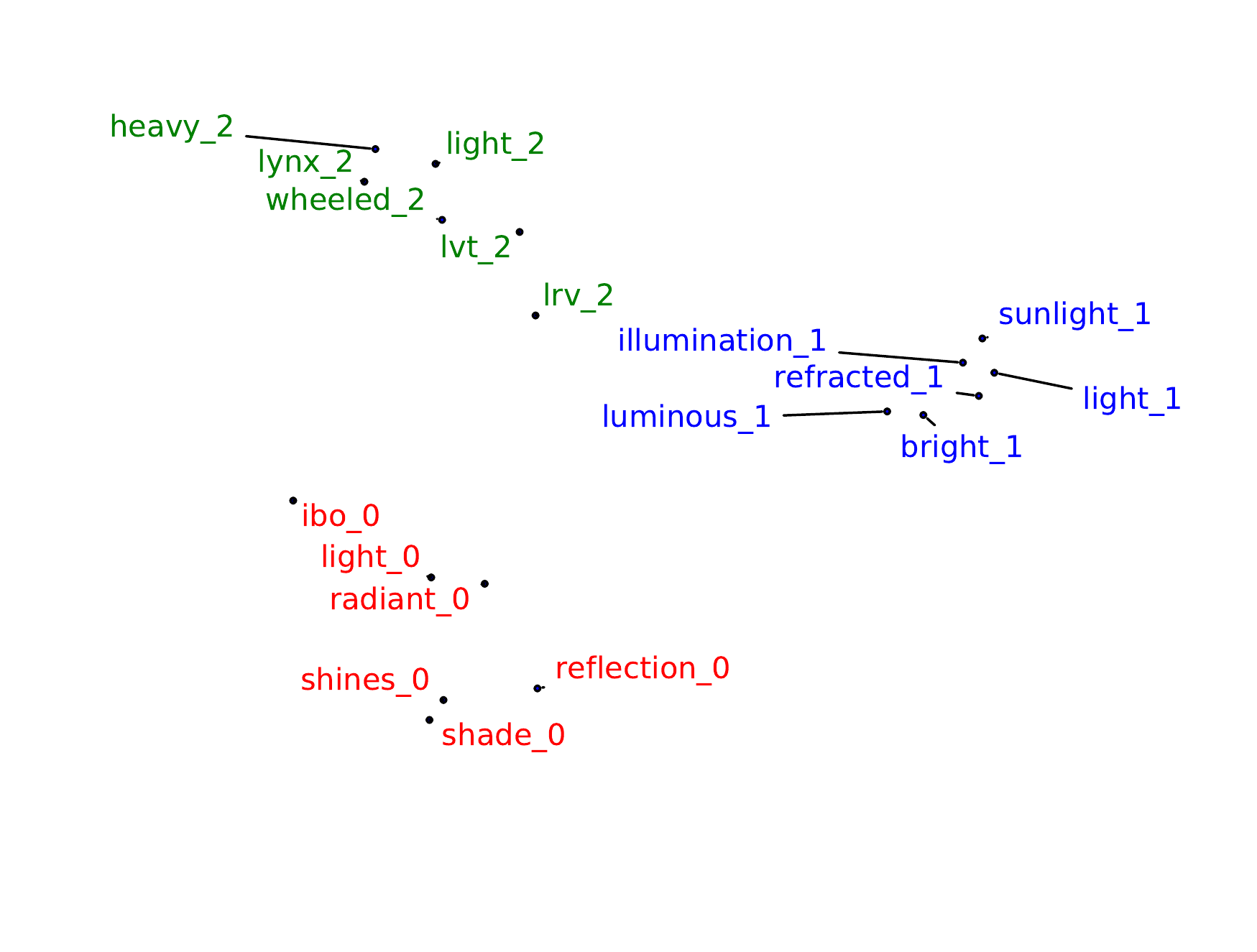}
    \caption{light}
  \end{subfigure}
  \hfill
  \begin{subfigure}[b]{0.495\linewidth}
    \centering
    \includegraphics[width=\textwidth]{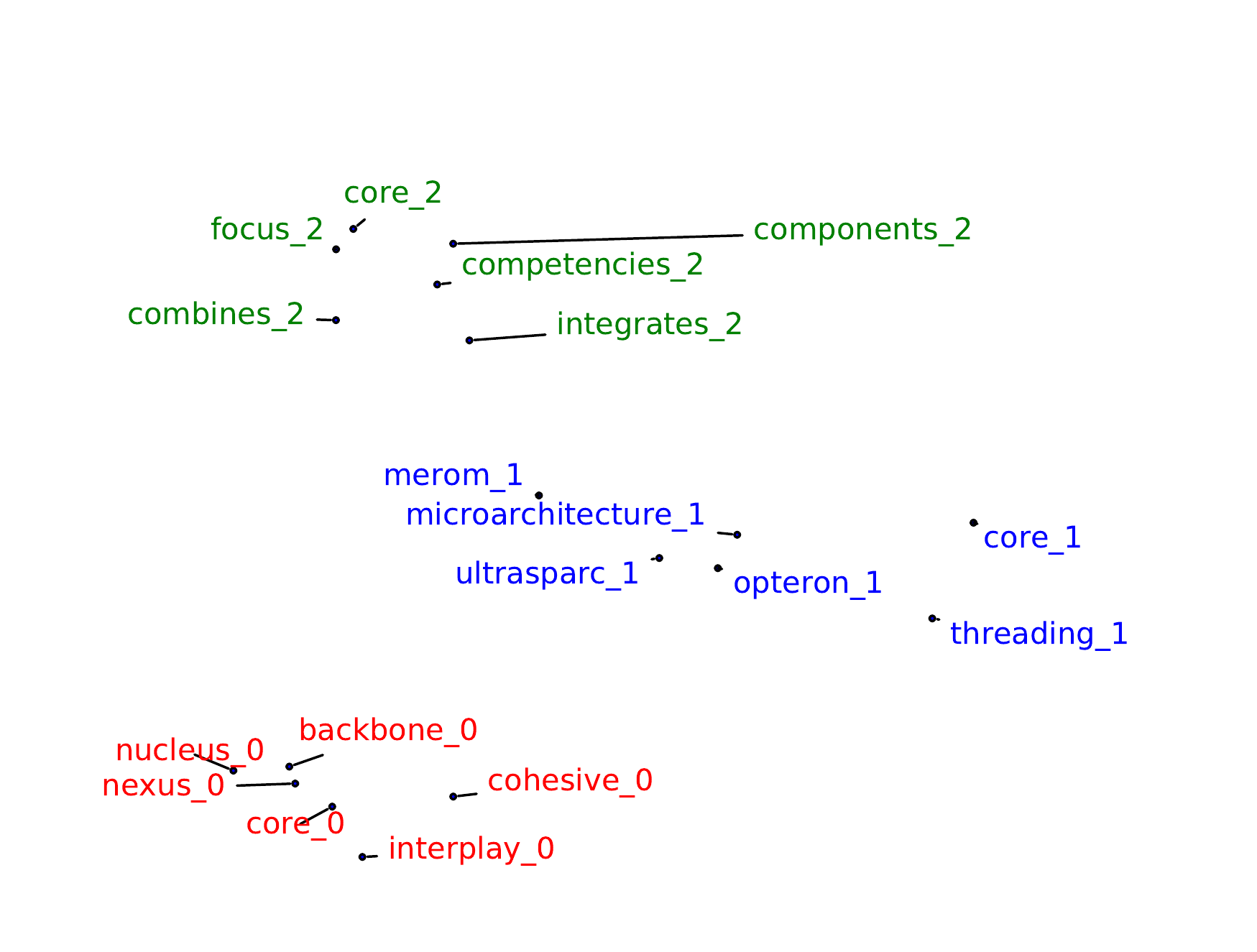}
    \caption{core}
  \end{subfigure}
  \disgramVisCaption{II}
  \label{fig:disgram_neighbors_pca_2}
\end{figure}
We used Principal Components Analysis to reduce the embedding dimensionality.

In most cases our model discovers senses accurately. For example, in the case of `fox', the first sense for the model without the entropy cost
is a broadcasting company, the second is an animal and the third is a family name. In the case of `mouse', the first sense is a cartoon character,
the second is a pointing device and the third is an animal.
We should note here that most of the words from~\tabref{disgram_neighbors_1} and~\tabref{disgram_neighbors_2}
have two dominating senses, rather than three. In those cases, one of the senses is usually split by our model into two.
However, the split is not random but corresponds to different contexts in which those senses occur. For example both sense 1 and sense 3 of
word `plant' in~\tabref{disgram_neighbors_2} describe a factory. However, sense 1 is a agricultural factory while sense 3 is
a petrochemical factory.

\begin{figure}[hbt!]
  \centering
  \includegraphics[width=0.5\linewidth]{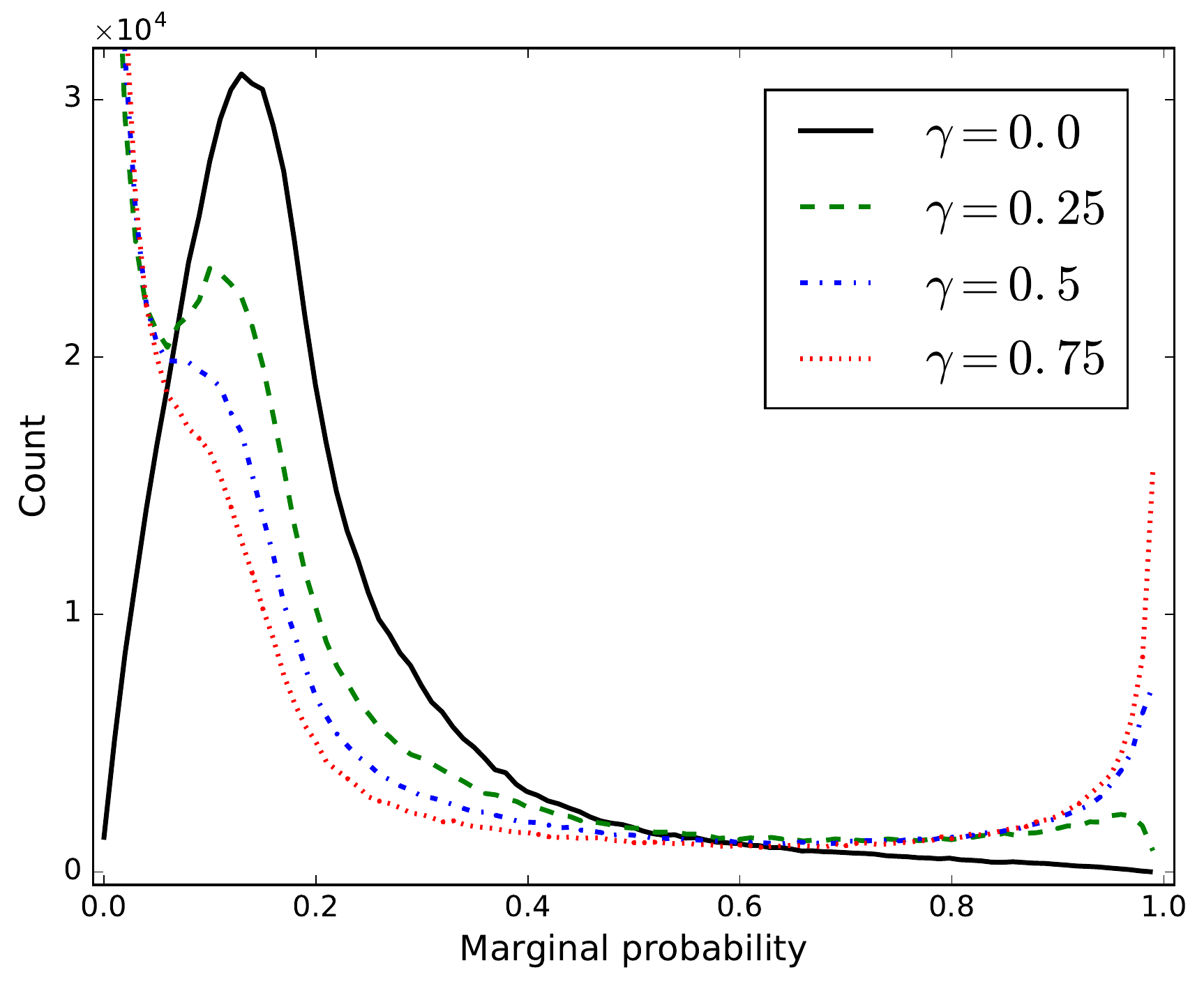}
  \longcaption{Histograms of marginal probabilities of word senses learned by Disambiguated Skip-gram models with different values
               of the entropy cost $ \gamma $.}{Experiments conducted for 5-sense 50-dimensional model.}
  \label{fig:up_influence_on_sense_number}
\end{figure}

To better understand how the entropy cost influences the number of senses, we fine-tuned 5-sense, 50-dimensional models with the entropy
cost $ \gamma $ ranging from $ 0.0 $ (no penalty) to $ 1.0 $, and counted senses having marginal probability $ p \geq 0.05 $.
In~\figref{up_influence_on_sense_number}
\begin{table}[htb]
  \centering
    \begin{tabular}{|c|c|c|c|c|c|c|}
      \hline
      $ \gamma $           & 0.0 & 0.1 & 0.25 & 0.5 & 0.75 & 1.0 \\ \hline
      Average sense number & 4.7 & 4.3 & 3.7  & 3.2 & 2.8  & 2.5 \\ \hline
    \end{tabular}
  \caption{Average number of senses per word with marginal probability $ p \geq 0.05 $, learned by Disambiguated
           Skip-gram models with different values of the entropy cost $ \gamma $.}
  \label{tab:up_influence_on_sense_number}
\end{table}
we present histograms of marginal probabilities of senses of all words from the vocabulary. As depicted, in the case there
is no entropy cost (black curve) marginal probabilites are distributed quite evenly: there are very few
word senses with probabilities close to $ 0.0 $ or $ 1.0 $. For higher values of the entropy cost, marginal probability
distributions are more polarized, i.e. there are more word senses with probabilities close to $ 0.0 $ or $ 1.0 $. Assuming that we prune
senses having probabilities smaller than $ 0.05 $, we get on average $ 4.7 $ active senses for model without a penalty and as low as $ 2.5 $
senses on average for model with a strong $ \gamma = 1.0 $ entropy cost (~\tabref{up_influence_on_sense_number}).

\subsection{Word sense induction experiments}

Word sense induction (WSI) is considered the most important task used to quantitatively assess multi-sense word embedding
models~\cite{bartunov2016breaking}. For WSI experiments, we use four datasets described in \sectionref{datasets_wsi}: SemEval-2007,
SemEval-2010, SemEval-2013 and Wikipedia Word-sense Induction (WWSI). To disambiguate words from those datasets we first calculate vector
representations of associated contexts by averaging all sense embeddings for all context words:
\begin{equation}
  \bar{\mathbf{c}}_w = \left( k \cdot \#C_w \right)^{-1}
                       \sum_{c \in \mathcal{C}_w} \sum_{s=1}^k \mathbf{v}_{cs}.
\end{equation}
Then, we select a sense of the center word whose vector representation has the smallest cosine distance to the context representation:
\begin{equation} \label{eq:word_sense_inference}
  s_w = \argmax_j \cos \left( \mathbf{v}_{wj}, \bar{\mathbf{c}}_w \right).
\end{equation}
To compare sense assignments with gold standard we use adjusted
rand index \cite{hubert1985comparing}. Results for different dimensionalities, number of senses and different model regularizers
are presented in~\tabref{disgram_wsi}.
\begin{table}[htb]
  \centering
  \setlength{\tabcolsep}{3pt}
    \begin{tabular}{|c|c|c|c|c|c|c|}
      \hline
      \multirow{2}{*}{Dim.} & Sense              & \multirow{2}{*}{Penalty} & \multirow{2}{*}{SemEval-2007} & \multirow{2}{*}{SemEval-2010} & \multirow{2}{*}{SemEval-2013} & \multirow{2}{*}{WWSI} \\ 
                            & num.               &                          &                               &                               &                               &                       \\ \hline \hline
      \multirow{10}{*}{50}  & \multirow{5}{*}{3} & No penalty               & 0.0711                        & 0.1                           & 0.0501                        & 0.272                 \\ \cline{3-7}
                            &                    & $ \gamma = 0.25 $        & 0.0803                        & 0.0898                        & 0.0435                        & 0.243                 \\ \cline{3-7}
                            &                    & $ \gamma = 0.5 $         & \textbf{0.0853}               & 0.0795                        & 0.0406                        & 0.161                 \\ \cline{3-7}
                            &                    & $ \delta = 0.0001 $      & 0.0729                        & 0.0966                        & \textbf{0.0545}               & 0.271                 \\ \cline{3-7}
                            &                    & $ \delta = 0.005 $       & 0.074                         & 0.102                         & 0.0515                        & 0.28                  \\ \cline{2-7}
                            & \multirow{5}{*}{5} & No penalty               & 0.0638                        & 0.107                         & 0.04                          & 0.304                 \\ \cline{3-7}
                            &                    & $ \gamma = 0.25 $        & 0.0826                        & \textbf{0.116}                & 0.0432                        & 0.244                 \\ \cline{3-7}
                            &                    & $ \gamma = 0.5 $         & 0.0637                        & 0.0908                        & 0.0449                        & 0.182                 \\ \cline{3-7}
                            &                    & $ \delta = 0.0001 $      & 0.0617                        & 0.11                          & 0.04                          & \textbf{0.306}        \\ \cline{3-7}
                            &                    & $ \delta = 0.005 $       & 0.0594                        & 0.108                         & 0.0413                        & 0.289                 \\ \hline \hline

      \multirow{10}{*}{300} & \multirow{5}{*}{3} & No penalty               & 0.0799                        & 0.098                         & \textbf{0.0583}               & 0.273                 \\ \cline{3-7}
                            &                    & $ \gamma = 0.25 $        & 0.0807                        & 0.091                         & 0.0487                        & 0.235                 \\ \cline{3-7}
                            &                    & $ \gamma = 0.5 $         & 0.0818                        & 0.0781                        & 0.0497                        & 0.169                 \\ \cline{3-7}
                            &                    & $ \delta = 0.0001 $      & \textbf{0.0916}               & 0.0976                        & 0.0546                        & 0.277                 \\ \cline{3-7}
                            &                    & $ \delta = 0.005 $       & 0.084                         & 0.104                         & 0.0538                        & 0.275                 \\ \cline{2-7}
                            & \multirow{5}{*}{5} & No penalty               & 0.0765                        & 0.117                         & 0.0445                        & 0.292                 \\ \cline{3-7}
                            &                    & $ \gamma = 0.25 $        & 0.0795                        & 0.113                         & 0.0454                        & 0.259                 \\ \cline{3-7}
                            &                    & $ \gamma = 0.5 $         & 0.0653                        & 0.0911                        & 0.0495                        & 0.183                 \\ \cline{3-7}
                            &                    & $ \delta = 0.0001 $      & 0.0761                        & 0.114                         & 0.0441                        & 0.291                 \\ \cline{3-7}
                            &                    & $ \delta = 0.005 $       & 0.0762                        & \textbf{0.121}                & 0.0433                        & \textbf{0.296}        \\ \hline

    \end{tabular}
  \longcaption{Adjusted rand index for the Disambiguated Skip-gram model with different dimensionalities,}
              {sense numbers and different regularization terms, evaluated on different test datasets.
               The best results for each dataset and for each dimensionality are highlighted.}
  \label{tab:disgram_wsi}
\end{table}
The best results for each dataset and for each dimensionality are highlighted.
In most cases adding a small entropy cost or parallel penalty in fine-tuning has a positive effect.
Comparison with other multi-sense word embedding models is presented in~\tabref{disgram_wsi_compare}.
\begin{table}[htb]
  \centering
  \setlength{\tabcolsep}{3pt}
    \begin{tabular}{|c|c|c|c|c|}
      \hline
      Model                   & SemEval-2007   & SemEval-2010   & SemEval-2013   & WWSI           \\ \hline
      MSSG                    & 0.048          & 0.085          & 0.033          & 0.194          \\ \hline
      NP-MSSG                 & 0.033          & 0.044          & 0.033          & 0.110          \\ \hline
      MPSG                    & 0.044          & 0.077          & 0.014          & 0.160          \\ \hline
      AdaGram                 & 0.069          & 0.097          & \textbf{0.061} & 0.286          \\ \hline
      Disambiguated Skip-gram & \textbf{0.077} & \textbf{0.117} & 0.045          & \textbf{0.292} \\ \hline
    \end{tabular}
  \longcaption{Adjusted rand index for different 300-dimensional multi-sense word embedding models.}{Disambiguated Skip-gram was trained
               with 5 senses, without neither entropy cost nor parallel penalty. MSSG and MPSG were trained with 3 senses. AdaGram was
               trained with $ \alpha = 0.15 $. Results for all models except Disambiguated Skip-gram are taken from~\cite{bartunov2016breaking}.}
  \label{tab:disgram_wsi_compare}
\end{table}
Results for competing state-of-the-art models were taken from~\cite{bartunov2016breaking}. Our model outperforms all those models on three
out of four benchmark test sets.

\subsection{Word-similarity experiments}

The aim of the word-similarity task is to tell how two given words are similar to each other. Word-similarity datasets consist of word pairs
with numerical similarity measures specified by experts. Predictions made by the model are compared against those gold standards.
Unfortunately, many word-similarity datasets do not provide contexts and therefore are not suited to evaluate multi-sense word embeddings.
As described in \sectionref{datasets_ws} one popular dataset that does provide contexts is SCWS~\cite{huang2012improving}.
We use it as the main word-similarity benchmark in this work. Following~\cite{bartunov2016breaking} we use two metrics to assess contextual
word-similarity:
\begin{equation}
\begin{aligned}
avgSimC(w_1, & w_2) = \\
             & \sum_{s_1=1}^k \sum_{s_2=1}^k P(s_1 \mid w_1,C_{w_1}) P(s_2 \mid w_2,C_{w_2})
               \cos ( \vect{v}_{w_1, s_1}, \vect{v}_{w_2, s_2} )
\end{aligned}
\end{equation}
and
\begin{equation}
maxSimC(w_1, w_2) = \cos ( \vect{v}_{w_1, s_1}, \vect{v}_{w_2, s_2} ),
\end{equation}
where:
\begin{equation}
\begin{aligned}
s_1 & = \argmax_{s=1,\ldots,k} P(s \mid w_1,C_{w_1})    \\
s_2 & = \argmax_{s=1,\ldots,k} P(s \mid w_2,C_{w_2}).
\end{aligned}
\end{equation}
Performance of Disambiguated Skip-gram in this task and comparison with other word embedding models is presented in~\tabref{disgram_spearman}.
\begin{table}[htb]
  \centering
    \begin{tabular}{|c|c|c|}
      \hline
      Model                   & avgSimC       & maxSimC            \\ \hline 
      Skip-gram               & 65.2          & \textbf{65.2}      \\ \hline
      MSSG                    & \textbf{69.3} & 57.3               \\ \hline
      NP-MSSG                 & 69.1          & 59.8               \\ \hline
      MPSG                    & 65.4          & 63.6               \\ \hline
      AdaGram                 & 61.2          & 53.8               \\ \hline
      Disambiguated Skip-gram & 64.4          & 62.0               \\ \hline
    \end{tabular}
  \longcaption{Spearman's correlation coefficient multiplied by 100 for different 300-dimensional models evaluated on the SCWS dataset.}
              {Disambiguated Skip-gram, MSSG and MPSG were trained with 3 senses. AdaGram was trained with $ \alpha = 0.15 $.
              Metrics definitions and results for all models except Disambiguated Skip-gram taken from~\cite{bartunov2016breaking}.}
  \label{tab:disgram_spearman}
\end{table}
Note that the vanilla skip-gram model gives the best results according to \emph{maxSimC} metric, despite the fact that it models
only single sense per word and consequently ignores the context during prediction. This observation indicate that even contextual
word-similarity datasets are not well suited to evaluate multi-sense embedding models.

For the sake of completeness we also evaluate Disambiguated Skip-gram using one popular context-less datasets, namely
WordSim353~\cite{finkelstein2001placing}. Since this dataset does not provide contexts, we simply average similarities between
embeddings of all senses of compared words:
\begin{equation}
avgSim(w_1, w_2) = \frac{1}{k^2} \sum_{s_1=1}^k \sum_{s_2=1}^k \cos ( \vect{v}_{w_1, s_1}, \vect{v}_{w_2, s_2} )
\end{equation}
Results for this benchmark are presented in~\tabref{disgram_spearman_single_sense}.
\begin{table}[!htb]
  \centering
    \begin{tabular}{|c|c|c|c|}
      \hline
      Model                   & avgSim \\ \hline
      Skip-gram               & 70.4   \\ \hline
      MSSG                    & 70.9   \\ \hline
      NP-MSSG                 & 68.6   \\ \hline
      Disambiguated Skip-gram & 70.1   \\ \hline
    \end{tabular}
  \longcaption{Spearman's correlation coefficient multiplied by 100 for different 300-dimensional models evaluated on WordSim353 dataset.}
              {Disambiguated Skip-gram and MSSG were trained with 3 senses. Metric definition and results for all models except Disambiguated
               Skip-gram are taken from~\cite{neelakantan2015efficient}.}
  \label{tab:disgram_spearman_single_sense}
\end{table}

\tabref{disgram_spearman} and \tabref{disgram_spearman_single_sense} demonstrate that Disambiguated Skip-gram retain most of the word
similarity features. The fact that Disambiguated Skip-gram performs a bit worse than some of the earlier models should not worry, because,
as mentioned, this is not the most important task for evaluation of multi-sense word embedding models.

\section{Conclusions}

In this chapter we presented a novel neural model for learning multi-sense word embeddings. We have shown that our method outperforms
competing state-of-the-art approaches on the word sense induction task on three out of four benchmark datasets. We also performed
qualitative evaluation of our model by querying a few nearest neighbors for ten popular ambiguous words. In contextual word similarity
task our model performs slightly worse than some other multi-sense word embeddings. However, as pointed out by
Bartunov~et~al.~\cite{bartunov2016breaking}, word similarity task is not the best way to evaluate this kind of models.

It is worth noting that the problem of ambiguity goes beyond text data. For example, a dimensionality reduction technique proposed by Hinton and
Roweis~\cite{hinton2002stochastic} and later extended by van der Maaten and Hinton~\cite{maaten2012visualizing} is able to place any ambiguous
high-dimensional object in multiple separated regions of the low-dimensional space.

As a direction for future research we believe that the Disambiguated Skip-gram model can be enriched with subword (character n-gram)
information. This strategy has recently turned out to be effective in the case of single-sense word embeddings~\cite{bojanowski2016enriching}.
To adapt this idea to multi-sense models we can either learn character n-gram embeddings separately for each sense or we can share subword
embeddings among senses. Disadvantage of the first approach is a significant increase in memory consumption and training time.
In the second case, we would most probably need to introduce and learn some subword weights, which will allow us to mix subword into
word-senses.

\chapter{Conclusions and directions for future research} \label{ch:conclusions}

Advances in machine learning revolutionize the world and the way we live. Learning algorithms are an becoming essential part of information
systems. Text data, including output from speech recognition systems, is one of the most popular modalities used to train those algorithms.
Most of the learning algorithms require input data to be presented in a form of relatively short, fixed-size vectors.
Therefore, being able to represent text in a dense distributed way is a crucial step in developing well-performing algorithms for text
understanding and processing. This thesis contributes new algorithms and models for learning such representations.

\section{Contributions}

In this dissertation we focused on building vector representations of text data on two levels,
namely a document level and a word level. Some vector representations are generic and can be used for different machine learning tasks.
Other are more tailored for specific needs. When it comes to document representations we focused on those used for fast text retrieval.
Ideal information retrieval system would have both high precision and high recall. In practice, however, information retrieval systems
often need to trade off between precision and recall. When a user is not willing to accept omission of any relevant documents, the recall is of
primary importance. However, if the precision of top results is of higher importance than the recall, we can use locality preserving hashing
techniques to retrieve documents via fast approximate nearest neighbor search. This kind of search have one important design consideration:
it can have some false negatives but should return as few false positives as possible. Those conditions are acceptable in many real world
applications: users of search or recommendation engines will not complain if some relevant items are missing, as long as returned items are
relevant to their query or context.

Probably the most popular approach to approximate nearest neighbor search in text data is to learn binary codes from some real-valued
document representations. In this dissertation we proposed Binary Paragraph Vector: a neural network models for learning high-quality binary
codes from text data without using an intermediate representation (e.g. bag-of-words). We compared Binary PV models with the seminal
\emph{semantic hashing} technique and demonstrated their superiority on three popular datasets. Also, we tested these models against
methods that first learn real-valued representations and use them to infer binary codes. We showed that there is no clear benefit from
using these indirect approaches instead of Binary PV models.

Second part of the dissertations revolved around word level embeddings. Learning high quality word embeddings is important because they are
used for many downstream NLP tasks, like sentiment analysis, language modeling, question answering, part of speech tagging,
neural machine translation or text summarization. Most of the leading word embedding techniques learn only one vector per word, consequently
ignoring the fact that many words are ambiguous. Recently, however, there is an increasing interest in methods that learn separate vectors
for distinct senses of words. We review and compare existing multi-sense word embedding solutions and then we propose Disambiguated Skip-gram:
a new neural network that learns high-quality multi-sense word embeddings. Disambiguated Skip-gram outperforms state-of-the-art competing
methods in the word sense induction task on three out of four benchmark test sets we used in evaluation. Furthermore, it has a elegant
probabilistic interpretation. Finally, unlike previously proposed probabilistic multi-sense word embeddings models, Disambiguated Skip-gram
is end-to-end differentiable and, therefore, can be easily trained with backpropagation.

\section{Future research directions}

This thesis opens several interesting directions for future research. In the case of binary document embeddings one idea could be
to investigate binary document embedding models that takes word order into account and outperform simpler models, like Binary PV-DBOW.
Another line of research could focus on speeding up code inference for new documents. An inherent limitation of our solution, and original
Paragraph Vector as well, is the need for iterative numerical optimization in the inference phase. This optimization is much faster than
training, because most of the model parameters are fixed in this phase. However, some time for convergence is still needed, which makes our
models poorly suited for streaming or real-time processing. It would be interesting and valuable to extend Binary
Paragraph Vector models to this kind of applications.

In \sectionref{bin_units_comparison} we compared different strategies for binarization of activation in Binary PV models. It would be
interesting to add \emph{relaxed bernoulli} distribution to this comparision. Relaxed bernouli is a special case of the
Gumbel-Sofmax~\cite{jang2016categorical,maddison2016concrete} distribution, which we successfully used in Disambiguated Skip-gram.

In the case of Disambiguated Skip-gram, we believe that the model can be enriched with subword information, following the technique proposed
in~\cite{bojanowski2016enriching}. The idea is to predict context words using not only a given sense of a center word, but also vector
representation of character n-grams in this sense. To this end, we could either store separate subword embeddings for each modeled sense or
learn a model that would tell us how a given subword expresses itself in a given sense. Interestingly, character n-grams could also improve
document embeddings, so it may be worth trying to also incorporate them into the Binary Paragraph Vector models.

Disambiguated Skip-gram is a parametric model, i.e. we need to specify the number of senses to be learned. We suggested using sense pruning
to account for the fact that different words have different numbers of meanings. The assumption is that when a sense is pruned, its
instances are distributed among remaining senses. An alternative approach could be to merge similar senses of a given word. We believe that
this approach is worth investigation.

Another research direction could be to evaluate Disambiguated Skip-gram in downstream tasks. Most multi-sense word embedding models,
including ours, are evaluated using intrinsic evaluation methods, i.e. on some intermediate task with standardized benchmarks.
We believe that it would be interesting to also evaluate Disambiguated Skip-gram in an extrinsic way, i.e. on selected real-word tasks.
Since one of the most common architectures that are used with word embeddings is a recurrent neural network, we believe that
it would be interesting to implement an LSTM network which takes multi-sense word embeddings as inputs and test it in downstream NLP tasks,
e.g. language modeling.

The Gumbel-Softmax gradient estimator used in Disambiguated Skip-gram is a biased estimator, i.e. its expected value differs
from the true gradient. After we carried out the experiment with Disambiguated Skip-gram, \cite{tucker2017rebar} proposed \emph{REBAR}:
an unbiased, but nevertheless low-variance, gradient estimator for models with discrete random variables. They report a good performance
of this estimator in several models. It would be interesting to check whether replacing Gumbel-Softmax in Disambiguated Skip-gram with REBAR
would improve learned embeddings.

Yet another interesting direction for future research is multi-task learning in context of multi-sense word embeddings. Traditionally,
learning models are developed for specific problems and therefore perform specific tasks. However, in some domains, e.g. computer vision,
it is not unusual to share neural network architectures between different tasks. It is more difficult to implement this idea in natural
language processing. Nevertheless, some models were recently proposed that address this issue, e.g.~\cite{hashimoto2017joint}.
It would be interesting to see how multi-sense word embeddings can enrich such architectures.

\appendix

\chapter{Datasets and experimental setup} \label{ch:software_datasets}

In this appendix we describe benchmark datasets and software libraries that we used to conduct the experiments.

\section{Datasets} \label{sec:datasets}

Experiments in~\chapterref{dsh} and~\appendixref{improving_bow} were conducted based on two popular English language text datasets,
namely 20 Newsgroups and RCV1.
Some basic test preprocessing was applied to the datasets.
All characters were converted to lower case and stopwords (very frequent words
that do not convey any specific informations) were removed.
In case of some experiments stemming or lemmatization was applied on corpora.
For stemming we used Porter's algorithm~\cite{porter1980algorithm}.
Below we briefly describe the 20 Newsgroups and RCV1 datasets.

\subsection{20 Newsgroups}

The 20 Newsgroups\footnote{Available at \url{http://qwone.com/~jason/20Newsgroups}} dataset consist of
$ 1.13 \times 10^4 $ train documents and $ 7.5 \times 10^3 $ test documents.
Documents were collected in the mid 90's from \emph{Usenet} discussion groups.
They are written in an informal style and belong to one of the 20 topics.
Documents are evenly distributed among topics. The full set of topics is presented in \tabref{tng_groups}.
\begin{table}[htb]
  \centering
    \begin{tabular}{|l|l|}
      \hline
      talk.politics.guns    & comp.graphics            \\
      talk.politics.mideast & comp.os.ms-windows.misc  \\
      talk.politics.misc    & comp.sys.ibm.pc.hardware \\
      talk.religion.misc    & comp.sys.mac.hardware    \\
      soc.religion.christ   & comp.windows.x           \\
      alt.atheism           & rec.sport.hockey         \\
      sci.med               & rec.sport.baseball       \\
      sci.space             & rec.motorcycles          \\
      sci.electronics       & rec.autos                \\
      sci.crypt             & misc.forsale             \\ \hline
    \end{tabular}
  \caption{All the groups from the 20 Newsgroups dataset}
  \label{tab:tng_groups}
\end{table}

\subsection{Reuters corpus volume 1} \label{sec:datasets_rcv1}

Reuters Corpus, Volume~1~(RCV1)\footnote{Available at \url{http://trec.nist.gov/data/reuters/reuters.html}}
is a collection of more than $ 8 \times 10^5 $ professional news bulletins and articles
written in years 1996-1997 in English and published by Reuters news agency.
The language used is more formal than in the 20 Newsgroups corpus.
The articles are annotated by one or more topics that form a hierarchy.
The taxonomy of categories is convoluted. There are four top categories in the hierarchy:
\begin{itemize}  
\item markets,
\item economics,
\item government/social,
\item corporate/industrial.
\end{itemize}
There exist two variants of the RCV1 corpus. The first variant is an original one published by Reuters company.
Unfortunately it contains some erroneous data. For example some documents are not assigned to any of the topics.
Other documents' topic assignment violate the topic taxonomy.
Therefore David D. Lewis at al. proposed~\cite{lewis2004rcv1} a set of corrections and cleansing techniques.
Corrected corpus is reffered as RCV1-v2. We use this variant in our experiments described later on in the dissertation.
There is not official split into training and test data. The common choices are 50/50 and~90/10.

\subsection{English Wikipedia} \label{sec:datasets_enwiki}

As of May 2017, English Wikipedia contains $ 5.4 \times 10^6 $ articles written by an open community of editors and contributors.
The Wikimedia Foundation, the body which governs Wikipedia, offers free download of backup files for each language separately.
Articles are exported in a form of XML files containing not only text of articles but also categorization, metadata and hyperlinks.
However, no external media, like graphics or videos, are included.
Compressed dump file for English Wikipedia has size almost 14~GB.
Due to its size and wide variety of topics, this corpus is often used to train general purpose word vectors.
Wikipedia articles can have one or more categories assigned to them. Categories form a hierarchy but not a tree.
Categories can have multiple parent categories and cyclic dependencies.
In 2014 over 100 teams took part in a Kaggle competition\footnote{Available at \url{https://www.kaggle.com/c/lshtc}}
aiming at classifying documents from English Wikipedia.

Often in order to be able to compare results obtained using Wikipedia researchers do not download the latest snapshot but instead use older dumps
previously reported in research papers. One of those reference dumps of English Wikipedia is the 2010 Westbury Lab Wikipedia
corpus\footnote{Available~at \url{http://www.psych.ualberta.ca/~westburylab/downloads/westburylab.wikicorp.download.html}}.
It is provided in raw text format. All XML tags were removed but no further text cleansing was applied.
In our experiments we also use more recent snapshot from April 5th, 2016.



\subsection{Word similarities datasets} \label{sec:datasets_ws}

To assess quality of word embeddings word similarity datasets are used.
The most popular word similarity dataset is WordSim353~\cite{finkelstein2001placing}.
The datasets consists of word-word pairs with similarity scores ranked by humans.
Words do not need to be synonymous to have high score.
Words with different meanings, but which are often used together also have high scores assigned in WordSim353.
For example, the following word-word pairs have high scores in WordSim353 datasets:
weather-forecast, hotel-reservation and psychology-psychiatry.
In contrast to WordSim353, SimLex-999~\cite{hill2016simlex} is a datasets which provides `genuine' similarity measures between words.
Only word that are synonyms will get the high similarity values.
There are both monosemous and polysemous words in both of those datasets.
Due to the lack of contexts, there is no way to tell which sense of polysemous word is used.
To overcome this limitation,
Huang et al. proposed~\cite{huang2012improving} the Contextual Word Similarities (SCWS) dataset.
In addition to word-word pairs, the dataset provides textual context of each of the words.
In contrast to the majority of word similarity datasets,
Stanford Rare Word (RW) Similarity Dataset~\cite{luong2013better} contains infrequent or morphologically complex words.

\subsection{Word sense induction and disambiguation datasets} \label{sec:datasets_wsi}

Word sense induction (WSI) datasets are used to assess and compare WSI systems.
Those datasets consist of sets of words and for each word set of contexts.
Each context is hand-annotated with sense assignment forming so-called \emph{gold standards}.
The WSI systems are given unannotated contexts and are asked to induce senses and disambiguate given words in contexts.
The results are compared against the gold standards by using some metric like V-Measure~\cite{rosenberg2007vmeasure},
F-Score~\cite{artiles2009role} or adjusted rand index~(ARI)~\cite{hubert1985comparing}.

Some popular WSI datasets frequently used as benchmarks were initially used to compare systems presented at the \emph{SemEval} workshops
organized annually by the Association for Computational Linguistics. Among them is the test dataset prepared for the task 2 of the 2007
edition of SemEval, described in~\cite{agirre2007semeval}. It consists of over $ 2.7 \times 10^4 $ short texts from the Wall Street Journal.
Each text has one selected word (noun or verb) which sense is hand-annotated. There are 100 distinct sense-annotated words, having 3.68 senses
on average. Other popular dataset was prepared for the task 14 of the 2010 edition of SemEval and is described in~\cite{manandhar2010semeval}.
It consists of approximately $ 9 \times 10^3 $ texts from multiple news sources. As in the case of the above, each text has one selected word
which sense is hand-annotated and there are 100 distinct sense-annotated words having on average 3.79 senses. Dataset introduced as a part
of the task 13 of the SemEval 2013 workshop~\cite{jurgens2013semeval} contains over $ 4 \times 10^3 $ contexts for 50 distinct words
with a number of contexts per word ranging from 22 to 100. An average the number of senses for this dataset is much higher than for the
two previous and equals 6.02.

An example of a bigger WSI dataset is the Wikipedia Word-sense Induction (WWSI) dataset described in~\cite{bartunov2016breaking}.
It has over $ 3.6 \times 10^4 $ sense-annotated texts for 188 distinct words having 2.2 senses on average. The contexts were extracted from
\emph{ambiguous} English Wikipedia pages, i.e. those that have the following text on top of them: \emph{For other uses, see X
(disambiguation)}, where X is a placeholder for the current page name.

\section{Software}

One of the factors that enabled deep learning to took off was increased popularity
of general-purpose computing on graphics processing units~(GPU).
Writing software which efficiently runs on GPU is considered to be more challenging
than writing software targeting a central processing unit~(CPU).
Therefore, new software packages easing deep learning research have been created in recent years.

For couple of years, three most popular libraries were Theano~\cite{bastien2012theano}, Torch~\cite{collobert2011torch7}
and Caffe~\cite{jia2014caffe}.
They all share some common characteristics.
Computations are defined in a form of a \emph{computation graph} using some high level language
like Python (in the case of Theano and Caffe) or Lua (in the case of Torch).
Computations are executed on either CPU or GPU.
If they are executed on Nvidia GPU, then they can take advantage of Nvidia CuDNN,
an efficient low level GPU-accelerated library providing implementations of activation functions
and other operations common in neural networks.
All the software libraries mentioned above provide some extension points.
Users are welcome to implement custom activation functions or optimization algorithms. 

More recent deep learning libraries are TensorFlow~\cite{abadi2016tensorflow}, CNTK~\cite{yu2014introduction}
and MXNet~\cite{chen2015mxnet}. The first one is supported by Google, the second by Microsoft
and the third is an effort of a community of academia and industry contributors.
Since TensorFlow was used to conduct experiments described in this dissertation
the more elaborate description of this library is provided in a subsequent section.

The other part of experiments described in this dissertation was carried out using AGH deep learning library~\cite{dlcuda2015},
which is described below as well. Finally, it is worth mentioning that some researches prefer using higher-level neural networks APIs
like Keras\footnote{\url{https://keras.io/}}, which enables build models from predefined blocks.

\subsection{TensorFlow} \label{sec:tensorflow}

According to~\cite{clark2016tensorflow}, TensorFlow in just a half year became by far the most popular
and versatile deep learning library leaving competition behind.
TensorFlow was crated as a way to overcome limitations
of an earlier deep learning library called \emph{DistBelief}~\cite{dean2012large}.
As in the case of most of the deep learning libraries, computations in TensorFlow are defined in a form of a data flow graph.
The graph can be seen as a blueprint for the computations.
It is not immediately executed but only when passed to a given TensorFlow \emph{session}.
The graph consists of operations (dubbed \emph{ops} for short) that are invoked on tensors.
There are three types of tensors: \emph{constants}, \emph{variables} and \emph{placeholders}. 
Constants are not modified during training and can be used to store model hyperparameters.
Variables are often used to store model parameters. They can be modified during the training phase.
Placeholders are tensors which are filled with data at execution time and are often used for passing training examples to the graph.

After the graph is defined and the session is created a user can fill placeholders with some data
and request evaluation of a certain tensor.
Often this operation triggers many others dependent operations in the graph to be executed. 

User can decide where the tensors should be stored and evaluated.
It is possible to run the graph on CPU, GPU, a distributed computing cluster or even on a mobile or embedded device.
Regardless of execution platform TensorFlow tries to parallelize computations as much as possible. 
In the case of CPU and GPU platforms, computation are executed concurrently on multiple cores,
while in the case of the distributed computing cluster on multiple machines.
Moreover, the TensorFlow session enables requesting operation execution from multiple client threads.
That way, huge dataset can be fed to the model concurrently by multiple workers.

Probably the most appealing feature of TensorFlow is its ability to automatically compute gradients for any loss function.
In order to compute gradients one need to select one of the many provided optimizers
and invoke \emph{compute\_gradients} method on it, passing the loss function to it.
There are many standard loss functions available in TensorFlow but users are welcome to implement their own as well.
As a result, gradients for all parameters depending on the loss are computed.
Having gradients one can pass them to the \emph{apply\_gradients} method, which will apply them to the relevant model parameters.
Since those two steps often come together, there is a \emph{minimize} method, which internally performs those two steps.

\begin{figure}[htb!]
  \centering
  \includegraphics[width=1.0\linewidth]{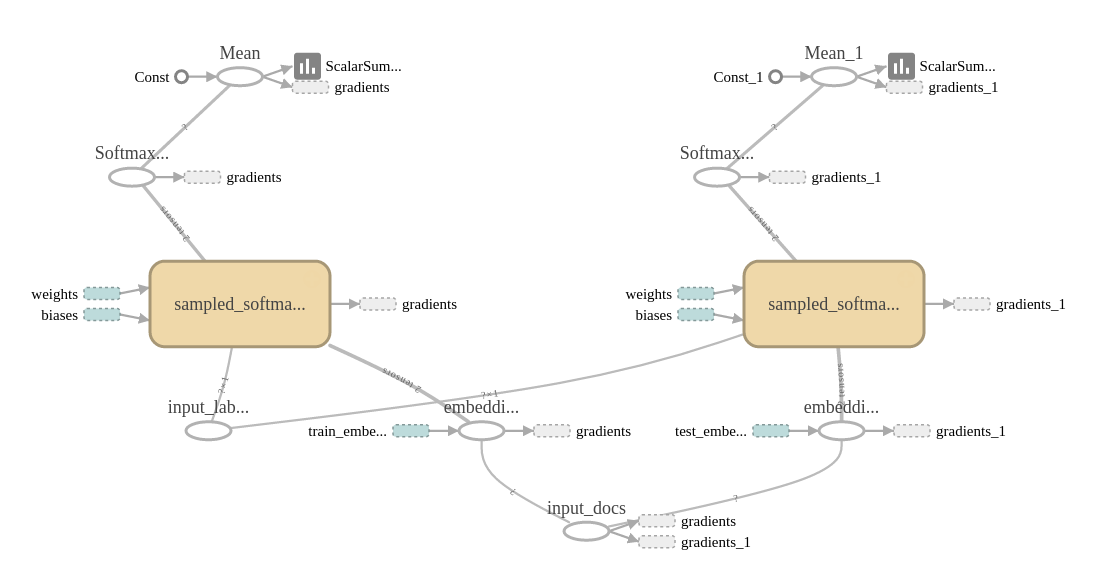}
  \caption{TensorBoard visualization of a PV-DBOW model computation graph.}
  \label{fig:tensorboard}
\end{figure}

TensorFlow is shipped with \emph{TensorBoard}, a user-friendly web application for monitoring,
analysis and debugging of models written using TensorFlow.
In TensorBoard one can display and browse the computation graph, analyze various statistics
and monitor how certain model parameters are changing over time.
For example in~\figref{tensorboard} visualization of PV-DBOW model computation graph is presented.
Two parallel computation flows are visible on the figure.
It is a consequence of the fact that PV-DBOW has two separate learning phases, namely training and inference.
Each computation flow has its own embeddings, cost function and its own set of gradients.
Softmax weight, as well as \emph{input\_docs} and \emph{input\_labels} placeholders are shared among those two phases.

\subsection{AGH deep learning library} \label{sec:dlcuda}

AGH deep learning library is developed at Department of Computer Science, AGH University of Science and Technology.
It supports three types of neural networks models, namely deep belief networks, deep autoencoders and multilayer perceptrons.
The networks can be build of one of the five available layer types: sigmoid, rectified linear,
linear (with optional Gaussian noise), softmax (in the output layer) and constrained Poisson (for the bag-of-words input data).
The library is written in C++11 and exclusively targets Nvidia CUDA platform.
Most of linear algebra operations are delegated to the highly optimized Nvidia CUDA BLAS (cuBLAS) library.
Operations that are not provided in cuBLAS are implemented in a form of \emph{kernel} functions.
Random numbers are generated using Nvidia cuRAND library.
During training, all of the model parameters and network activations are stored in the GPU device memory. 
To ease definition of experiments, a high level Python API is provided. 
Comprehensive description of the library together with performance evaluation is presented in~\cite{dlcuda2015}.
An example of research carried out based on this library is reported in~\cite{sparseInit2015} and~\cite{orthogonality2016}.
We used this library to implement deep autoencoders used to test two simple document representations introduced in \appendixref{improving_bow}.

\chapter{Supplementary material} \label{ch:improving_bow}

Although distributed representations of text data are becoming more and more popular we believe
there is still place for the traditional bag-of-words model.
Its main advantage is simplicity. Unlike dense document representations, word counts can be computed in almost no time.
The other advantage is that some machine learning algorithms or models explicitly require discrete input values.
An example of such a model is the constrained Poisson model~\cite{salakhutdinov2009semantic}.
Also many topic models are build from simple word counts (e.g.~\cite{hinton2009replicated}).
Therefore, we believe that improvements to the bag-of-words model is still a vital research area.
In this appendix we describe preliminary results towards developing two such methods.

\section{Improving the multi-prototype vector-space model with transfer learning} \label{sec:mp_vsm_tl}

One of the limitations of the bag-of-words model is that all senses of a given polysemous or homonymous word
are represented by a single feature. To rectify this, as we discuss in \sectionref{wsd},
a \emph{multi-prototype vector-space model} (MP-VSM)~\cite{reisinger2010multi} can be constructed.
This model uses word sense induction~\cite{schutze1998automatic} to discover word senses.
By and large, word sense induction requires big corpus to work well.
However, this is not always the case. Sometimes we want to create MP-VSM for a relatively small dataset, like 20 Newsgroups.

To address this problem, in this section we suggest to enrich MP-VSM with transfer learning.
Specifically, as in MP-VSM we induce word senses in a target dataset using word occurrences' contexts clustering.
However, unlike in MP-VSM we generate a vectorized representation of context by averaging context's words embeddings
generated on a big external corpus.
After induction, we represent a target dataset in a form of a multiset of those senses.
We call this representation \emph{bag-of-senses}.

Our representation is tested with deep autoencoder.
Deep autoencoders are useful for dimensionality reduction and data compression tasks~\cite{hinton2006reducing}.
After compression, we evaluate the resultant representation on an information retrieval task.
We use two popular text benchmark datasets, namely 20 Newsgroups and RCV1-v2.

\subsection{The bag-of-senses model}

To create the bag-of-senses representation of a target dataset
we first need to generate high-quality word embeddings on an external big text corpus.
We do this using the continuous bag of words~(CBOW)~\cite{mikolov2013efficient} model, but other models could be used as well.
Then, we select some number of popular words, we randomly select some number of occurrences of those words in the target dataset
and we generate vectorized representations of occurrences' contexts by averaging contexts' word vectors scaled by inverse document
frequencies (IDF, defined by \equationref{idf} on page~\pageref{eq:idf}) computed based on the target dataset.
Then, we cluster contexts for each word separately. We use agglomerative hierarchical clustering with the \emph{complete} linkage method.
We have to limit a number of words for which clustering is carried out
and a number of contexts for each word because clustering is a time-consuming operation.
After clustering, branches of a resultant dendrogram are cut off at some level.
The cut-off level can be considered as a model hyperparameter.
For simplicity, we use the same level for all the words.
The cut-off level can be seen as a knob by which we regulate how many senses on average there are.
After that, we build a sense dictionary. For each word in the dictionary we store a list of its senses.
Each sense is represented by a globally unique identifier and a cluster centroid vector.

Having the word sense dictionary constructed we can represent target dataset documents as counts of word senses from the dictionary.
To this end, for each word occurrence in a given document,
its context vectorized representation is generated by averaging context's word embeddings
(the same word embeddings that were used in the sense induction stage).
Then, the context vectorized representation is compared with vectorized representations of each sense of a given word.
The sense with the lowest cosine distance to the context is selected.
A counter for that sense identifier in a document representation is incremented.
This operation is repeated for all word occurrences in all the documents in a dataset.
In the case there is no entry in a dictionary for a given word, this word is omitted.
After processing all the documents, some number of the most frequent senses is selected (e.g. 2000)
and the final target representation is generated, which include only those selected senses.
It is likely that some words will be represented by multiple dimensions (features) in the target representation while others by none.
Finally, it is worth mentioning that we cluster context only based on the training set. To convert test set into bag-of-senses we use the
same cluster representations as for the training set. Also, since we can cluster context for each word separately it can be considered
an embarrassingly parallel problem.

\subsection{Experiments}

We evaluate our approach on two popular text datasets, namely 20 Newsgroups and RCV1-v2.
The datasets are described in \sectionref{datasets}.
Word embeddings are trained on English Wikipedia dump downloaded on April 5th 2016 using the \emph{gensim}~\cite{rehurek2010software} library.
Before embedding generation, stopwords were removed from the corpus.
For both datasets, occurrence contexts were selected for 5000 globally most frequent words.
In the case of 20 Newsgroups, for each word we randomly selected up to 200 documents containing that word
(on average just 90 documents because less popular words occur in less than 100 documents)
and from each document we randomly selected just one context.
By context we mean 5 words before and 5 words after a given word.
Since the RCV1 corpus is much bigger than 20 Newsgroups and all 5000 globally most frequent words
occur in at least 100 documents we decided to randomly select only 100 documents for each word,
which gives us a similar number of contexts to be clustered as in the case of 20 Newsgroups.
Then, we clustered contexts using \emph{hclust} package from R software environment. We used version 3.2.1 of R.
Optimal cut-off level for 20 Newsgroups selected on the validation set is as high as 0.85, which gives only 5199 unique meanings.
For RCV1 it is 0.5, which gives 26037 meanings. Validation set for 20 Newsgroups was build by holding out a quarter of the training set,
which in the case of 20 Newsgroups is given explicitly.
Validation set for RCV1 was constructed by holding out $ 10^4 $ documents from the training set. The training set, in turn, was created
by holding out $ 10^4 $ documents for testing from the entire document collection.
Finally, for both datasets we prepared target representations of datasets having 2000 dimensions each,
corresponding to 2000 globally most common senses.

\begin{figure}[htb!]
  \centering
  \includegraphics[width=1.0\linewidth]{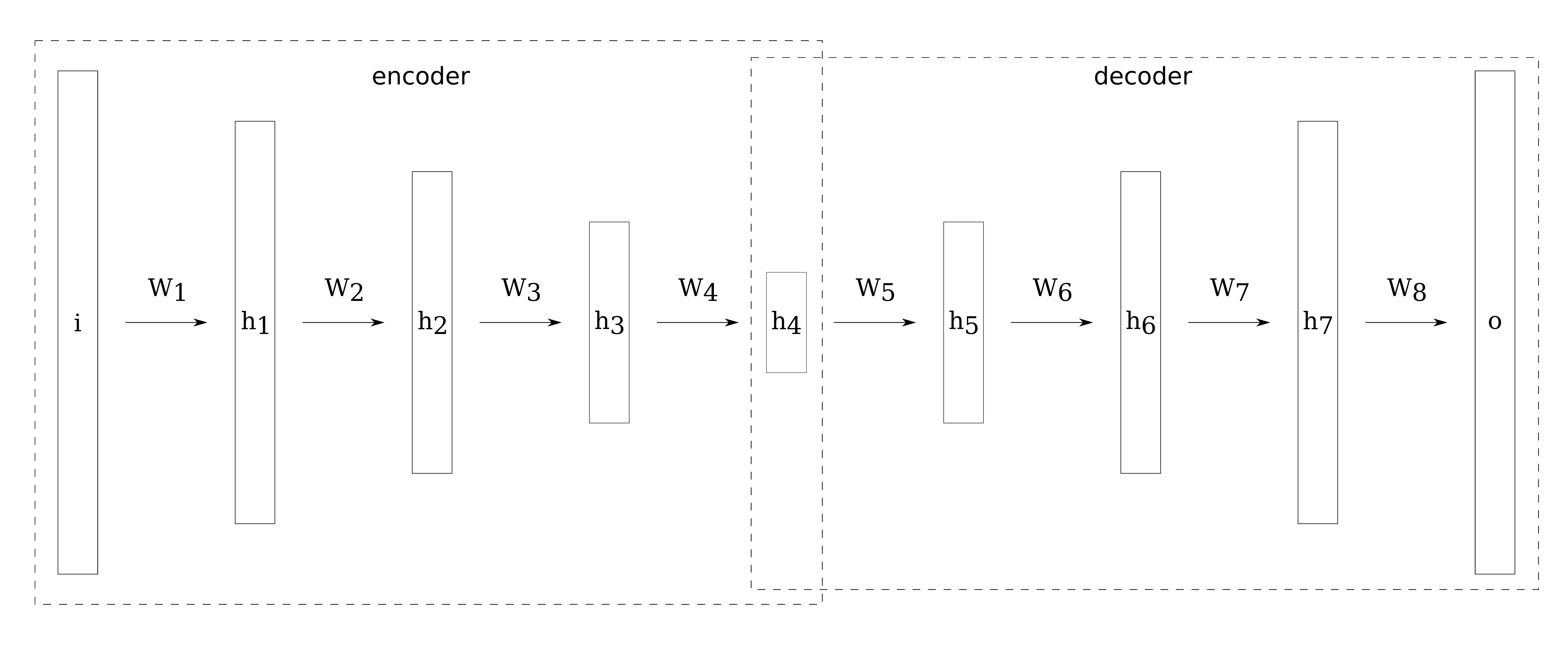}
  \longcaption{A deep autoencoder with hidden layers $ h_1 $ to $ h_7 $ and weight matrices $ W_1 $ to $ W_8 $.}
              {The aim of training is to restore output $ o $ to resemble input $ i $ as closely as possible.
               After the training, an encoder part of the autoencoder can be used to generate
               a low-dimensional representation $ h_4 $ of input data $ i $.}
  \label{fig:deep_autoencoder_big}
\end{figure}

To evaluate the resultant bag-of-senses representation we conducted dimensionality reduction task,
and then performed information retrieval experiments. We used deep autoencoder for dimensionality reduction.
For 20 Newsgroups we used network architecture depicted in~\figref{deep_autoencoder_big} and described in~\cite{orthogonality2016}.
Sizes of encoding layers are: 2000, 500, 250, 125 and 32. All of network hyperparameters were taken from that publication as well.

Deep autoencoder architecture for RCV1 has been taken from~\cite{salakhutdinov2009semantic}.
It is depicted in~\figref{deep_autoencoder} on page~\pageref{fig:deep_autoencoder}. It has one less encoding and decoding layers than
the one used for 20 Newsgroups. The sizes are as follow: 2000, 500, 500 and 128.
Both autoencoders were pre-trained using deep belief network.
Networks were implemented based on the GPU-accelerated AGH deep learning library, described in \sectionref{dlcuda}.

After generating low-dimensional representations of the datasets we perform information retrieval experiments
following the procedure described in \sectionref{dsh_ir}.
All the experiments were carried out on the HP Apollo XL750f Gen9 liquid cooled
HPC machines equipped with two Intel Xeon E5-2680v3 processors, 128 GB RAM and
two Nvidia Tesla K40 GPUs. It takes approximately 30 seconds to cluster contexts for a single word on this machine.

\subsection{Results}

\begin{figure}[htb!]
  \centering
  \begin{subfigure}[b]{0.495\linewidth}
    \centering
    \includegraphics[width=\textwidth]{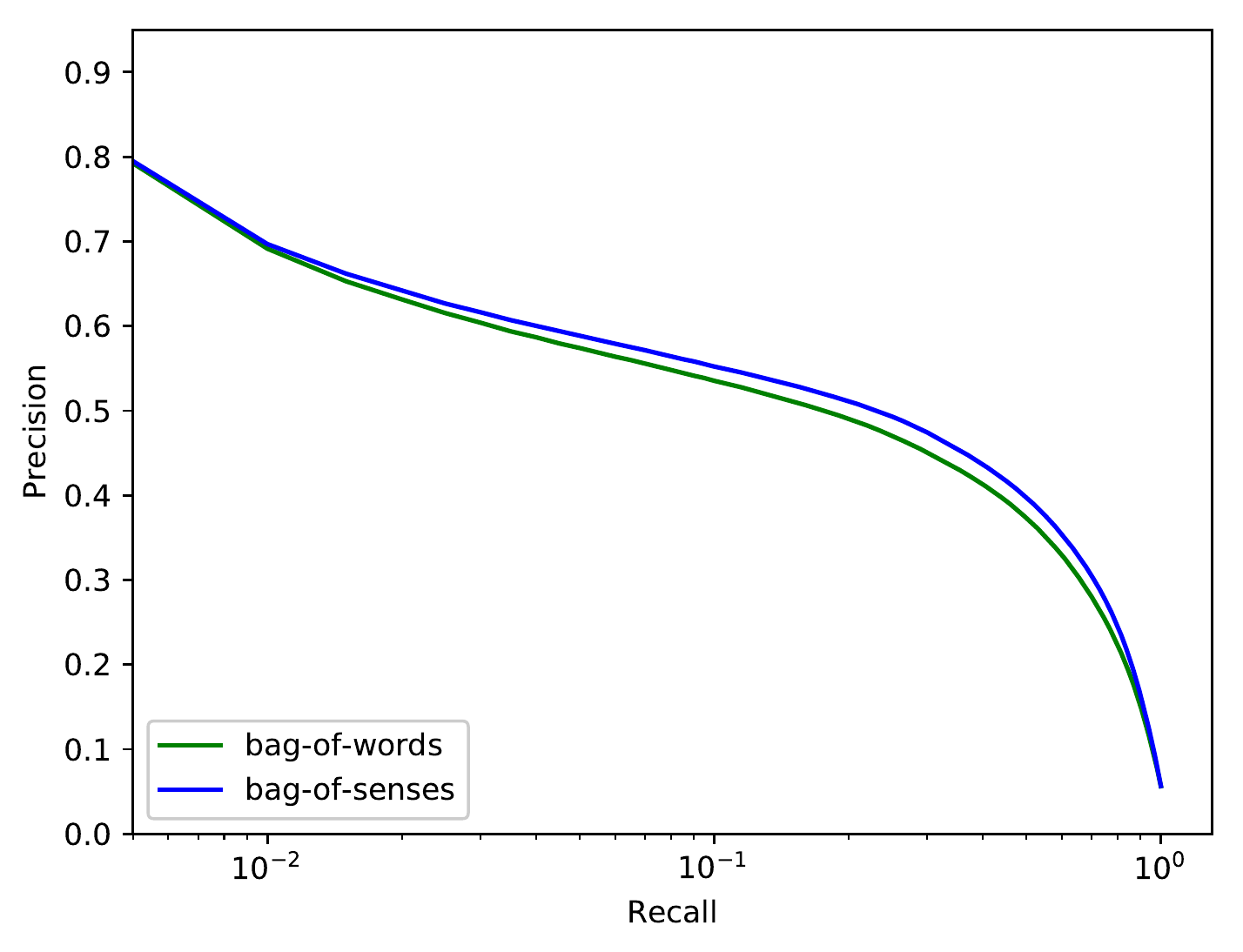}
    \caption{20 Newsgroups}
  \end{subfigure}
  \hfill
  \begin{subfigure}[b]{0.495\linewidth}
    \centering
    \includegraphics[width=\textwidth]{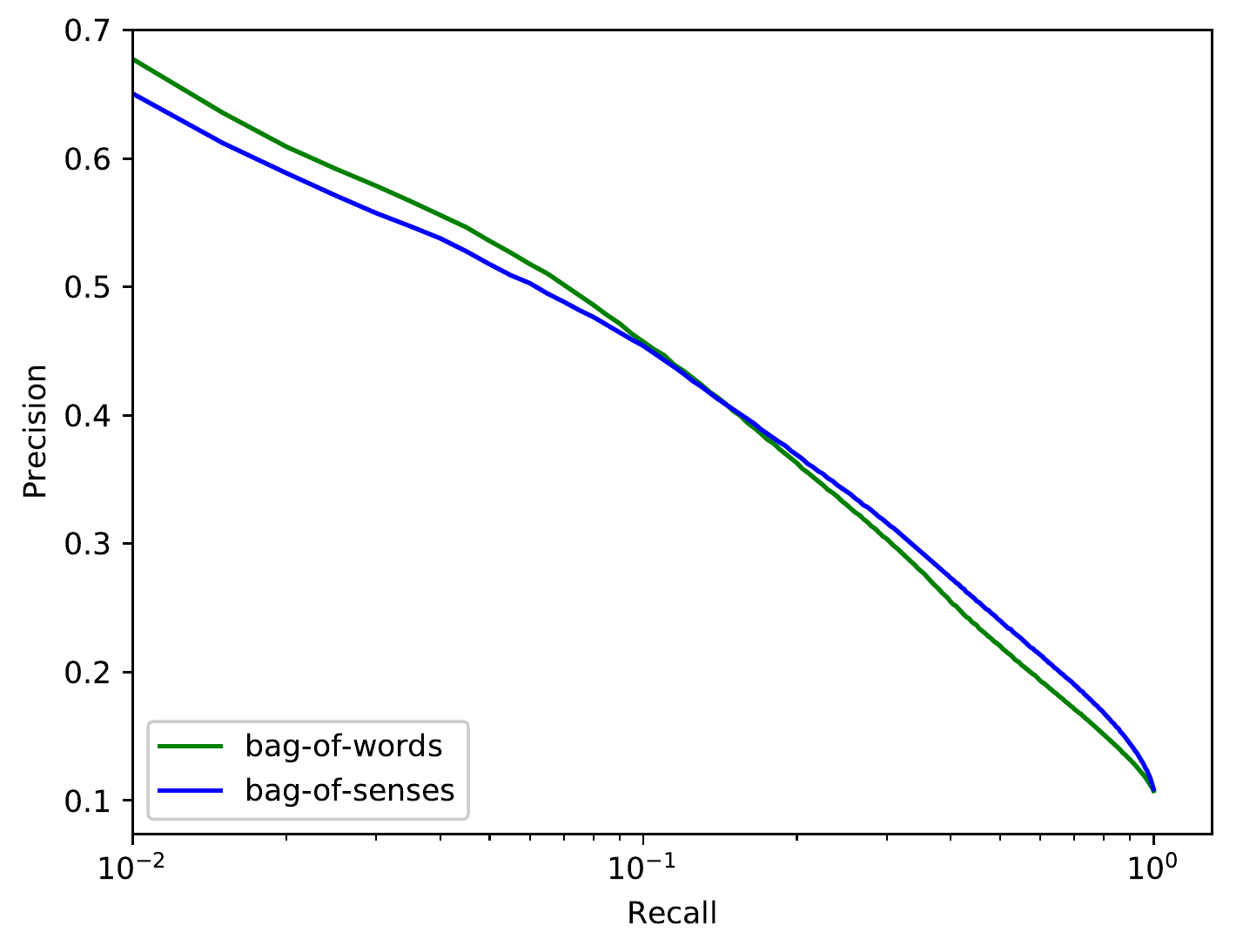}
    \caption{RCV1}
  \end{subfigure}
  \caption{Precision-recall curves for bag-of-senses experiments.}
  \label{fig:bos_precision_recall}
\end{figure}

To evaluate results we use two popular metrics, namely mean average precision (MAP) and normalized discounted cumulative gain
(NDCG)~\cite{jarvelin2002cumulated}. Results for 20 Newsgroups are reported in~\tabref{bos_tng_results} while for RCV1 in~\tabref{bos_rcv1_results}.
\begin{table}[htb]
  \centering
    \begin{tabular}{|c|c|c|c|}
      \hline
      Model         & MAP           & NDCG@10 \\ \hline
      bag-of-words  & 0.36          & 0.64    \\ \hline
      bag-of-senses & \textbf{0.38} & 0.64   \\ \hline
    \end{tabular}
  \longcaption{Results for the 20 Newsgroups dataset represented using 2000-dimensional bag-of-senses model compared against bag-of-words model.}
              {Both models were compressed to 32 dimensions using the same deep autoencoder.}
  \label{tab:bos_tng_results}
\end{table}
\begin{table}[htb]
  \centering
    \begin{tabular}{|c|c|c|c|}
      \hline
      Model         & MAP           & NDCG@10       \\ \hline
      bag-of-words  & 0.22          & 0.74          \\ \hline
      bag-of-senses & \textbf{0.23} & 0.72          \\ \hline
    \end{tabular}
  \longcaption{Results for the RCV1 dataset represented using 2000-dimensional bag-of-senses model compared against bag-of-words model.}
              {Both models were compressed to 128 dimensions using the same deep autoencoder.}
  \label{tab:bos_rcv1_results}
\end{table}
Precision-recall curves are depicted in~\figref{bos_precision_recall}.
We compare our results with a simple bag-of-words representation, where each document is encoded as a multiset of 2000 most frequent words.
In the case of bag-of-words as well as in the case of bag-of-senses we considered simple word sense counters. We did not apply any
normalization or weighting. In the case of both datasets MAP metric value is slightly higher for bag-of-senses than bag-of-words.
In the case or RCV1 a few top results are more accurate in the case of bag-of-words but overall AUC is higher for bag-of-senses.
Since only one of the two important information retrieval metrics yielded systematically superior results, we decided not to declare results
from this section as a part of contribution of this dissertation.

\section{Scaled-up TF-IDF representation}

As we discuss in \sectionref{vsm}, the bag-of-words representation can be enriched by taking into account the fact that some words occur in
a small number of documents and therefore are more meaningful than words that occur in almost all the documents. However, weighting word
counts with IDF shifts document vectors from integer into real-valued space, which sometimes can be considered a limitation.
In this section we propose a simple trick that remedy this. Specifically, we suggest encoding documents as term frequencies with
$ L_2 $ normalization (TF) multiplied by inverse document frequencies (IDF) and by the ratio of the mean (MR) of all word counts
(TF without any normalization) to the mean of all TF-IDF values:
\begin{equation}
MR = \frac{avg(\text{\emph{word\_count}})}{avg(\text{\emph{TF-IDF}})} = \frac{ \sum\limits_{i=1}^{N} \sum\limits_{j=1}^{D} \emph{word\_count} }{ \sum\limits_{i=1}^{N} \sum\limits_{j=1}^{D} \text{\emph{TF-IDF} } },
\end{equation}
where $ N $ is a number of documents in a dataset and $ D $ is a number of dimensions.
In addition, resultant values are discretized (rounded to integers).
This way we obtain a representation in which globally frequent terms are penalized but at the same time
the mean of all the inputs equals a simple word count representation mean.
Therefore, we can feed such document to a network that requires discrete inputs
(e.g. constrained Poisson model~\cite{salakhutdinov2009semantic} or replicated softmax \cite{hinton2009replicated}).
It should be noted that obtained model contains almost the same amount of information as the standard TF-IDF model.
However, scaling it up to the level of word counts make it possible to use it in models which require word counts as an input. 
We call this representation \emph{TF-IDF-MR}.
Since scaling factor MR is common for the whole dataset, computation complexity of preparation of TF-IDF-MR equals TF-IDF.

\subsection{Results}

\begin{figure}[htb!]
  \centering
  \begin{subfigure}[b]{0.495\linewidth}
    \centering
    \includegraphics[width=\textwidth]{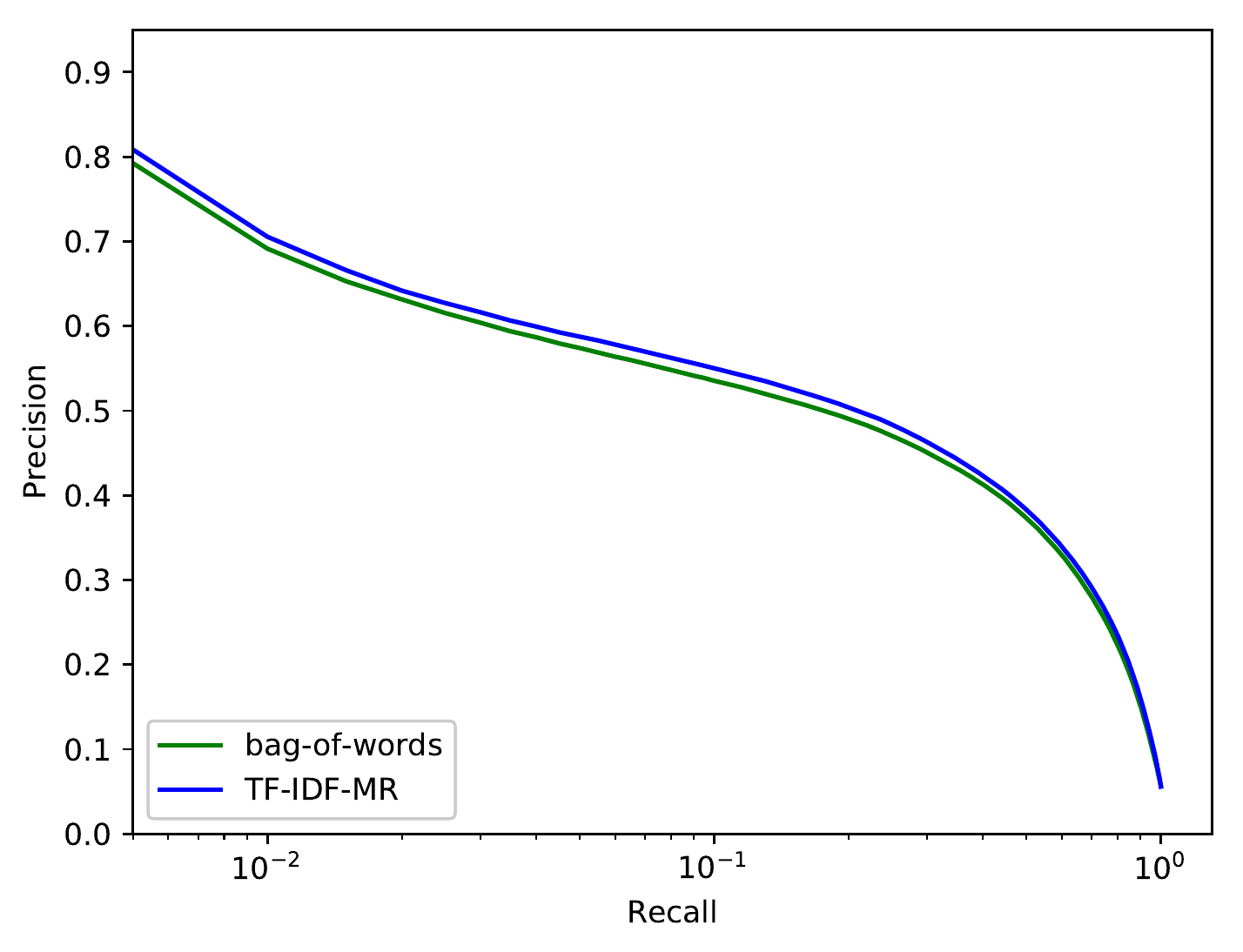}
    \caption{20 Newsgroups}
  \end{subfigure}
  \hfill
  \begin{subfigure}[b]{0.495\linewidth}
    \centering
    \includegraphics[width=\textwidth]{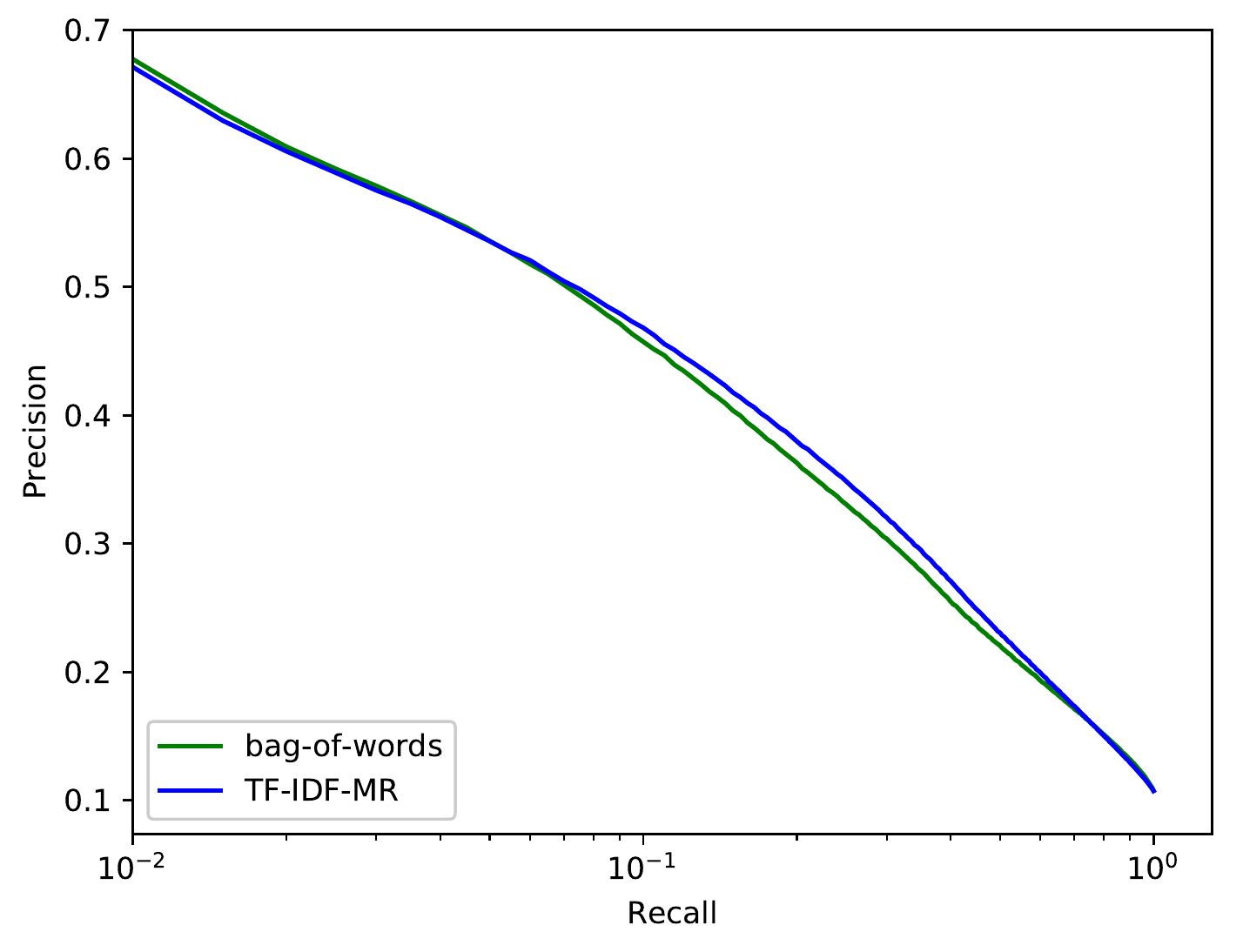}
    \caption{RCV1}
  \end{subfigure}
  \caption{Precision-recall curves for TF-IDF-MR experiments.}
  \label{fig:mr_precision_recall}
\end{figure}

To evaluate the TF-IDF-MR representation we follow the same procedure as in \sectionref{mp_vsm_tl}.
Specifically, we crated 2000-dimensional representations of the 20 Newsgroups and RCV1 datasets,
and then we compressed them to 32 and 128 dimensions, respectively.
We used the same autoencoder architectures with the same hyperparmeters values. Then, we performed information retrieval task.
Results for 20 Newsgroups are reported in~\tabref{mr_tng_results} while for RCV1 in~\tabref{mr_rcv1_results}.
\begin{table}[htb]
  \centering
    \begin{tabular}{|c|c|c|c|}
      \hline
      Model         & MAP           & NDCG@10       \\ \hline
      bag-of-words  & 0.36          & 0.64          \\ \hline
      TF-IDF-MR     & \textbf{0.37} & \textbf{0.65} \\ \hline
    \end{tabular}
  \caption{Results for 20 Newsgroups dataset represented using 2000-dimensional TF-IDF-MR model and compressed to 32 dimensions using deep autoencoder.}
  \label{tab:mr_tng_results}
\end{table}
\begin{table}[htb]
  \centering
    \begin{tabular}{|c|c|c|c|}
      \hline
      Model         & MAP           & NDCG@10       \\ \hline
      bag-of-words  & 0.22          & 0.74          \\ \hline
      TF-IDF-MR     & \textbf{0.23} & 0.74          \\ \hline
    \end{tabular}
  \caption{Results for the RCV1 dataset represented using 2000-dimensional TF-IDF-MR model and compressed to 128 dimensions using deep autoencoder.}
  \label{tab:mr_rcv1_results}
\end{table}
Precision-recall curves are depicted in~\figref{mr_precision_recall}.

As revealed by these figures, for both datasets, TF-IDF-MR yields slightly better results
than simple frequent word multiset representation. To some extent this observation could be considered obvious
because the TF-IDF representation is, by and large, considered a stronger representation than unweighted BoW.
Our contribution lies in showing how TF-IDF can be simply adopted to be used by algorithms which require discrete inputs.
However, the difference in performance between bag-of-words and TF-IDF-MR is so small, that it could be withing margin of error.
Therefore, we decided not to declare those results as a contribution of this dissertation.

\section{Automated blog author profiling}

As discussed many times in this dissertation, dense vectorized representations of text data are finding more and more applications.
One specific problem that, to the best of our knowledge, has not been tackled with the embedding methods is automated blog author profiling.
Blog posts and social media posts are written at a massive scale. Being able to profile authors of those short and informal
pieces of text is crucial. It can give some insight into the population structure and enables selection of user groups for microtargeted marketing and political
campaigns~\cite{schwartz2013personality}.

As a part of this dissertation we wanted to employ the Paragraph Vector models to predict some basic personal information of authors
of blog posts based solely on their content. To this end, we used The Blog Authorship Corpus \footnote{Available
at~\url{http://u.cs.biu.ac.il/~koppel/BlogCorpus.htm}}, which consists of almost $ 7 \times 10^5 $ posts written by almost $ 2 \times 10^4 $ authors.
The posts are gender and age annotated. $ 43 \% $ of the authors are teenagers, $ 42 \% $ are in their 20s and the rest are in their 30s.
Both sexes are equally represented. We wanted to predict gender and an age group. We started with generation of 300-dimensional document
embeddings using Paragraph Vector. Then we trained logistic regression classifier to correctly predict gender and age. Results are presented
in~\tabref{blogs}.
\begin{table}[htb]
  \centering
    \begin{tabular}{|l|c|c|}
      \hline
                                               & Gender                & Age                   \\ \hline
      Baseline (predicting a majority class)   & 0.51                  & 0.47                  \\ \hline
      Prediction based on handcrafted features & \multirow{2}{*}{0.73} & \multirow{2}{*}{0.74} \\
      (from \cite{schler2006effects})          &                       &                       \\ \hline 
      Prediction based on document embeddings  & 0.65                  & 0.61                  \\ \hline
    \end{tabular}
  \caption{Classification accuracy.}
  \label{tab:blogs}
\end{table}
To improve the accuracy we used stratified 3-folds cross-validation. The results are compared against baseline (predicting a majority class)
and reference results from~\cite{schler2006effects}, where the authors used handcrafted features to describe genders and age groups.
Our results are superior to baseline but unfortunately inferior to the reference. The conclusion could be that the task is relatively hard
and in such settings handcrafted features are crucial to describe subtle differences in language style between age groups and genders.

\begin{figure}[htb!]
  \centering
  \begin{subfigure}[b]{0.495\linewidth}
    \centering
    \includegraphics[width=\textwidth]{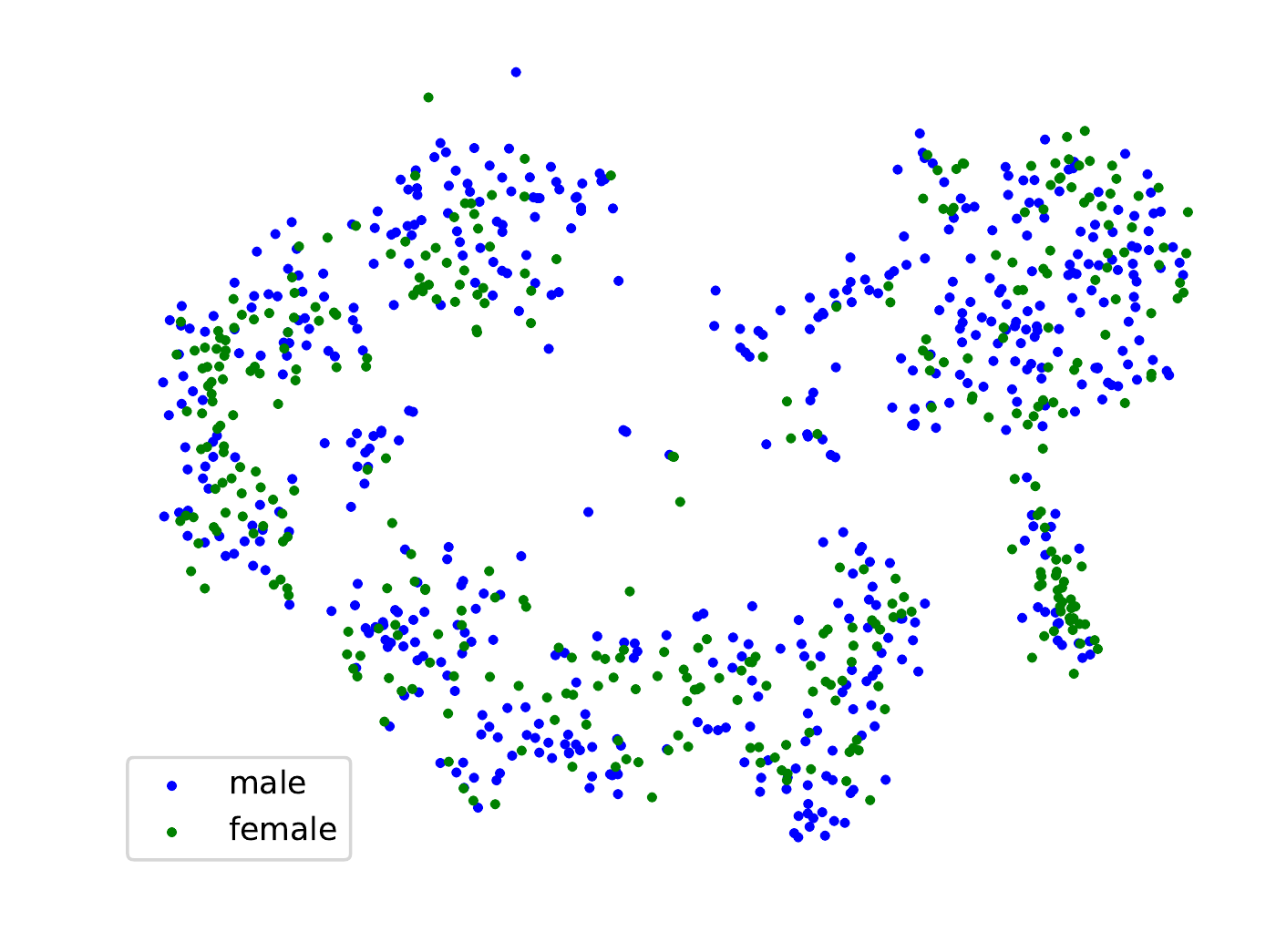}
    \caption{Separated by gender}
  \end{subfigure}
  \hfill
  \begin{subfigure}[b]{0.495\linewidth}
    \centering
    \includegraphics[width=\textwidth]{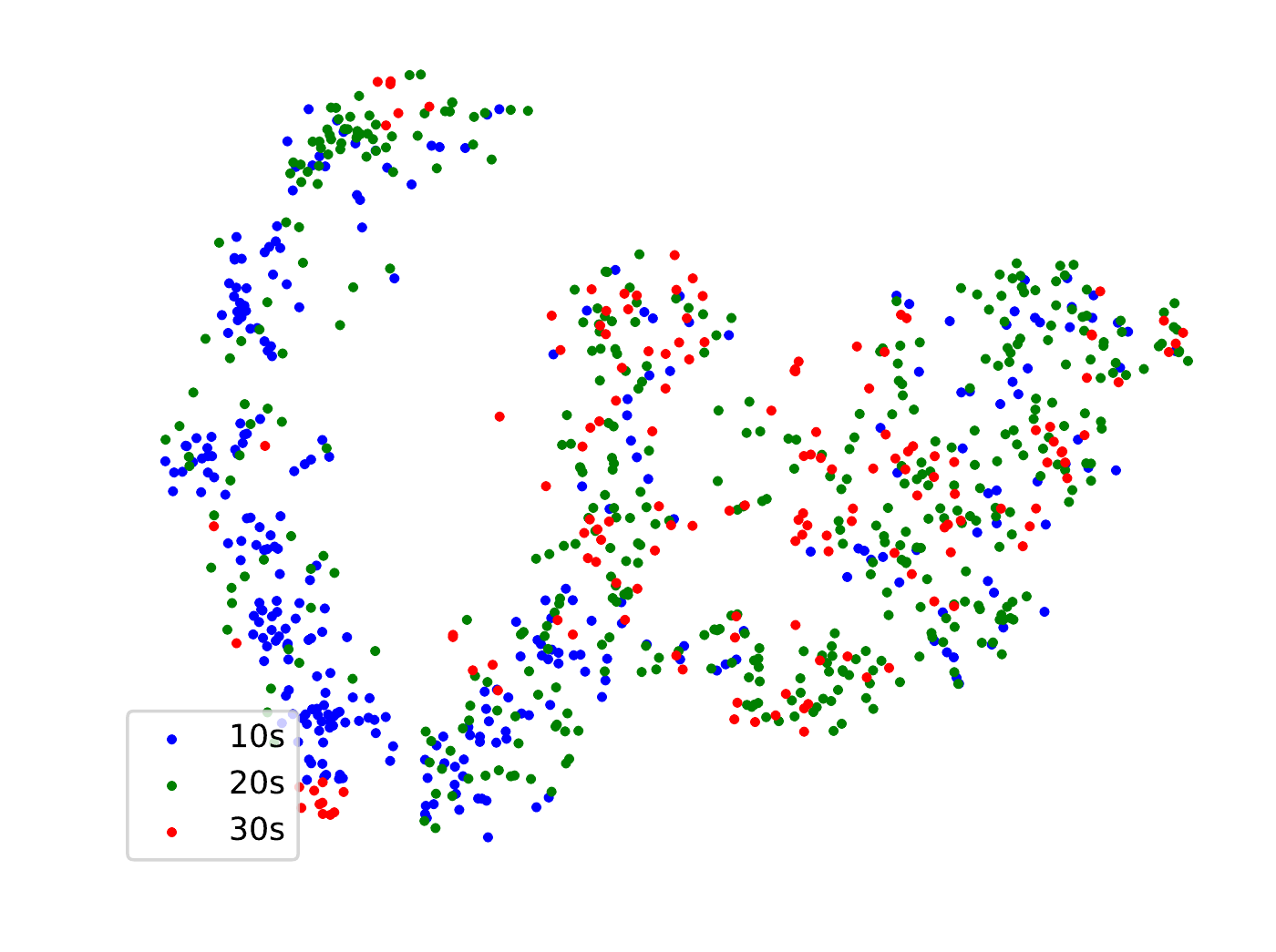}
    \caption{Separated by an age group.}
  \end{subfigure}
  \caption{t-SNE visualizations of blog posts embeddings.}
  \label{fig:blogs}
\end{figure}

To qualitatively evaluate the generated blog embeddings we compressed them to 2D using t-SNE and placed them on the plane. Visualizations
are depicted in~\figref{blogs}.
There is not much difference in gender. There is some difference in age. Teenage authors are separated from those in their 20s and 30s.

\section{Conclusions}

In this appendix we presented two simple tricks that lead to slight improvement in experiments relying on the bag-of-word representation.
Those tricks could be used in processing pipeline downstream algorithms requiring discrete input. Important exemplars of those algorithm are
topic models.

\bibliography{bibliography}

\end{document}